\documentclass{sig-alternate-2013}
\permission{}
\conferenceinfo{}{}
\copyrightetc{\noindent\hrulefill\\\\
The work of Ping Li and Anshumali Shrivastava was partially supported by NSF-III-1360971, AFOSR-FA9550-13-1-0137, and
NSF-Bigdata-1419210.}
\crdata{ }

\usepackage{times,url,mathrsfs,color}
\usepackage{amsmath,wrapfig,color,bigfoot,enumerate}
\usepackage{amsfonts}
\usepackage{graphicx}
\usepackage{subfigure}
\usepackage{epstopdf}
\newtheorem{theorem}{Theorem}

\begin{document}
%

\title{2-Bit Random Projections, NonLinear Estimators, and Approximate Near Neighbor Search}
%
%
%
%
%

\numberofauthors{3} 
%
\author{
%
%
\alignauthor
{Ping Li}\\
       \affaddr{Department of Statistics}\\
       \affaddr{Dept. of Computer Science}\\
       \affaddr{Rutgers University}\\
       \affaddr{Piscataway, NJ 08854, USA}\\
       \email{pingli@stat.rutgers.edu}
\alignauthor
Michael Mitzenmacher\\
       \affaddr{School of Engineering and Applied Sciences}\\
       \affaddr{Harvard University}\\
       \affaddr{Cambridge, MA 02138, USA}\\
       \email{michaelm@eecs.harvard.edu}
\alignauthor
{Anshumali Shrivastava}\\
       \affaddr{Dept. of Computer Science}\\
       \affaddr{Rice University}\\
      \affaddr{Houston, TX 77005, USA}\\
       \email{anshumali@rice.edu}
}

\maketitle
\begin{abstract}

\noindent The method of random projections has become a standard tool for machine learning, data mining, and search with massive data at Web scale. The effective use of random projections requires efficient coding schemes for quantizing (real-valued) projected data into integers. In this paper, we focus on  a simple 2-bit coding scheme. In particular, we develop accurate \textbf{nonlinear estimators} of  data similarity based on the 2-bit strategy. This work will have important practical applications. For example, in the task of near neighbor search, a crucial step (often called {\bf re-ranking})  is  to compute or estimate  data similarities once a set of candidate data points have been identified  by hash table techniques. This re-ranking step can take advantage of the proposed coding scheme and estimator.

\vspace{0.07in}

\noindent As a related task, in this paper, we also study a simple uniform quantization scheme for the purpose of building hash tables with projected data. Our analysis shows that typically only a small number of bits are needed. For example, when the target similarity level is high,  2 or 3 bits might be  sufficient. When the target similarity level is not so high, it is preferable to use only 1 or 2 bits. Therefore, a 2-bit scheme appears to be  overall a good choice for the task of sublinear time approximate near neighbor search via hash tables.

\vspace{0.07in}

\noindent Combining these results, we  conclude that 2-bit random projections should be recommended for approximate near neighbor search and similarity estimation. Extensive experimental results are  provided.

\end{abstract}

\vspace{-0.05in}
\section{Introduction}

Computing (or estimating) data similarities is a fundamental  task in numerous practical applications. The popular method of random projections provides a potentially effective strategy for estimating data similarities (correlation or Euclidian distance) in massive high-dimensional datasets, in a memory-efficient manner. Approximate near neighbor search is a typical example of those applications.

The  task of {near neighbor search} is to identify a set of data points which are ``most similar'' (in some measure of similarity) to a query data point. Efficient algorithms for near neighbor search  have  numerous applications in search, databases, machine learning,  recommender systems, computer vision, etc. Developing efficient algorithms for finding near neighbors has been an active  research topic since the early days of modern computing~\cite{Article:Friedman_75}. Near neighbor search with extremely high-dimensional data (e.g., texts or images) is still a challenging task and an active research problem.

In the specific setting of the World Wide Web, the use of hashing and random projections for applications such as detection of near-duplicate Web pages dates back to (e.g.,)~\cite{Proc:Broder_WWW97,Proc:Charikar}.  The work in this area has naturally continued, improved, and expanded;  see, for example, ~\cite{Proc:Casey_ISMIR06,Proc:Henzinger_SIGIR06,Proc:Bayardo_WWW07,Proc:Hajishirzi_SIGIR10,Proc:Duan_ISWC12,Proc:Kong_SIGIR12,Proc:Kulis_ICCV09,Proc:Li_Konig_WWW10,Proc:Leng_SIGIR14,Proc:Mitzenmacher_WWW14} for research  papers with newer results on the theoretical frameworks, performance, and applications for such methods.  In particular, such techniques have moved beyond near-duplicate detection and retrieval to detection and retrieval for more complex data types, including images and videos.  Our work continues on this path;  specifically, we seek to obtain accurate similarity scores using very small-memory random projections, for applications where the goal is to determine similar objects, or equivalently nearest neighbors in a well-defined space.\\

\vspace{-0.11in}
\subsection{Data Correlation}

Among many types of similarity measures, the (squared) Euclidian distance (denoted by $d$) and the correlation (denoted by $\rho$) are most commonly used. Without loss of generality, consider two high-dimensional data vectors $u, v\in\mathbb{R}^D$. The squared Euclidean distance and  correlation are defined as follows:
\begin{align}\notag
d = \sum_{i=1}^D |u_i - v_i|^2, \hspace{0.2in} \rho =  \frac{\sum_{i=1}^Du_iv_i}{\sqrt{\sum_{i=1}^D u_i^2} \sqrt{\sum_{i=1}^D v_i^2} }
\end{align}

The correlation  $\rho$ is nicely normalized between -1 and 1. For convenience, this study will assume that the marginal $l_2$ norms $\sum_{i=1}^D |u_i|^2$ and $\sum_{i=1}^D |v_i|^2$ are known. This is a often reasonable assumption~\cite{Proc:Li_Hastie_Church_COLT06}, as computing the marginal $l_2$ norms only requires scanning the data once, which is anyway needed during the data collection process. In machine learning practice, it is  common  to first normalize the data before feeding the data to classification (e.g., SVM) or clustering (e.g., K-means) algorithms. Therefore, for convenience, throughout this paper, we  assume  unit $l_2$ norms:
\begin{align}\notag
&\rho =  \sum_{i=1}^D u_iv_i,\hspace{0.3in} \text{where } \ \ \sum_{i=1}^D u_i^2 = \sum_{i=1}^D v_i^2 = 1
\end{align}

\vspace{-0.13in}
\subsection{Random Projections and   Quantization}

As an effective tool for dimensionality reduction, the idea  of random projections is to multiply the data, e.g.,   $u, v\in\mathbb{R}^D$, with a random normal projection matrix $\mathbf{R}\in\mathbb{R}^{D\times k}$, to generate:
\begin{align}\notag
&x = u\times \mathbf{R} \in\mathbb{R}^k,\hspace{0.2in} y = v\times \mathbf{R} \in\mathbb{R}^k, \\\notag
&\mathbf{R} = \{r_{ij}\}{_{i=1}^D}{_{j=1}^k}, \hspace{0.2in} r_{ij} \sim N(0,1) \text{ i.i.d. }
\end{align}
This method has become popular for large-scale machine learning applications such as classification, regression, matrix factorization, singular value decomposition, near neighbor search, bio-informatics, etc.~\cite{Proc:Papadimitriou_PODS98,Proc:Dasgupta_FOCS99,Proc:Bingham_KDD01,Article:Buher_Tompa,Proc:Fradkin_KDD03,Book:Vempala,Proc:Dasgupta_UAI00,Article:JL84,Proc:Wang_Li_SDM10}.

The projected data  ($x_j = \sum_{i=1}^D u_i r_{ij}$, $y_j = \sum_{i=1}^D v_i r_{ij}$) are real-valued. For many applications it is however crucial to quantize them into integers. The quantization step is  in fact mandatary if the projected data are used for the purpose of indexing  and/or sublinear time near neighbor search (e.g.,) in the framework of {\em locality sensitive hashing (LSH)}~\cite{Proc:Indyk_STOC98}.

Another strong motivation for quantization is for reducing memory consumption. If only a few (e.g., 2) bits suffice for producing accurate estimate of the similarity, then we do not need to store the entire (e.g., 32 or 64 bits)  real-valued projection data. This would be a very significant cost-saving in storage as well as computation.

%

\vspace{0.08in}

In this paper, we focus on 2-bit coding and estimation for multiple reasons.  As analyzed in Section~\ref{sec_LSH}, the 2-bit coding appears to provide an overall good scheme for building hash tables in near neighbor search. The focus of this paper is on developing accurate nonlinear estimators, which are typically computationally quite expensive. Fortunately, for 2-bit coding, it is still feasible to find the numerical solution fairly easily, for example, by tabulation.

\section{2-Bit Random Projections}

Given two (high-dimensional) data vectors $u, v\in\mathbb{R}^D$, we generate two projected values $x$ and $y$ as follows:
\begin{align}\notag
x = \sum_{i=1}^D x_i r_i,\hspace{0.2in} y = \sum_{i=1}^D x_i r_i,\hspace{0.2in} r_i \sim N(0,1)\hspace{0.2in} i.i.d.
\end{align}
Assuming that the original data $u$, $v$ are normalized to  unit $l_2$ norm, the projected data $(x,y)$ follow a bivariate normal distribution:
\begin{align}
\left[
\begin{array}{c}
x\\
y
\end{array}
\right] \sim N\left(
\left[
\begin{array}{c}
0\\
0
\end{array}
\right],\ \  \Sigma =
\left[
\begin{array}{cc}
1 &\rho\\
\rho &1
\end{array}
\right]
\right)
\end{align}
Note that when using random projections in practice, we will need  (e.g.,) $k=200\sim 2000$ independent projections, depending on applications; and we will use $x_j$, $y_j$, $j=1$ to $k$, to denote them.

\begin{figure}[h!]
\begin{center}
\includegraphics[width = 3.5in]{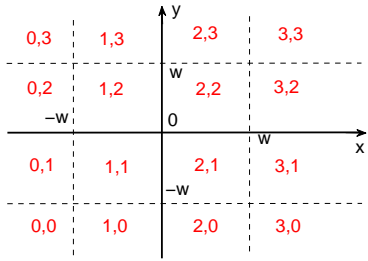}
\end{center}
\vspace{-0.3in}
\caption{2-bit random projections.}\label{fig_16region}
\end{figure}

As the projected data $(x,y)$ are real-valued, we will have to quantize them either for  indexing or for achieving compact storage.  Figure~\ref{fig_16region} pictures the 2-bit coding scheme after random projections. Basically, a random projection value $x$ is mapped to an integer $\in \{0, 1, 2, 3\}$ according to a threshold $w$ (and $-w$).

As shown in Figure~\ref{fig_16region}, the  space is divided into 16 regions according to the pre-determined threshold $w$.  To fully exploit the  information, we  need to jointly analyze the probabilities in all 16 regions.  We will see that  the analysis  is quite involved.

\vspace{0.08in}

The first step of the analysis is to compute the probability of each region. Fortunately, due to symmetry (and asymmetry), we just need to conduct the computations for three regions:
\begin{align}\notag
&P_{2,2}(\rho,w) = \mathbf{Pr}\left\{\text{Region (2,2)}\right\}= \int_0^w \int_0^w f(x,y) dx dy,\\\notag
&P_{2,3}(\rho,w) = \mathbf{Pr}\left\{\text{Region (2,3)}\right\}= \int_0^w \int_w^\infty f(x,y) dx
dy,\\\notag
&P_{3,3}(\rho,w) = \mathbf{Pr}\left\{\text{Region (3,3)}\right\}= \int_w^\infty \int_w^\infty f(x,y) dx
dy.
\end{align}
Due to symmetry, the probabilities of other regions are simply
\begin{align}\notag
&P_{3,2}(\rho,w) = P_{0,1}(\rho,w) = P_{1,0}(\rho,w) = P_{2,3}(\rho,w),\\\notag
&P_{2,0}(\rho,w) = P_{3,1}(\rho,w) = P_{0,2}(\rho,w) = P_{1,3}(\rho,w) =P_{2,3}(-\rho,w),\\\notag
&P_{1,1}(\rho,w) = P_{2,2}(\rho,w), \hspace{0.05in} P_{1,2}(\rho,w) = P_{2,1}(\rho,w) = P_{2,2}(-\rho,w),\\\notag
&P_{0,0}(\rho,w) = P_{3,3}(\rho,w), \hspace{0.05in} P_{0,3}(\rho,w) = P_{3,0}(\rho,w) = P_{3,3}(-\rho,w).
\end{align}

\subsection{Region Probabilities and Their Derivatives}

We use the following standard notation for the normal  distribution pdf $\phi(x)$ and cdf $\Phi(x)$:
\begin{align}\notag
\phi(x) = \frac{1}{\sqrt{2\pi}} e^{-\frac{x^2}{2}}, \hspace{0.3in} \Phi(x) = \int_{-\infty}^x \phi(x) dx.
\end{align}
After some tedious calculations (which are skipped), the probabilities of the three regions are
\begin{align}\notag
&P_{2,2}(\rho,w)=\int_{0}^{w}\phi(x)\left[ \Phi\left(\frac{w-\rho x}{\sqrt{1-\rho^2}}\right)- \Phi\left(\frac{-\rho x}{\sqrt{1-\rho^2}}\right)\right]dx\\\notag
&P_{2,3}(\rho,w)=\int_{0}^{w}\phi(x)\Phi\left(\frac{-w+\rho x}{\sqrt{1-\rho^2}}\right)dx,\\\notag
&P_{3,3}(\rho,w) = \frac{1}{4}+\frac{\arcsin\rho}{2\pi}-P_{22}(\rho,w)-2P_{23}(\rho,w)
\end{align}
Their first derivatives (with respect to $\rho$) are
\begin{align}\notag
&P_{2,2}^\prime=\frac{\partial P_{2,2}(\rho,w)}{\partial \rho}=\frac{1}{2\pi}\frac{1}{\sqrt{1-\rho^2}}\left[1 - 2e^{-\frac{w^2}{2(1-\rho^2)}} + e^{-\frac{w^2}{1+\rho}}\right]\\\notag
&P_{2,3}^\prime=\frac{\partial P_{2,3}(\rho,w)}{\partial \rho} =\frac{1}{2\pi}\frac{1}{\sqrt{1-\rho^2}}\left[e^{-\frac{w^2}{2(1-\rho^2)}} - e^{-\frac{w^2}{1+\rho}}
 \right]\\\notag
 &P_{3,3}^\prime=\frac{\partial P_{3,3}(\rho,w)}{\partial \rho} =\frac{1}{2\pi}\frac{1}{\sqrt{1-\rho^2}}e^{-\frac{w^2}{1+\rho}}
 \end{align}
 Their second derivatives are
 \begin{align}\notag
&P_{2,2}^{\prime\prime}=\frac{\partial^2 P_{2,2}(\rho,w)}{\partial \rho^2} =\frac{1}{2\pi}\frac{\rho}{(1-\rho^2)^{3/2}}\\\notag
&\hspace{0.5in}-\frac{1}{2\pi}\frac{2\rho}{(1-\rho^2)^{3/2}}e^{-\frac{w^2}{2(1-\rho^2)}}\left[1-\frac{w^2}{1-\rho^2}\right]\\\notag
&\hspace{0.5in}+\frac{1}{2\pi}\frac{1}{\sqrt{1-\rho^2}}e^{-\frac{w^2}{1+\rho}}
\left[\frac{\rho}{1-\rho^2}+\frac{w^2}{(1+\rho)^2}\right]\\\notag
&P_{2,3}^{\prime\prime}=\frac{\partial^2 P_{2,3}(\rho,w)}{\partial \rho^2} =\frac{1}{2\pi}\frac{\rho}{(1-\rho^2)^{3/2}}e^{-\frac{w^2}{2(1-\rho^2)}}\left[1-\frac{w^2}{1-\rho^2}\right]\\\notag
&\hspace{1in}-\frac{1}{2\pi}\frac{1}{\sqrt{1-\rho^2}}e^{-\frac{w^2}{1+\rho}}
\left[\frac{\rho}{1-\rho^2}+\frac{w^2}{(1+\rho)^2}\right]\\\notag
&P_{3,3}^{\prime\prime}=\frac{\partial^2 P_{3,3}(\rho,w)}{\partial \rho^2} =\frac{1}{2\pi}\frac{1}{\sqrt{1-\rho^2}}e^{-\frac{w^2}{1+\rho}}
\left[\frac{\rho}{1-\rho^2}+\frac{w^2}{(1+\rho)^2}\right]
\end{align}

Because $\rho$ is bounded, we can tabulate the above probabilities and their derivatives for the entire range of $\rho$ and selected $w$ values. Note that in practice, we anyway have to first specify a $w$. In other words, the computations of the probabilities and derivatives are a simple matter of efficient table look-ups.

\subsection{Likelihood}

Suppose we use  in total $k$ projections. Due to symmetry (as shown in Figure~\ref{fig_16region}), the log-likelihood is a sum of 6 terms (6 cells).
\begin{align}\notag
&l(\rho,w) = \sum_{i,j} k_{i,j}\log P_{i,j}\left(\rho,w\right)\\\notag
&=\left(k_{2,2}+k_{1,1}\right)\log{P_{2,2}(\rho,w)}\\\notag
& + \left(k_{2,3}+k_{3,2}+k_{0,1}+k_{1,0}\right)\log{P_{2,3}(\rho,w)}\\\notag
&+\left(k_{3,3}+k_{0,0}\right)\log{P_{3,3}(\rho,w)}
+ \left(k_{1,2}+k_{2,1}\right)\log{P_{2,2}(-\rho,w)}\\\notag
&+ \left(k_{0,2}+k_{1,3}+k_{2,0}+k_{3,1}\right)\log{P_{2,3}(-\rho,w)}\\\notag
& +  \left(k_{0,3}+k_{3,0}\right)\log{P_{3,3}(-\rho,w)}
\end{align}
Corresponding to Figure~\ref{fig_16region}, $k_{1,1}$ is the number of observations (among $k$ observations) in the region (1,1). $k_{0,0}$, $k_{0,1}$ etc are defined similarity. Note that there is a natural constraint:
\begin{align}\notag
k =& \left(k_{2,2}+k_{1,1}\right)+ \left(k_{2,3}+k_{3,2}+k_{0,1}+k_{1,0}\right)
+\left(k_{3,3}+k_{0,0}\right)\\\notag
+& \left(k_{1,2}+k_{2,1}\right)
+ \left(k_{0,2}+k_{1,3}+k_{2,0}+k_{3,1}\right)+  \left(k_{0,3}+k_{3,0}\right)
\end{align}
In other words, this 6-cell problem only has 5 degrees of freedom. In fact, we can also choose to collapse some cells together to reduce this to an even smaller problem. For example, later we will show that if we reduce the 6-cell problem to a  5-cell problem, the estimation accuracy will not be affected  much.

There are more than one way to solve the MLE which maximizes the likelihood $l(\rho,w)$, for finding $\rho$.  Note that this is merely a one-dimensional optimization problem (at a fixed $w$) and we can tabulate all the probabilities (and their derivatives). In other words, it is not a difficult problem. We can do binary search, gradient descent, Newton's method, etc. Here we provide the first and second derivatives of $l(\rho,w)$. The first derive is
\begin{align}\notag
&l^\prime(\rho,w) = \frac{\partial l(\rho,w)}{\partial \rho}=\left(k_{2,2}+k_{1,1}\right)\frac{P_{2,2}^\prime(\rho,w)}{P_{2,2}(\rho,w)}\\\notag
 &+ \left(k_{2,3}+k_{3,2}+k_{0,1}+k_{1,0}\right)\frac{P_{2,3}^\prime(\rho,w)}{P_{2,3}(\rho,w)}\\\notag
&+\left(k_{3,3}+k_{0,0}\right)\frac{P_{3,3}^\prime(\rho,w)}{P_{3,3}(\rho,w)}
- \left(k_{1,2}+k_{2,1}\right)\frac{P_{2,2}^\prime(-\rho,w)}{P_{2,2}(-\rho,w)}\\\notag
&-\left(k_{0,2}+k_{1,3}+k_{2,0}+k_{3,1}\right)\frac{P_{2,3}^\prime(-\rho,w)}{P_{2,3}(-\rho,w)}\\\notag
& -  \left(k_{0,3}+k_{3,0}\right)\frac{P_{3,3}^\prime(-\rho,w)}{P_{3,3}(-\rho,w)}
\end{align}
and the second derivative is
\begin{align}\notag
&l^{\prime\prime}\left(\rho\right) =
\left(k_{2,2}+k_{1,1}\right)\frac{P_{2,2}^{\prime\prime}(\rho,w)P_{2,2}(\rho,w) - \left(P_{2,2}^{\prime}(\rho,w)\right)^2}{\left(P_{2,2}(\rho,w)\right)^2} \\\notag
&+ \left(k_{2,3}+k_{3,2}+k_{0,1}+k_{1,0}\right)\frac{P_{2,3}^{\prime\prime}(\rho,w)P_{2,3}(\rho,w) - \left(P_{2,3}^{\prime}(\rho,w)\right)^2}{\left(P_{2,3}(\rho,w)\right)^2} \\\notag
&+\left(k_{3,3}+k_{0,0}\right)\frac{P_{3,3}^{\prime\prime}(\rho,w)P_{3,3}(\rho,w) - \left(P_{3,3}^{\prime}(\rho,w)\right)^2}{\left(P_{3,3}(\rho,w)\right)^2} \\\notag
&+\left(k_{1,2}+k_{2,1}\right)\frac{P_{2,2}^{\prime\prime}(-\rho,w)P_{2,2}(-\rho,w)-\left(P_{2,2}^\prime(-\rho,w)\right)^2}{\left(P_{2,2}(-\rho,w)\right)^2}\\\notag
&+\left(k_{0,2}+k_{1,3}+k_{2,0}+k_{3,1}\right)\frac{P_{2,3}^{\prime\prime}(-\rho,w)P_{2,3}(-\rho,w) - \left(P_{2,3}^{\prime}(-\rho,w)\right)^2}{\left(P_{2,3}(-\rho,w)\right)^2} \\\notag
&+\left(k_{0,3}+k_{3,0}\right)\frac{P_{3,3}^{\prime\prime}(-\rho,w)P_{3,3}(-\rho,w) - \left(P_{3,3}^{\prime}(-\rho,w)\right)^2}{\left(P_{3,3}(-\rho,w)\right)^2}
\end{align}

If we use  Newton's method, we can find the solution iteratively from $\rho^{(t)} = \rho^{(t-1)} - \frac{l^\prime(\rho)}{l^{\prime\prime}(\rho)}$, by starting from a good guess, e.g., the estimate using 1-bit information. Normally a small number of iterations will be sufficient. Recall that these derivatives and second derivatives are pre-computed and stored in look-up tables.

\vspace{0.08in}

For this particular 2-bit coding scheme, it is possible to completely avoid the numerical procedure by further exploiting look-up table tricks.  Suppose we tabulate the MLE results for each $k_{i,j}/k$, spaced at 0.01. Then a 6-cell scheme would only require $O\left(10^{10}\right)$ space, which is not too large.  (Recall there are only 5 degrees of freedom). If we adopt a 5-cell scheme, then the space would be reduced to $O(10^8)$.  Of course, if we hope to use more than 2 bits, then we can not avoid  numerical computations.

\subsection{Fisher Information and\\ Asymptotic Variance of the MLE}

The asymptotic (for large $k$) variance of the MLE (i.e., the $\rho$ which maximizes the log likelihood $l(\rho,w)$) can be computed from classical statistical estimation theory. Denote the MLE by $\hat{\rho}_{2,MLE}$. Then its asymptotic variance should be
\begin{align}
Var\left(\hat{\rho}_{2,MLE}\right) = \frac{1}{I_{2,\rho,w}} + O\left(\frac{1}{k^2}\right)
\end{align}
where $I_{2,\rho,w} = -E(l^{\prime\prime}(\rho))$ is the {\em Fisher Information}.
\begin{theorem}\label{thm_2bit_Info}
The Fisher Information is
\begin{align}\label{eqn_Fisher2}
&I_{2,\rho,w}
= 2k\left[ A\right],\hspace{0.2in}\text{ where}\\\notag
&A = \frac{\left(P_{2,2}^\prime(\rho,w)\right)^2}{P_{2,2}(\rho,w)} + 2\frac{\left(P_{2,3}^\prime(\rho,w)\right)^2}{P_{2,3}(\rho,w)}
+\frac{\left(P_{3,3}^\prime(\rho,w)\right)^2}{P_{3,3}(\rho,w)}\\\notag
& +  \frac{\left(P_{2,2}^\prime(-\rho,w)\right)^2}{P_{2,2}(-\rho,w)}
+ 2\frac{\left(P_{2,3}^\prime(-\rho,w)\right)^2}{P_{2,3}(-\rho,w)} +  \frac{\left(P_{3,3}^\prime(-\rho,w)\right)^2}{P_{3,3}(-\rho,w)}.
\end{align}
\textbf{Proof}:\hspace{0in} We need to compute $I_{2,\rho,w} = -E(l^{\prime\prime}(\rho))$.  Because the expectation $E\left(k_{2,2}+k_{1,1}\right) = 2P_{2,2}(\rho,w)$, the expression $E(l^{\prime\prime}(\rho))$ can be simplified substantially. Then we take advantage of the fact that $\sum_{i,j} P_{i,j}(\rho,w) =1 $, $ \sum_{i,j} P_{i,j}^\prime(\rho,w) = \sum_{i,j} P_{i,j}^{\prime\prime}(\rho,w)=0$, to obtain the desired result. $\hfill\Box$
\end{theorem}

While the expressions appear sophisticated, the Fisher Information and variance  can be  verified by simulations; see Figure~\ref{fig_simu}.

\subsection{The 2-Bit Linear Estimator}

A linear estimator  only uses the information  whether the code of $x$ equals the code of $y$. In other words, linear estimators only use the diagonal information in Figure~\ref{fig_16region}. With a 2-bit scheme, $\rho$ can be estimated from counts in collapsed cells, by solving for $\rho$ from
\begin{align}\notag
(k_{0,0}+k_{1,1}+ k_{2,2}+k_{3,3})/k  = P_{0,0}+P_{1,1}+P_{2,2}+P_{3,3},
\end{align}
which still requires a numerical procedure (or tabulation). The analysis of the linear estimator was done in~\cite{Proc:Li_ICML14}, and  can also be inferred from the analysis of the nonlinear estimator in this paper.

\subsection{The 1-Bit Estimator}

This special case can be derived from the results of 2-bit random projections by simply letting $w\rightarrow \infty$. The estimator, by counting the observations in each quadrant, has a simple closed-form~\cite{Article:Goemans_JACM95,Proc:Charikar}, i.e., $\mathbf{Pr}\left(sgn(x)=sgn(y)\right) = 1-\frac{1}{\pi}\cos^{-1}\rho$.     The Fisher Information of estimator, denoted by $I_{1,\rho}$,  is then\\
\begin{align}\notag
I_{1,\rho}
=&2k\left[\frac{\left(P_{2,2}^\prime(\rho,\infty)\right)^2}{P_{2,2}(\rho,\infty)} +
 \frac{\left(P_{2,2}^\prime(-\rho,\infty)\right)^2}{P_{2,2}(-\rho,\infty)}
\right]\\\notag
=&2k\frac{1}{4\pi^2(1-\rho^2)}\left[\frac{1}{\frac{1}{4}+\frac{\arcsin\rho}{2\pi}} + \frac{1}{\frac{1}{4}-\frac{\arcsin\rho}{2\pi}}
\right]
\end{align}
The ratio
\begin{align}\label{eqn_Rw}
R_{\rho,w} = \frac{I_{2,\rho,w}}{I_{1,\rho}}
\end{align}
characterizes the reduction  of variance by using the 2-bit scheme and the MLE,  as a function of $\rho$ and $w$.

\begin{figure}[h!]
\begin{center}
\includegraphics[width = 1.8in]{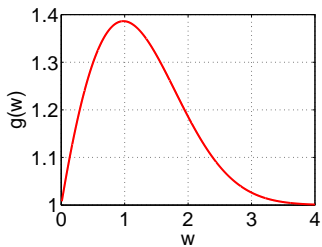}
\end{center}
\vspace{-0.25in}
\caption{The curve of $g(w)$ as defined in (\ref{eqn_R0}). 
}\label{fig_gs}
\end{figure}

We provide the following Theorem, to show that the ratio $R_{\rho,w}$ is  close to 2 when $\rho\rightarrow0$. Later we will see that, for high similarity regions, the ration can be substantially higher than 2.
\begin{theorem}\label{thm_R0}
For (\ref{eqn_Rw}) and $\rho\rightarrow 0$,  we have $R_{0,w}
 =\left[g(w)\right]^2$,
\begin{align}\label{eqn_R0}
\text{where}\hspace{0.2in} g(s) =\frac{1}{2}\left[\frac{\left[1-e^{-\frac{w^2}{2}}\right]^2}{\Phi(w)-\frac{1}{2}} + \frac{e^{-w^2}}{1- \Phi\left(w\right)}\right].\hspace{0.2in}\Box
\end{align}
\end{theorem}
Figure~\ref{fig_gs} shows that $g(w)$ has a unique maximum = 1.3863 (i.e., maximum of $\left[g(w)\right]^2$ is 1.9218), attained at $w = 0.9816$.

\vspace{-0.1in}
\subsection{The Choice of $w$}

\begin{figure}[h!]
\begin{center}
\mbox{\includegraphics[width = 1.75in]{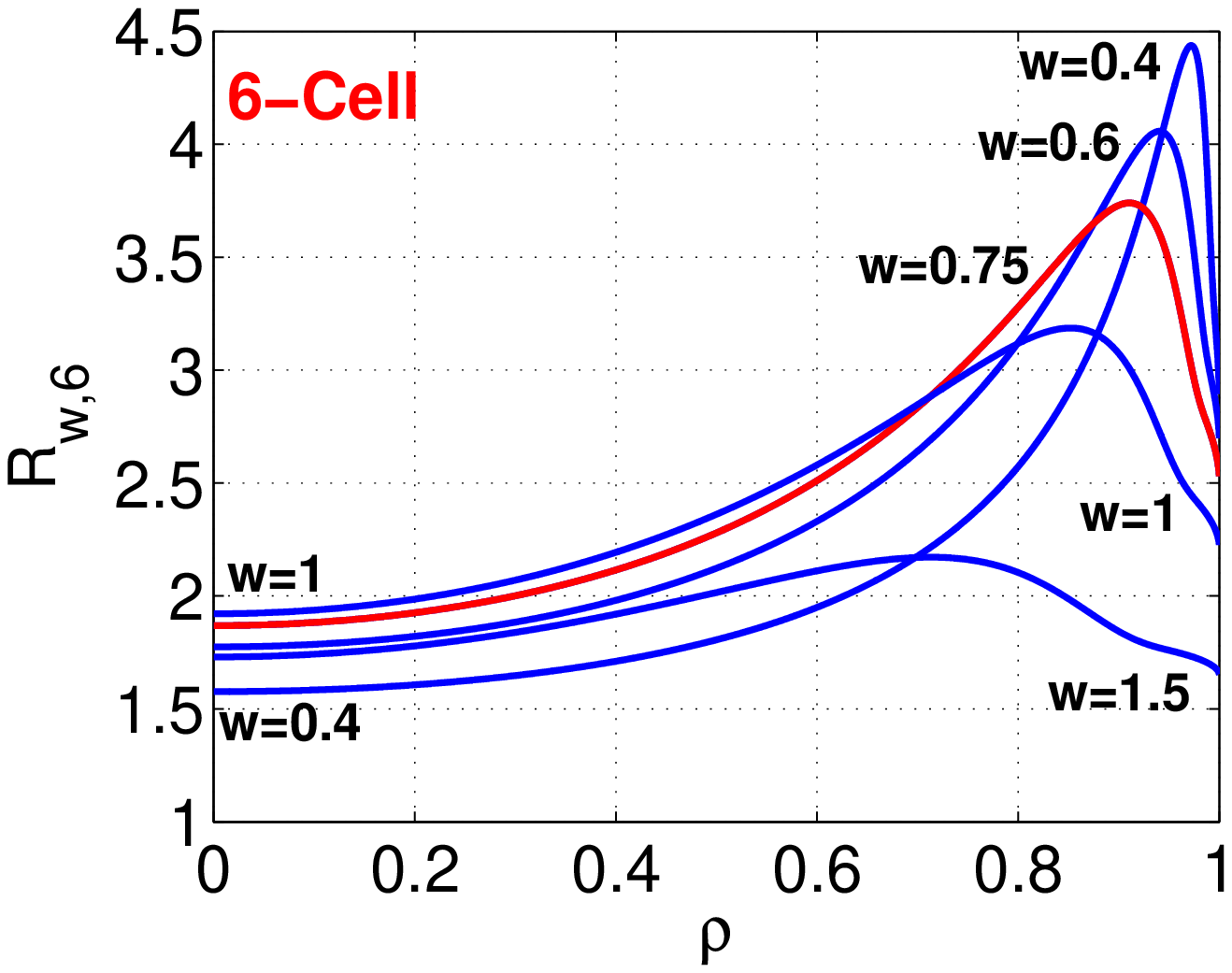}\hspace{-0.12in}
\includegraphics[width = 1.75in]{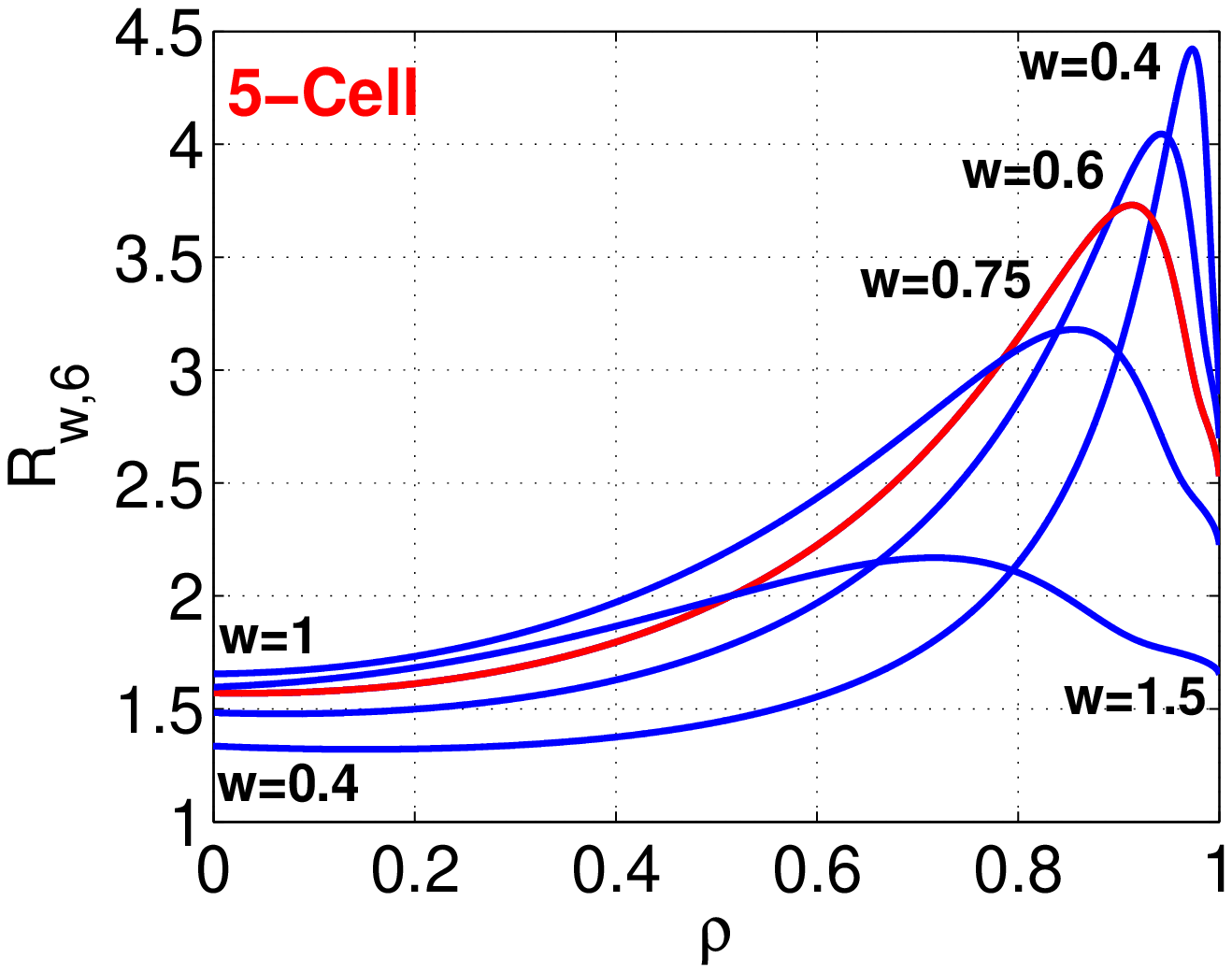}}

\end{center}
\vspace{-0.25in}
\caption{The ratio $R_{\rho,w}$ (\ref{eqn_Rw}) at $w=0.4, 0.6, 0.75, 1, 1.5$, which characterizes the improvement of the MLE ($\hat{\rho}_{2,\rho}$) over the 1-bit estimator $\hat{\rho}_1$. It looks $w=0.75$ provides an overall good trade-off.  The problem is a 6-cell (ie., left panel) contingency table estimation problem. To demonstrate the simplification of the  process  by using 5 cells (see the main text for the description of the procedure), we also include the same type of improvements for using the reduced 5-cell model in the right panel. }\label{fig_Rw}\vspace{-0.1in}
\end{figure}

\begin{figure}[h!]
\begin{center}
\mbox{\includegraphics[width = 1.75in]{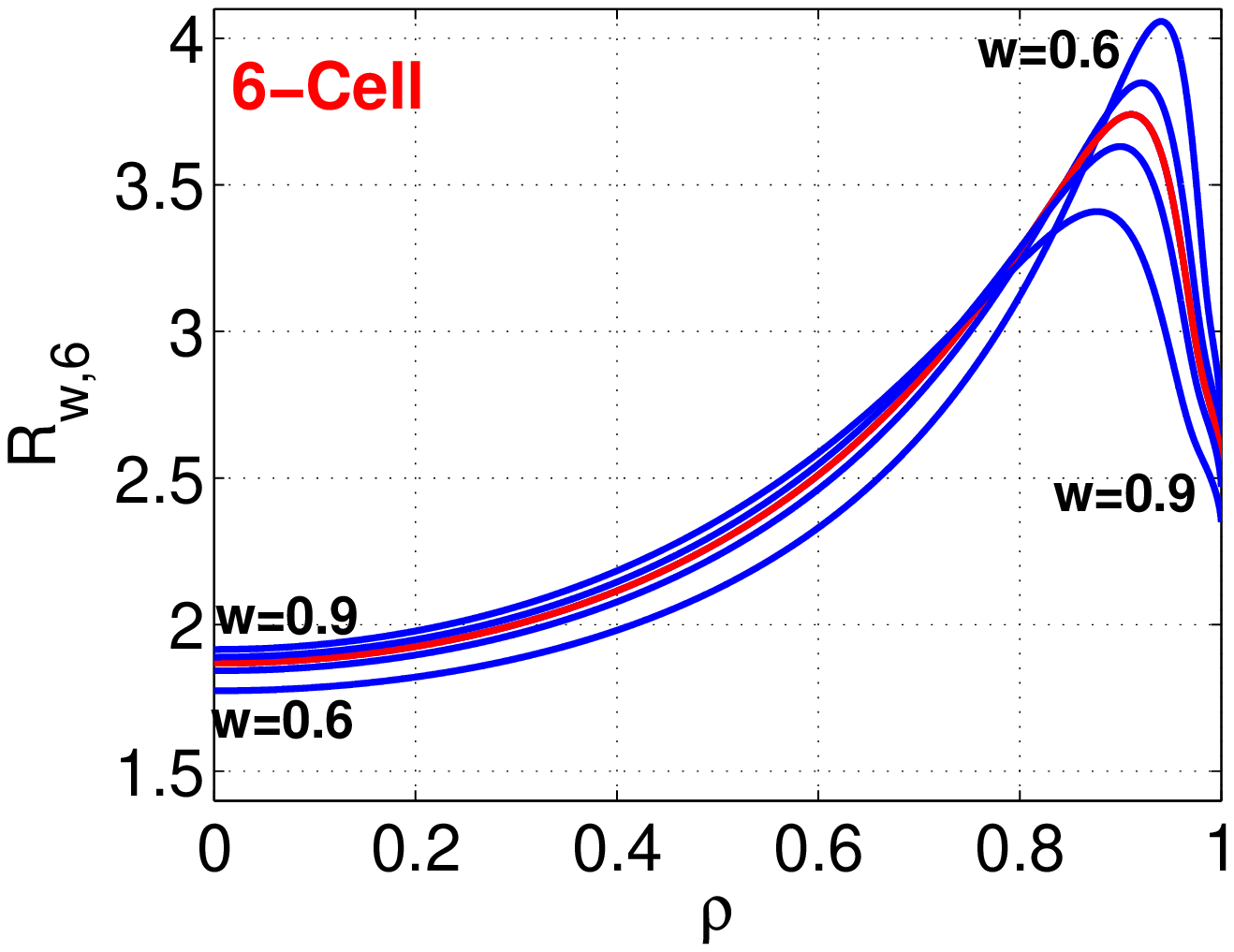}\hspace{-0.12in}
\includegraphics[width = 1.75in]{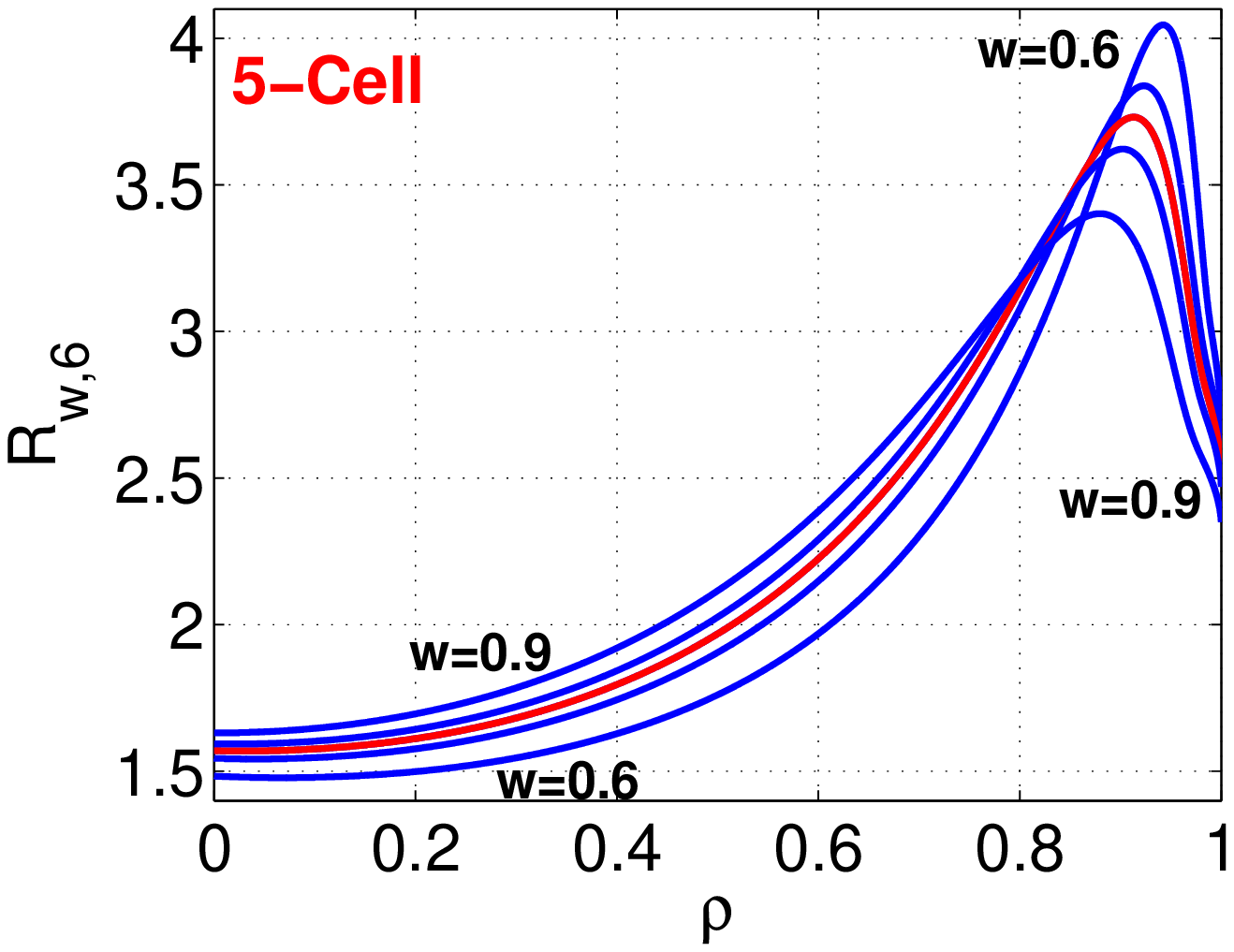}}

\end{center}
\vspace{-0.25in}
\caption{The ratio $R_{\rho,w}$ (\ref{eqn_Rw}) at $w=0.6, 0.7, 0.75, 0.8, 0.9$, to show $w=0.75$ is an overall good trade-off. There is no space to label $w=0.7, 0.75, 0.8$ but the order of curves should be a good indicator. We  plot $w=0.75$ in red, if color is available.}\label{fig_Rw_small}
\end{figure}

The performance depends on $w$ (and $\rho$). In practice, we need to  pre-specify a value of $w$ for random projections and we have to use the same $w$ for all data points because this coding process is non-adaptive.  Figure~\ref{fig_Rw} and Figure~\ref{fig_Rw_small} plot the ratio $R_{\rho,w}$ (left panels) for selected $w$ values, confirming that $w=0.75$  should be an overall good choice.  In addition, we present some additional work in the right panels of  Figure~\ref{fig_Rw} and Figure~\ref{fig_Rw_small} to show that if we collapse some cells appropriately (from a 6-cell model to a 5-cell model), the performance would not degrade much (not at all for high similarity region, which is often more interesting in practice).

According to Figure~\ref{fig_16region}, we collapse the three cells (0,3), (0,2), and (1,3) into one cell. Note that (0,2) and (1,3) have the same probabilities and are already treated as one cell. Due to symmetry, the other three cells (3,0), (2,0), and (3,1) are also collapsed into one. This way, we have in total 5 distinct cells. The intuition is that if we are mostly interested in high similar regions, most of the observations will be falling around the diagonals. This treatment simplifies the estimation process and does not lead to an obvious degradation of the accuracy at least for high similarity regions, according to Figure~\ref{fig_Rw} and Figure~\ref{fig_Rw_small}.

\subsection{Simulations}\label{sec_2bit_simu}

\begin{figure}[h!]

\hspace{-0.2in}
\mbox
{\includegraphics[width = 1.3in]{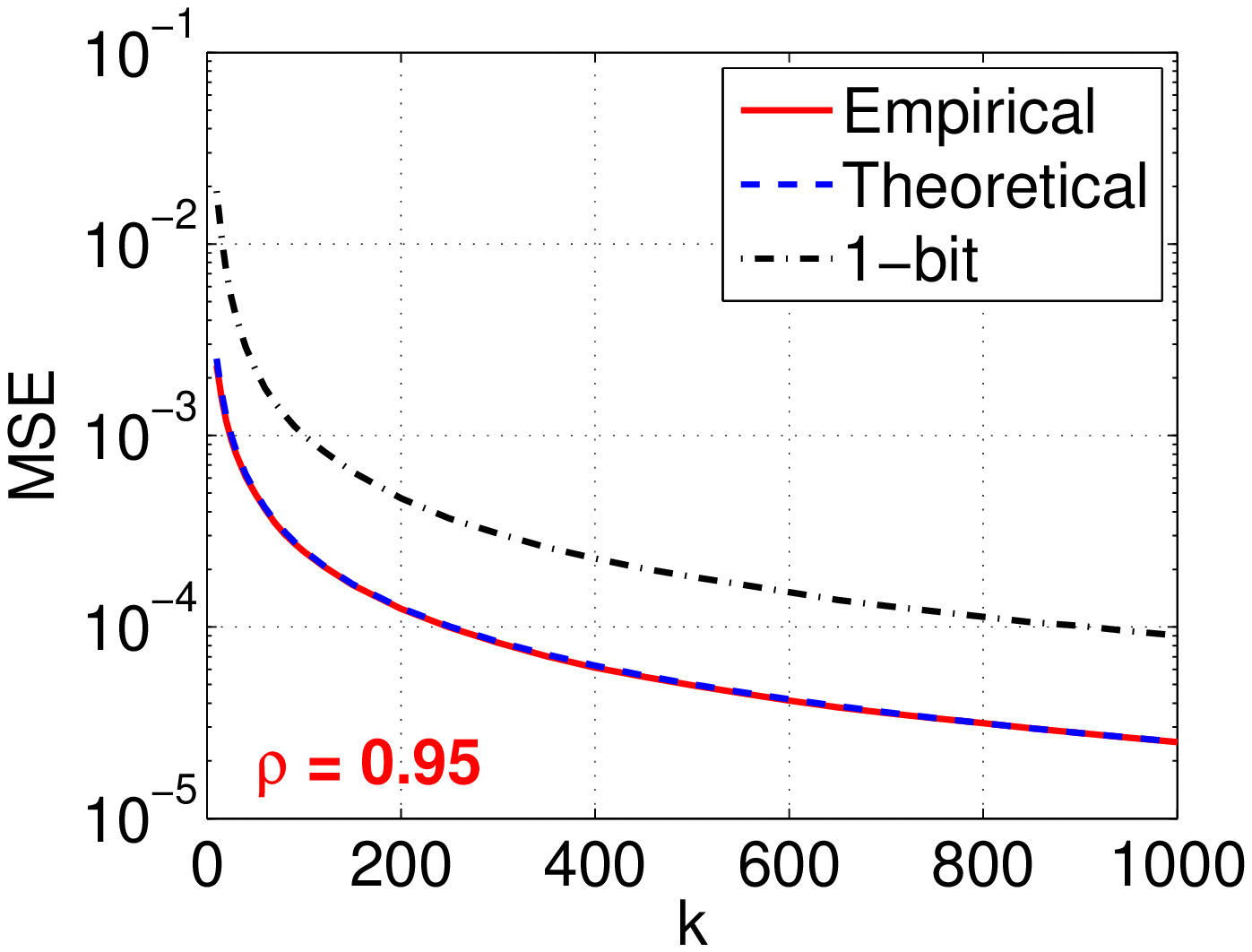}\hspace{-0.08in}
\includegraphics[width = 1.3in]{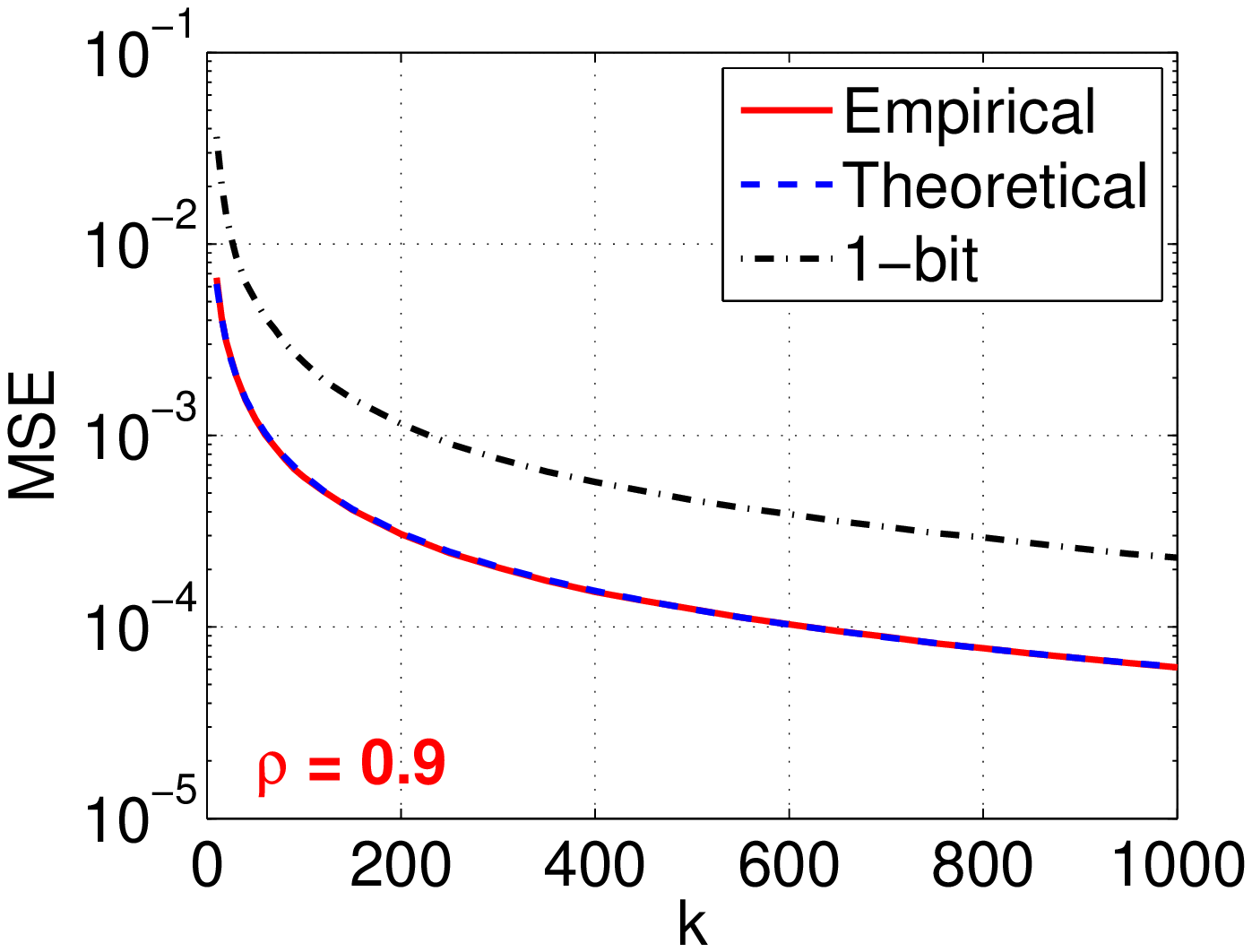}\hspace{-0.08in}
\includegraphics[width = 1.3in]{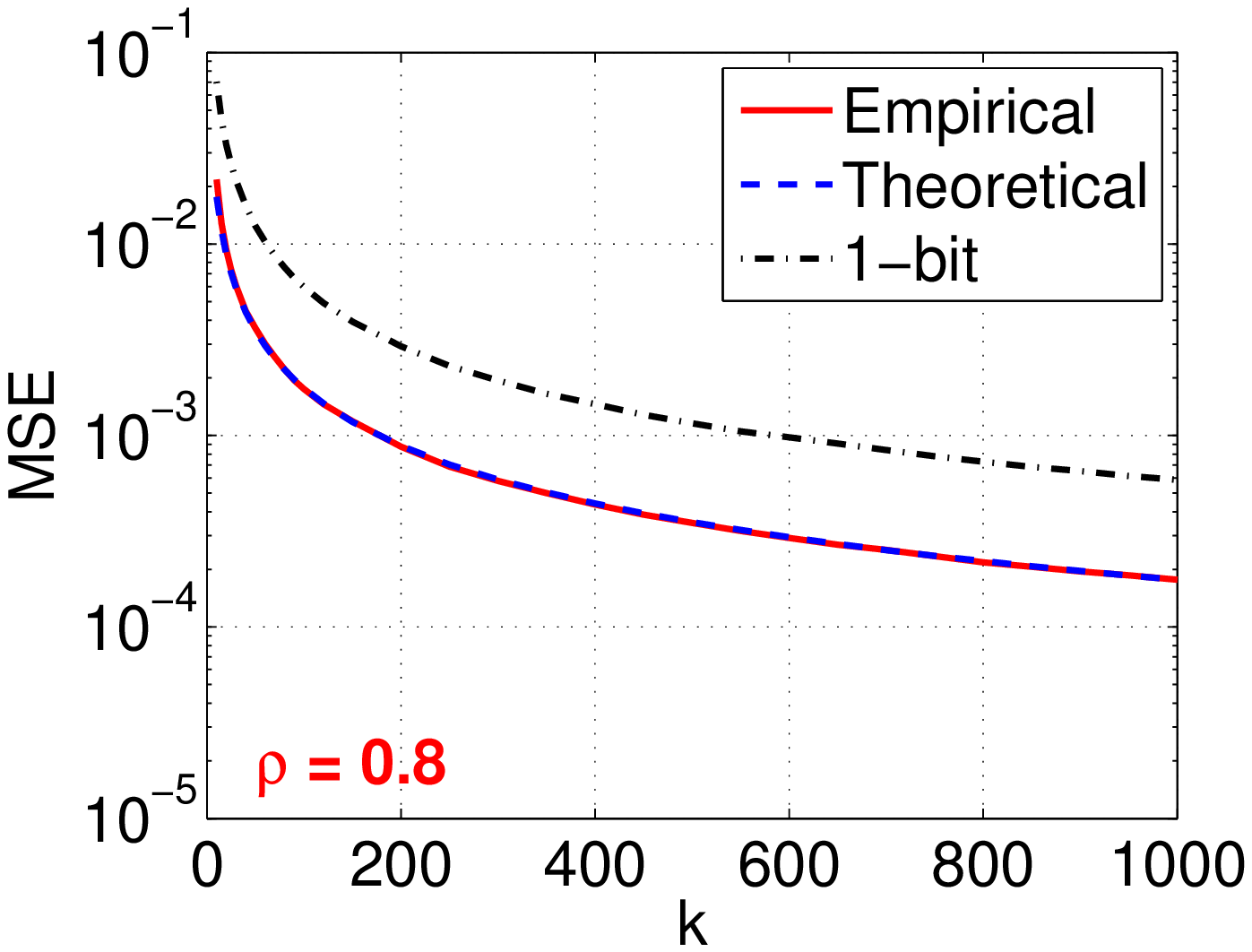}
}

\hspace{-0.2in}
\mbox{
\includegraphics[width = 1.3in]{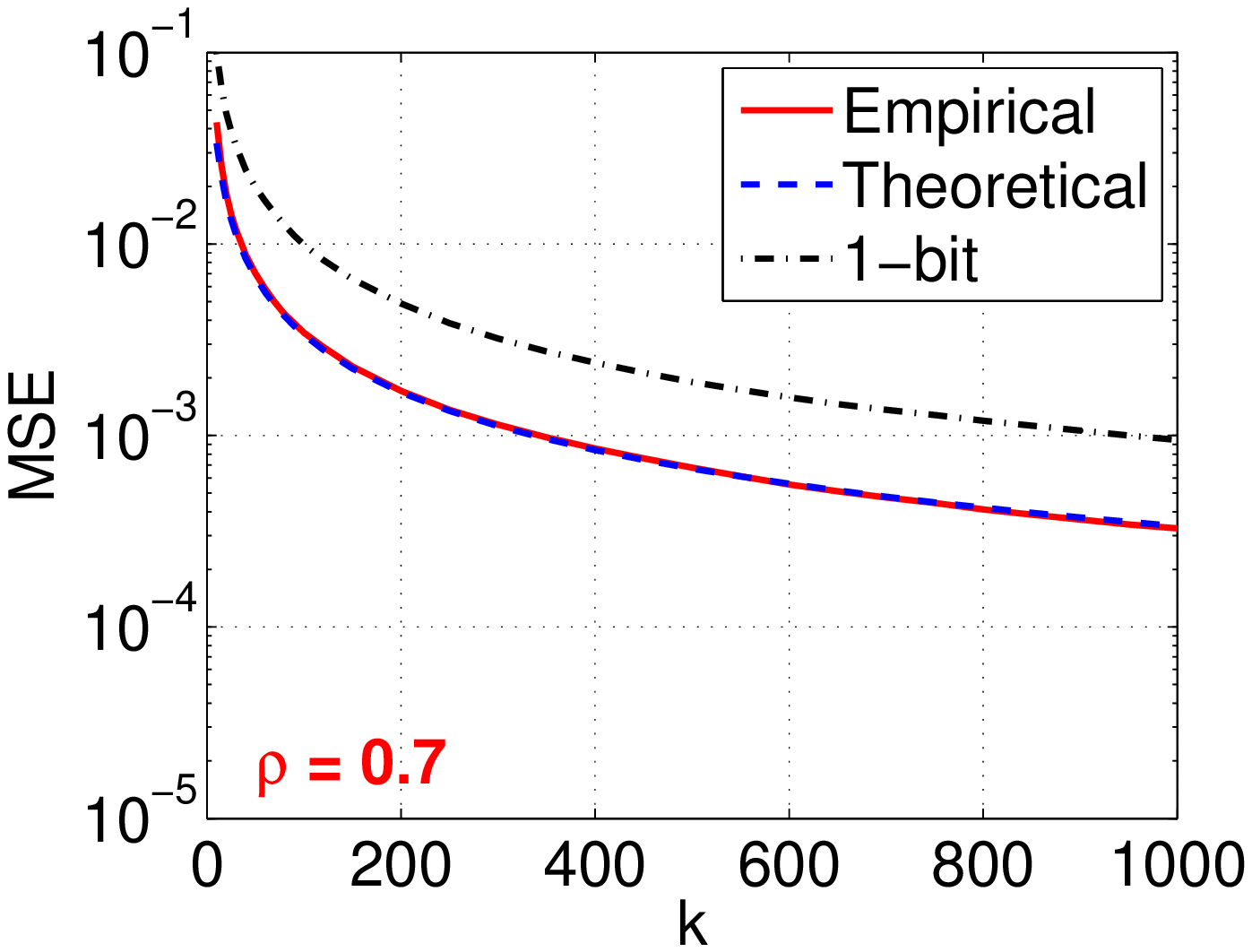}\hspace{-0.08in}
\includegraphics[width = 1.3in]{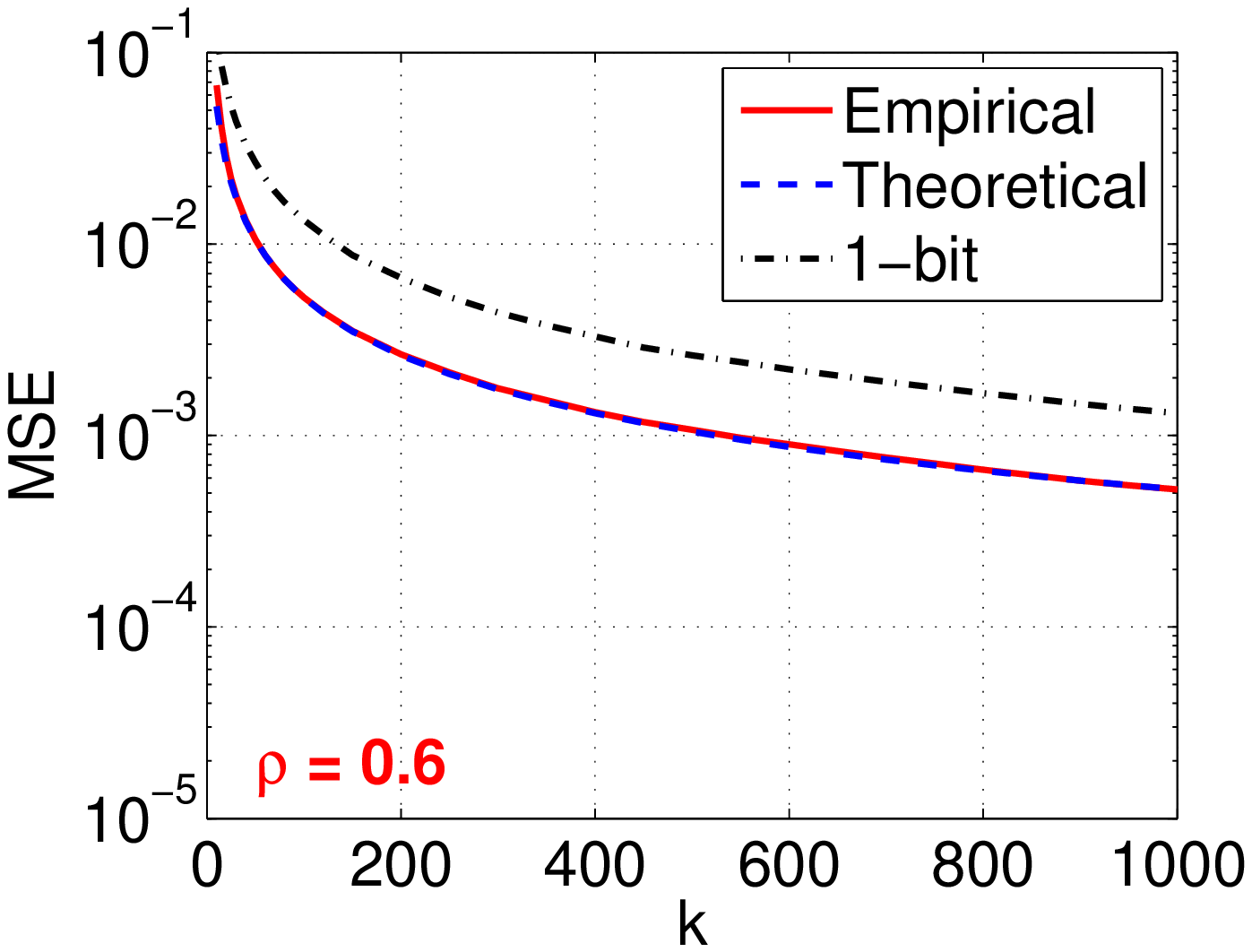}\hspace{-0.08in}
\includegraphics[width = 1.3in]{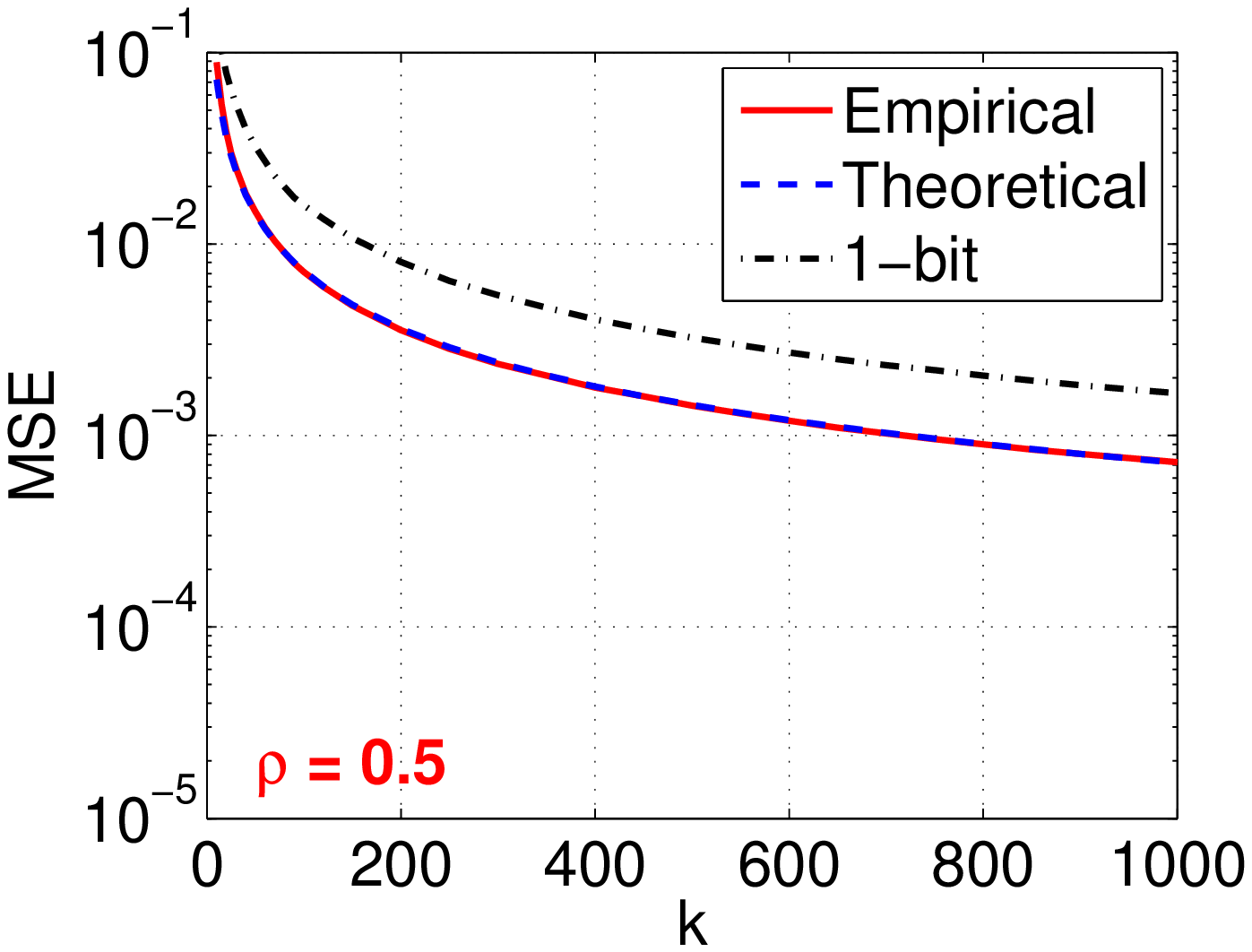}
}

\hspace{-0.2in}
\mbox
{\includegraphics[width = 1.3in]{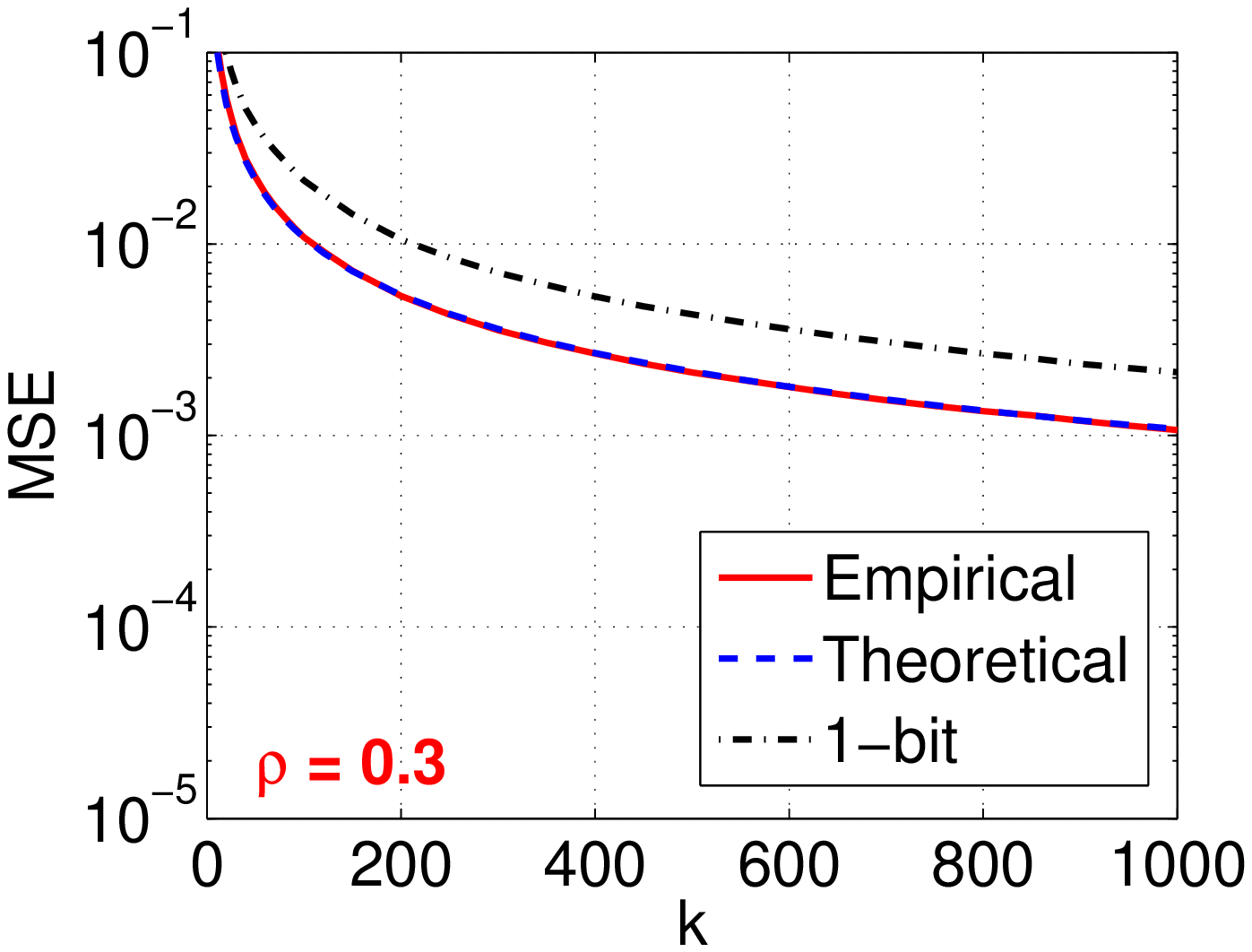}\hspace{-0.08in}
\includegraphics[width = 1.3in]{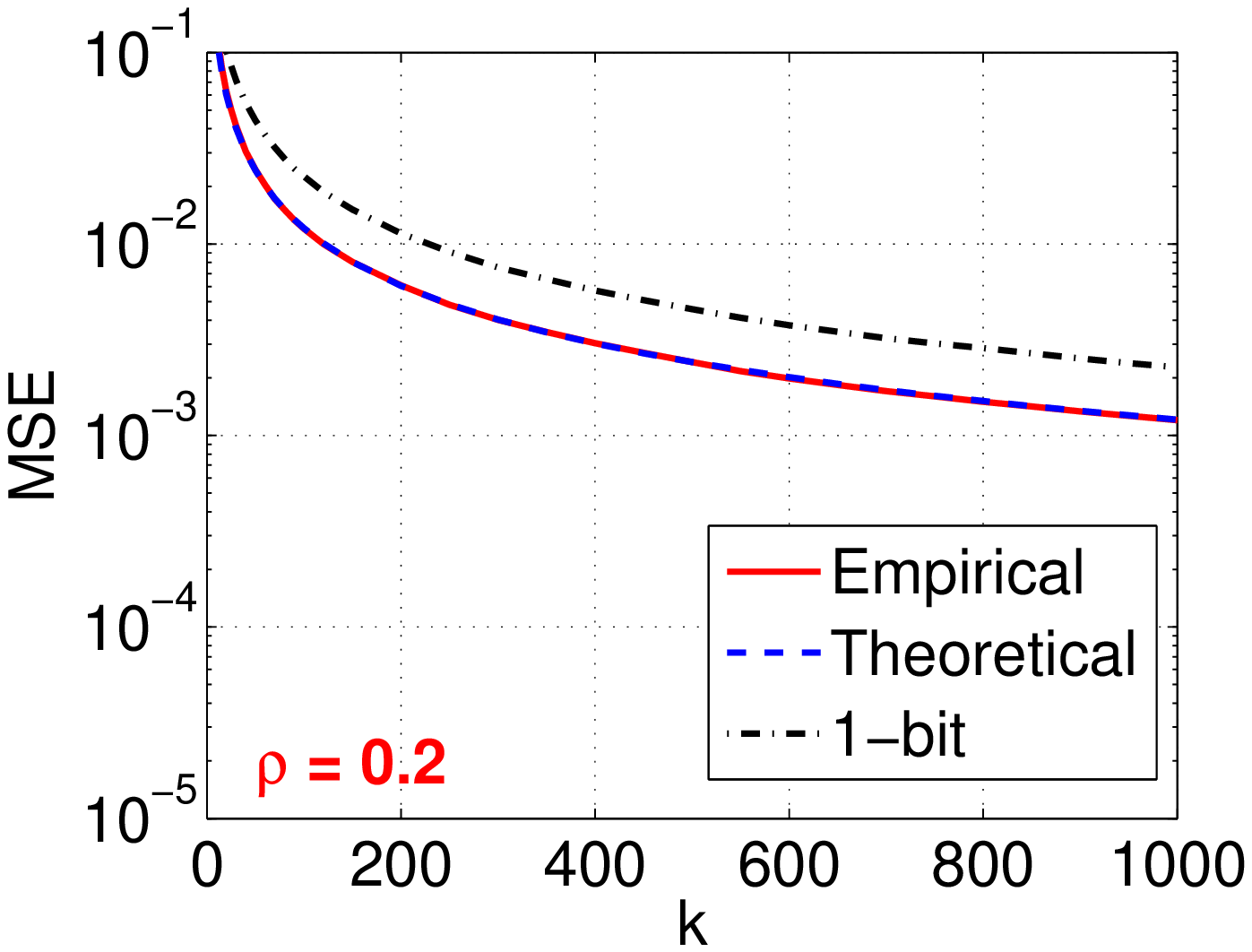}\hspace{-0.08in}
\includegraphics[width = 1.3in]{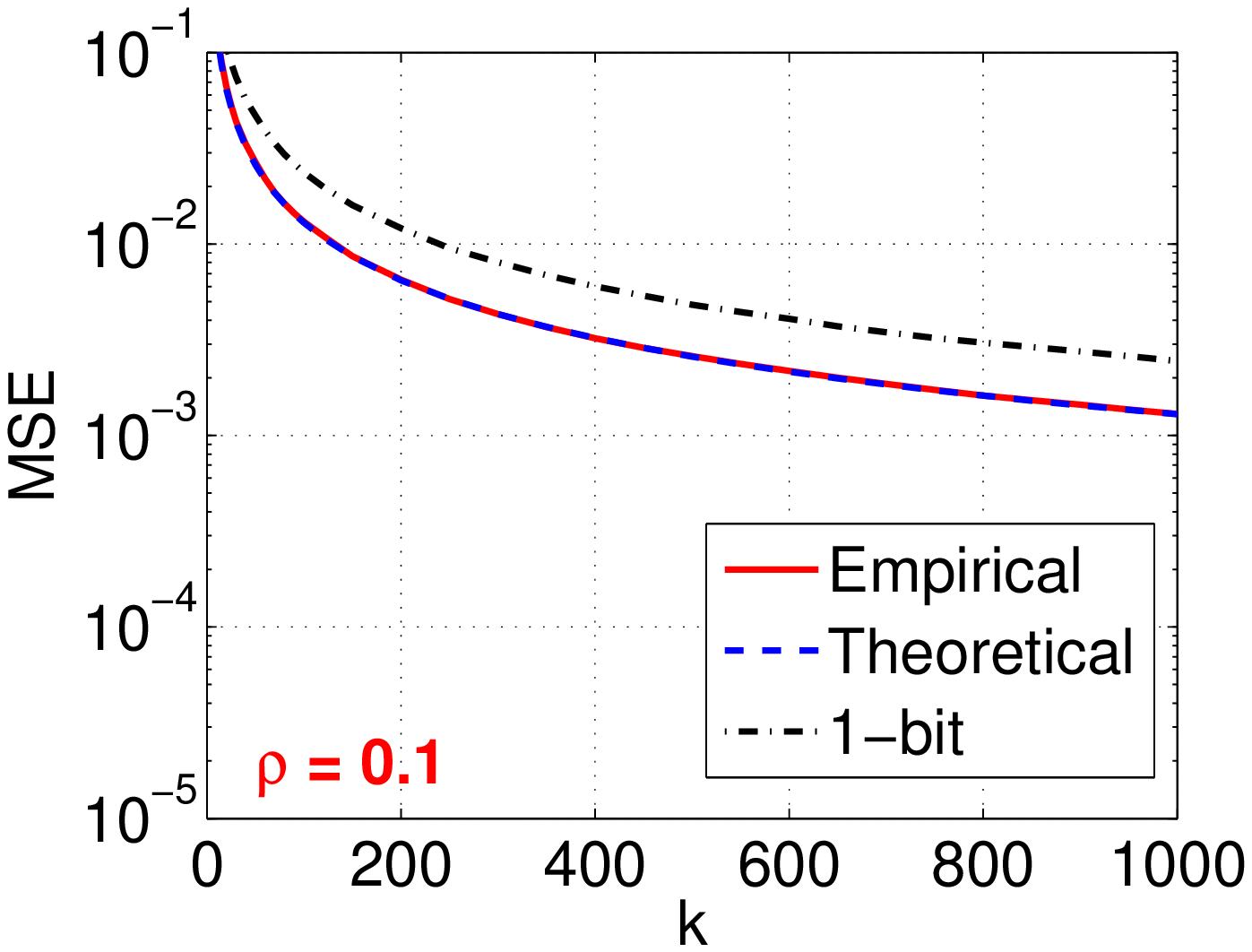}
}

\vspace{-0.15in}
\caption{Mean square errors (MSE) from the simulations to verify the nonlinear MLE. The empirical MSEs essentially overlap the asymptotic variances predicted by the Fisher information (\ref{eqn_Fisher2}), confirming the theoretical results. In addition, we also plot the empirical MSEs of the 1-bit estimator to verify the substantial improvement of the MLE. }\label{fig_simu}\vspace{-0.1in}
\end{figure}

As presented in Figure~\ref{fig_simu}, a simulation study is conducted to confirm the theoretical results of the MLE, for a wide range of $\rho$ values.  The plots confirm that the MLE substantially improves the 1-bit estimator, even at low similarities. They also verify that the theoretical asymptotic variance predicted by the Fisher Information (\ref{eqn_Fisher2}) is accurate, essentially no different from the empirical mean square errors. We hope this experiment might help readers who are less familiar with the classical  theory of Fisher Information.

\newpage
\section{Other Common Coding Schemes}\label{sec_others}

In this section, we review two common coding strategy: (i) the scheme based on windows + random offset;  (ii) the scheme based on simple uniform quantization. Note that both of them are strictly speaking infinite-bit coding schemes, although (ii) can be effectively viewed as a finite-bit scheme.

\subsection{Quantization with Random Offset}

\cite{Proc:Datar_SCG04} proposed the following well-known coding scheme, which uses
windows and a random offset:
\begin{align}\label{eqn_hwq}
h_{w,q}^{(j)}(u) = \left\lfloor\frac{x_j + q_j}{w}\right\rfloor,\hspace{0.3in} h_{w,q}^{(j)}(v) = \left\lfloor\frac{y_j + q_j}{w}\right\rfloor
\end{align}
where $q_j\sim uniform(0,w)$,  $w>0$ is the bin width and $\left\lfloor . \right\rfloor$ is the standard floor operation.
\cite{Proc:Datar_SCG04} showed that the collision probability can be written as a monotonic function of the Euclidean  distance:
\begin{align}\notag
P_{w,q} = &\mathbf{Pr}\left(h_{w,q}^{(j)}(u) = h_{w,q}^{(j)}(v)\right)
= \int_0^w\frac{1}{\sqrt{d}}2\phi\left(\frac{t}{\sqrt{d}}\right)\left(1-\frac{t}{w}\right)dt
\end{align}
where $d = ||u-v||^2= 2(1-\rho)$ is the distance between $u$ and $v$.

\subsection{Uniform Quantization without Offset}

A simpler (and in fact better) scheme than (\ref{eqn_hwq}) is based on  uniform quantization without offset:
\begin{align}\label{eqn_hw}
h_{w}^{(j)}(u) = \left\lfloor x_j/w\right\rfloor,\hspace{0.4in} h_{w}^{(j)}(v) = \left\lfloor y_j/w\right\rfloor
\end{align}

The collision probability for  (\ref{eqn_hw}) is
\begin{align}\notag
&P_{w} =\mathbf{Pr}\left(h_{w}^{(j)}(u) = h_{w}^{(j)}(v) \right)\\\notag
=& 2\sum_{i=0}^\infty\int_{iw}^{(i+1)w}\phi(z)\left[\Phi\left(\frac{(i+1)w-\rho z}{\sqrt{1-\rho^2}}\right)- \Phi\left(\frac{iw-\rho z}{\sqrt{1-\rho^2}}\right)\right]dz
\end{align}
$P_w$ is a monotonically increasing function of $\rho\geq0$. \\

The fact that $P_w$ is monotonically increasing in $\rho$ makes   (\ref{eqn_hw}) an appropriate coding scheme for approximate near neighbor search under the general framework of locality sensitive hashing (LSH). Note that while $P_{w}$ appears sophisticated, the expression is just for the analysis. Without  using the offset, the scheme (\ref{eqn_hw})  itself is operationally  simpler than the  popular scheme (\ref{eqn_hwq}).

In the prior work, \cite{Proc:Li_ICML14} studied the coding scheme (\ref{eqn_hw}) in the context of similarity estimation using linear estimators with application to building large-scale linear classifiers.  In this paper, we conduct  the study of (\ref{eqn_hw}) for sublinear time near neighbor search by building hash tables from coded projected data. This is a very different task from similarity estimation. Moreover, much of the space of the paper is allocated to the design and analysis of  nonlinear estimators which are very useful in the ``re-ranking'' stage of near neighbor search after the potentially similar data points are retrieved.

\vspace{0.08in}

There is another  important distinction between  (\ref{eqn_hw}) and (\ref{eqn_hwq}). By using a window and a random offset, (\ref{eqn_hwq}) is actually an ``infinite-bit'' scheme. On the other hand, with only a uniform quantization, (\ref{eqn_hw}) is essentially a finite-bit scheme, because the data are normalized and the Gaussian (with variance 1) density decays very rapidly at the tail. If we choose (e.g.,) $w\geq3$ (note that $1-\Phi(3)=1.3\times10^{-3}$), we essentially have a 1-bit scheme (i.e., by recording the signs of the projected data), because the analysis can show that using $w\geq 3$ is not essentially different from using $w=\infty$. Note that the 1-bit scheme~\cite{Article:Goemans_JACM95,Proc:Charikar} is also known as ``sim-hash'' in the literature.

\vspace{0.08in}

In this paper, we will show, through analysis and experiment, that often a 2-bit scheme (i.e., a uniform quantization with $w\geq 1.5$) is better for LSH (depending on the data similarity). Moreover, we have developed nonlinear estimators for 2-bit scheme which significantly improve the estimator using the 1-bit scheme as well as the linear estimator using the 2-bit scheme.

\section{Sublinear Time $c$-Approximate\\ Near Neighbor Search}\label{sec_LSH}

In this section, we compare the two coding schemes in Section~\ref{sec_others}:   (i) the scheme based on windows + random offset, i.e., (\ref{eqn_hwq});  (ii) the scheme based on simple uniform quantization, i.e., (\ref{eqn_hw}), in the setting of approximate near neighbor search.  We will show that (\ref{eqn_hw}) is more effective and in fact only a small number of bits are needed.

\vspace{0.08in}

Consider a data vector  $u$. Suppose there exists another vector whose Euclidian distance ($\sqrt{d}$) from $u$ is at most $\sqrt{d_0}$ (the target distance). The goal of {\em$c$-approximate $\sqrt{d_0}$-near neighbor} algorithms is to return data vectors (with high probability) whose Euclidian distances from $u$ are at most $c\times \sqrt{d_0}$ with $c>1$.

\vspace{0.08in}

Recall that, in our definition, $d = 2(1-\rho)$ is the squared Euclidian distance. To be consistent with the convention in~\cite{Proc:Datar_SCG04}, we present the results in terms of $\sqrt{d}$. Corresponding to the target distance $\sqrt{d_0}$,  the target similarity $\rho_0$ can be computed from $d_0 = 2(1-\rho_0)$ i.e., $\rho_0 = 1-d_0/2$. To simplify the presentation, we focus on $\rho\geq 0$ (as is common in practice), i.e., $0\leq d\leq 2$. Once we fix a target similarity $\rho_0$, $c$ can not exceed a certain value:
\begin{align}\notag
c\sqrt{2(1-\rho_0)}\leq \sqrt{2} \Longrightarrow c \leq \sqrt{\frac{1}{1-\rho_0}}
\end{align}
For example, when $\rho_0 =0.5$, we must have $1\leq c \leq \sqrt{2}$.

The performance of an LSH algorithm largely depends on the difference (gap) between the two collision probabilities $P^{(1)}$ and $P^{(2)}$ (respectively corresponding to $\sqrt{d_0}$ and $c\sqrt{d_0}$):
\begin{align}\notag
&P^{(1)}_w =  \mathbf{Pr}\left(h_w(u) = h_w(v) \right) \hspace{0.2in} \text{when } d = ||u - v||^2_2 = d_0\\\notag
&P^{(2)}_w =  \mathbf{Pr}\left(h_w(u) = h_w(v) \right) \hspace{0.2in} \text{when } d = ||u - v||^2_2 = c^2d_0
\end{align}
The  probabilities $P^{(1)}_{w,q}$ and $P^{(2)}_{w,q}$ are analogously defined for $h_{w,q}$.

\vspace{0.08in}

A larger difference between $P^{(1)}$ and $P^{(2)}$ implies a more efficient LSH algorithm. The following ``$G$'' values ($G_w$ for $h_w$ and $G_{w,q}$ for $h_{w,q}$, respectively) characterize the gaps:
\begin{align}\label{eqn_rho_M}
&G_w =\frac{\log 1/P_w^{(1)} }{\log 1/P_w^{(2)} },\hspace{0.5in} G_{w,q} =\frac{\log 1/P_{w,q}^{(1)} }{\log 1/P_{w,q}^{(2)} }
\end{align}
A smaller $G$ (i.e., larger difference between $P^{(1)}$ and $P^{(2)}$) leads to a potentially more efficient LSH algorithm and $\rho <\frac{1}{c}$ is particularly desirable~\cite{Proc:Indyk_STOC98}. The general theory of LSH says  the query time for $c$-approximate $d_0$-near neighbor is dominated by $O(N^G)$ distance evaluations, where $N$ is the total number of data vectors in the collection. This is better than $O(N)$, the cost of a linear scan.

\subsection{Theoretical Comparison of the Gaps}

\begin{figure}[h!]

\hspace{-0.15in}
\mbox{
\includegraphics[width = 1.25in]{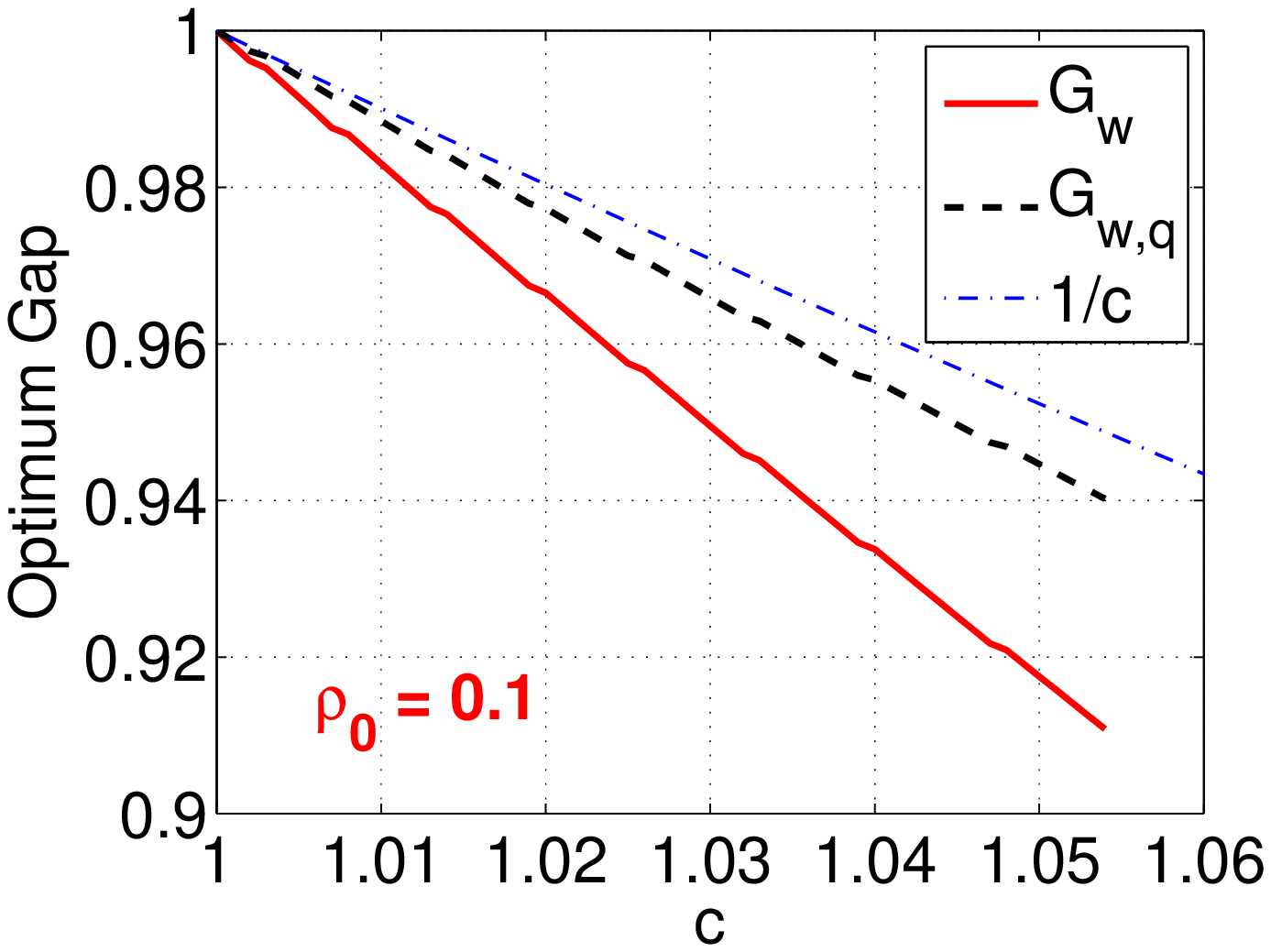}\hspace{-0.08in}
\includegraphics[width = 1.25in]{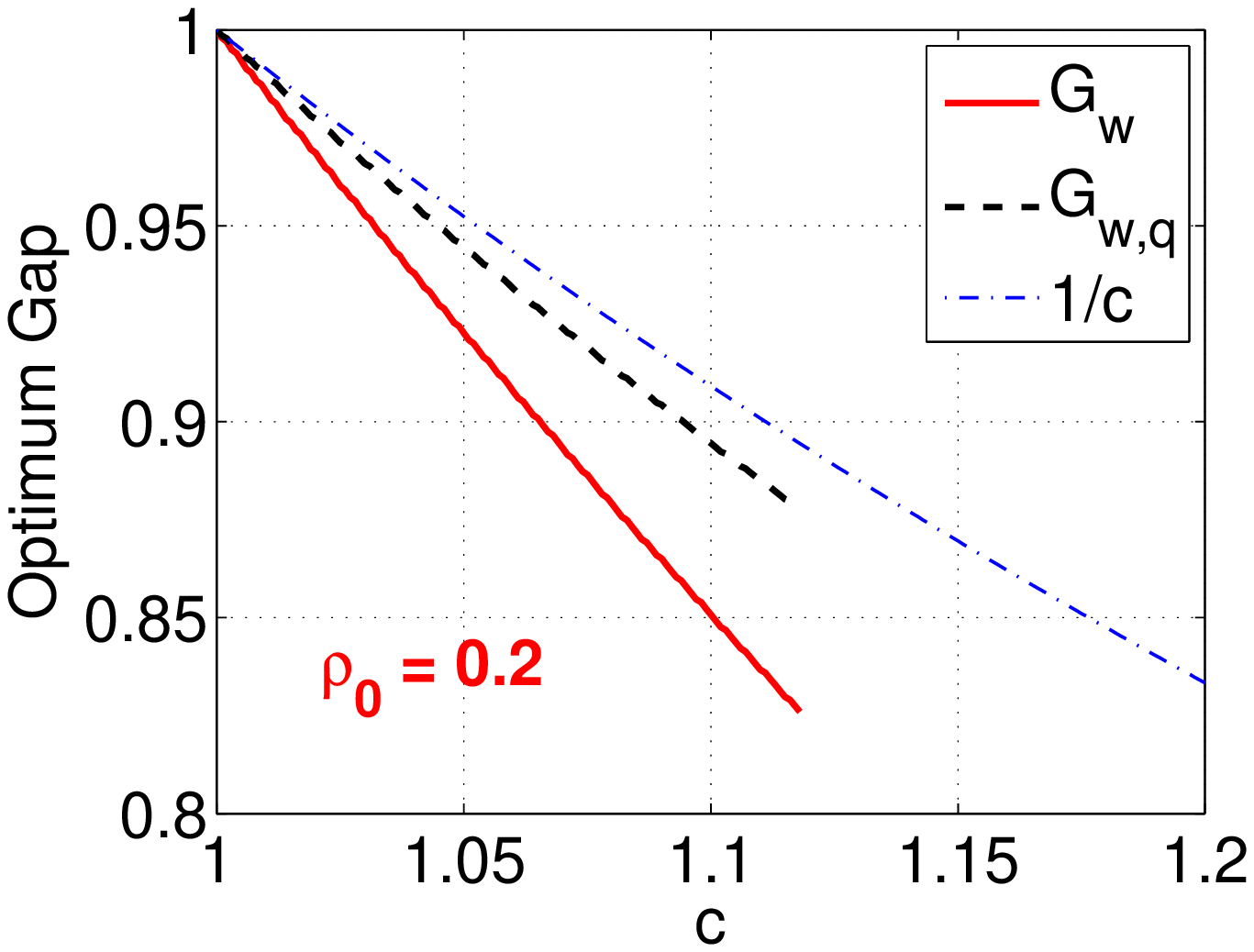}\hspace{-0.08in}
\includegraphics[width = 1.25in]{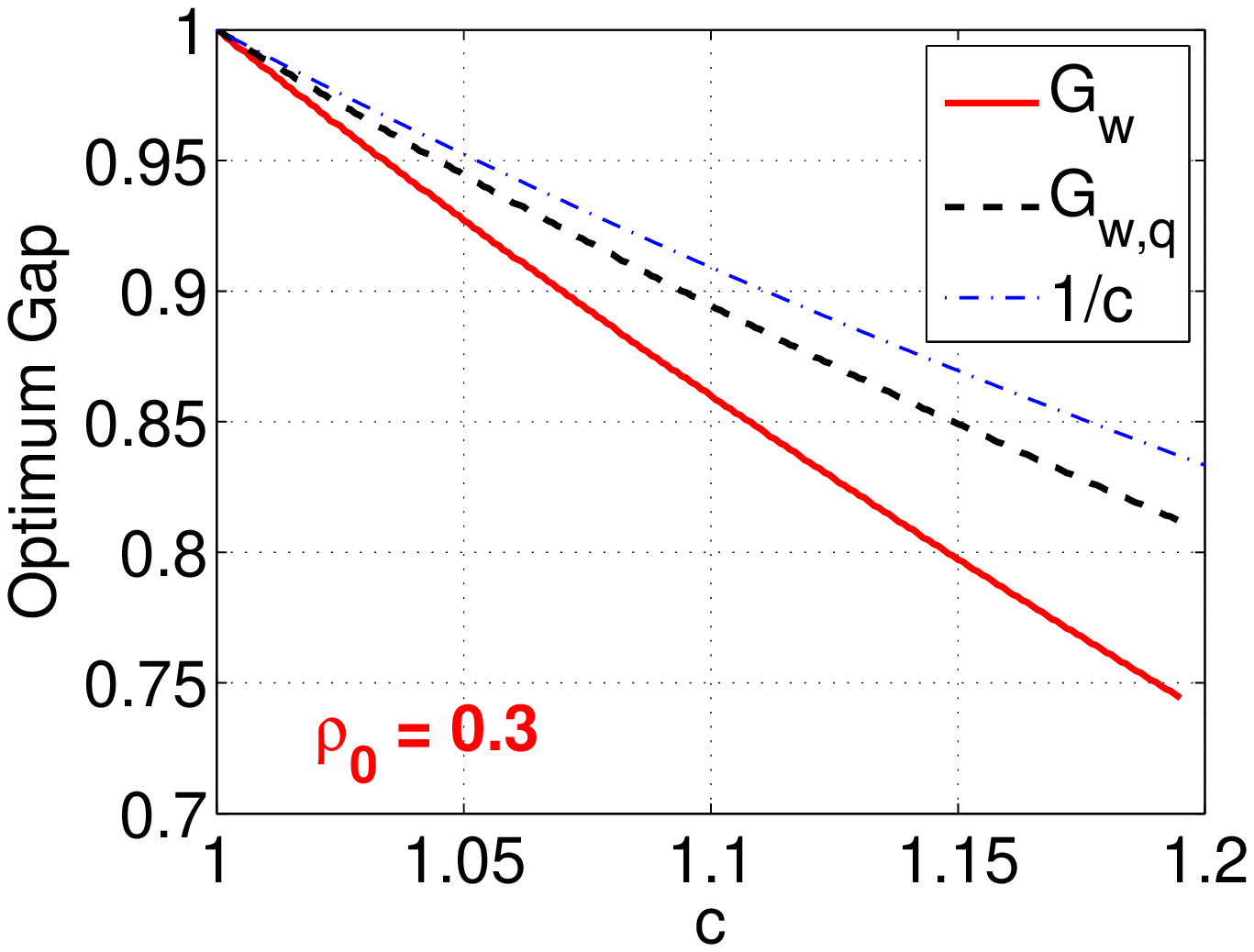}
}

\hspace{-0.15in}
\mbox{
\includegraphics[width = 1.25in]{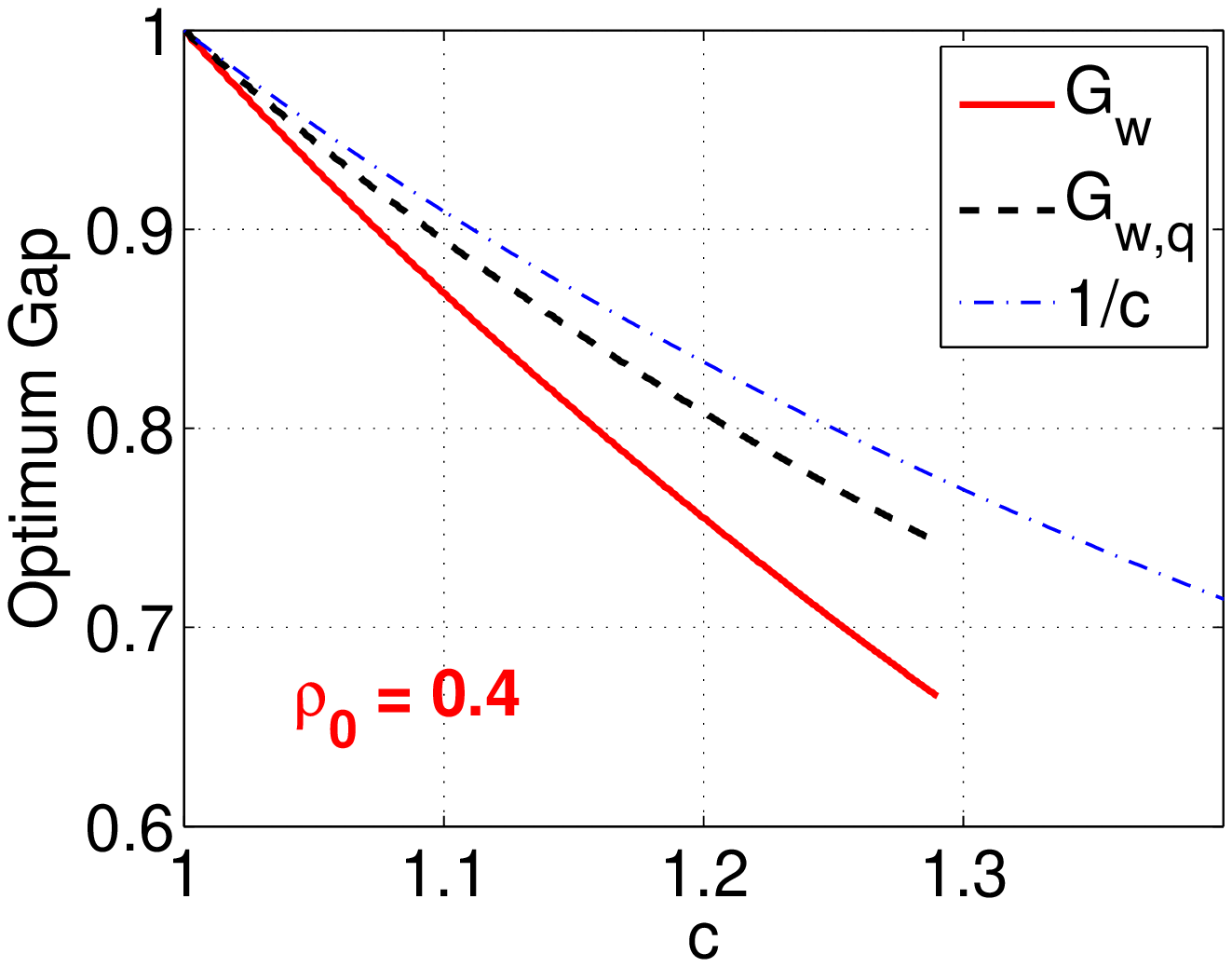}\hspace{-0.08in}
\includegraphics[width = 1.25in]{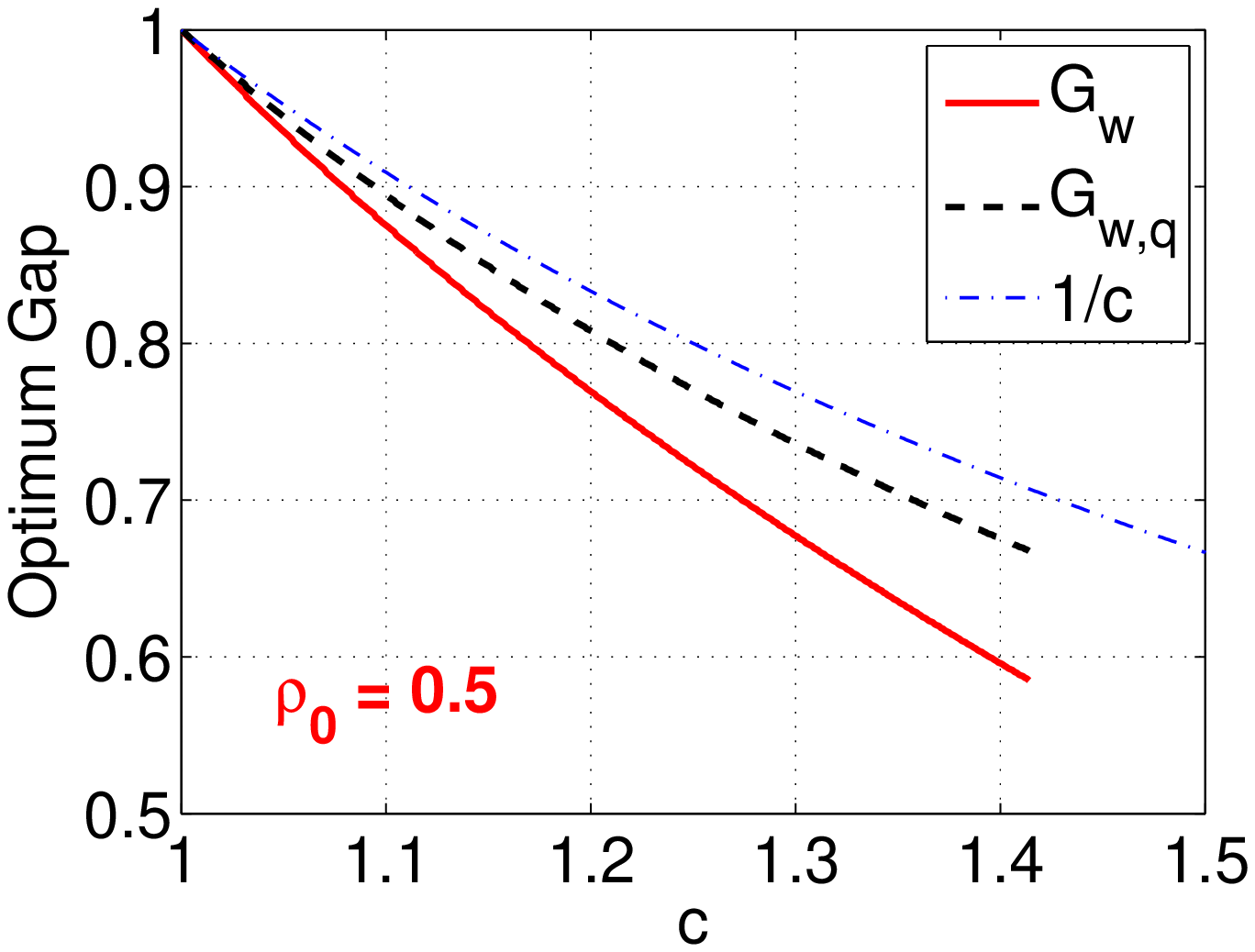}\hspace{-0.08in}
\includegraphics[width = 1.25in]{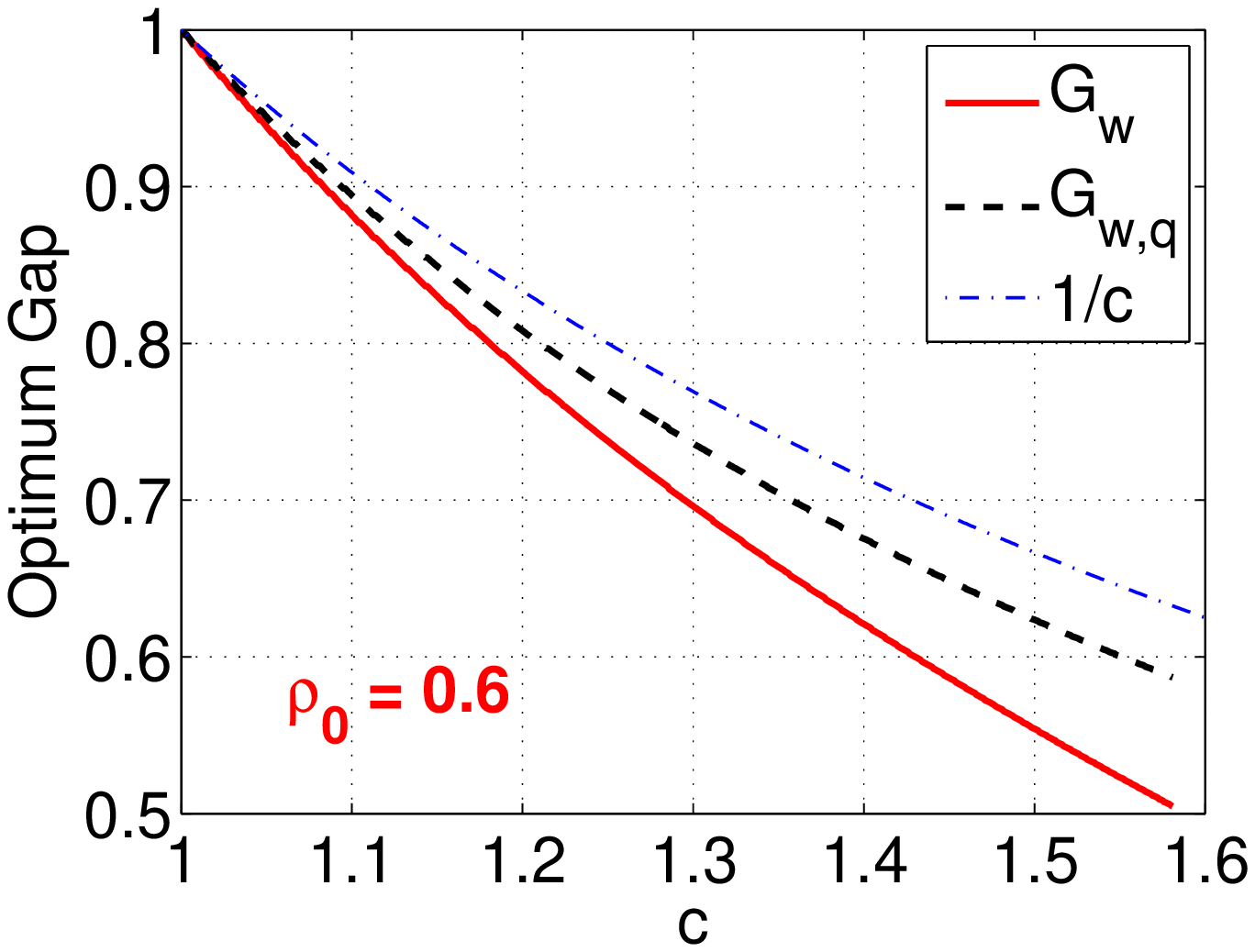}
}

\hspace{-0.15in}
\mbox{
\includegraphics[width = 1.25in]{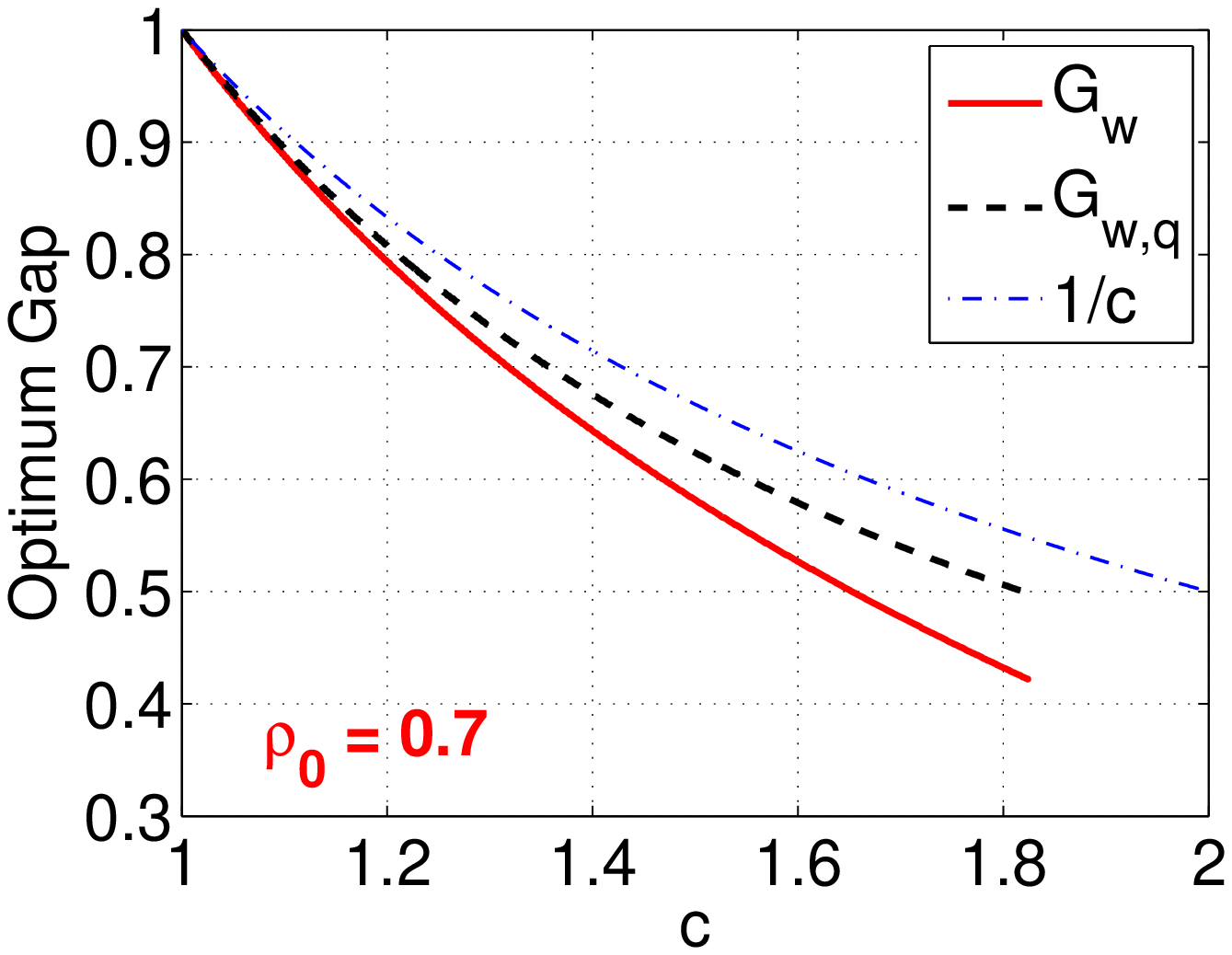}\hspace{-0.08in}
\includegraphics[width = 1.25in]{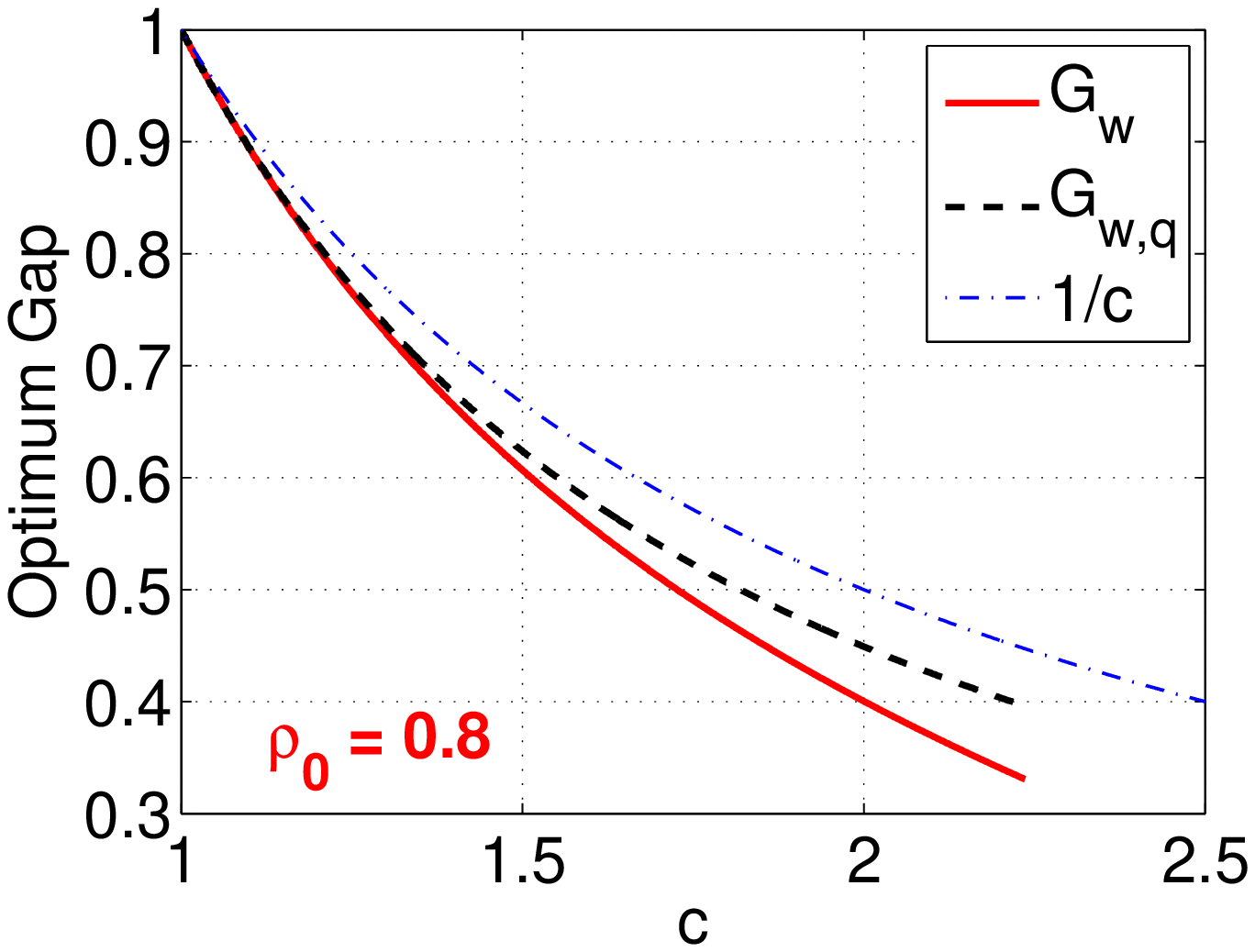}\hspace{-0.08in}
\includegraphics[width = 1.25in]{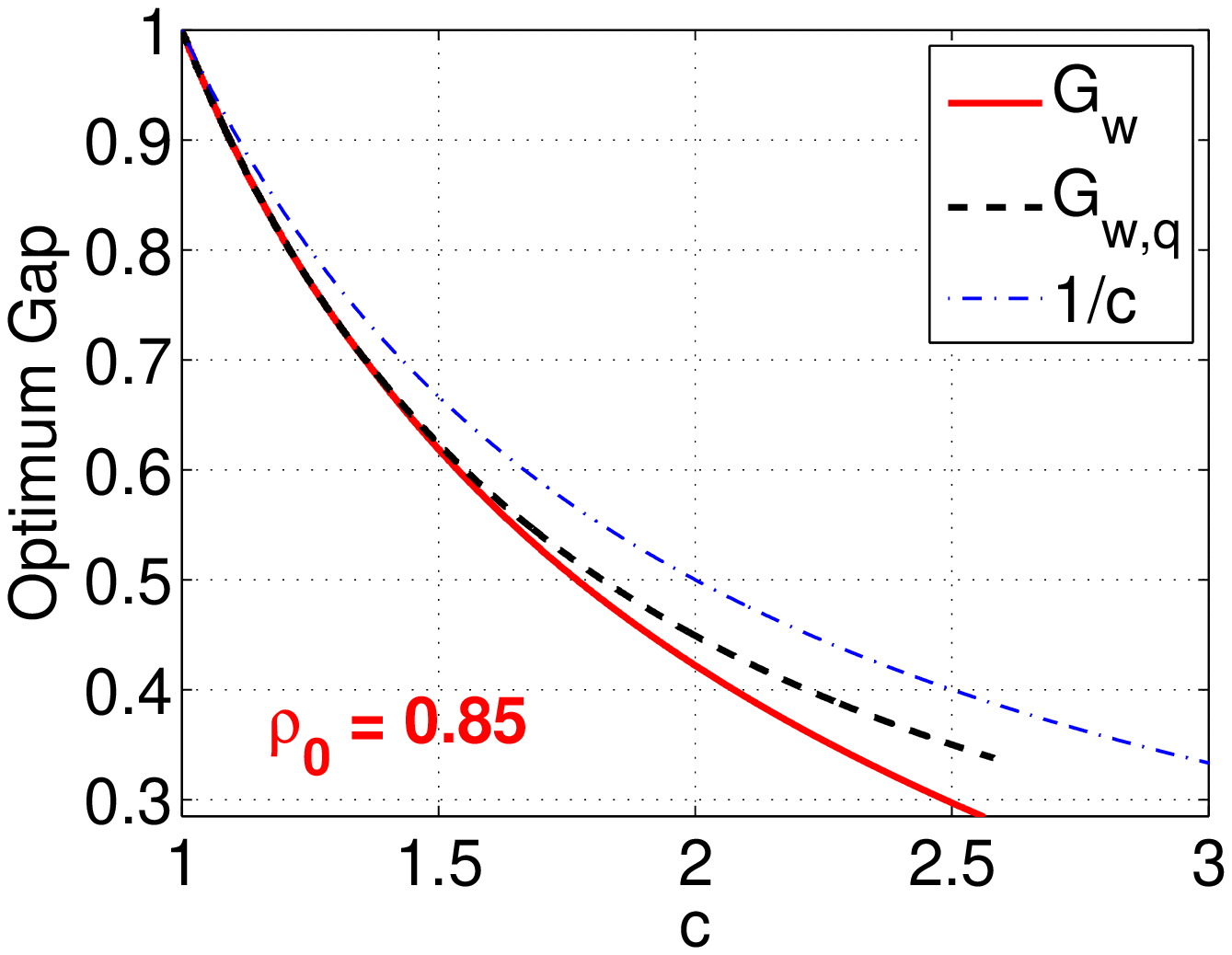}
}

\hspace{-0.15in}
\mbox{
\includegraphics[width = 1.25in]{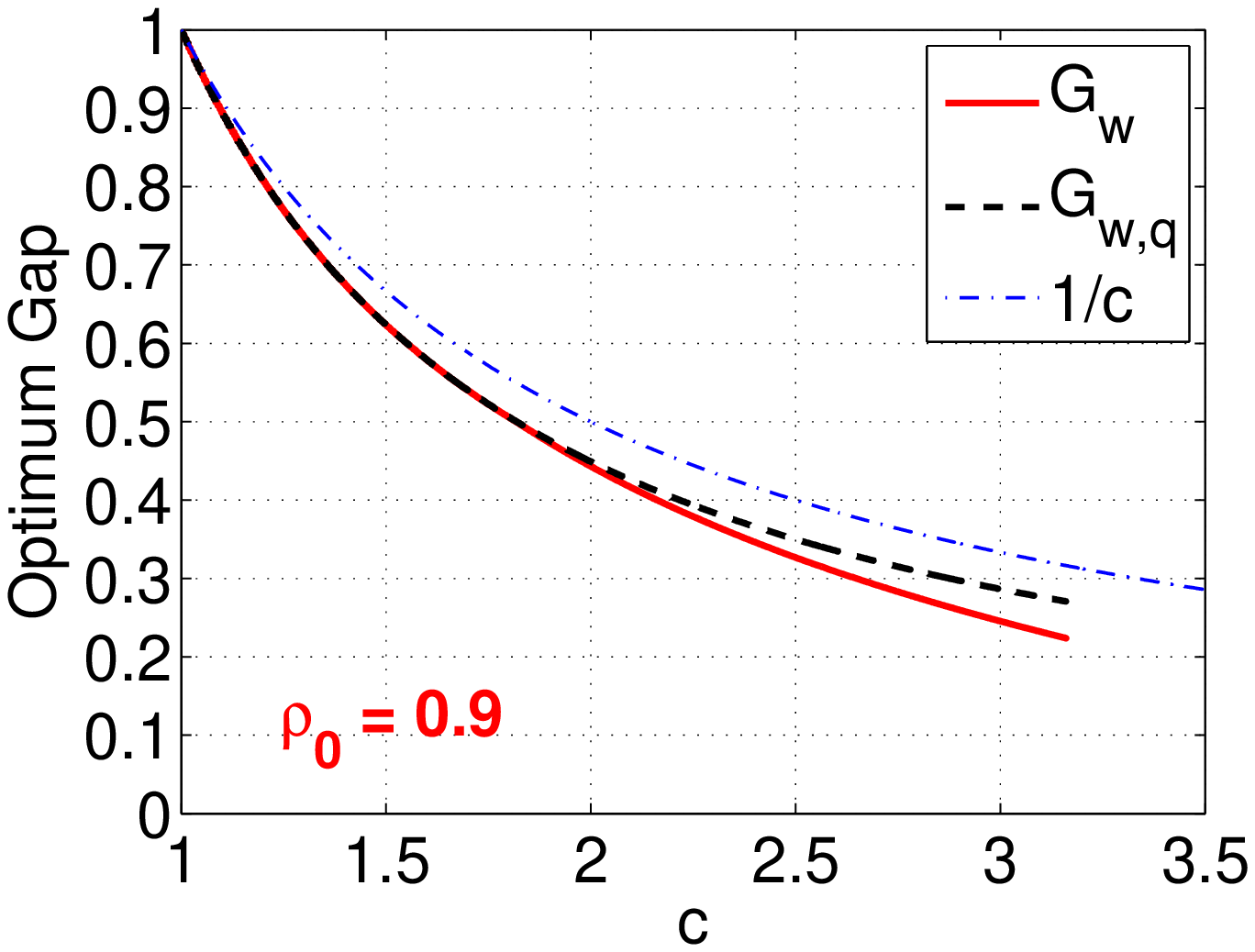}\hspace{-0.08in}
\includegraphics[width = 1.25in]{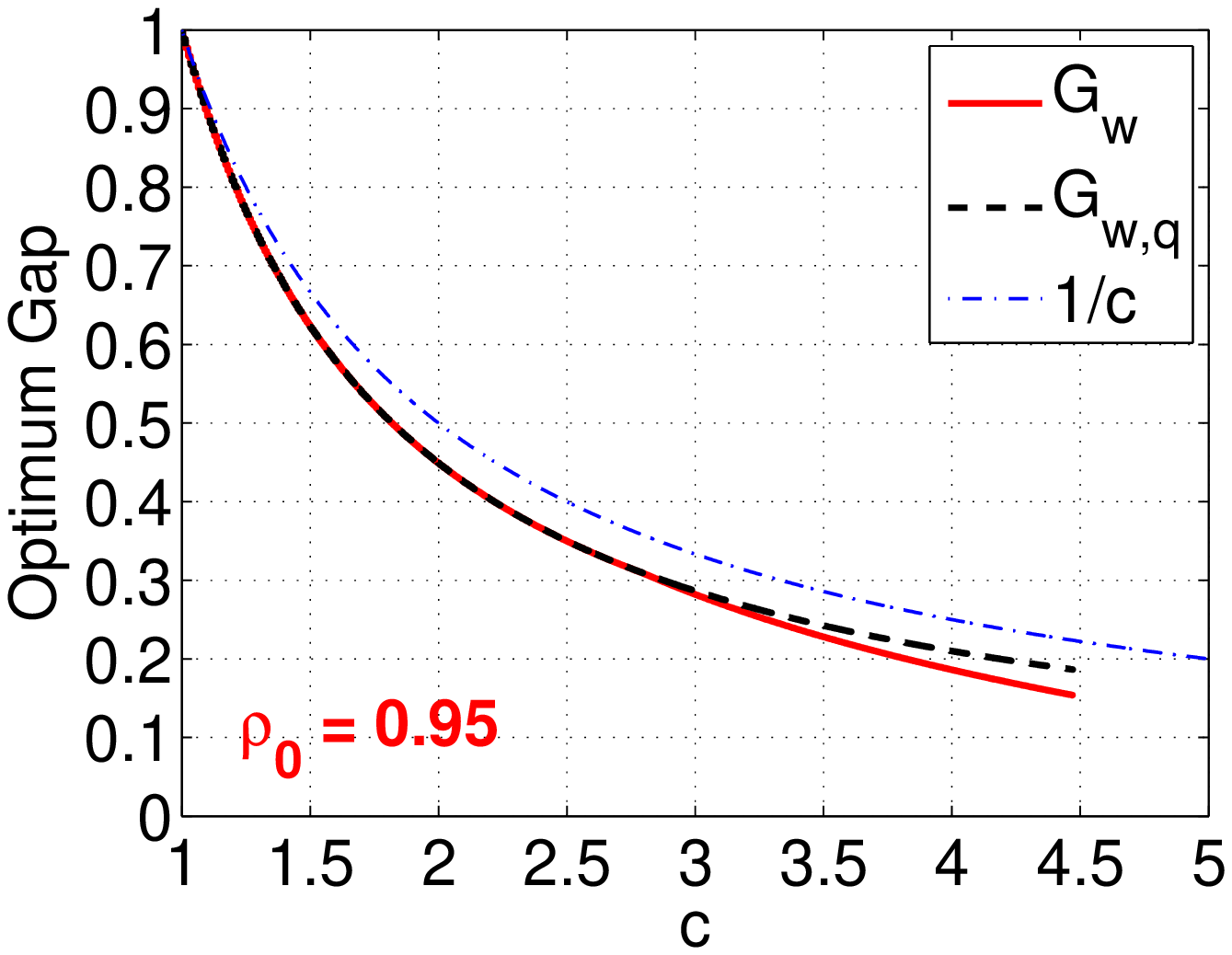}\hspace{-0.08in}
\includegraphics[width = 1.25in]{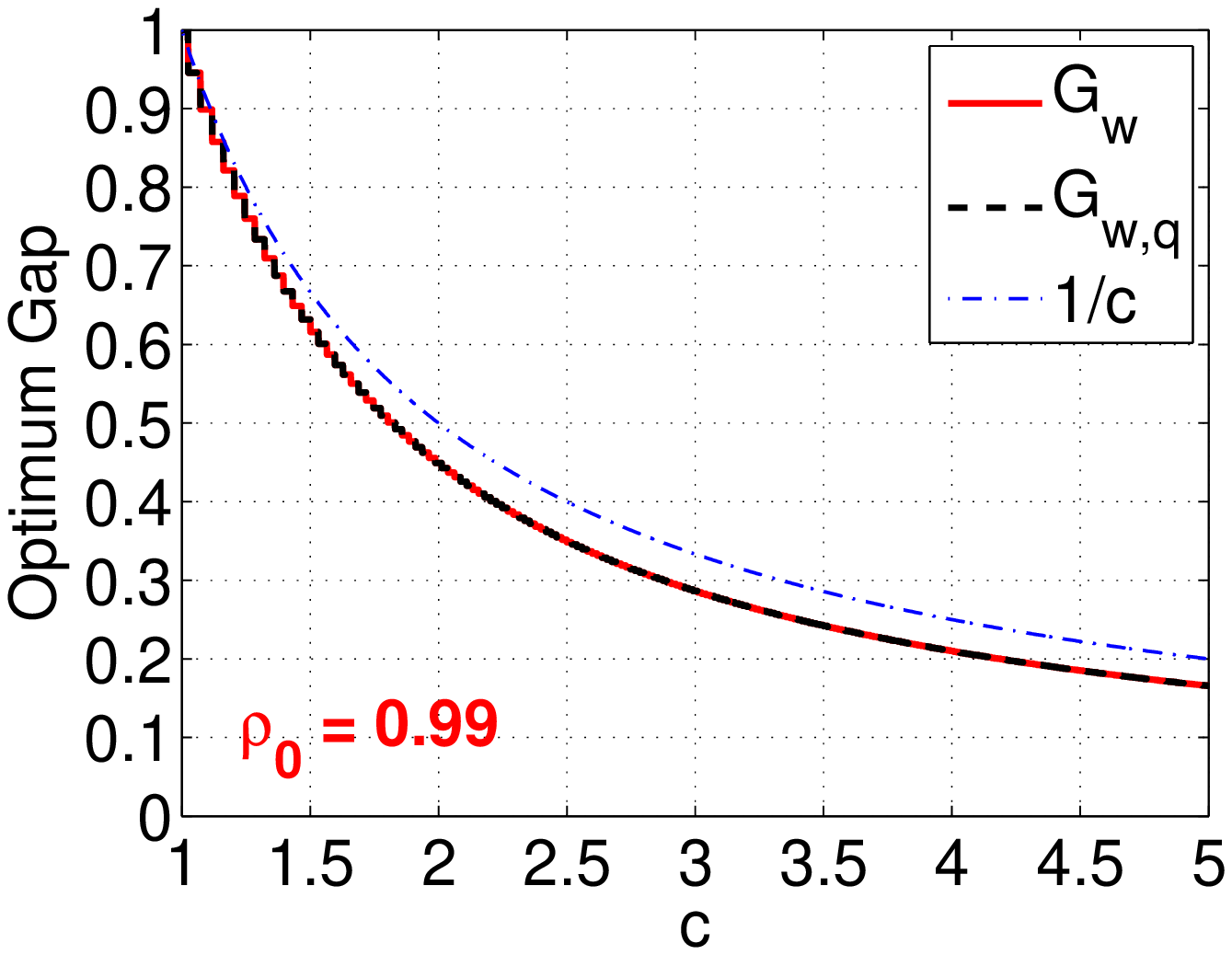}
}

\vspace{-.15in}
\caption{Comparison of the optimum gaps (smaller the better) for $h_w$ and $h_{w,q}$. For each $\rho_0$ and $c$, we can find the smallest gaps individually for $h_w$ and $h_{w,q}$, over the entire range of $w$. For all target similarity levels $\rho_0$, both $h_{w,q}$ and $h_w$ exhibit better performance than $1/c$. $h_w$ always has smaller gap than $h_{w,q}$, although in high similarity region both perform similarly. }\label{fig_GwqOpt}\vspace{-0.1in}
\end{figure}

Figure~\ref{fig_GwqOpt} compares $G_w$ with $G_{w,q}$ at their ``optimum'' $w$ values, as functions of $c$, for a wide range of target similarity $\rho_0$ levels. Basically, at each $c$ and $\rho_0$, we choose the $w$ to minimize $G_w$ and the  $w$ to minimize $G_{w,q}$. This figure illustrates that $G_w$ is smaller than $G_{w,q}$, noticeably so in the low similarity region.\\

\begin{figure}[h!]
\hspace{-.15in}
\mbox{
\includegraphics[width = 1.25in]{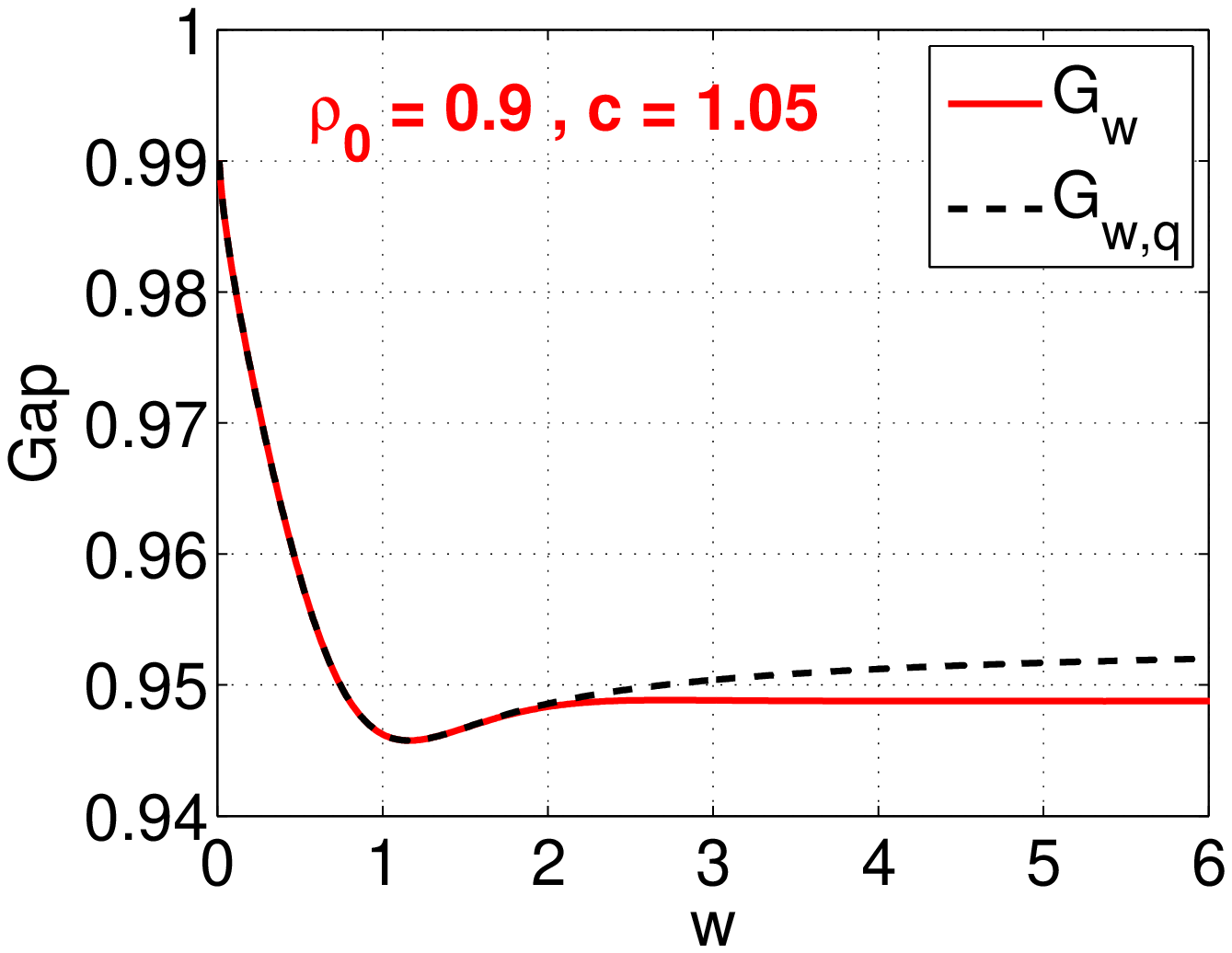}\hspace{-0.08in}
\includegraphics[width = 1.25in]{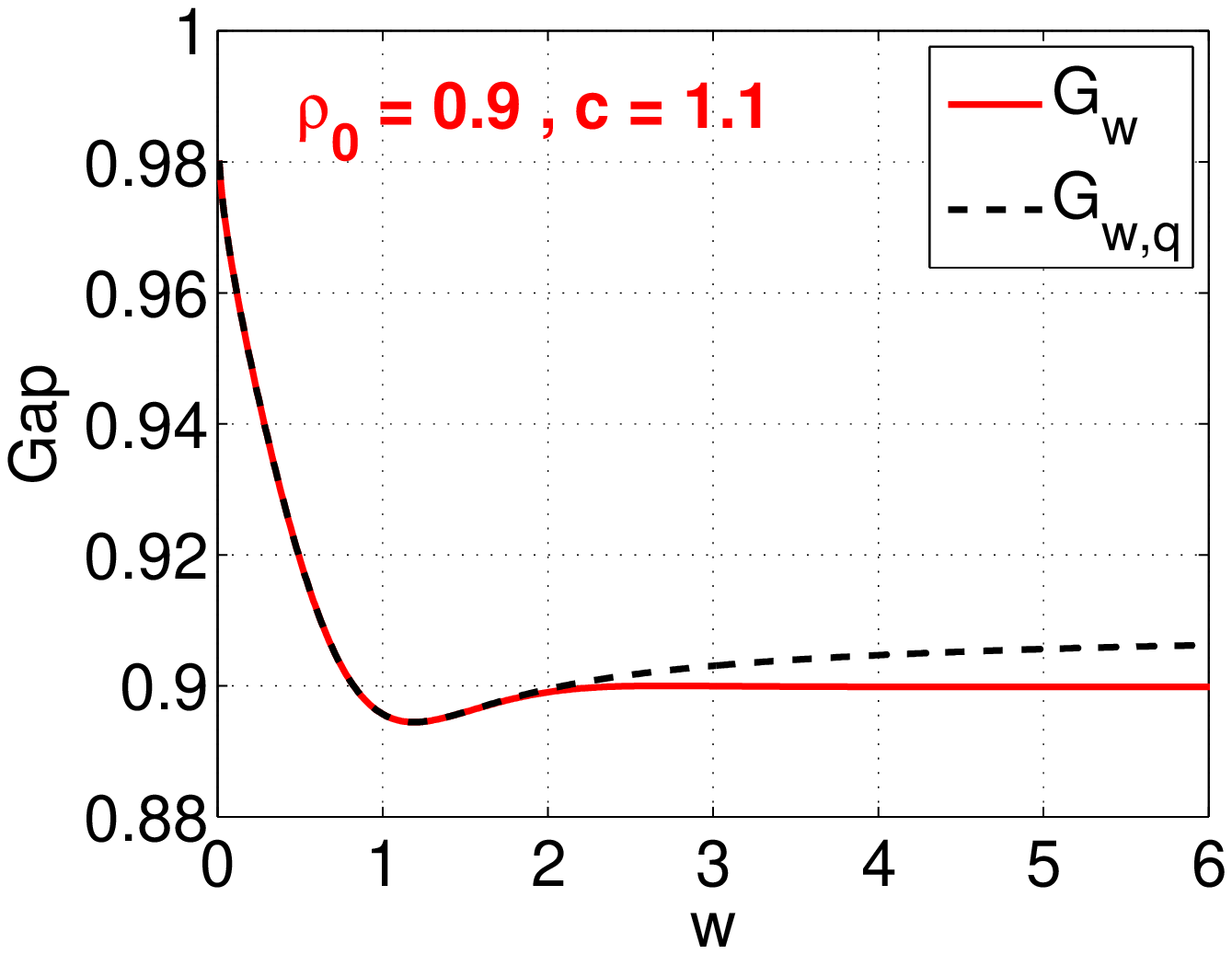}\hspace{-0.08in}
\includegraphics[width = 1.25in]{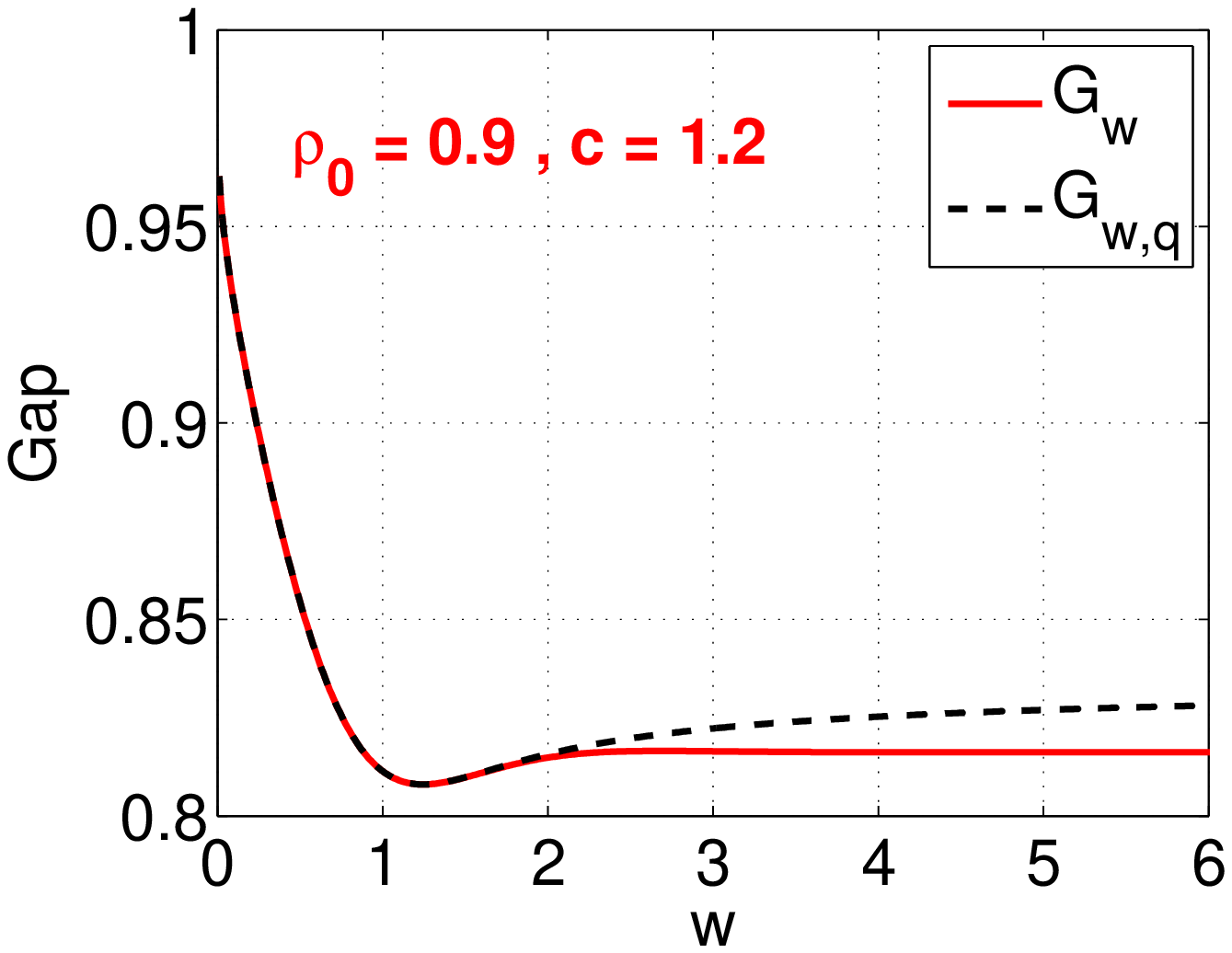}
}

\hspace{-.15in}
\mbox{
\includegraphics[width = 1.25in]{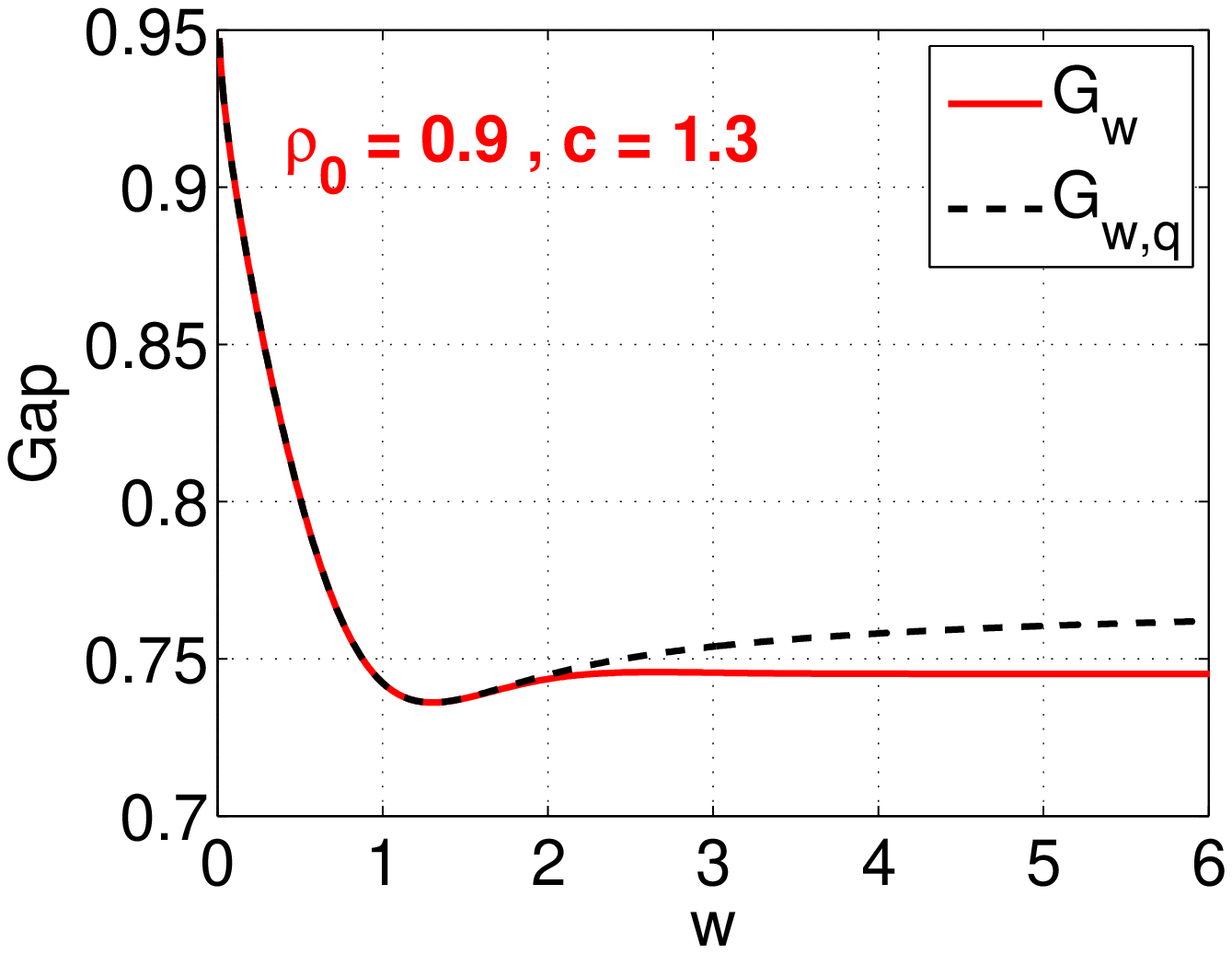}\hspace{-0.08in}
\includegraphics[width = 1.25in]{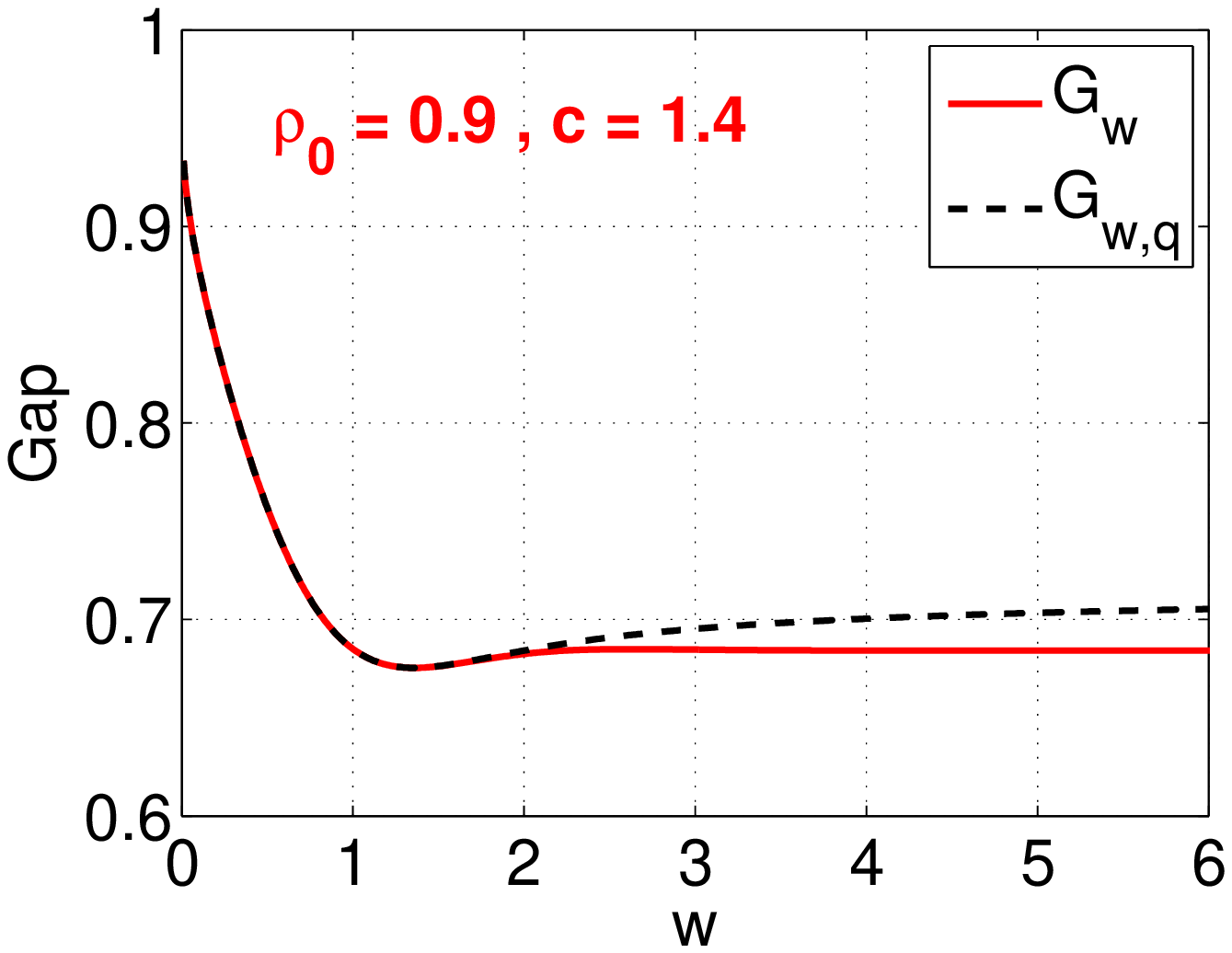}\hspace{-0.08in}
\includegraphics[width = 1.25in]{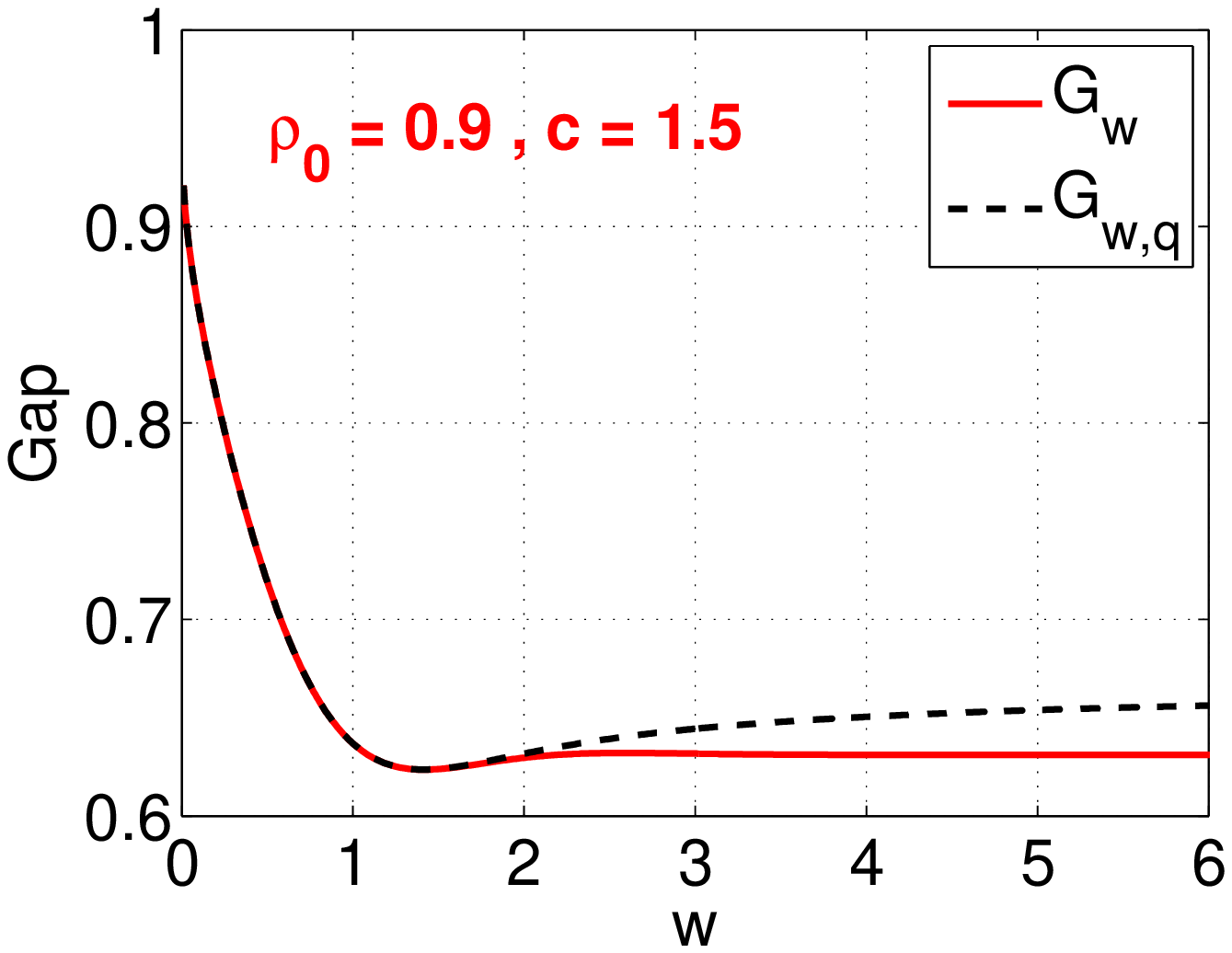}
}

\hspace{-.15in}
\mbox{
\includegraphics[width = 1.25in]{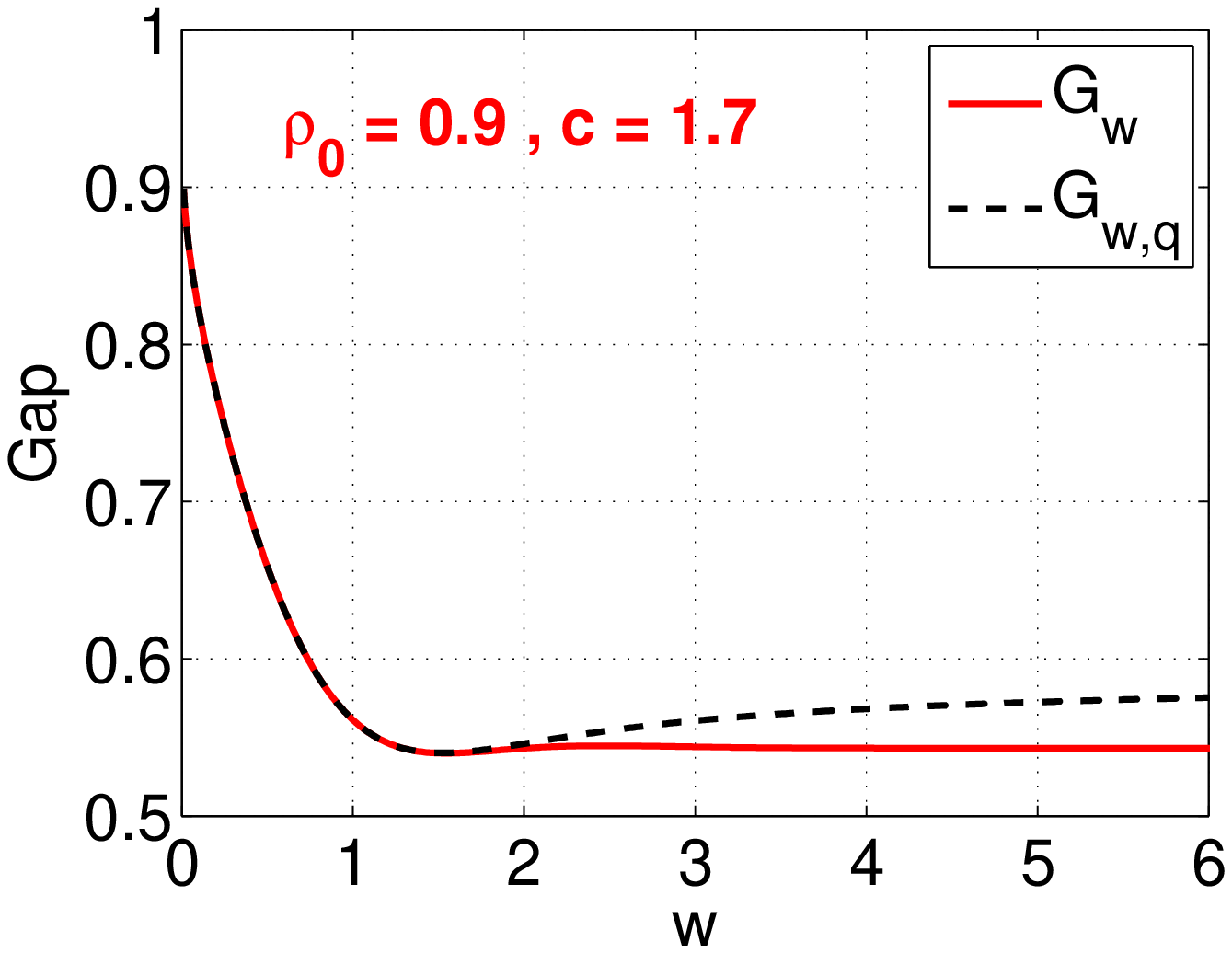}\hspace{-0.08in}
\includegraphics[width = 1.25in]{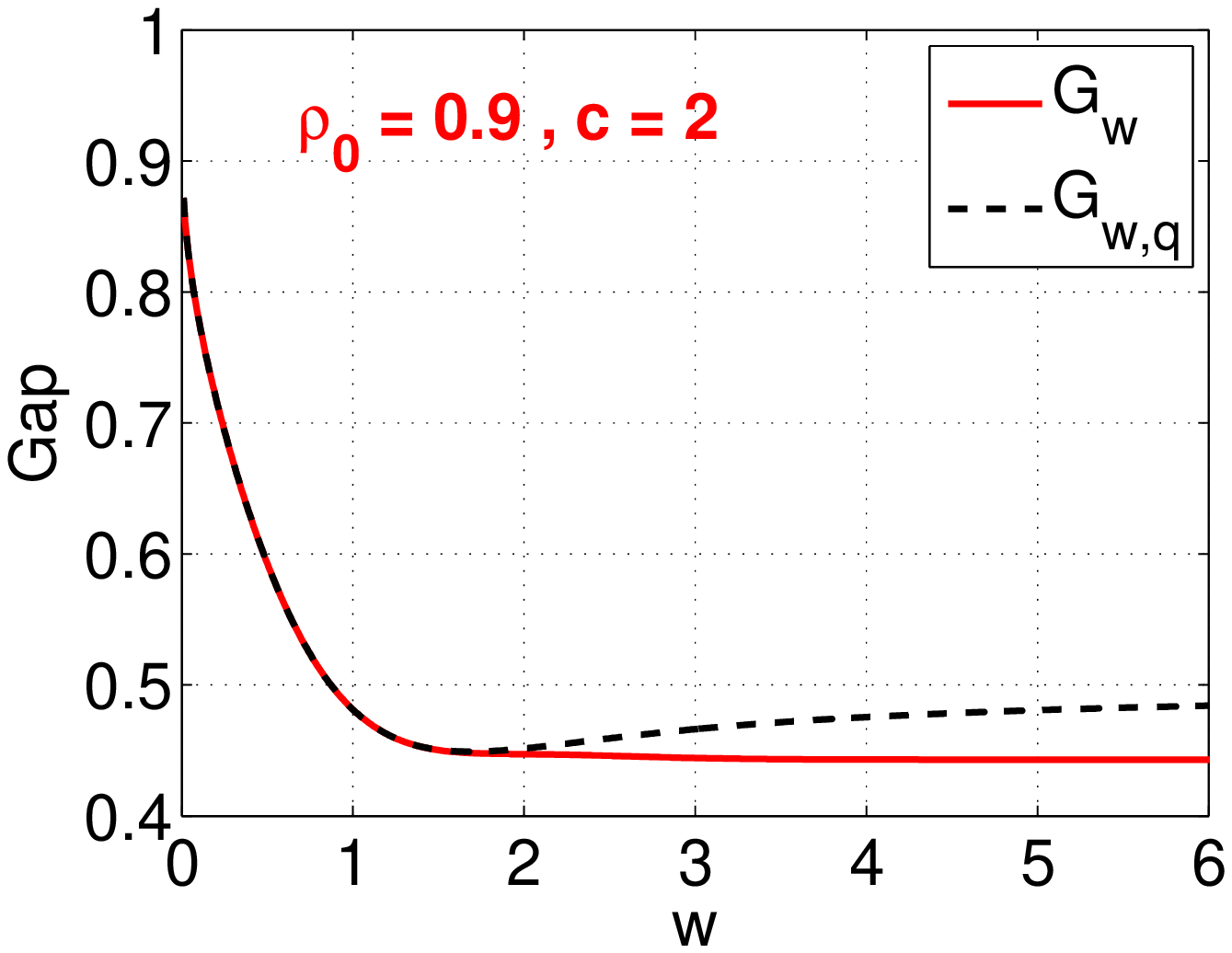}\hspace{-0.08in}
\includegraphics[width = 1.25in]{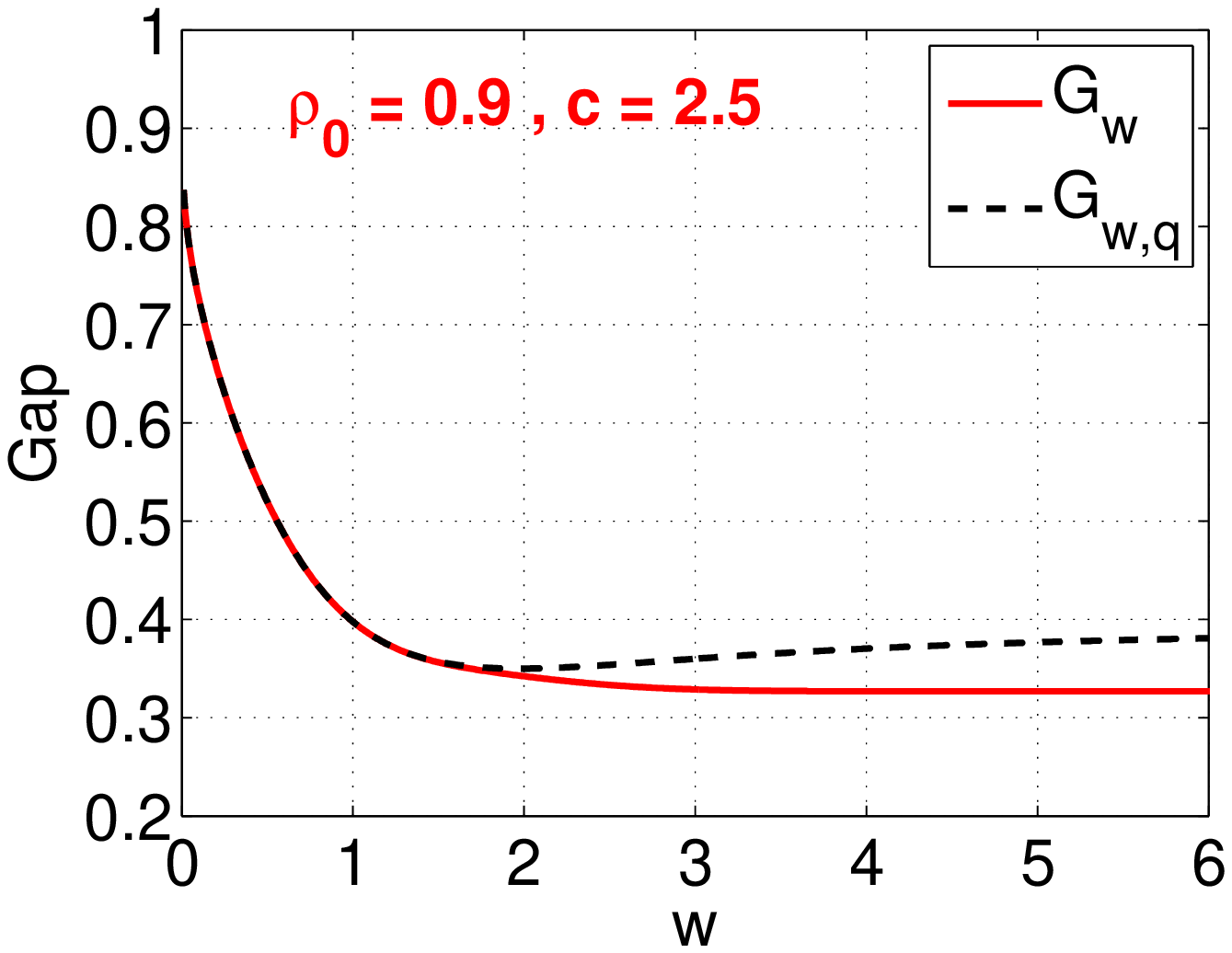}
}

\vspace{-.15in}
\caption{The gaps $G_w$ and $G_{w,q}$ as functions of $w$, for $\rho_0 = 0.9$. The lowest points on the curves are reflected in Figure~\ref{fig_GwqOpt}.}\label{fig_GwqR09C}\vspace{-0.1in}
\end{figure}

\begin{figure}[h!]

\hspace{-0.15in}
\mbox{
\includegraphics[width = 1.25in]{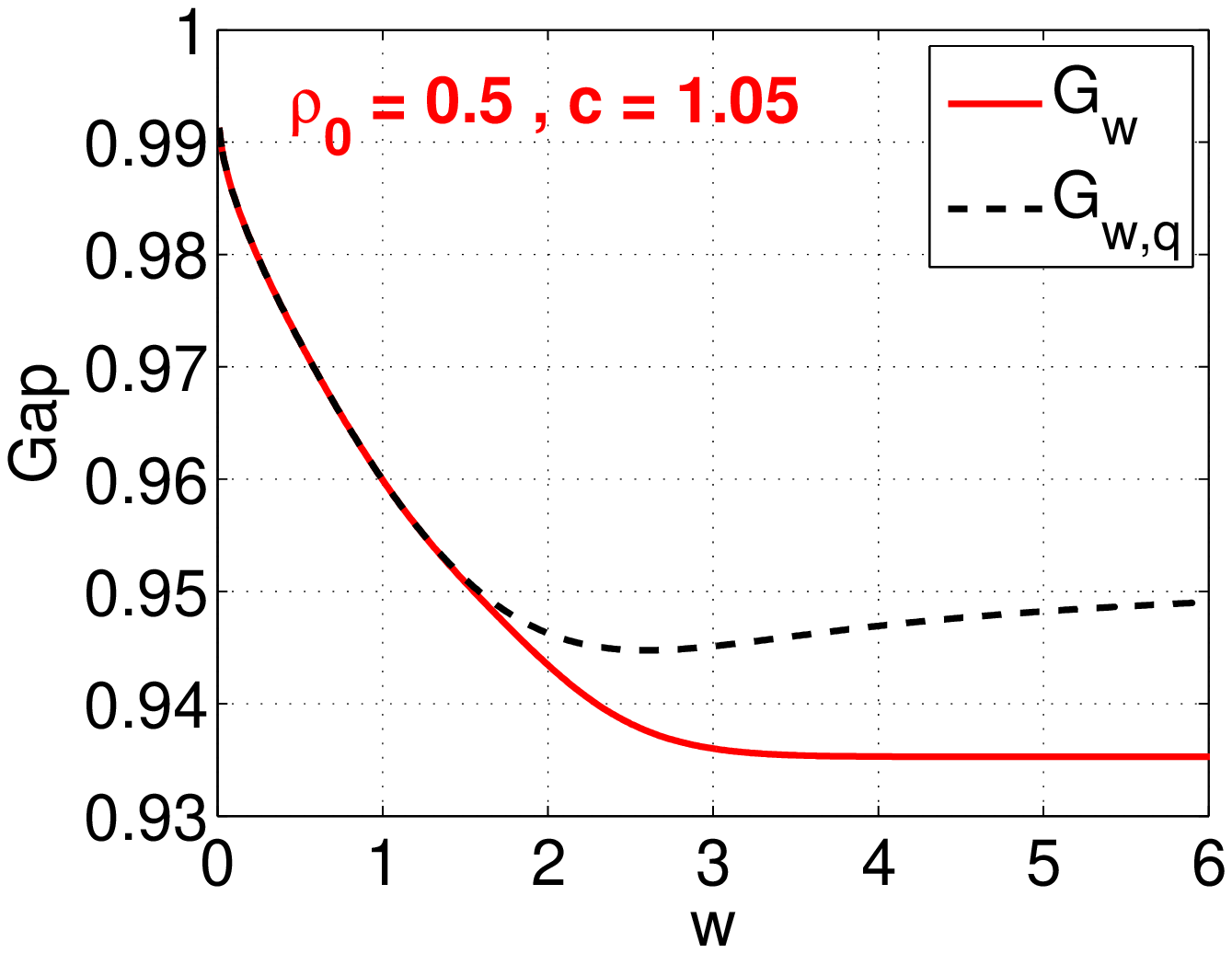}\hspace{-0.08in}
\includegraphics[width = 1.25in]{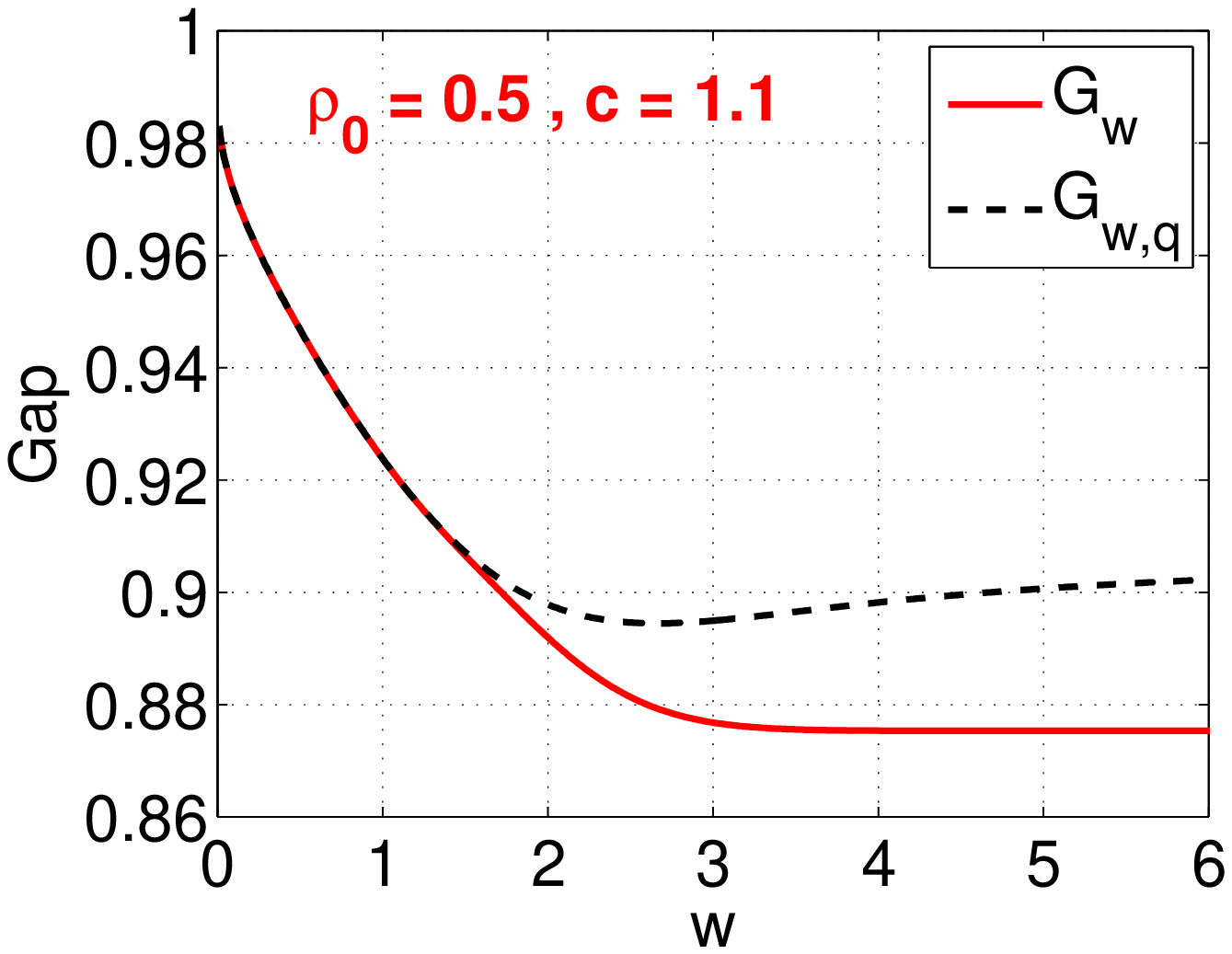}\hspace{-0.08in}
\includegraphics[width = 1.25in]{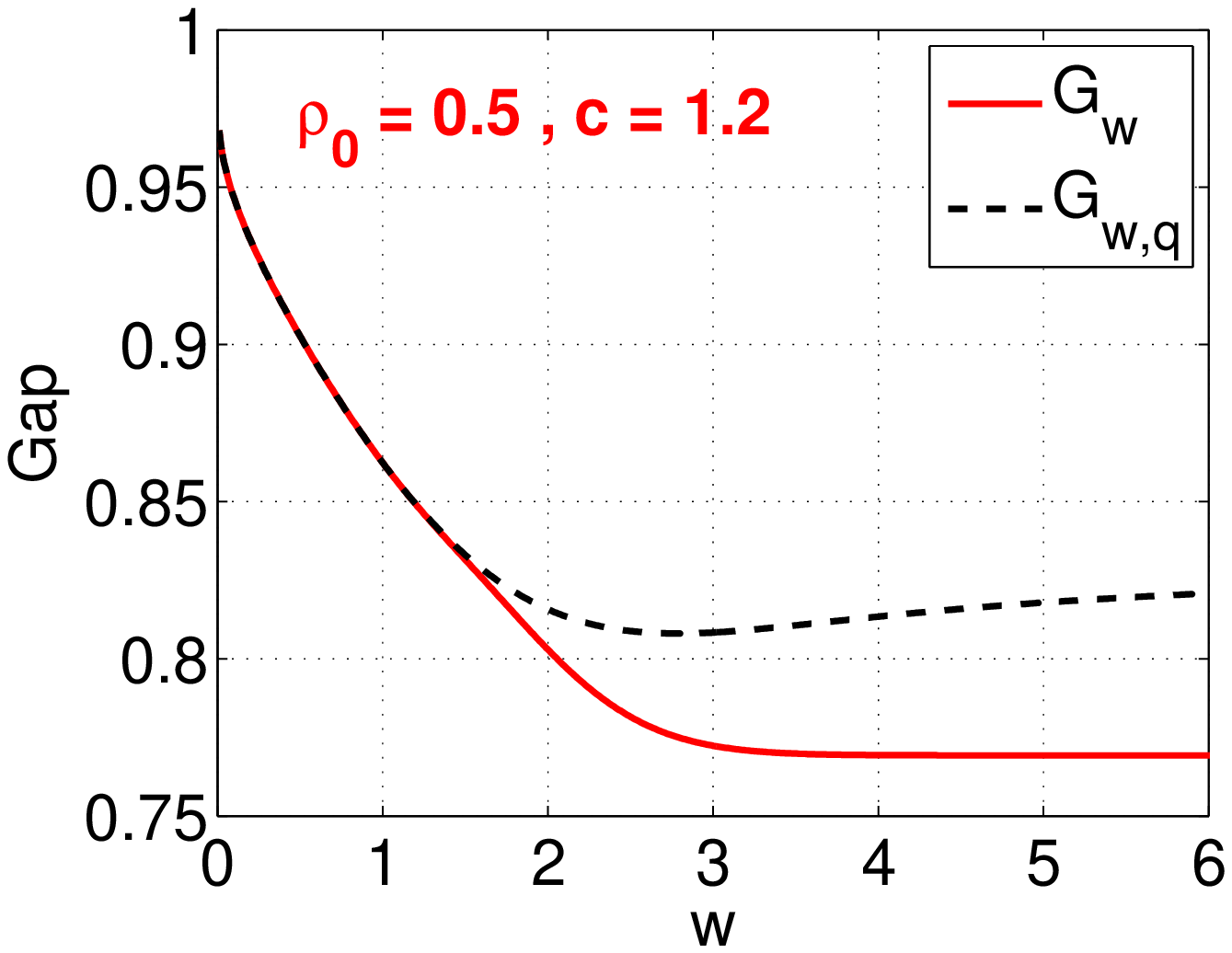}
}

\hspace{-0.15in}
\mbox{
\includegraphics[width = 1.25in]{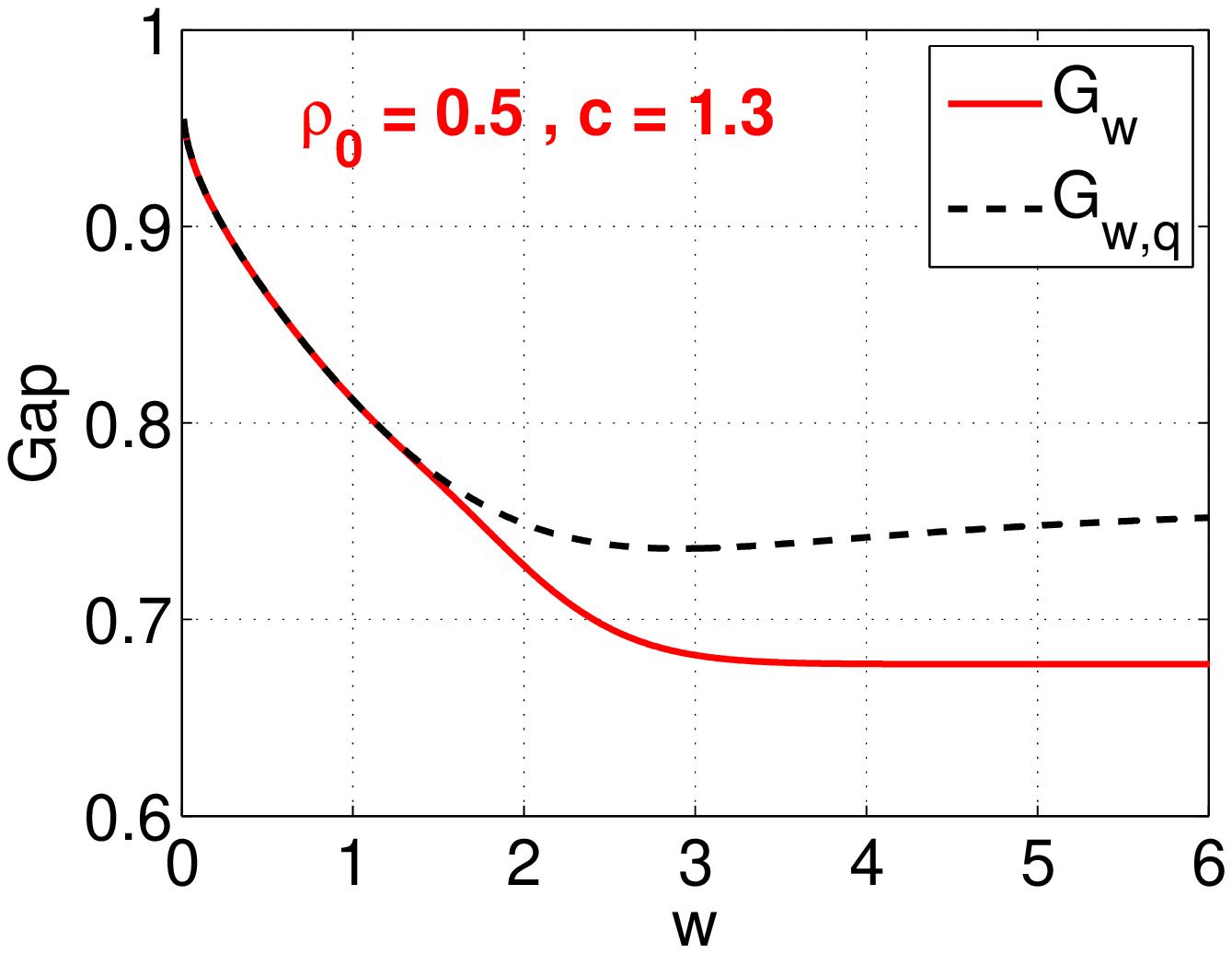}\hspace{-0.08in}
\includegraphics[width = 1.25in]{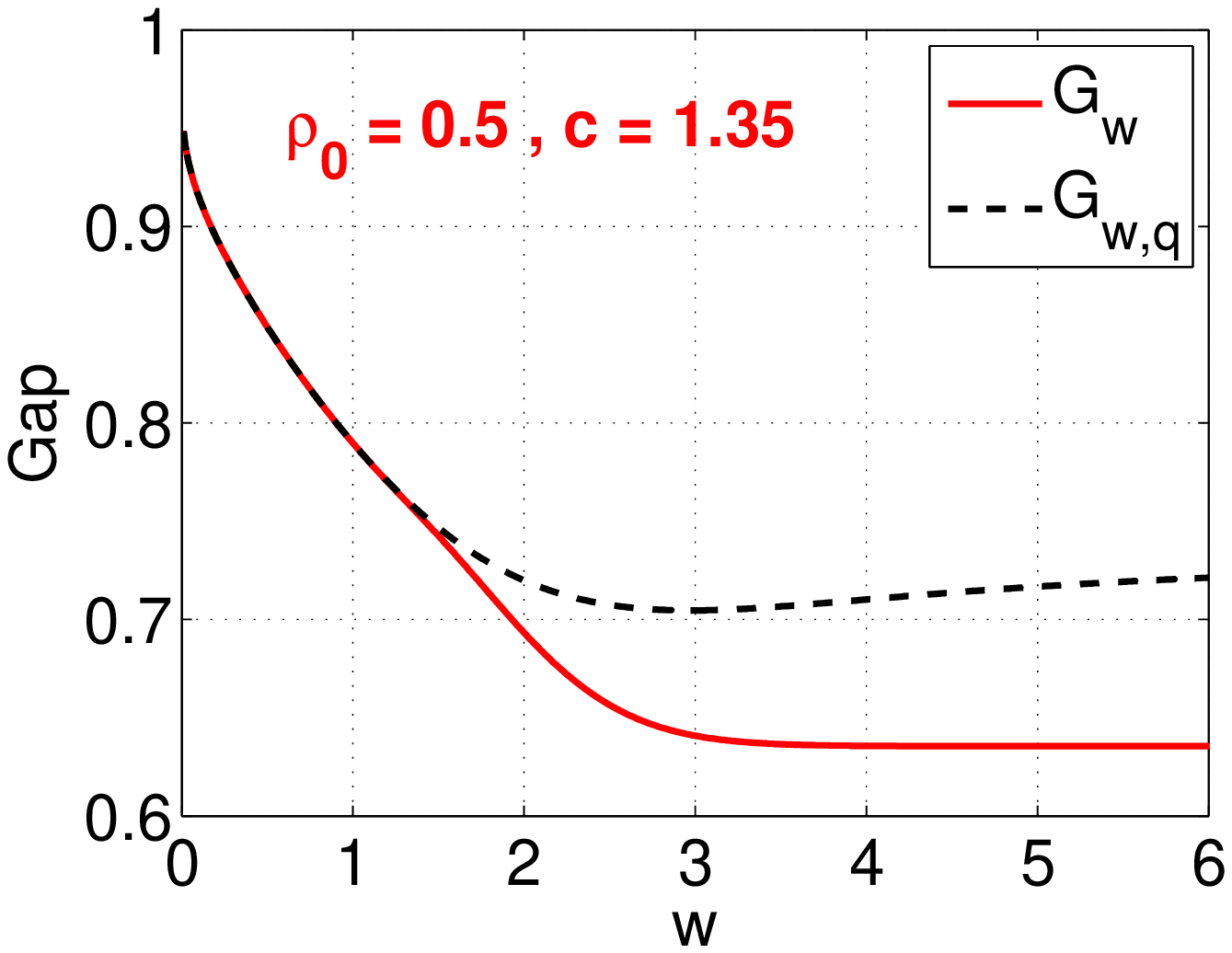}\hspace{-0.08in}
\includegraphics[width = 1.25in]{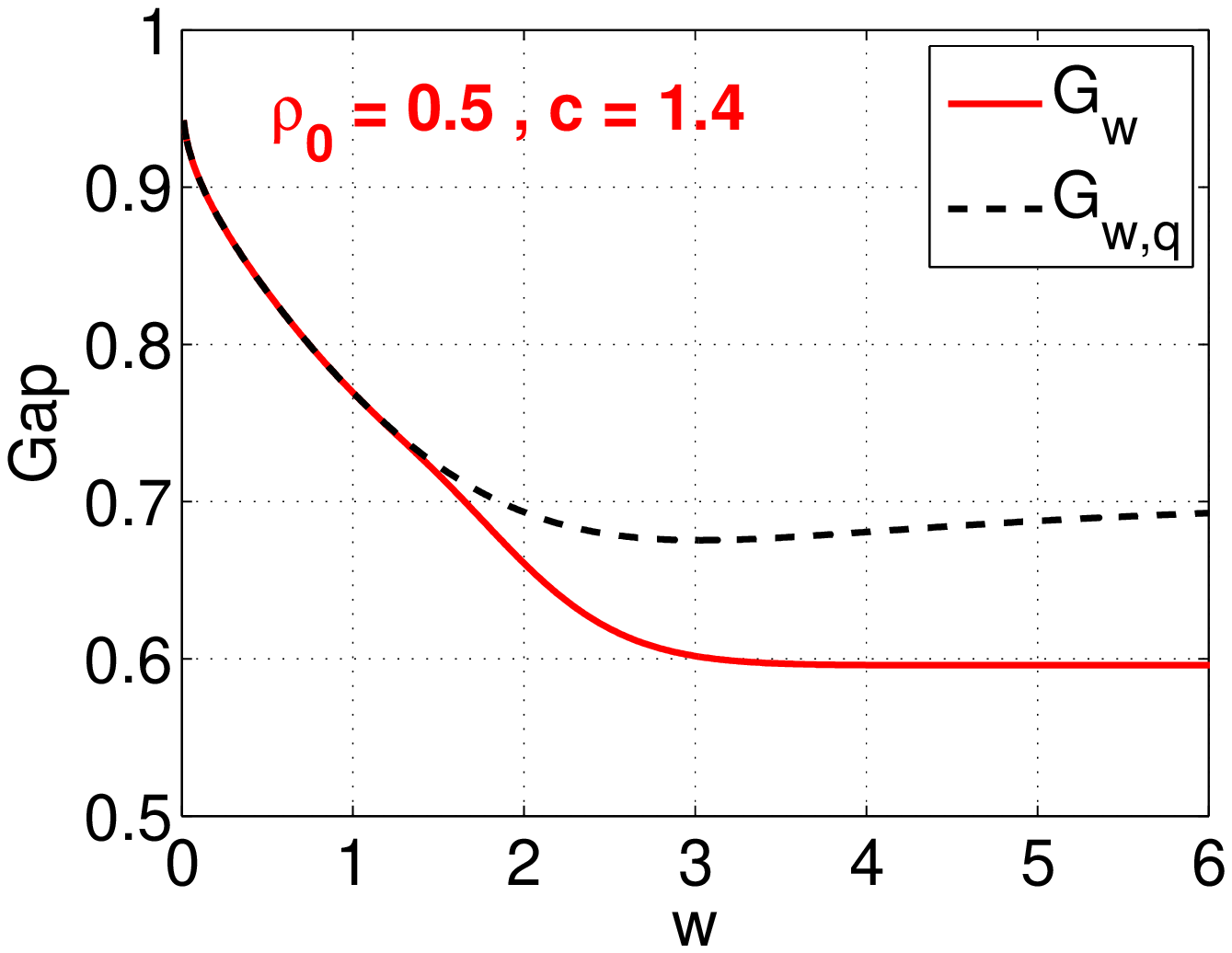}
}

\vspace{-.15in}
\caption{The gaps $G_w$ and $G_{w,q}$ as functions of $w$ for~$\rho_0~=~0.5$}\label{fig_GwqR05C}\vspace{-0.1in}
\end{figure}

Figure~\ref{fig_GwqR09C} and Figure~\ref{fig_GwqR05C} present $G_w$ and $G_{w,q}$ as functions of $w$, for $\rho_0 = 0.9$ and $\rho_0 = 0.5$, respectively. In each figure, we plot the curves for a wide range of $c$ values.  These figures illustrate where the optimum $w$ values are obtained. Clearly, in the high similarity region, the smallest $G$ values are obtained at low $w$ values, especially at small $c$. In the low (or moderate) similarity   region, the smallest $G$ values are usually attained at relatively large $w$.

\begin{figure}[h!]

\hspace{-0.15in}
\mbox{
\includegraphics[width = 1.25in]{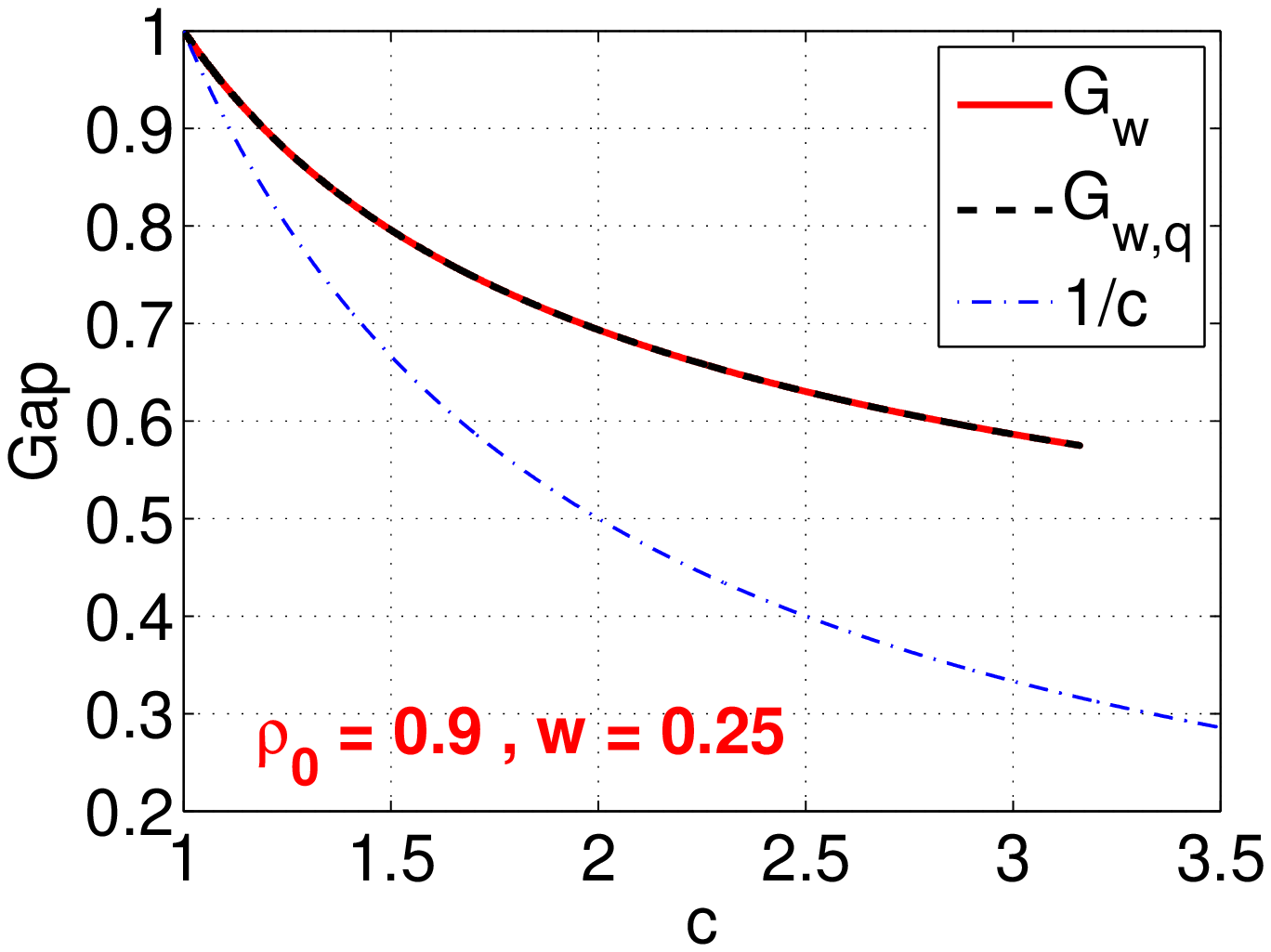}\hspace{-0.08in}
\includegraphics[width = 1.25in]{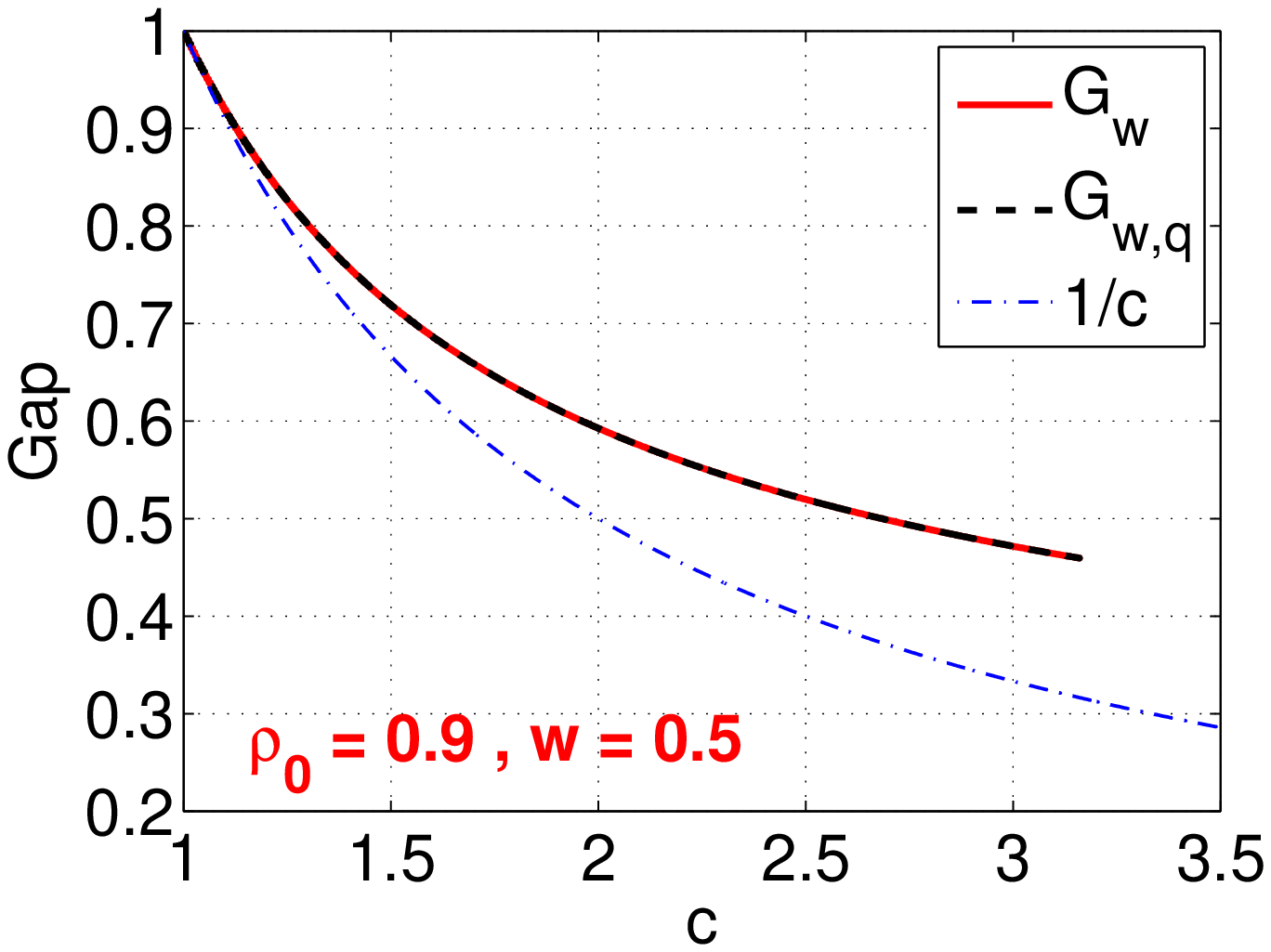}\hspace{-0.08in}
\includegraphics[width = 1.25in]{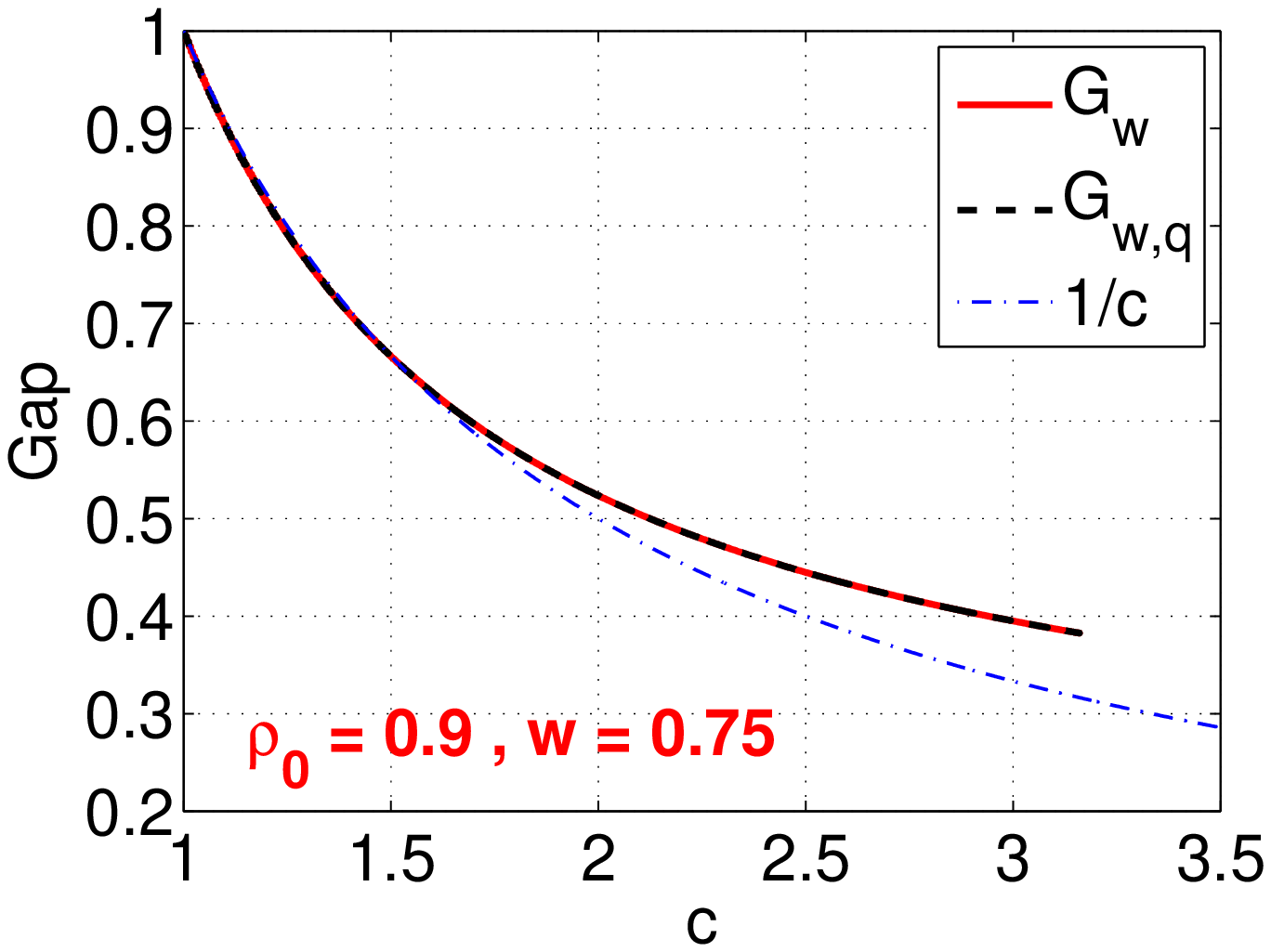}
}

\hspace{-0.15in}
\mbox{
\includegraphics[width = 1.25in]{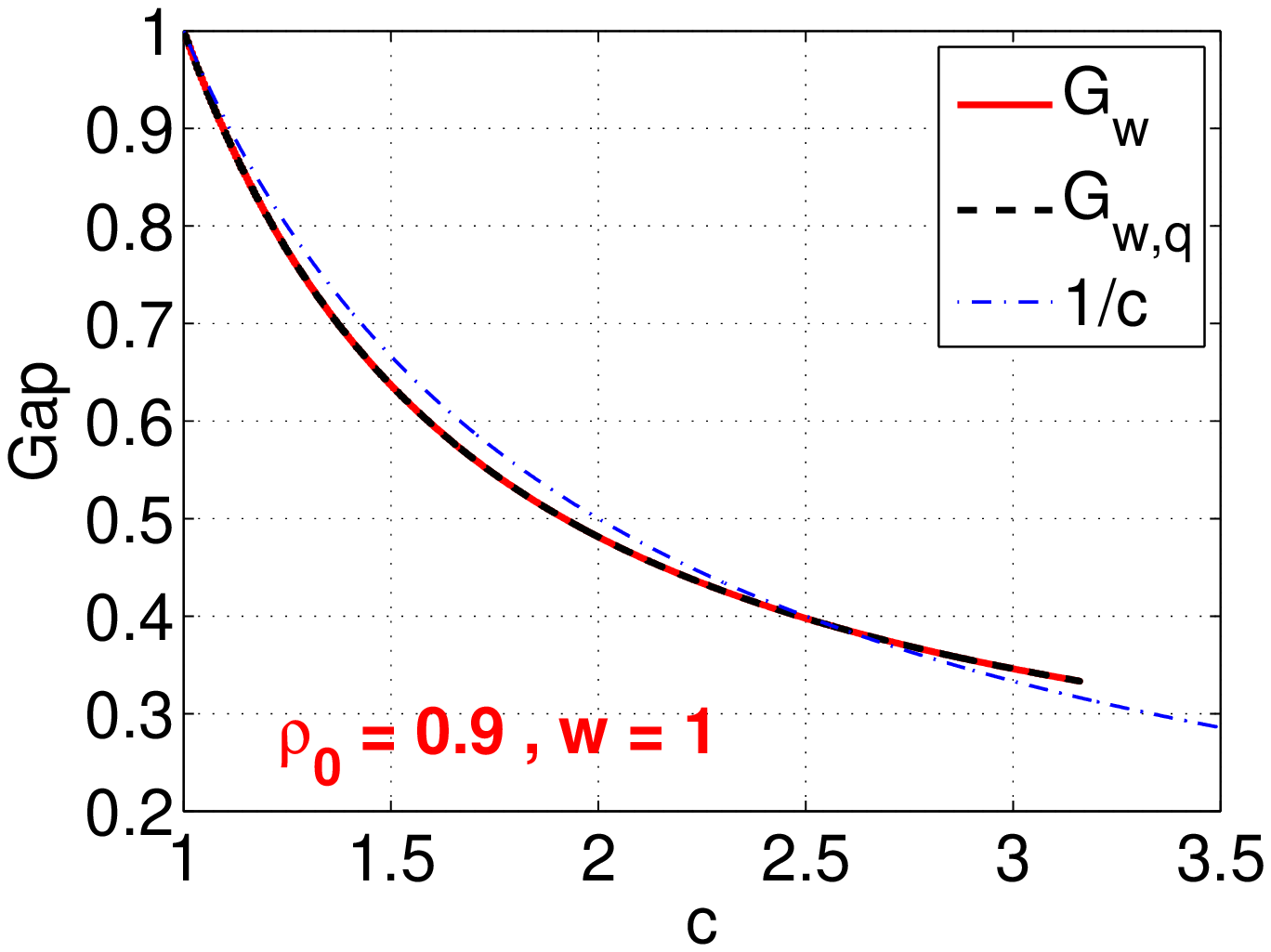}\hspace{-0.08in}
\includegraphics[width = 1.25in]{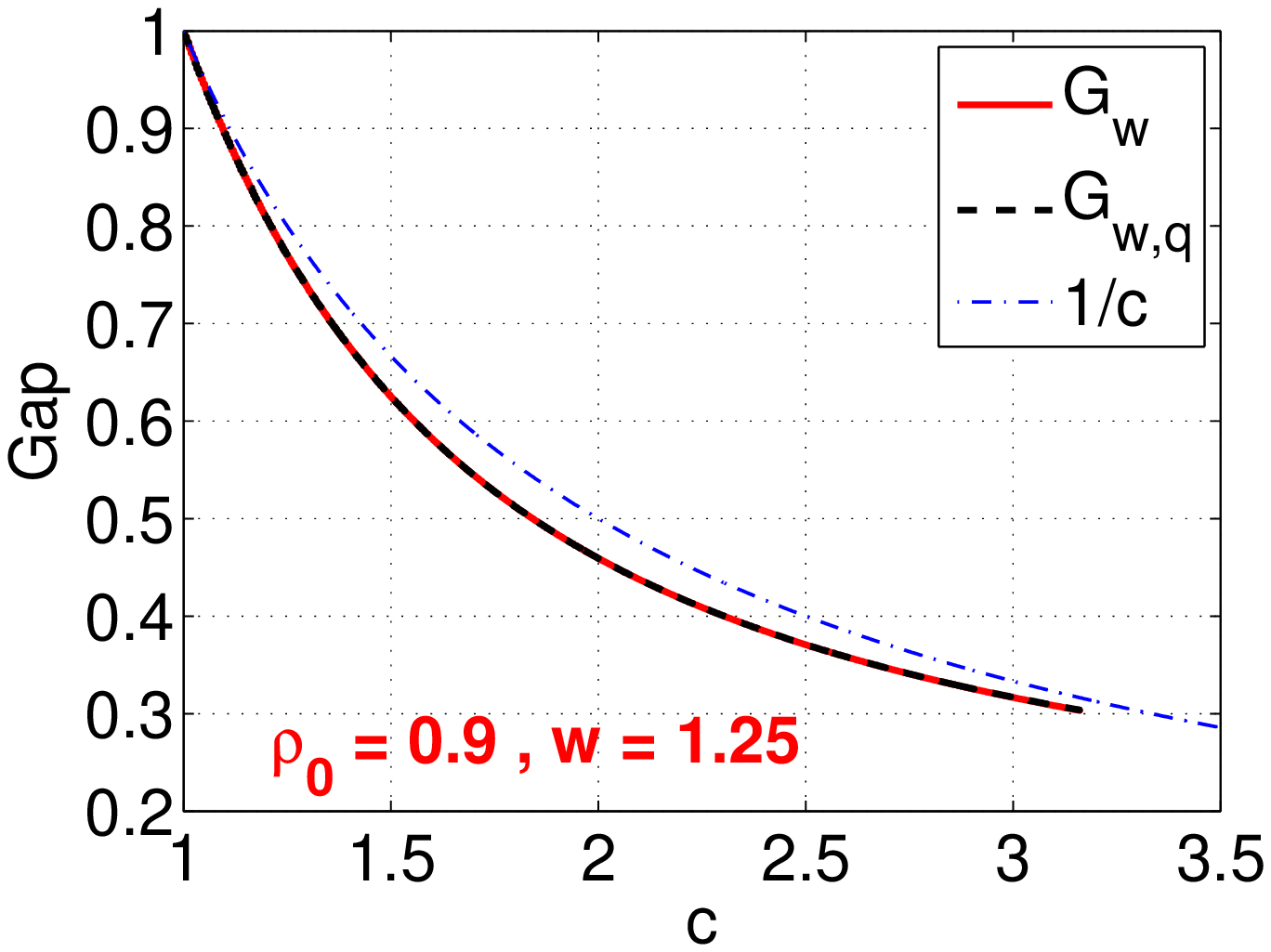}\hspace{-0.08in}
\includegraphics[width = 1.25in]{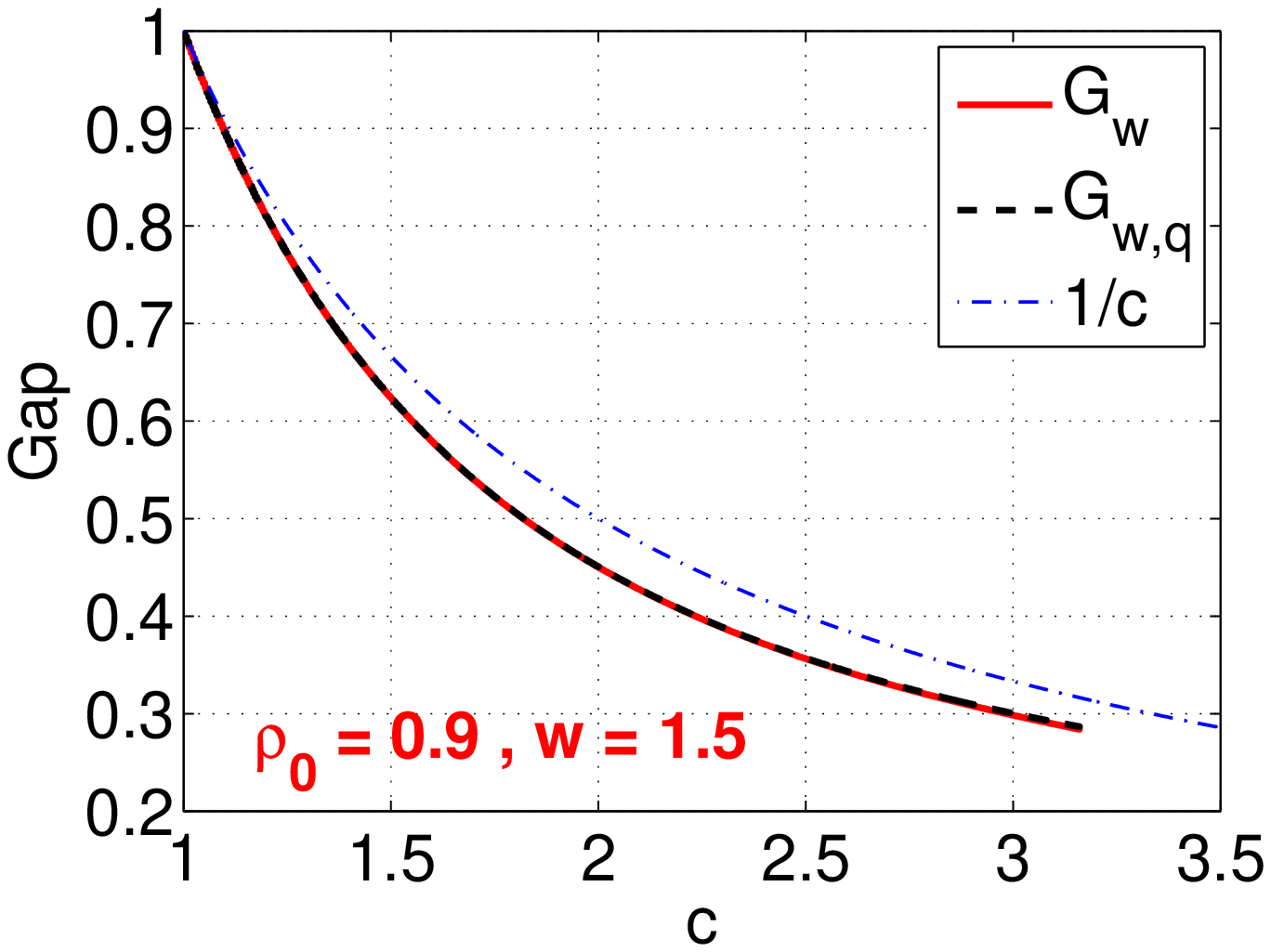}
}

\hspace{-0.15in}
\mbox{
\includegraphics[width = 1.25in]{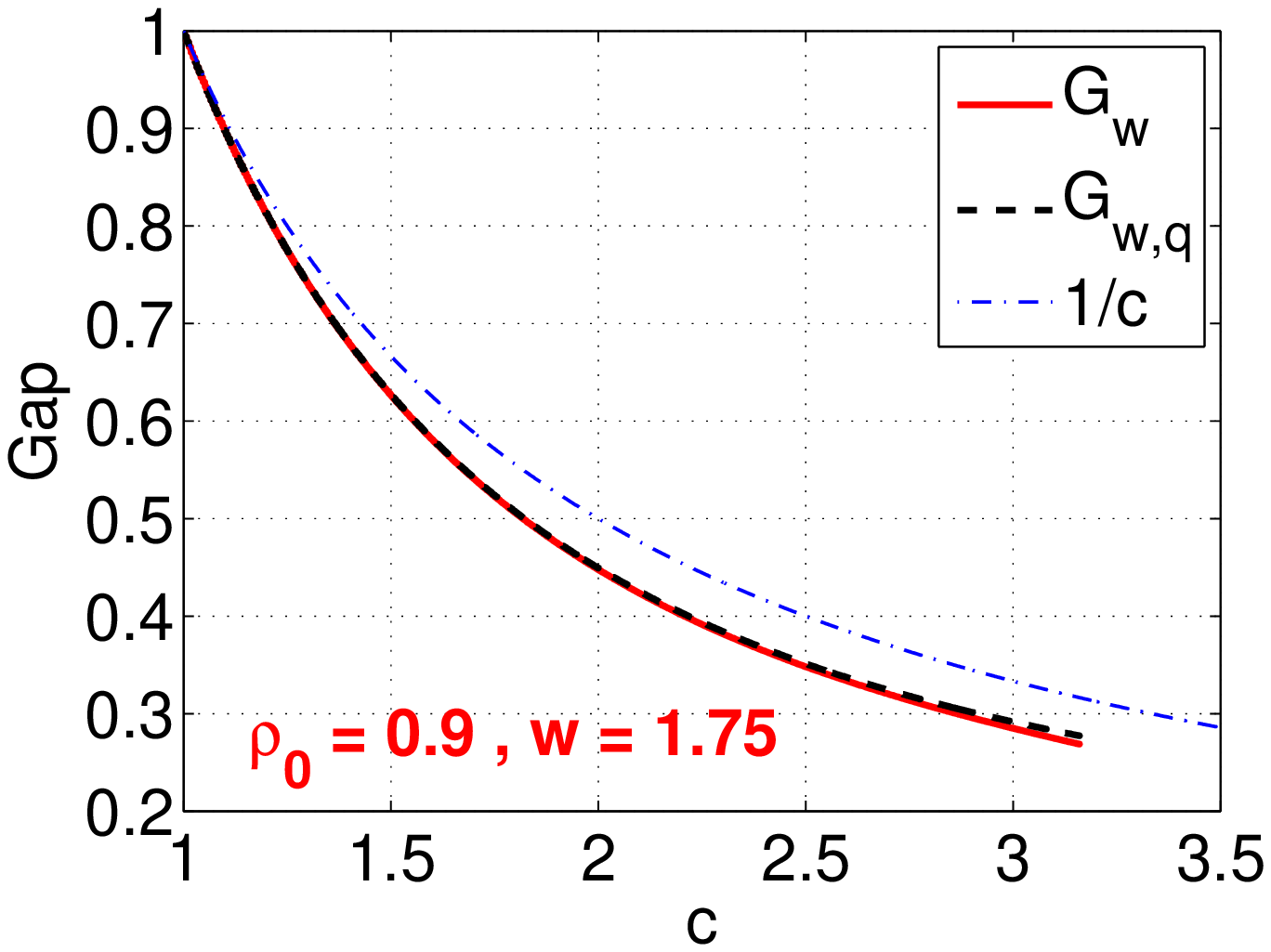}\hspace{-0.08in}
\includegraphics[width = 1.25in]{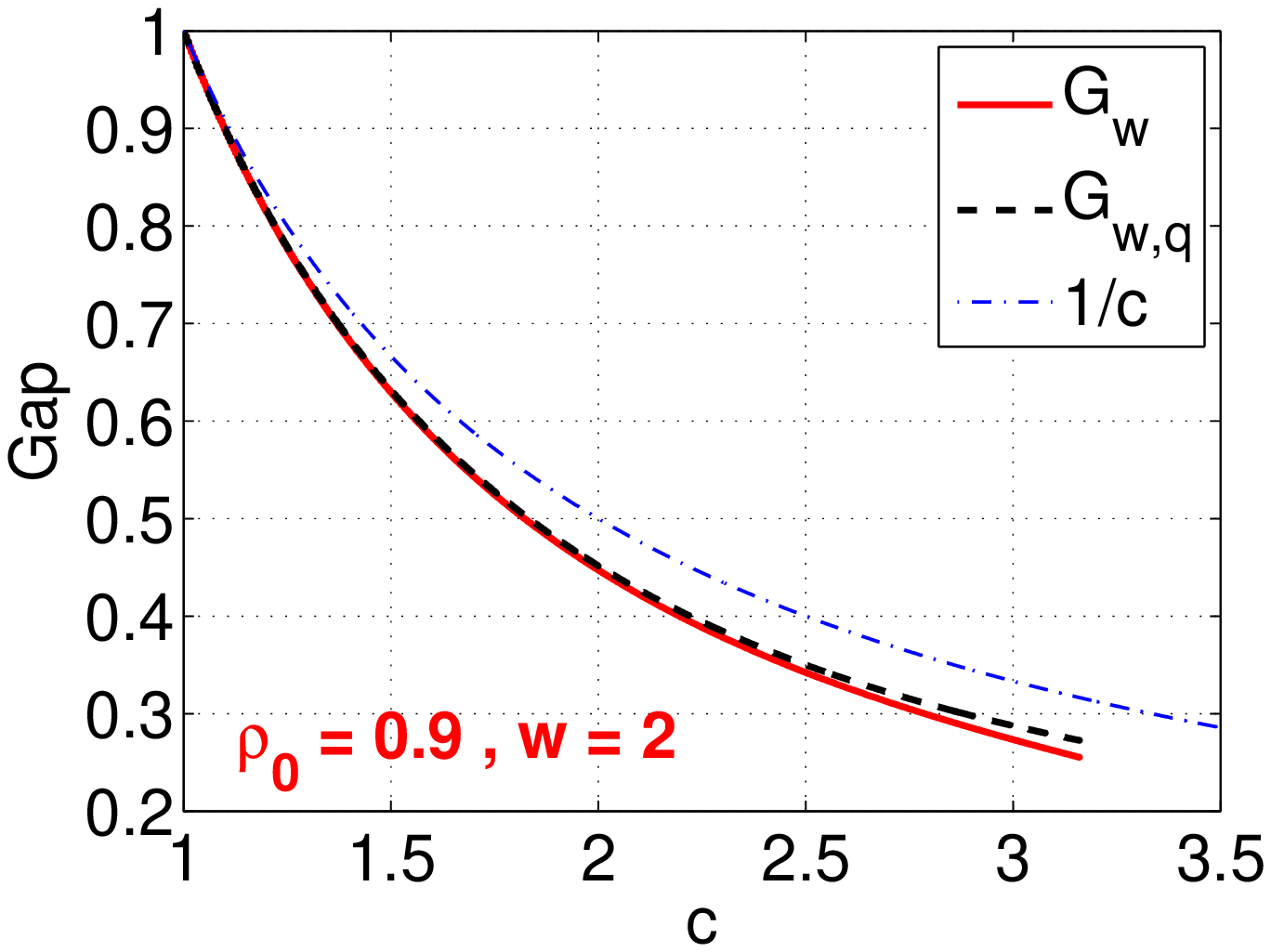}\hspace{-0.08in}
\includegraphics[width = 1.25in]{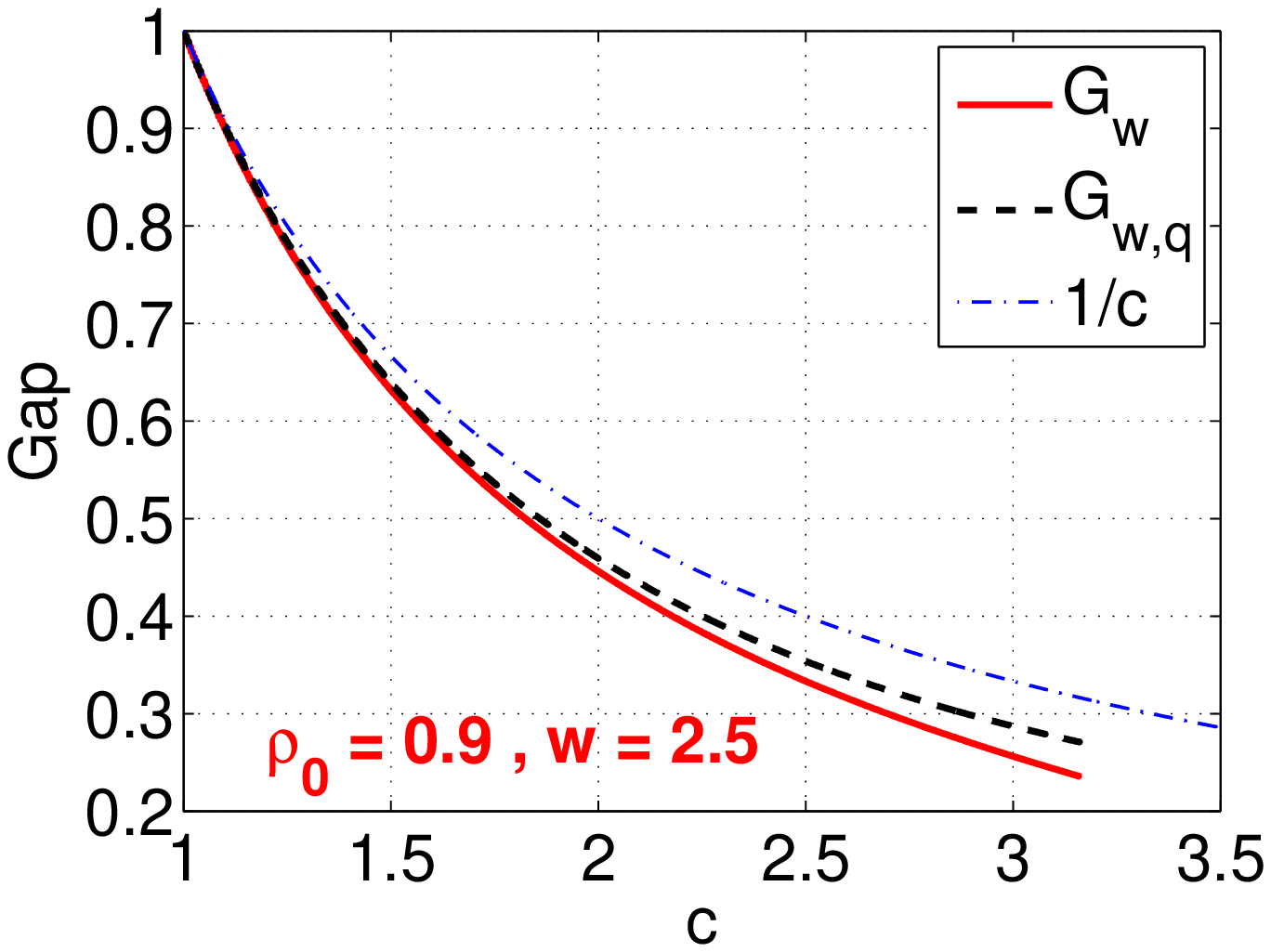}
}

\hspace{-0.15in}
\mbox{
\includegraphics[width = 1.25in]{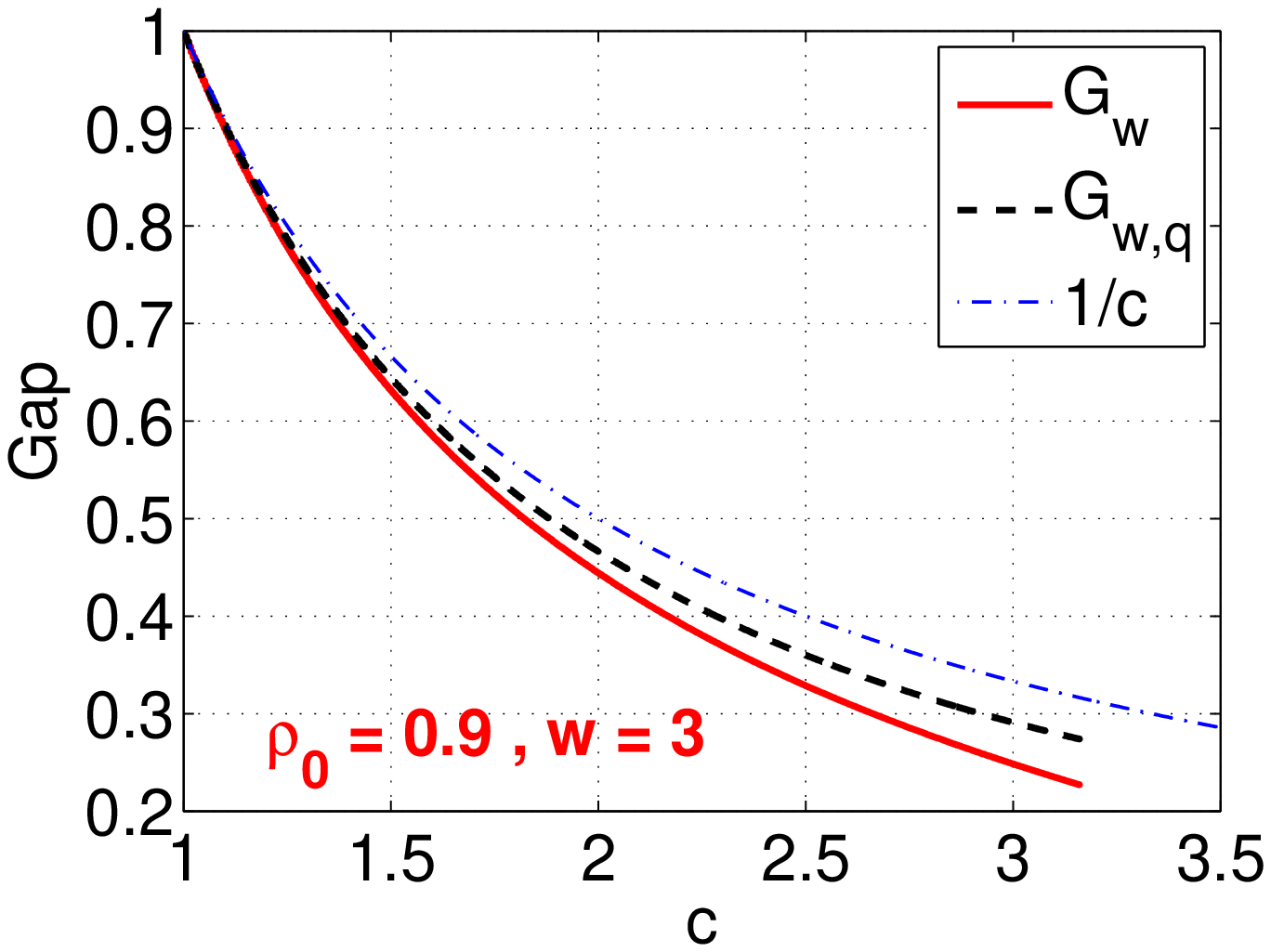}\hspace{-0.08in}
\includegraphics[width = 1.25in]{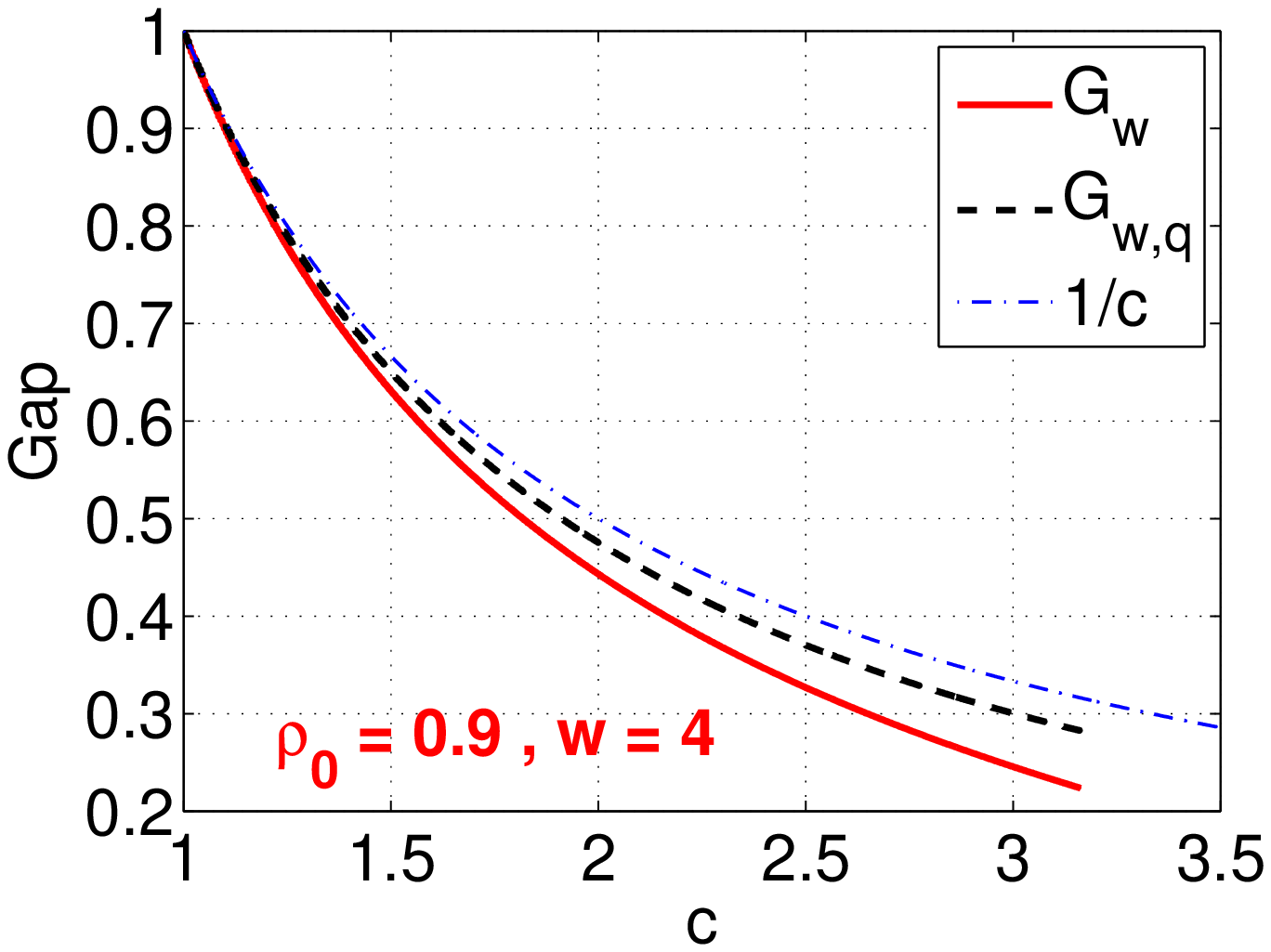}\hspace{-0.08in}
\includegraphics[width = 1.25in]{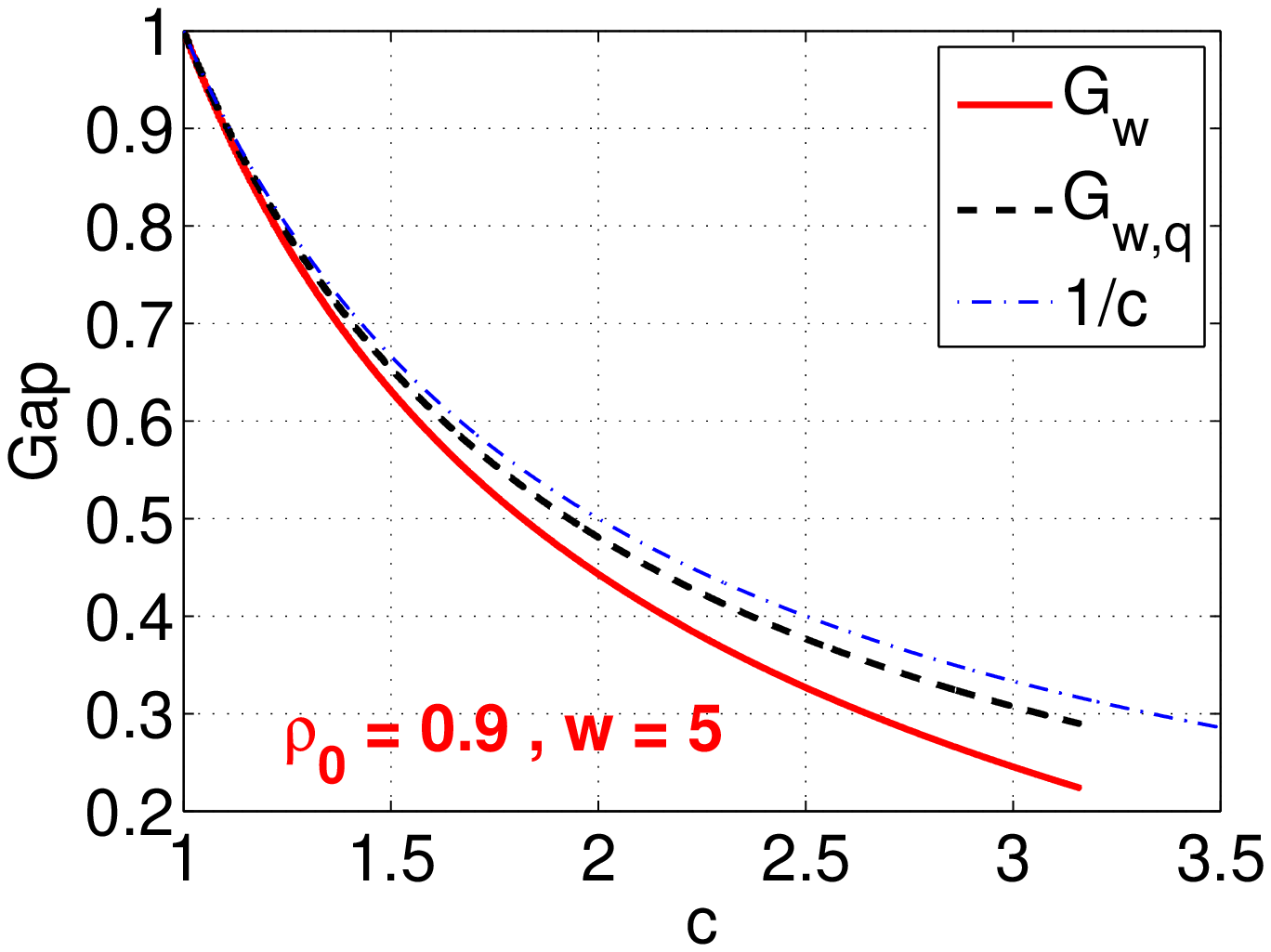}
}

\vspace{-.15in}
\caption{The gaps $G_w$ and $G_{w,q}$ as functions of $c$, for $\rho_0 = 0.9$. In each panel, we plot  $G_w$ and $G_{w,q}$ for one $w$ value. }\label{fig_GwqR09W}\vspace{-0.1in}
\end{figure}

\begin{figure}[h!]

\hspace{-0.15in}
\mbox{
\includegraphics[width = 1.25in]{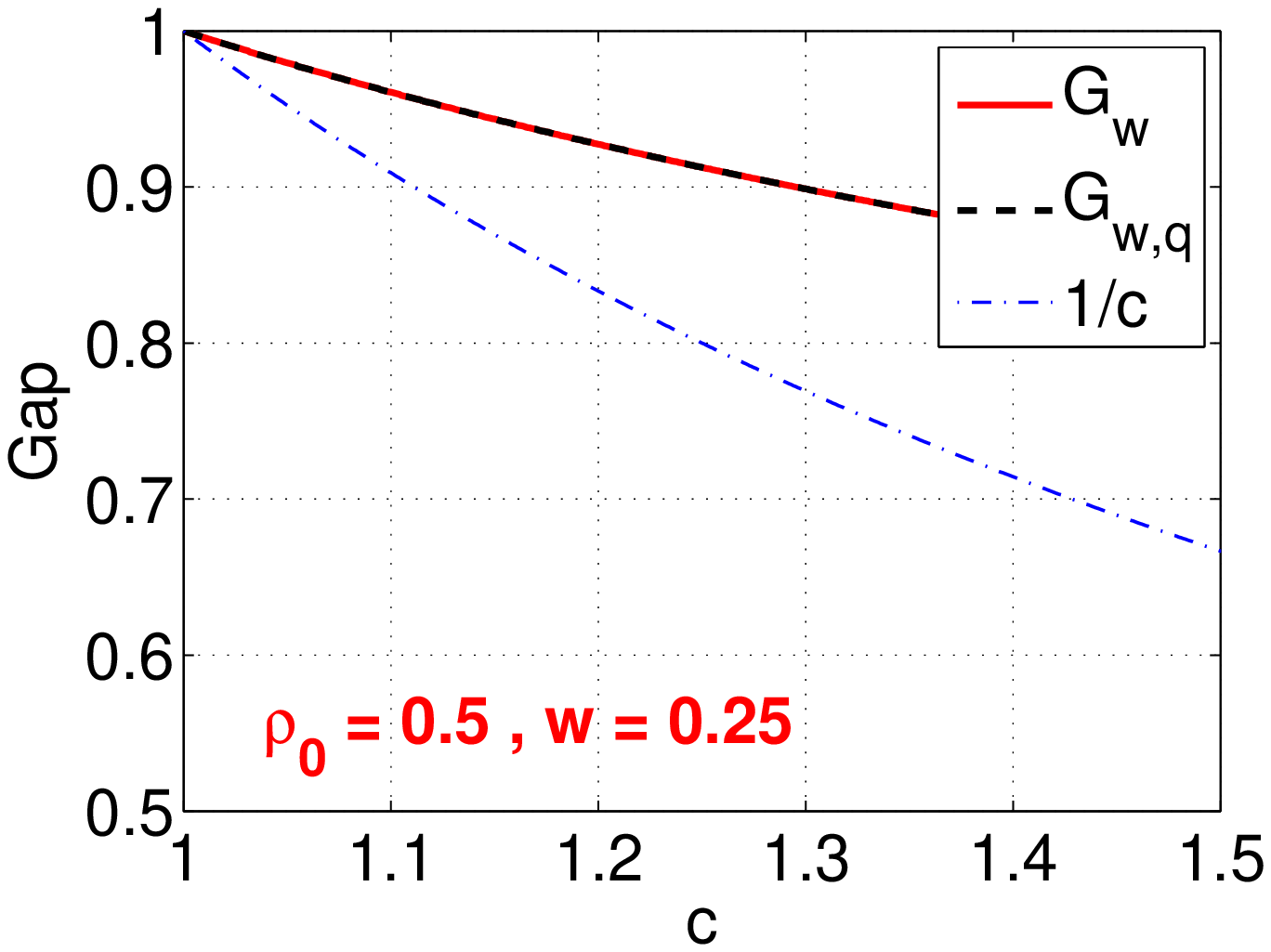}\hspace{-0.08in}
\includegraphics[width = 1.25in]{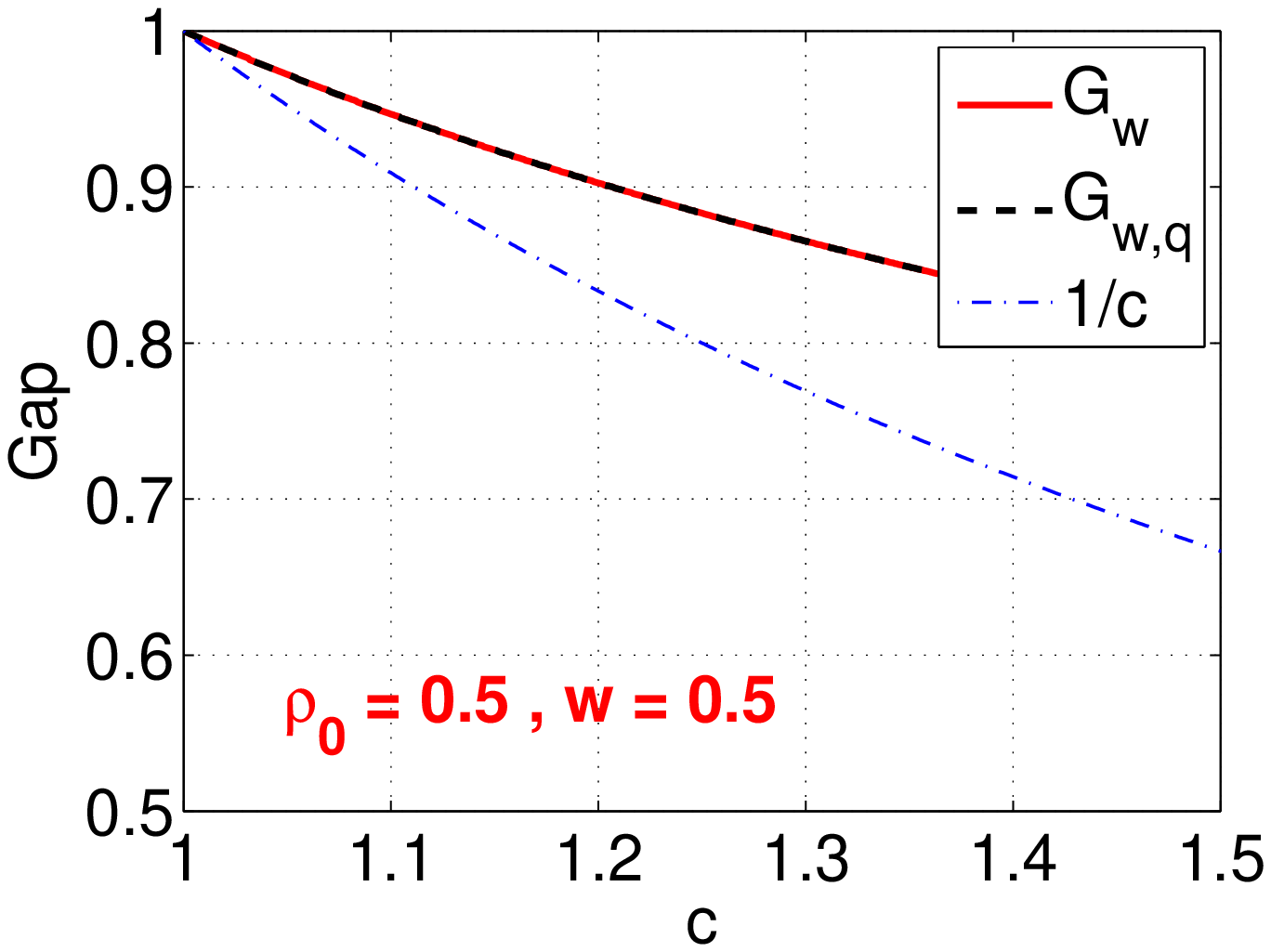}\hspace{-0.08in}
\includegraphics[width = 1.25in]{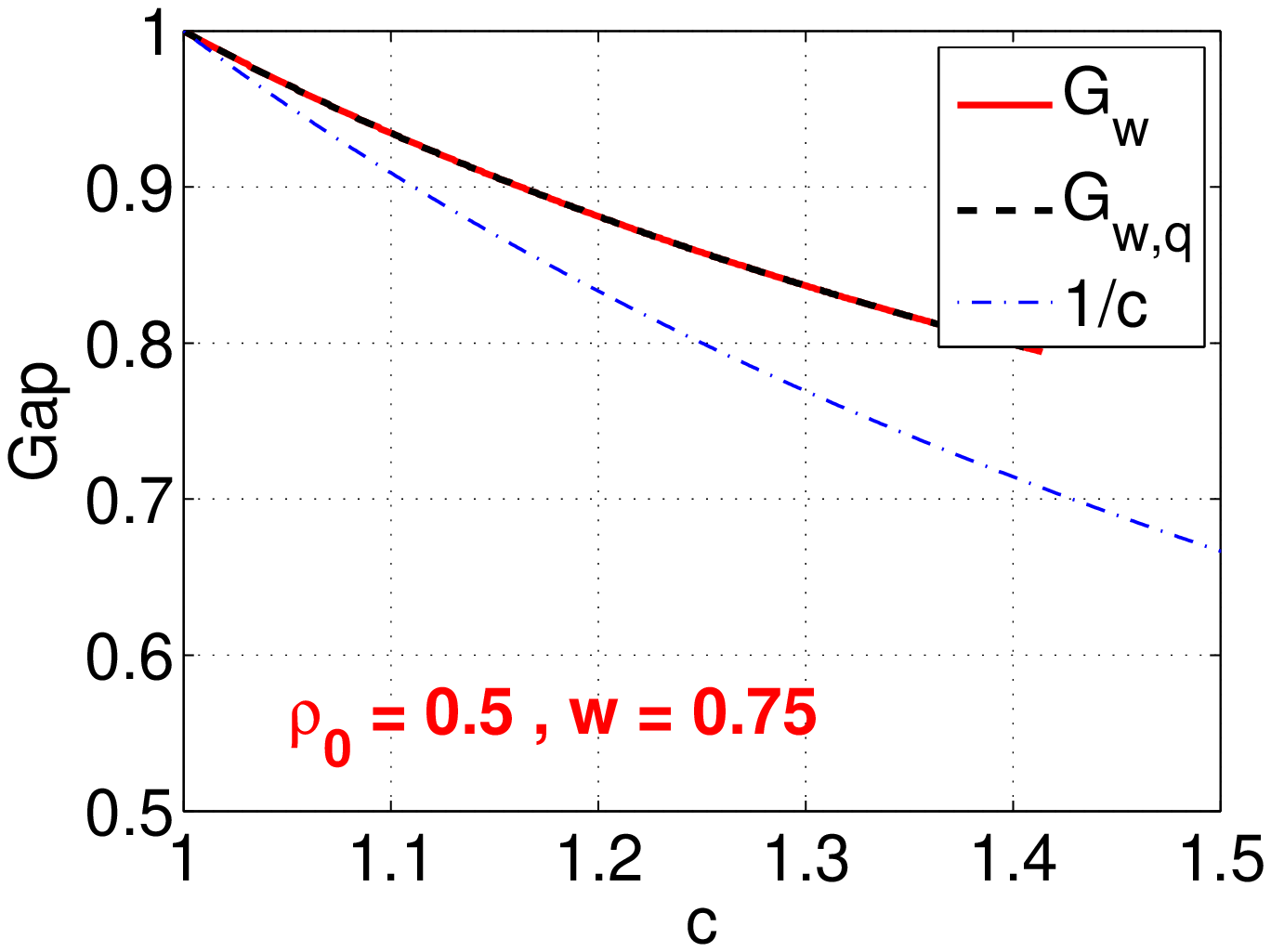}
}

\hspace{-0.15in}
\mbox{
\includegraphics[width = 1.25in]{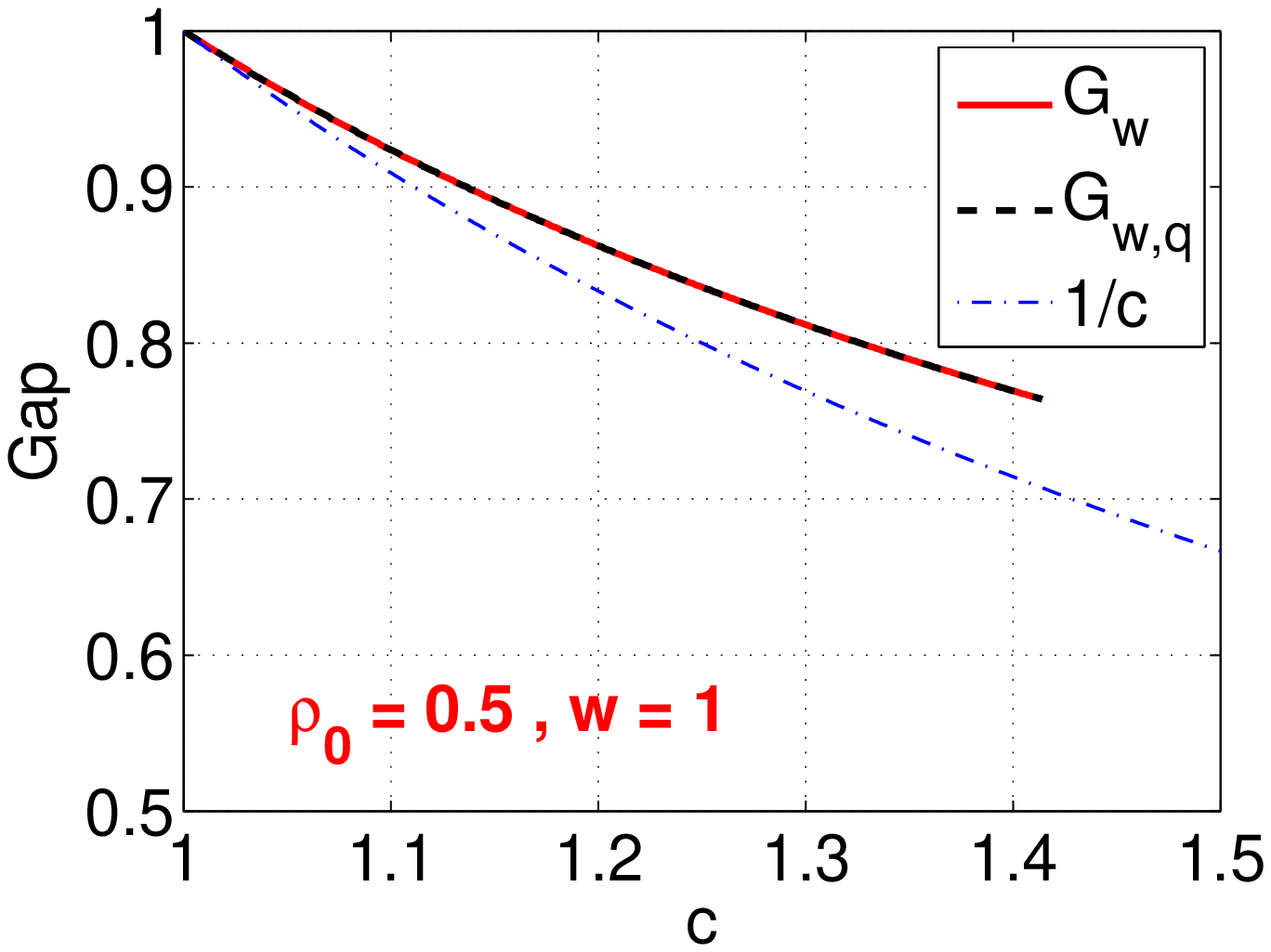}\hspace{-0.08in}
\includegraphics[width = 1.25in]{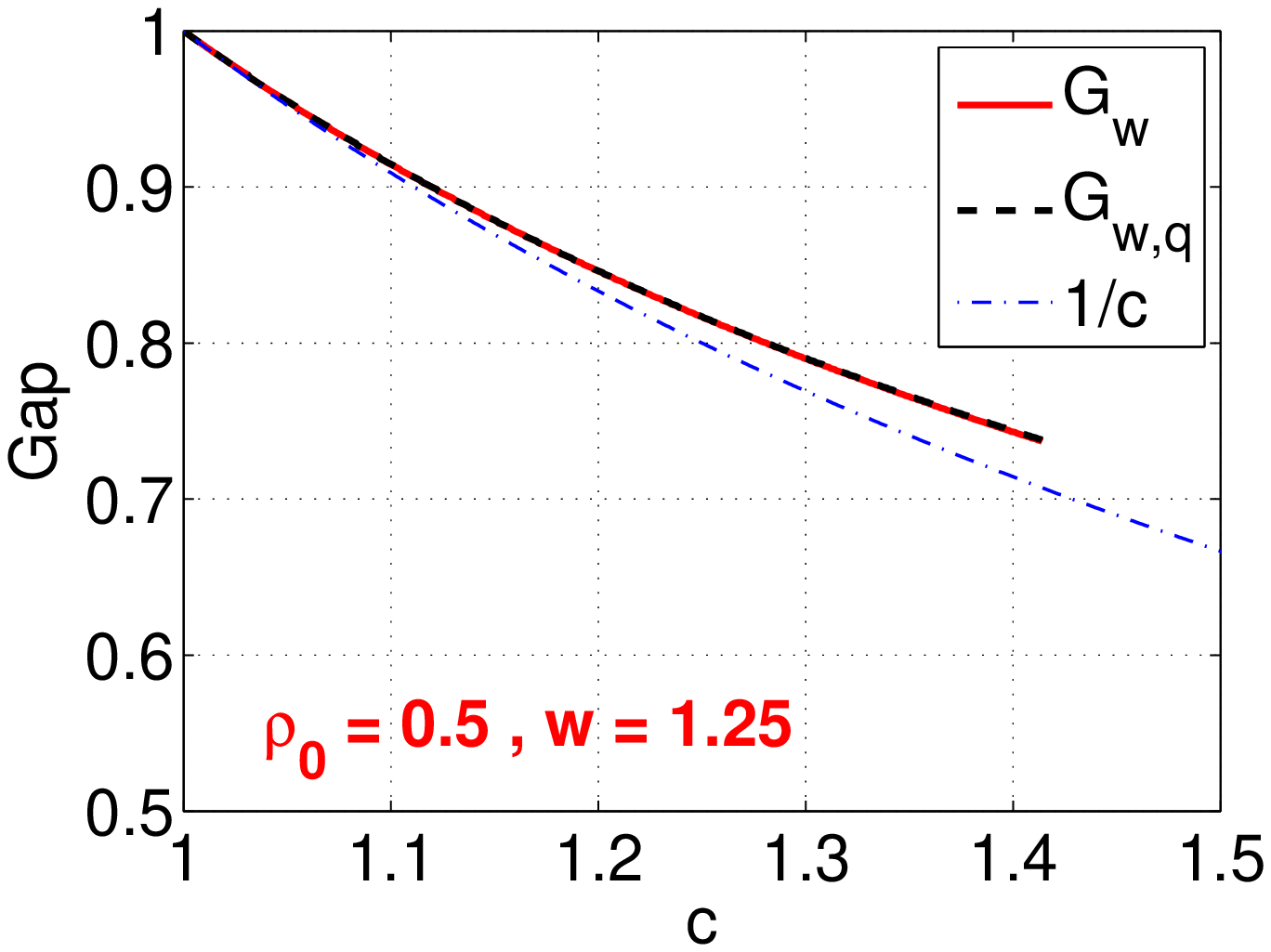}\hspace{-0.08in}
\includegraphics[width = 1.25in]{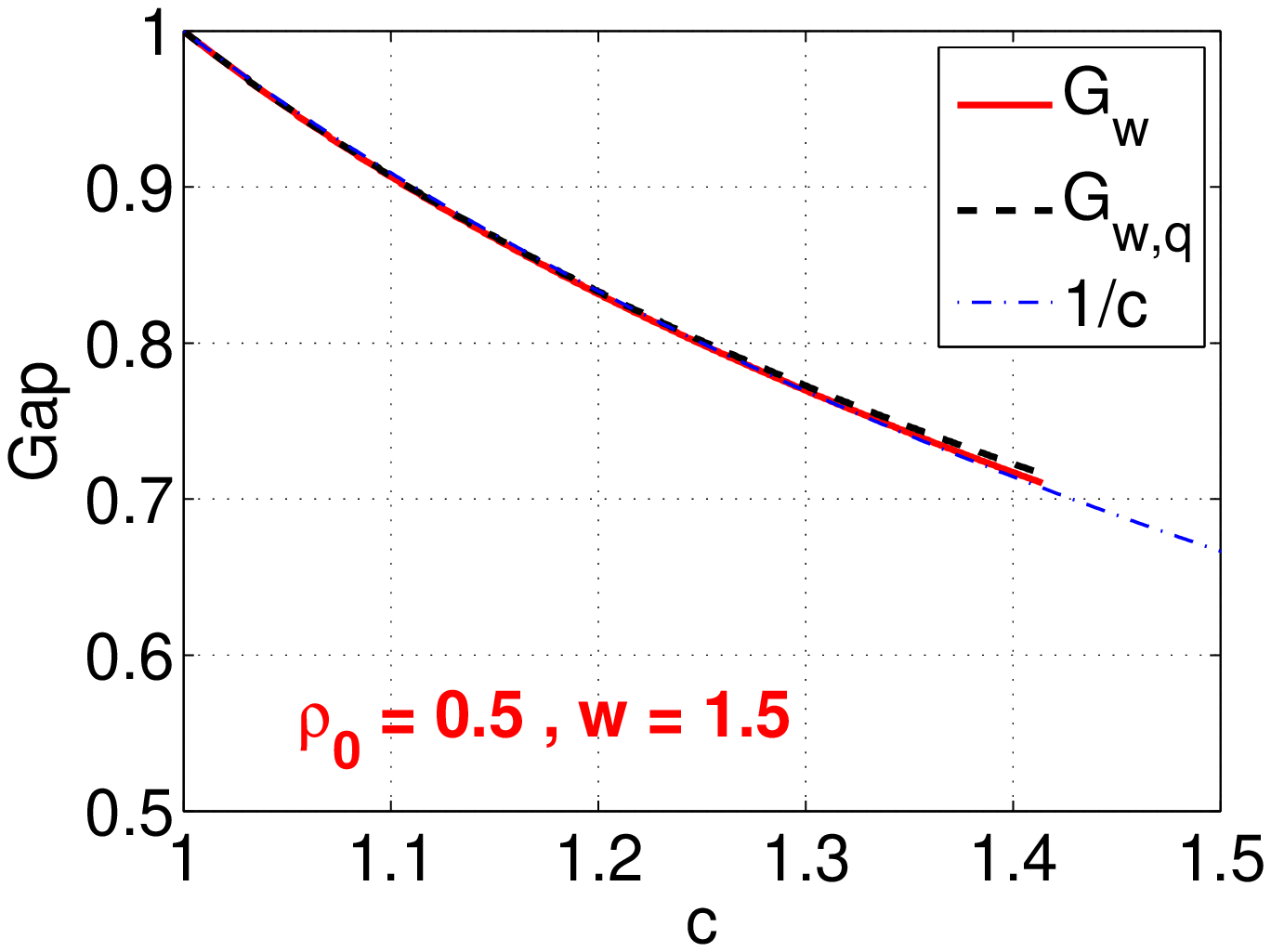}
}

\hspace{-0.15in}
\mbox{
\includegraphics[width = 1.25in]{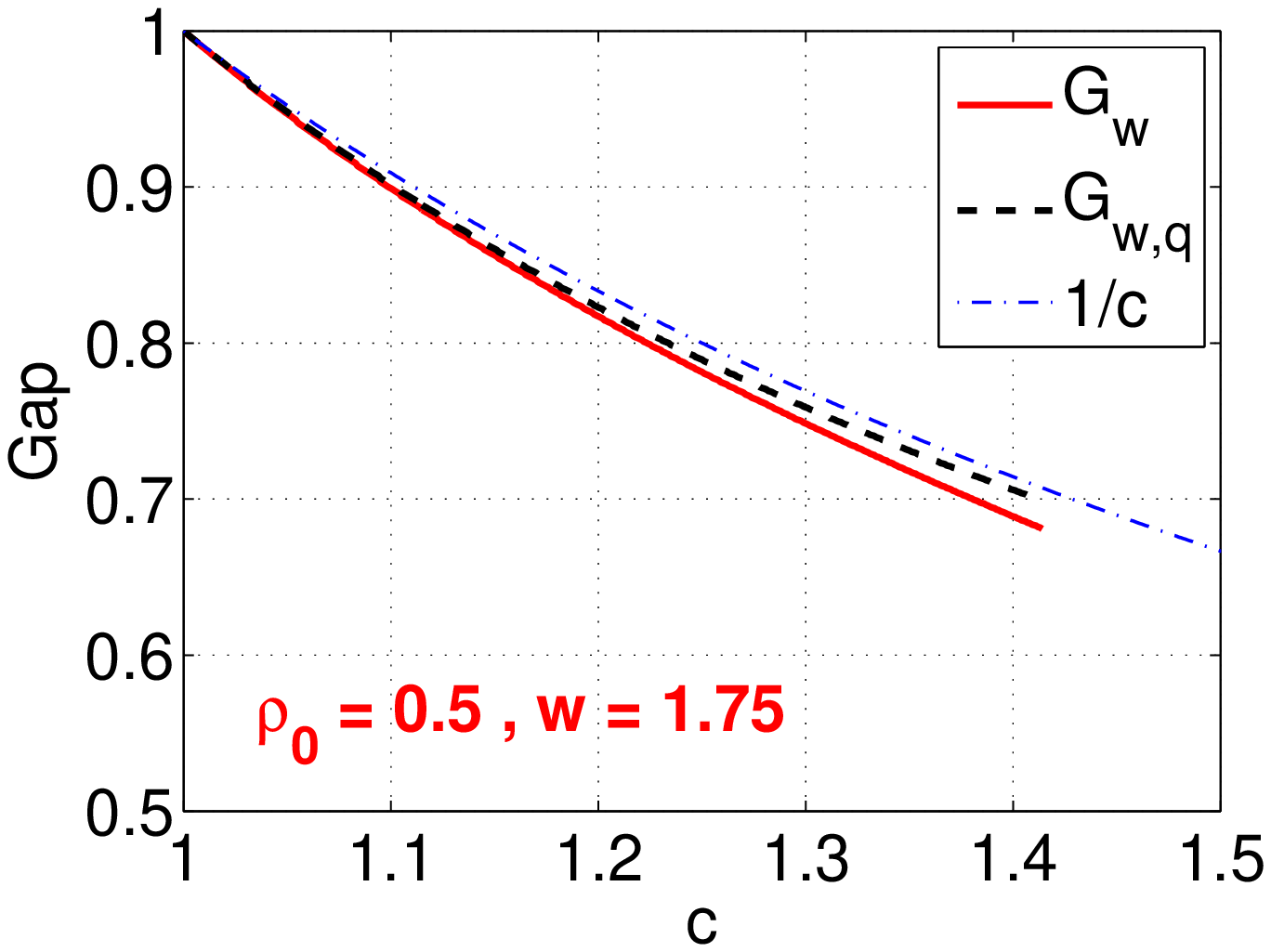}\hspace{-0.08in}
\includegraphics[width = 1.25in]{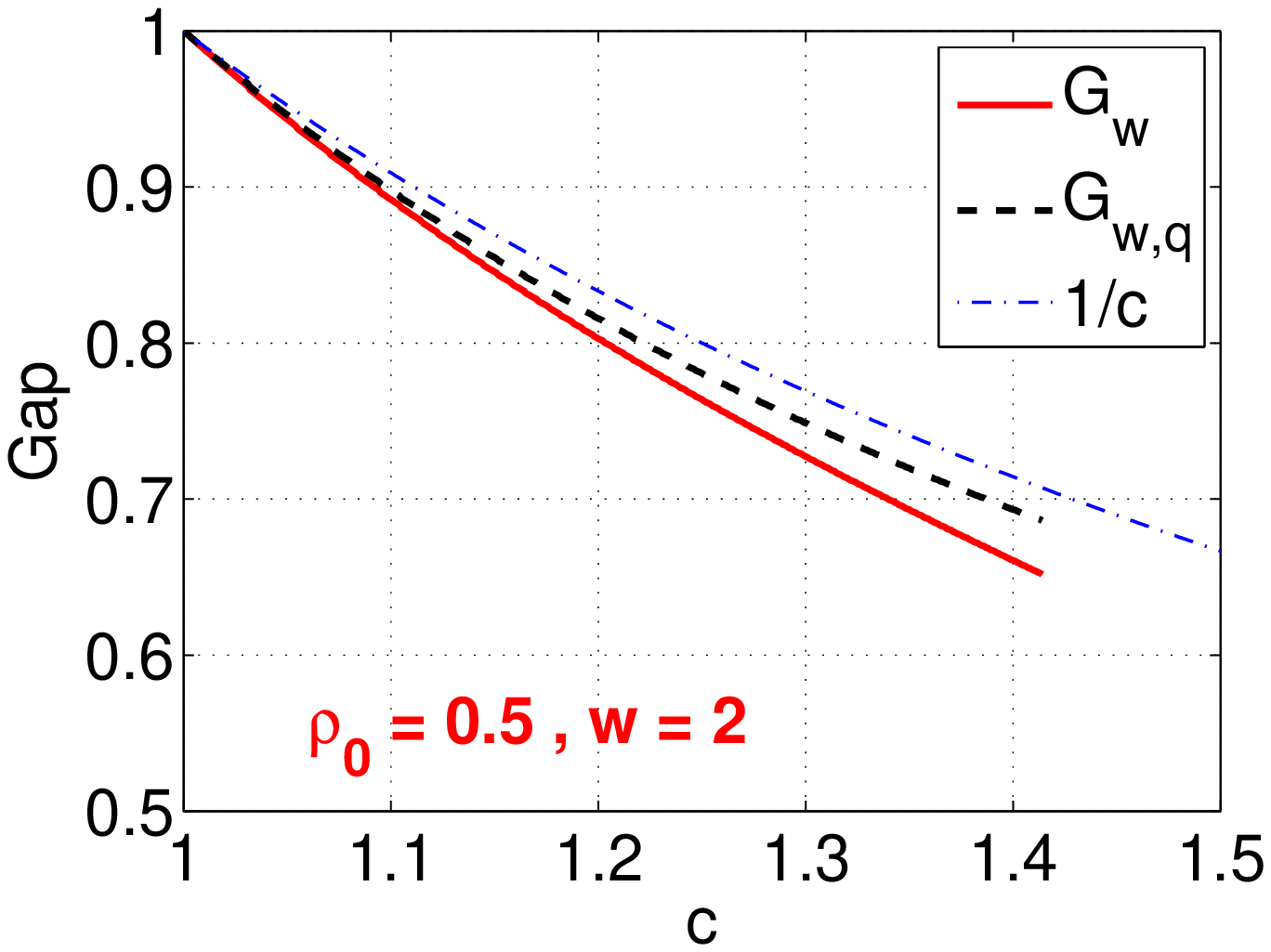}\hspace{-0.08in}
\includegraphics[width = 1.25in]{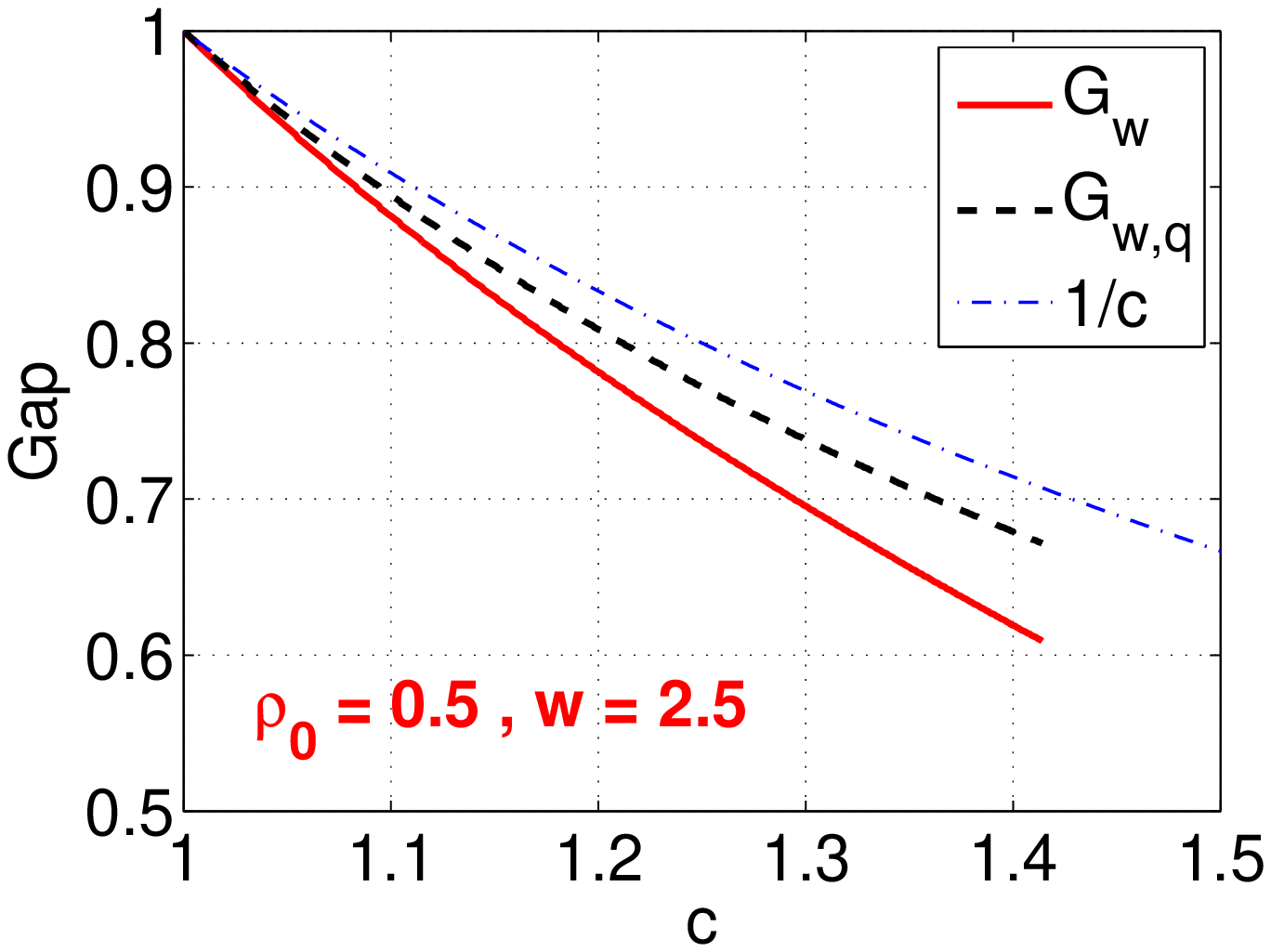}
}

\hspace{-0.15in}
\mbox{
\includegraphics[width = 1.25in]{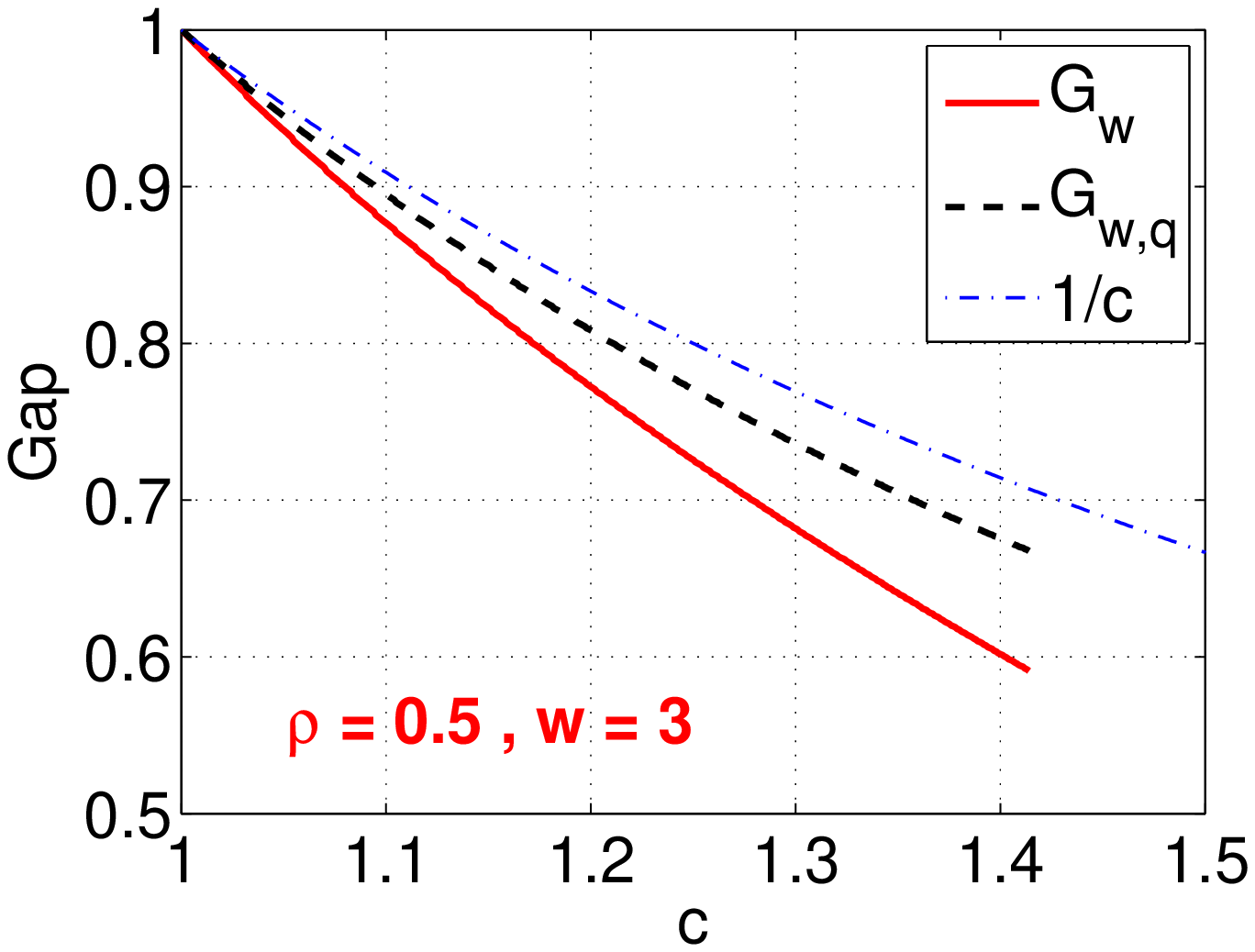}\hspace{-0.08in}
\includegraphics[width = 1.25in]{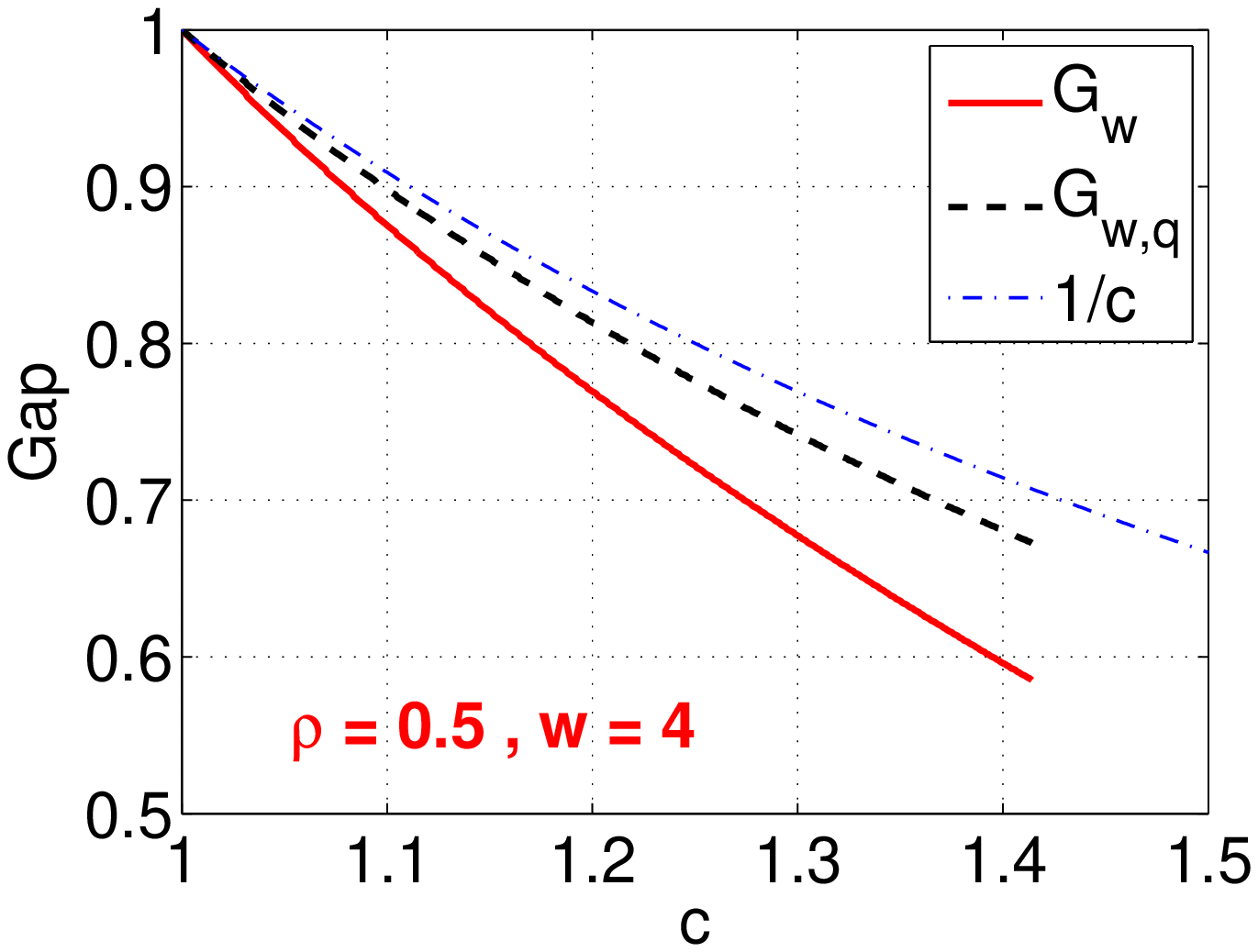}\hspace{-0.08in}
\includegraphics[width = 1.25in]{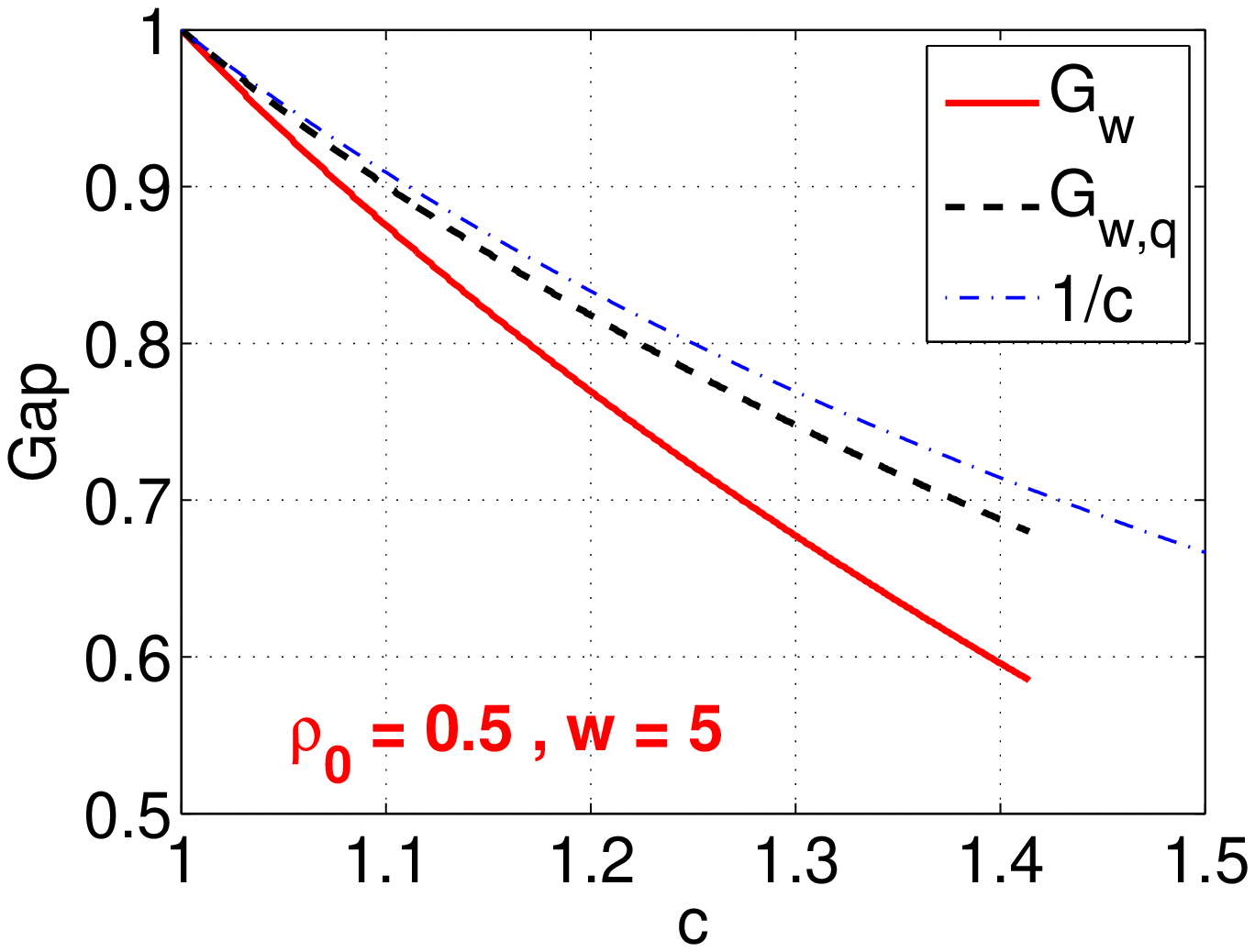}
}

\vspace{-.15in}
\caption{The gaps $G_w$ and $G_{w,q}$ as functions of $c$, for $\rho_0 = 0.5$. In each panel, we plot  $G_w$ and $G_{w,q}$ for one $w$ value. }\label{fig_GwqR05W}\vspace{-0.1in}
\end{figure}

\newpage

\begin{figure}[h!]

\hspace{-0.15in}
\mbox{
\includegraphics[width = 1.25in]{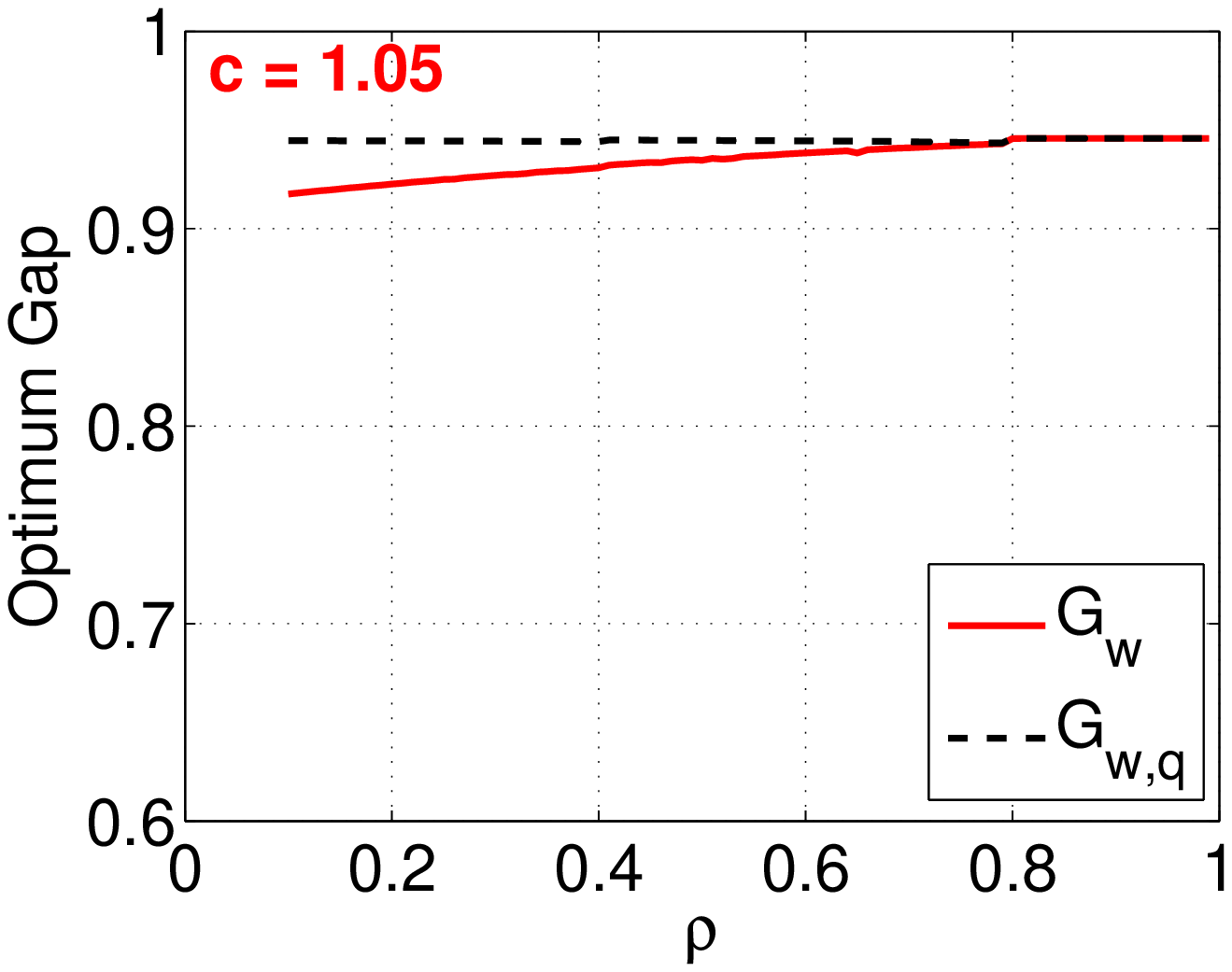}\hspace{-0.08in}
\includegraphics[width = 1.25in]{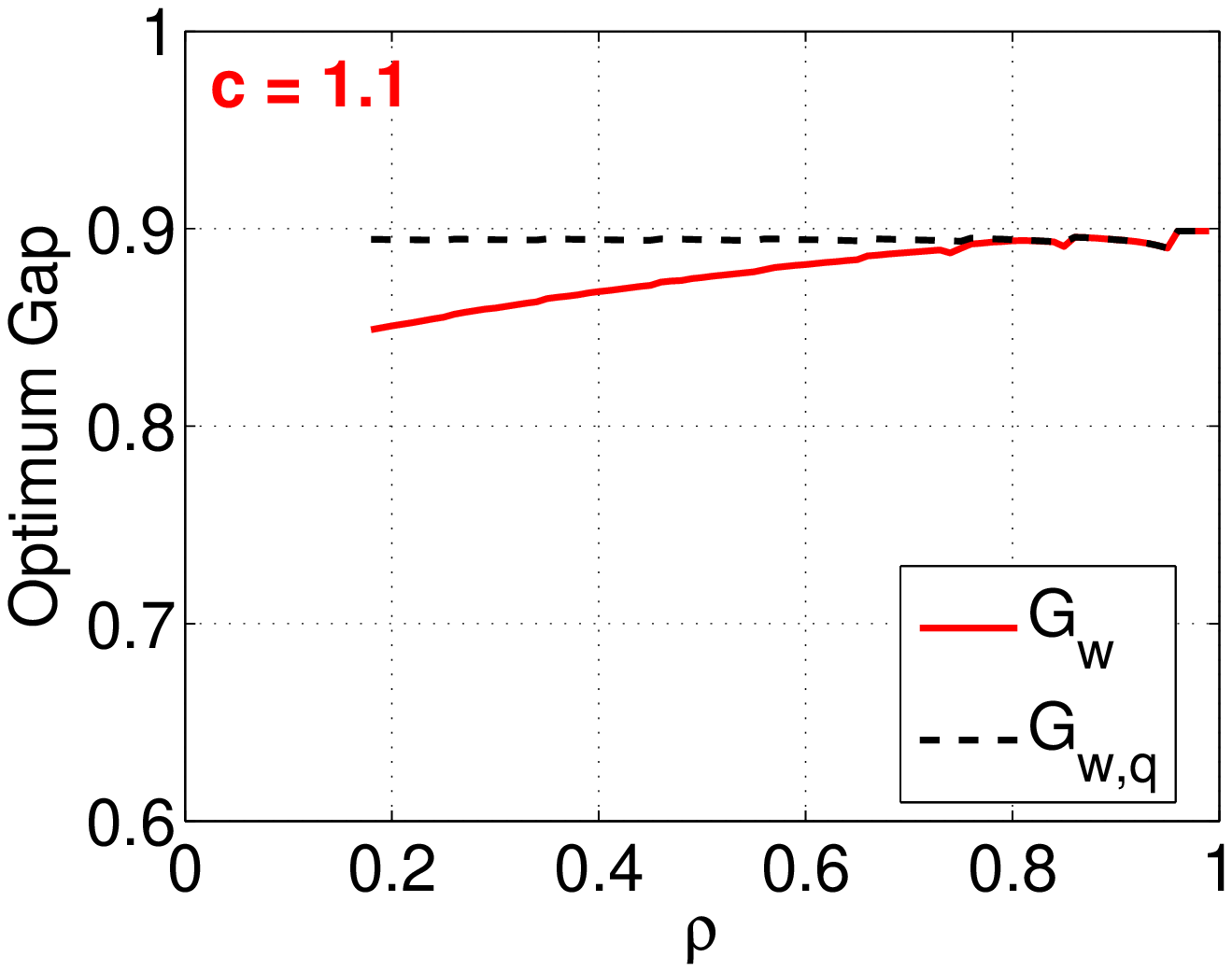}\hspace{-0.08in}
\includegraphics[width = 1.25in]{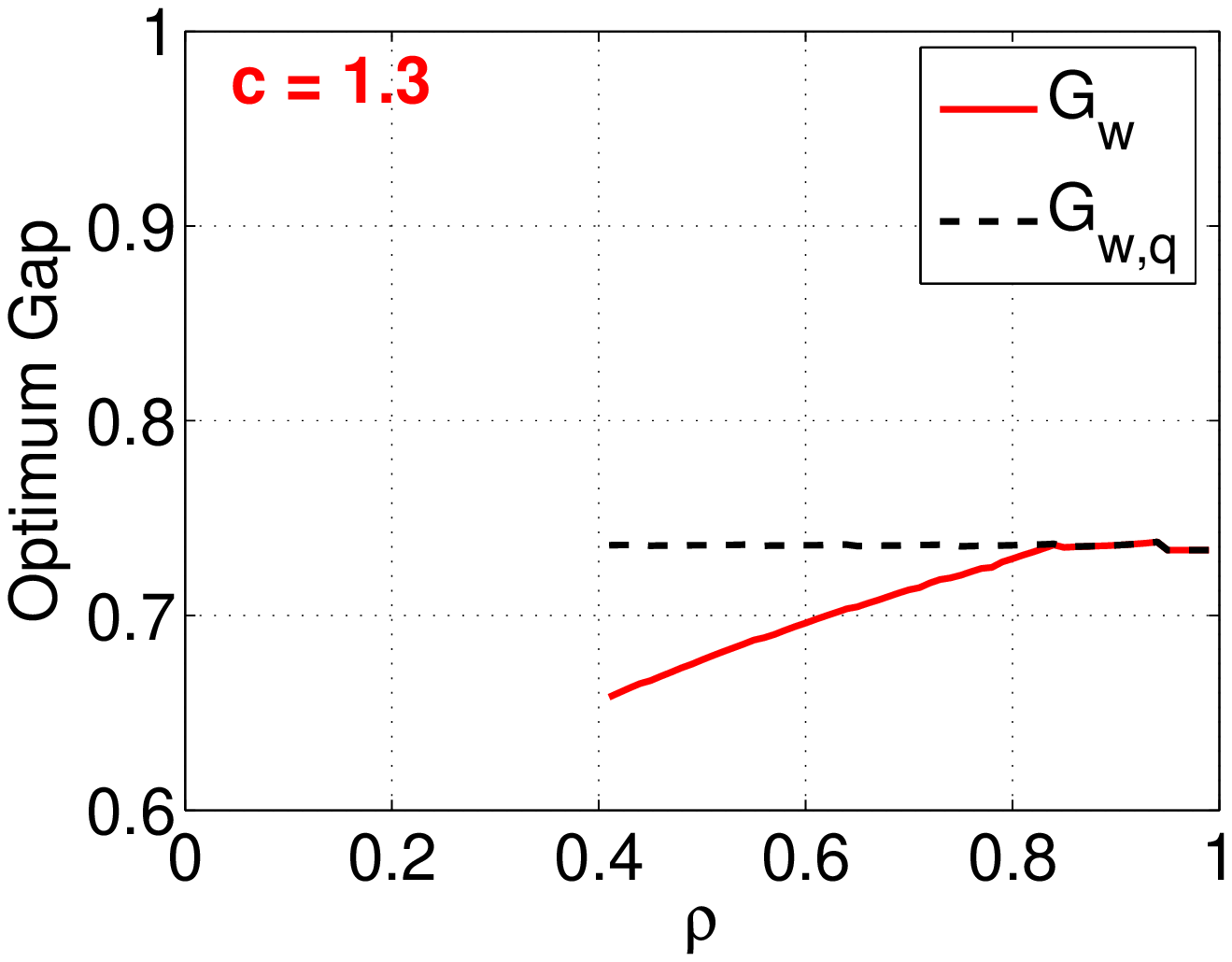}
}

\hspace{-0.15in}
\mbox{
\includegraphics[width = 1.25in]{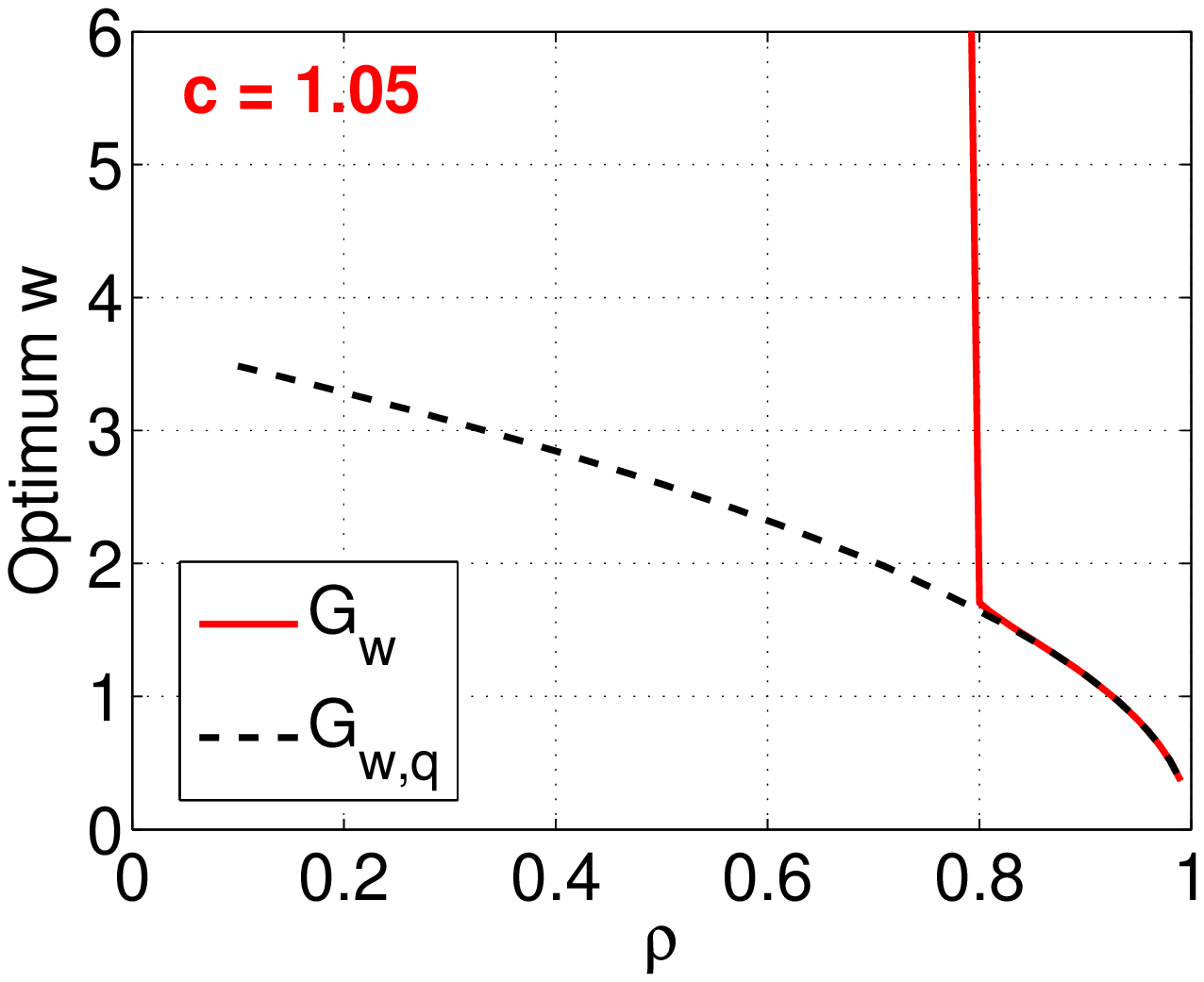}\hspace{-0.08in}
\includegraphics[width = 1.25in]{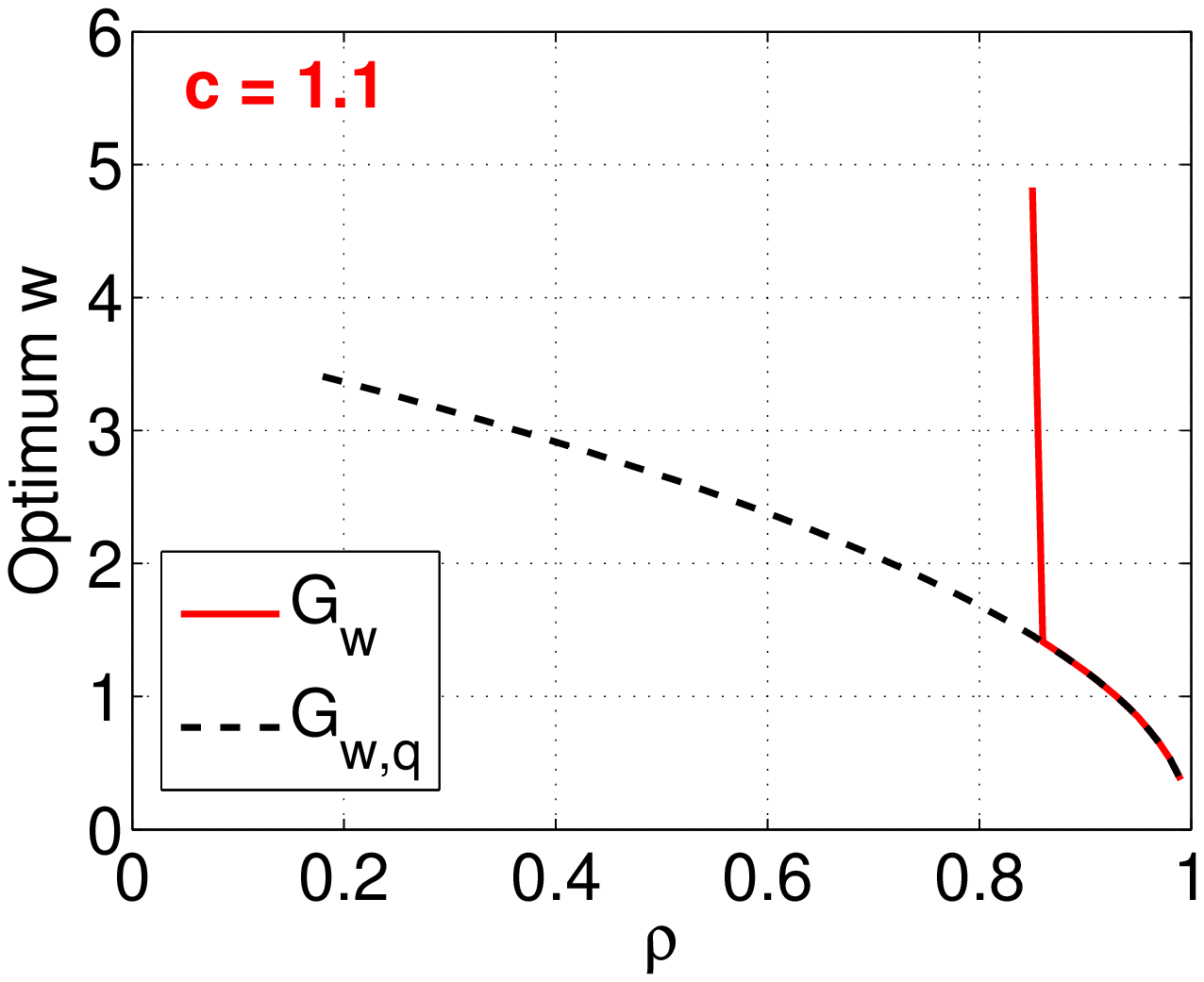}\hspace{-0.08in}
\includegraphics[width = 1.25in]{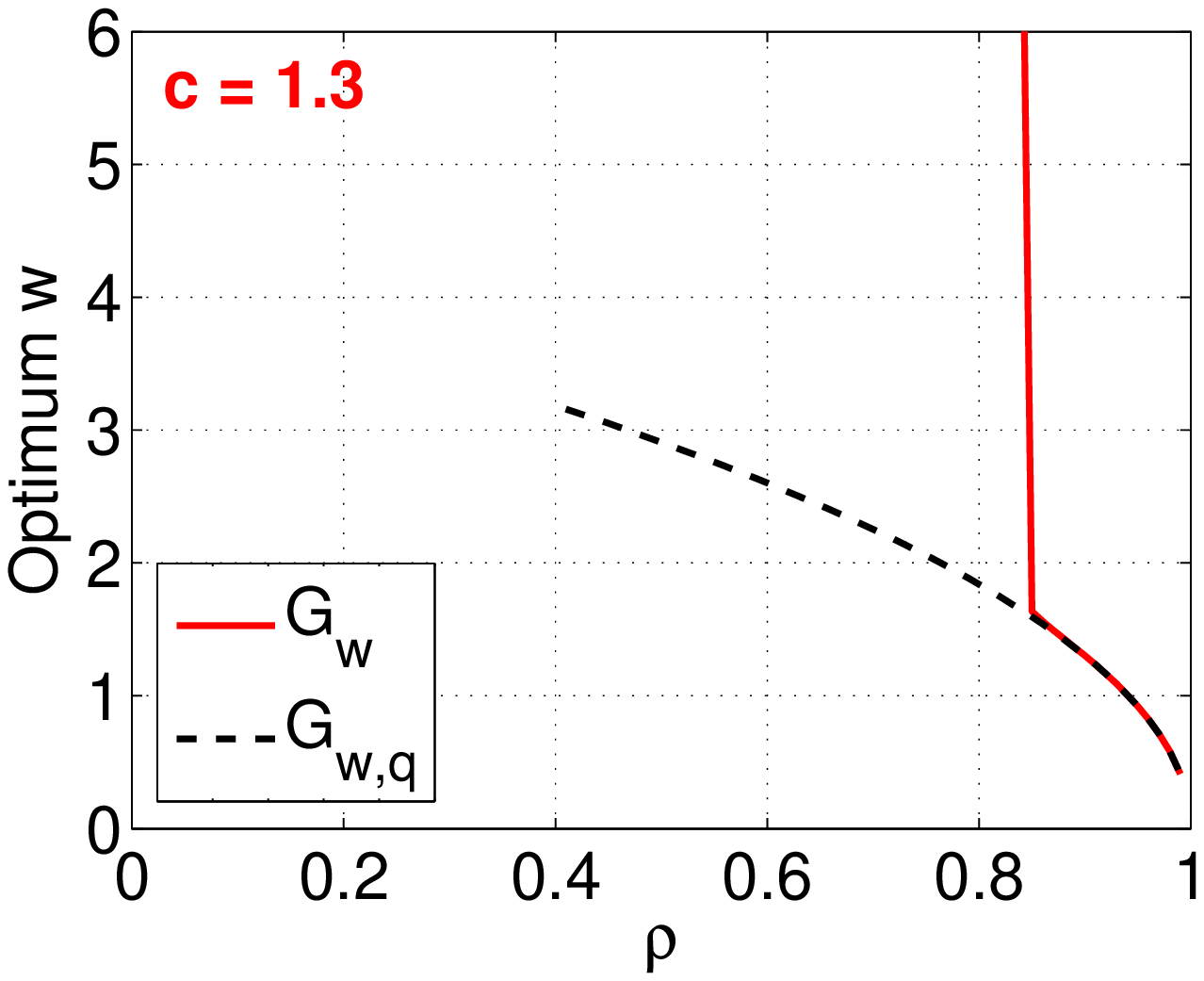}
}

\vspace{-0.15in}
\caption{\textbf{Upper panels}: the optimal (smallest) gaps at given $c$ values and the entire range of $\rho$. We can see that $G_{w,q}$ is always larger than $G_w$, confirming that it is better to use $h_w$ instead of $h_{w,q}$. \textbf{Bottom panels}: the optimal values of $w$ at which the optimal gaps are attained. When the target similarity $\rho$ is very high, it is best to use a relatively small $w$.  }\label{fig_OptGW1}\vspace{-0.1in}
\end{figure}

\subsection{The Optimal Gaps}
In practice, we normally have to  pre-specify the bin width $w$, for all $c$ and $\rho_0$ values. In other words, the ``optimum'' $G$ values presented in Figure~\ref{fig_GwqOpt} are in general not attainable. Thus,  Figure~\ref{fig_GwqR09W} and Figure~\ref{fig_GwqR05W} present $G_w$ and $G_{w,q}$ as functions of $c$, for  $\rho_0 = 0.9$ and $\rho_0 = 0.5$, respectively. In each figure, we plot the curves for a wide range of $w$ values. These figures again confirm that $G_w$ is smaller than $G_{w,q}$, i.e., the scheme  without offset (\ref{eqn_hw}) is better.

To view the optimal gaps more closely, Figure~\ref{fig_OptGW1}  plots the best gaps (upper panels) and the optimal $w$ values (bottom panels) at which the best gaps are attained, for selected values of $c$. These plots again confirm the previous comparisons:\vspace{-0.07in}
\begin{itemize}
\item We should always replace $h_{w,q}$ with $h_{w}$. At any $\rho$ and $c$, the optimal gap $G_{w,q}$ is  at least as large as the optimal gap $G_{w}$.  At relatively low similarities, the optimal $G_{w,q}$  can be substantially larger than the optimal $G_{w}$.\vspace{-0.07in}
\item If we use $h_w$ and target at very high similarity, a reasonable choice of the bin width $w$ might be $w=1\sim 1.5$.\vspace{-0.07in}
\item If we use $h_{w}$ and the target similarity is not too high, then we can safely use $w=2\sim3$.\vspace{-0.07in}
\end{itemize}

We should also mention that,  although the optimal $w$ values for $h_w$ appear to exhibit a ``jump'' in the right panels of  Figure~\ref{fig_OptGW1}, the choice of $w$ does not influence the performance much, as shown in previous plots. In Figures~\ref{fig_GwqR09C} and~\ref{fig_GwqR05C}, we have seen that even when the optimal $w$ appears to approach ``$\infty$'', the actual gaps are not much different between $w=3$ and $w\gg3$. In the real data evaluations in the next section, we will see the same phenomenon for $h_w$.

Note that the Gaussian density decays very rapidly at the tail, for example, $1-\Phi(3) = 1.3\times10^{-3}$ and $1-\Phi(6) = 9.9\times10^{-10}$. If we choose $w\geq 1.5$, then we practically just need (at most) 2 bits to code each hashed value, that is, we can simply  quantize the data according to $(-\infty,\ -w], (-w,\ 0], (0,\ w], [w, \infty)$ (see Figure~\ref{fig_16region}).

\vspace{-0.1in}

\section{Re-Ranking for LSH}

In the process of using hash tables for  sublinear time near neighbor search, there is an important step called ``re-ranking''.  With a good LSH scheme, the fraction of retrieved data points could be relatively  low (e.g., $1\%$). But the absolute  number of retrieved points can still be very large (e.g., $1\%$ of a billion points is still large). It is thus crucial to have a re-ranking mechanism, for which one will have to either compute or estimate the actual similarities.

When the original data are massive and high-dimensional, i.e., a data matrix in $\mathbb{R}^{n\times D}$ with both $n$ and $D$ being  large, it can be  challenging to evaluate the similarities. For example, it is often not  possible to load the entire dataset in the memory. In general, we can not store all pair-wise similarities at the cost of $O(n^2)$ space which is not practical even for merely $n=10^6$. In addition, the query  might be a new data point  so that we will have to compute the similarities on the fly anyway. If the data are high-dimensional, the computation itself of  the exact similarities can be too time-consuming.

A feasible solution is to estimate the similarities on the fly for re-ranking, from a small projected data stored in the memory.  This has motivated us to   develop \textbf{nonlinear estimators} for a 2-bit coding scheme, by exploiting full information of the  bits.

\vspace{0.08in}

There are other applications of  nonlinear estimators too. For example, we can use random projections and nonlinear estimators for computing nonlinear kernels for SVM. Another example is to find  nearest neighbors by random projections (to reduce the dimensionality and data size) and brute-force linear scan of the projected data, which is simple to implement and easy to run  in parallel.

\vspace{0.07in}

\noindent\textbf{Two-stage coding}. \ \ Note that the coding scheme for building hash tables should be separate from the coding scheme for developing accurate estimators. Once we have projected the data and place the  points into the buckets using a designated coding scheme, we can actually discard the codes. In other words, we can code the same projected data twice.  In the second time, we store the codes of (a fraction of) the projected data for the task of similarity estimation.

\vspace{-0.1in}

\section{Re-Ranking Experiments for LSH}

We conduct a set of experimental study for LSH and re-ranking to demonstrate the advantage of the proposed nonlinear estimator for the 2-bit coding scheme.  Again, we adopt the standard $(K,L)$-LSH scheme~\cite{Proc:Indyk_STOC98}. That is, we concatenate $K$ (independent) hash functions to build each hash table and we independently build  $L$ such hash tables. Note that here we  use the capital letter  $K$ to differentiate it from $k$, which we use for sample size (or number of projections) in the context of similarity estimation.

\vspace{0.07in}

We have showed that, for building hash tables,  it is good to use uninform quantization with bin width (e.g.,)  $w_1=1.5$ if the target similarity is high and $w_1\geq3$ if the target similarity is not so high.  Here we use $w_1$ to indicate that it is the bin width  for building hash tables.  For simplicity, we fix $w_1=1.5$ (for table building) and $w=0.75$ (for similarity estimation). We choose  $K=10$ and    $L\in\{50, 100\}$. The results (especially the trends) we try to present are not too sensitive to those parameters $K$ and $L$.

Once we have built the hash tables, we need to store a fraction of the coded projected data. To save space, we should store $k\ll K\times L$ projections. Here we choose $k=100$ and $k=200$, which appear to be sufficient to provide accurate estimates of the similarity for re-ranking of retrieved data points.

\vspace{0.07in}

We target at top-$T$ nearest neighbors, for $T\in\{10, 20, 50, 100\}$. We re-rank the retrieved  points according to estimated similarities based on 3 different estimators: (i) the MLE (nonlinear) for 2-bit coding as studied in this paper; (ii) the 2-bit linear estimator; (iii) the 1-bit estimator. We present the results in terms of precision-recall curves (higher is better) for retrieving the top-$T$ points.  That is, we first rank all  retrieved  points according to estimated similarities. Then for a particular $T$, we examine the top-$m$ of the list to compute one (precision, recall) tuple. By varying $m$, we obtain a precision-recall curve for each $T$, averaged over all query points.

\begin{figure*}[t]
\begin{minipage}[c][\textheight]{\textwidth}

\begin{center}
\mbox
{
\includegraphics[width = 1.7in]{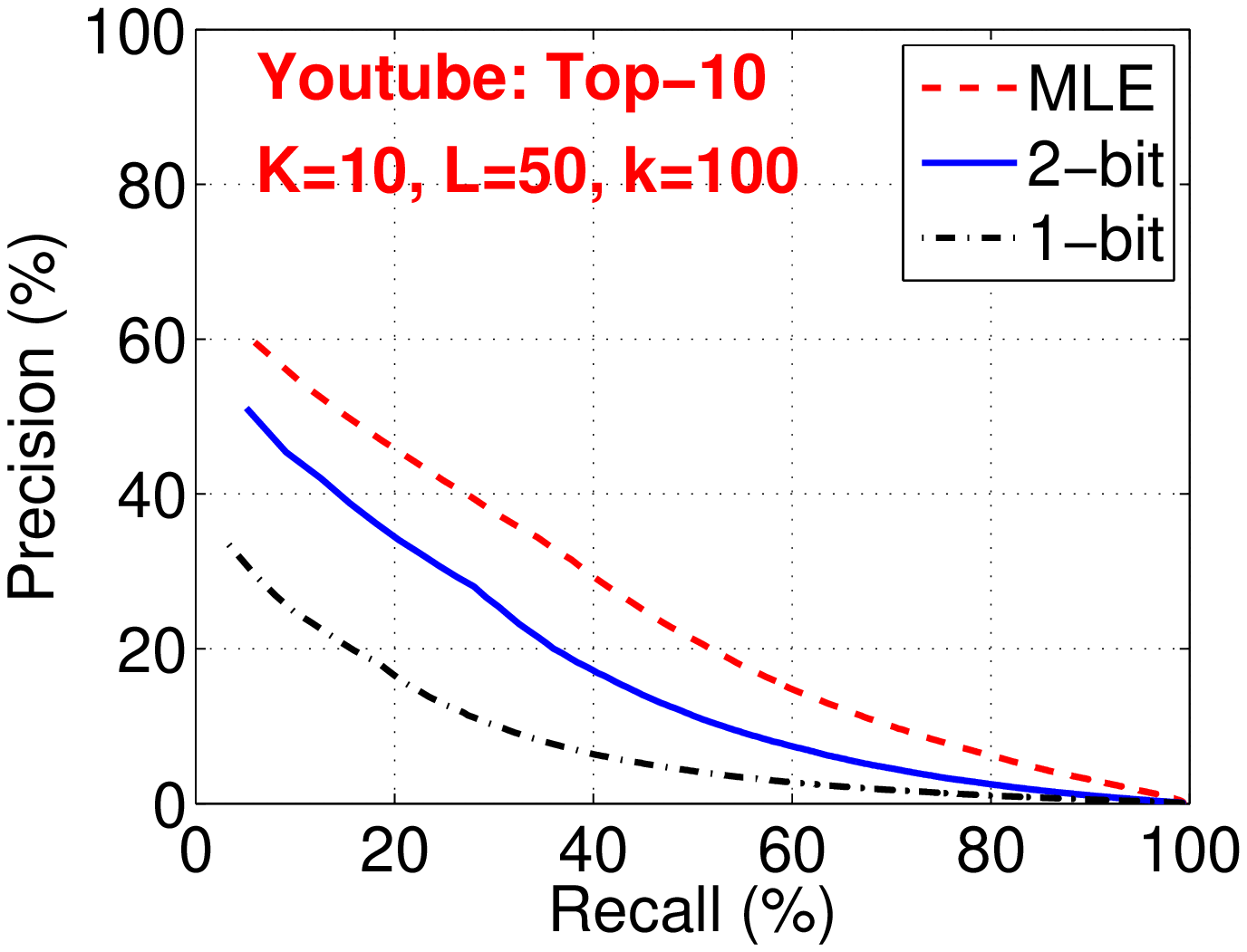}
\includegraphics[width = 1.7in]{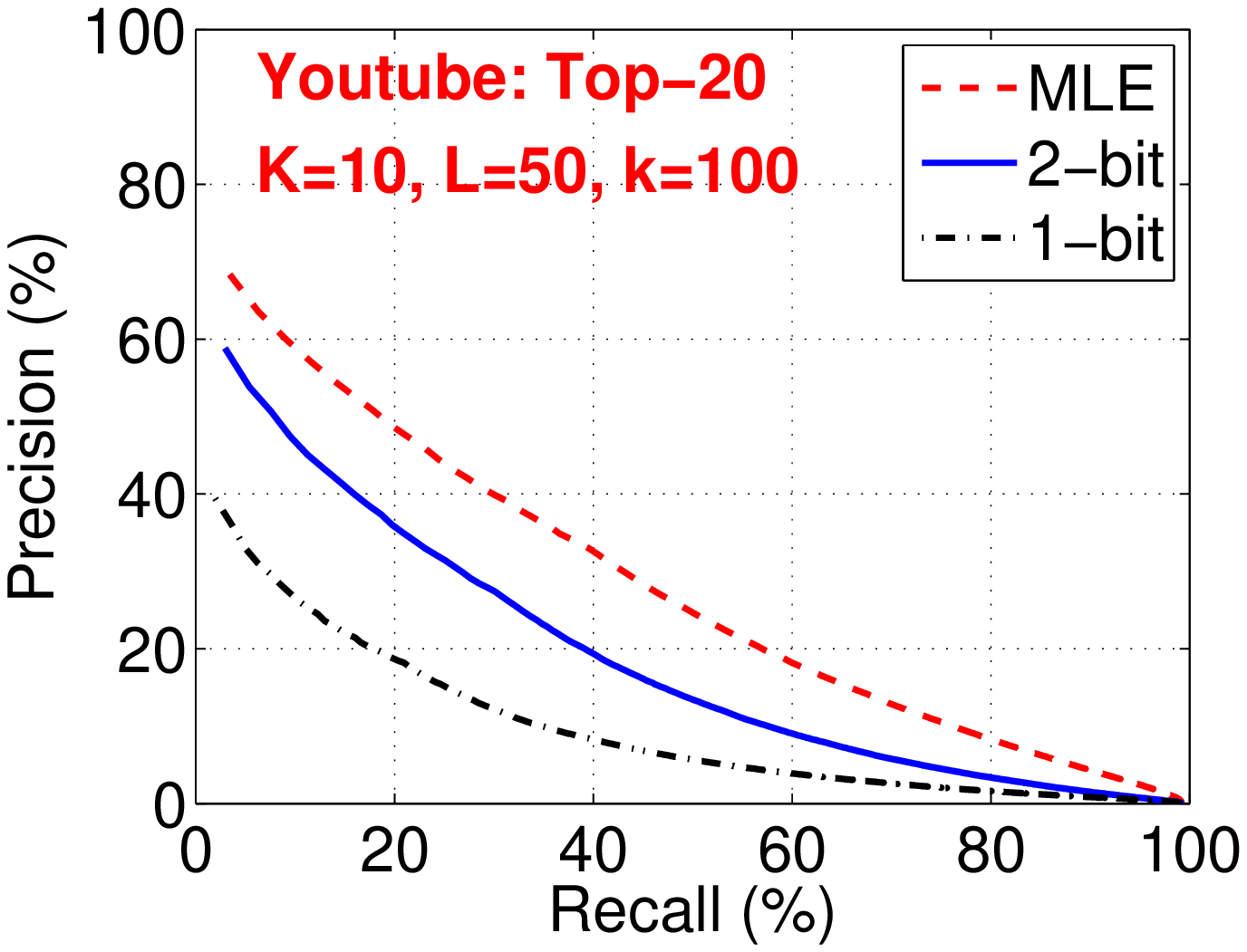}
\includegraphics[width = 1.7in]{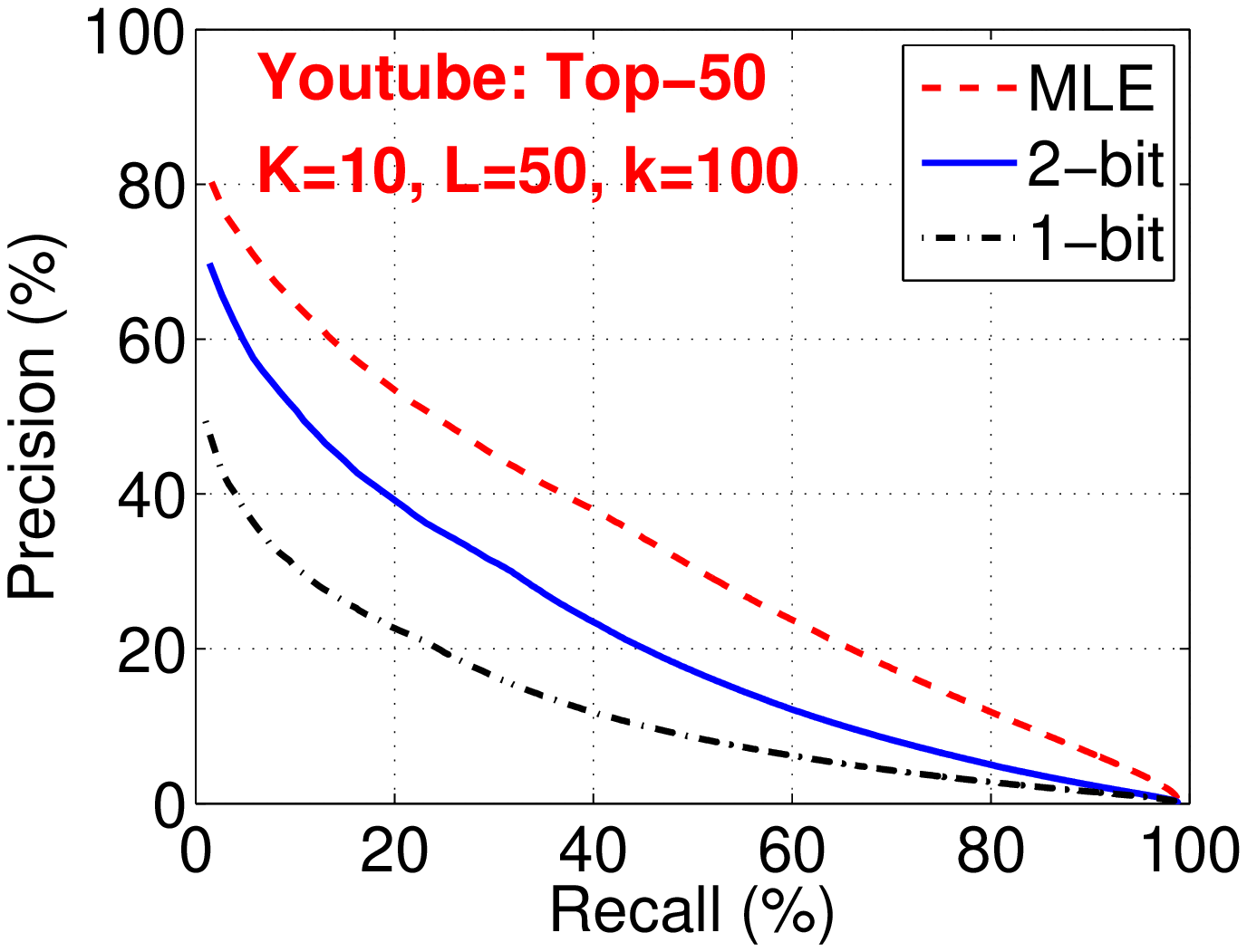}
\includegraphics[width = 1.7in]{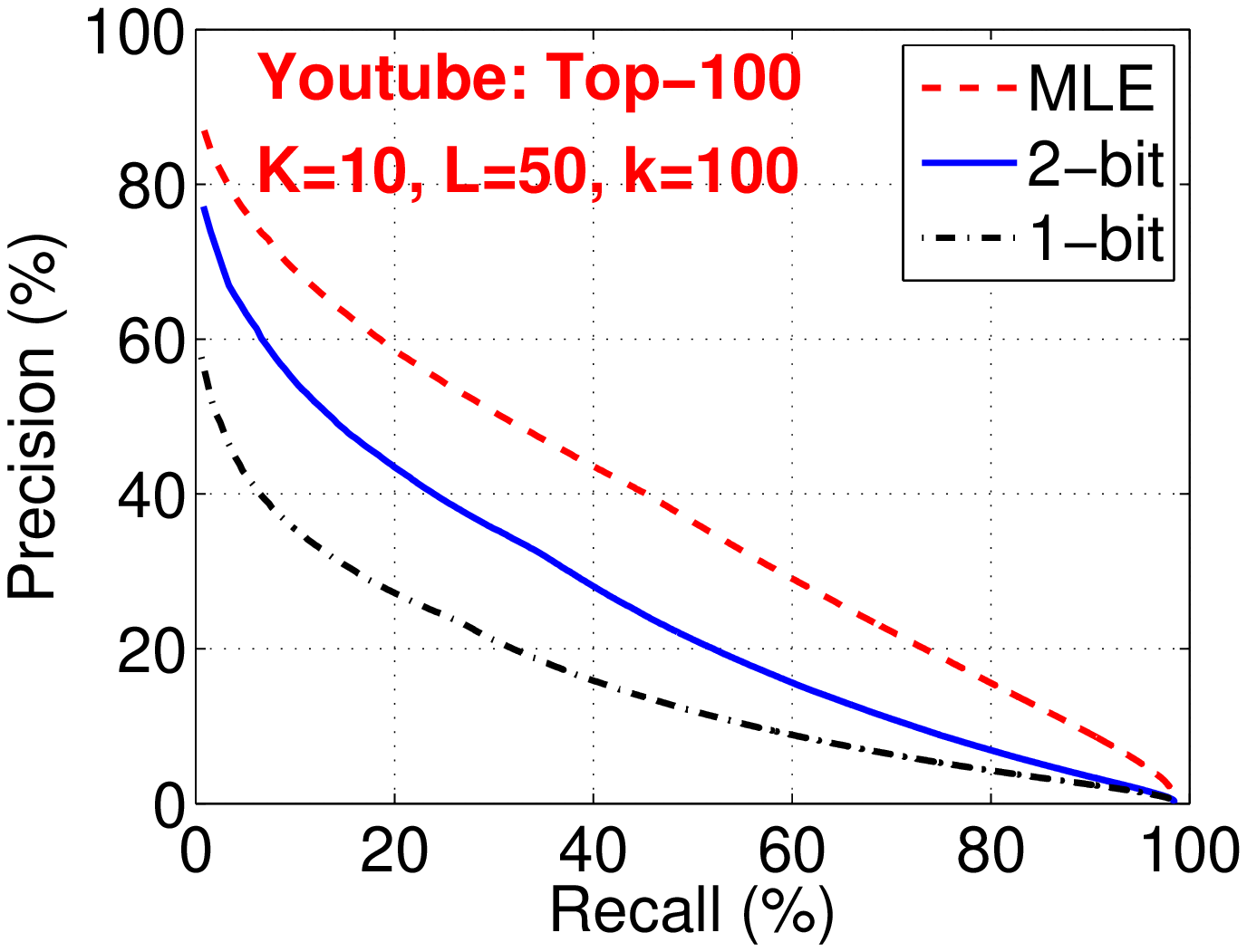}
}

\mbox{
\includegraphics[width = 1.7in]{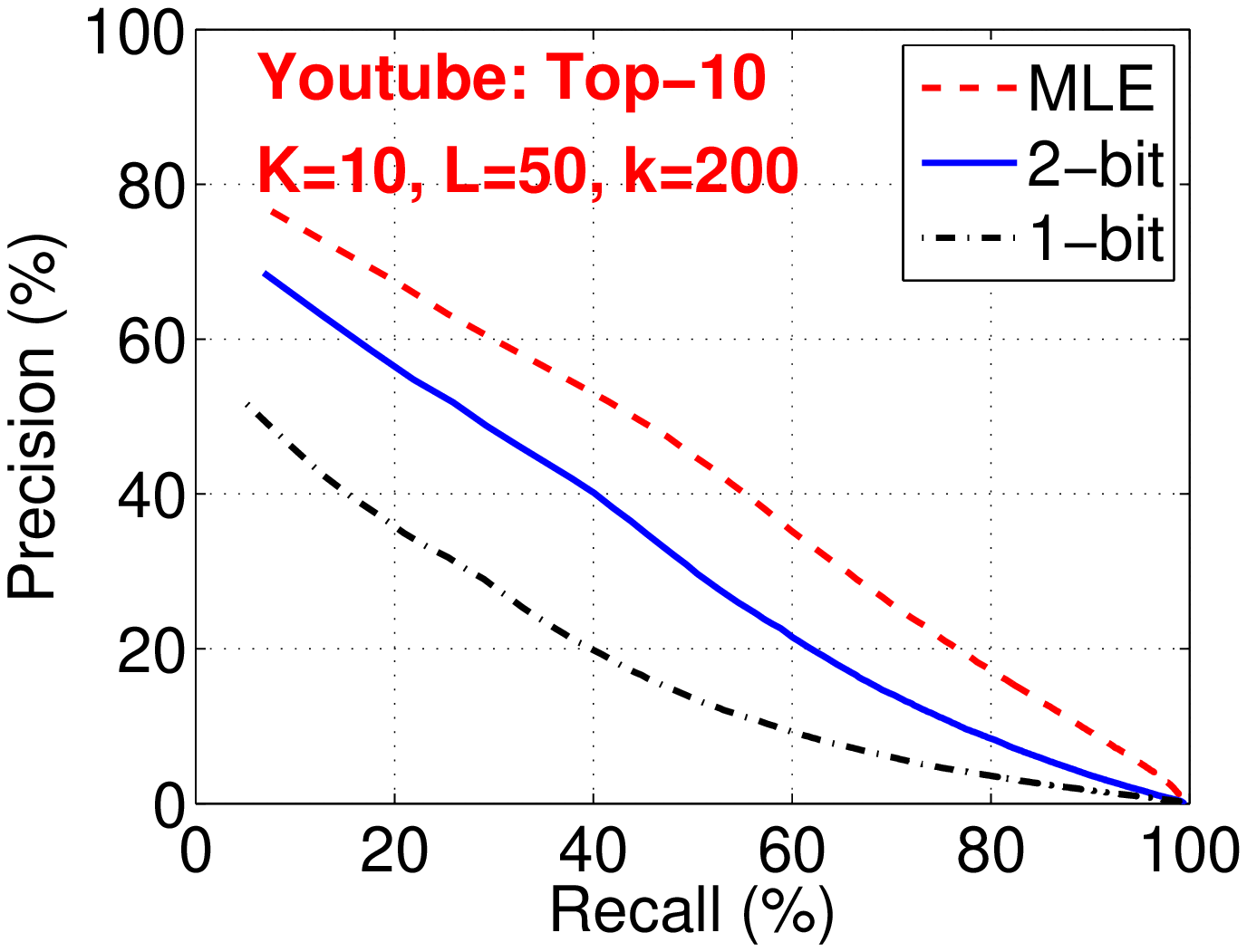}
\includegraphics[width = 1.7in]{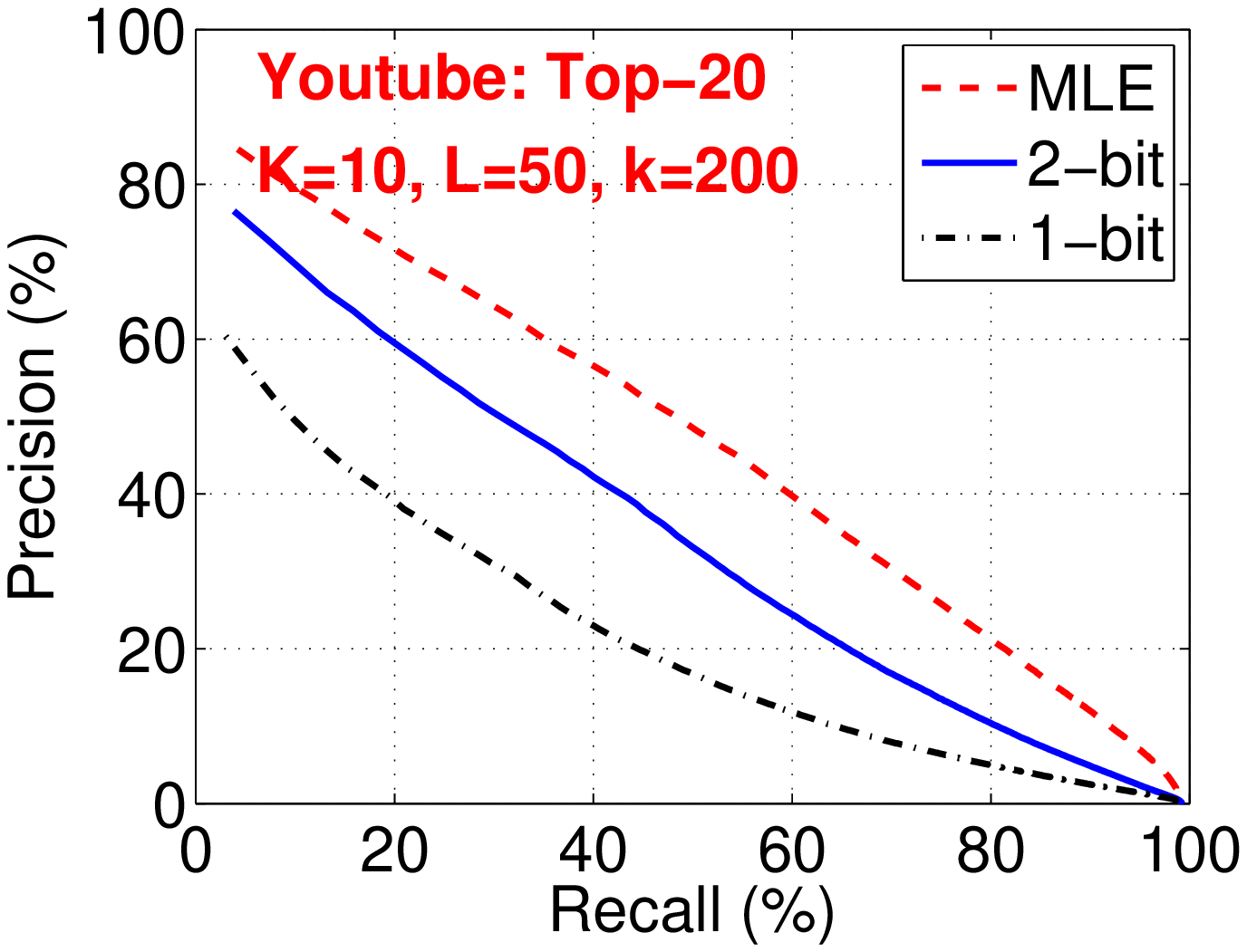}
\includegraphics[width = 1.7in]{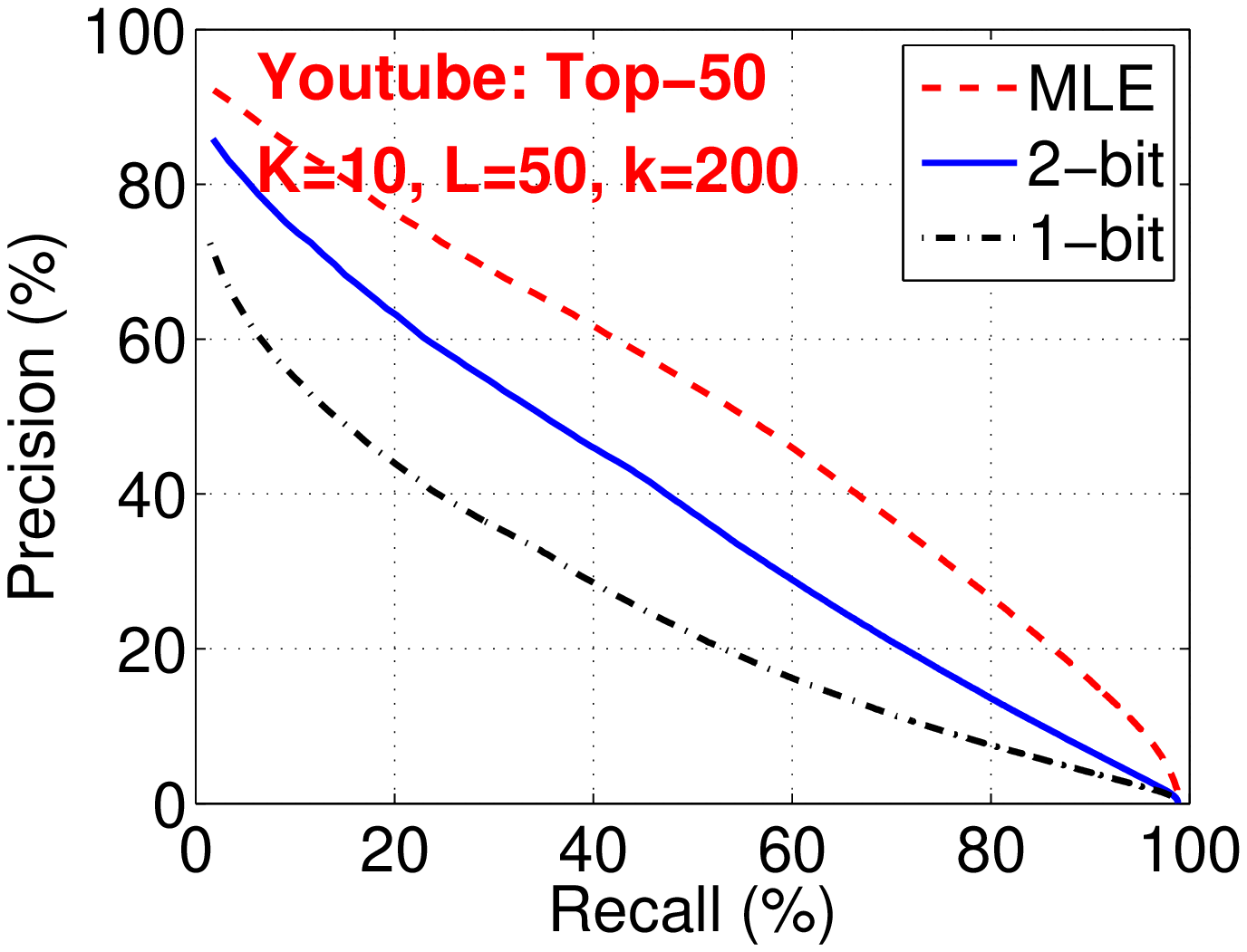}
\includegraphics[width = 1.7in]{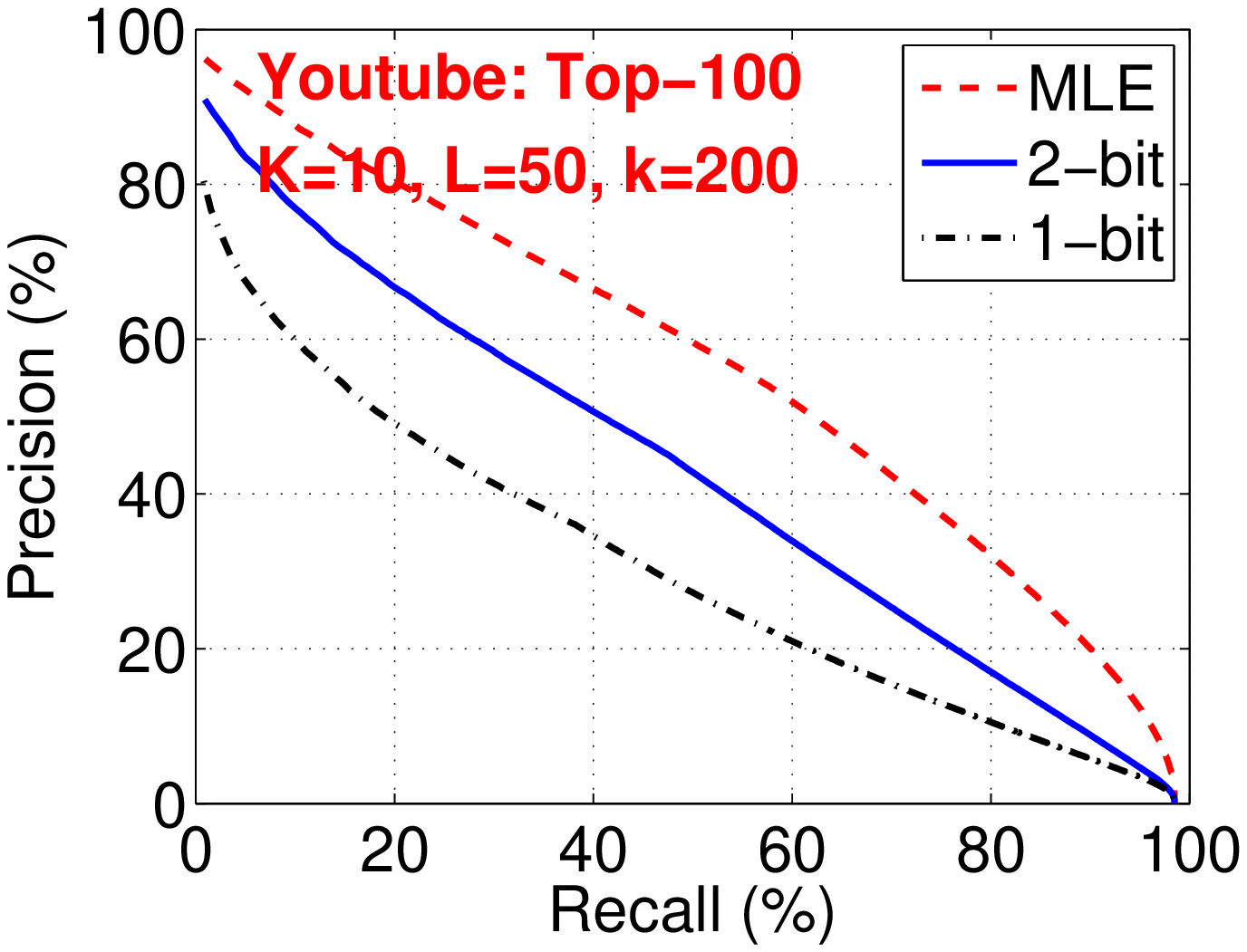}
}

\mbox
{
\includegraphics[width = 1.7in]{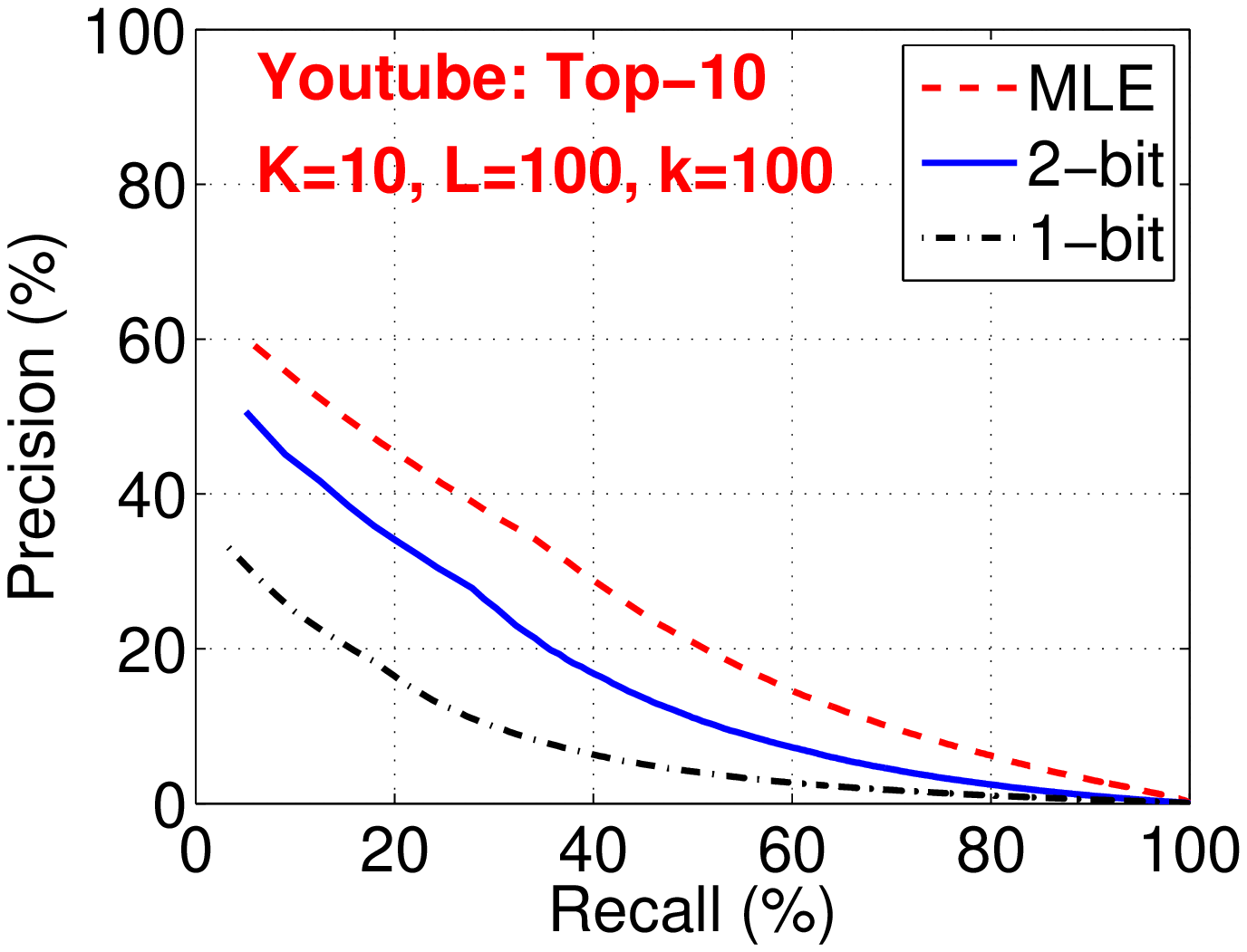}
\includegraphics[width = 1.7in]{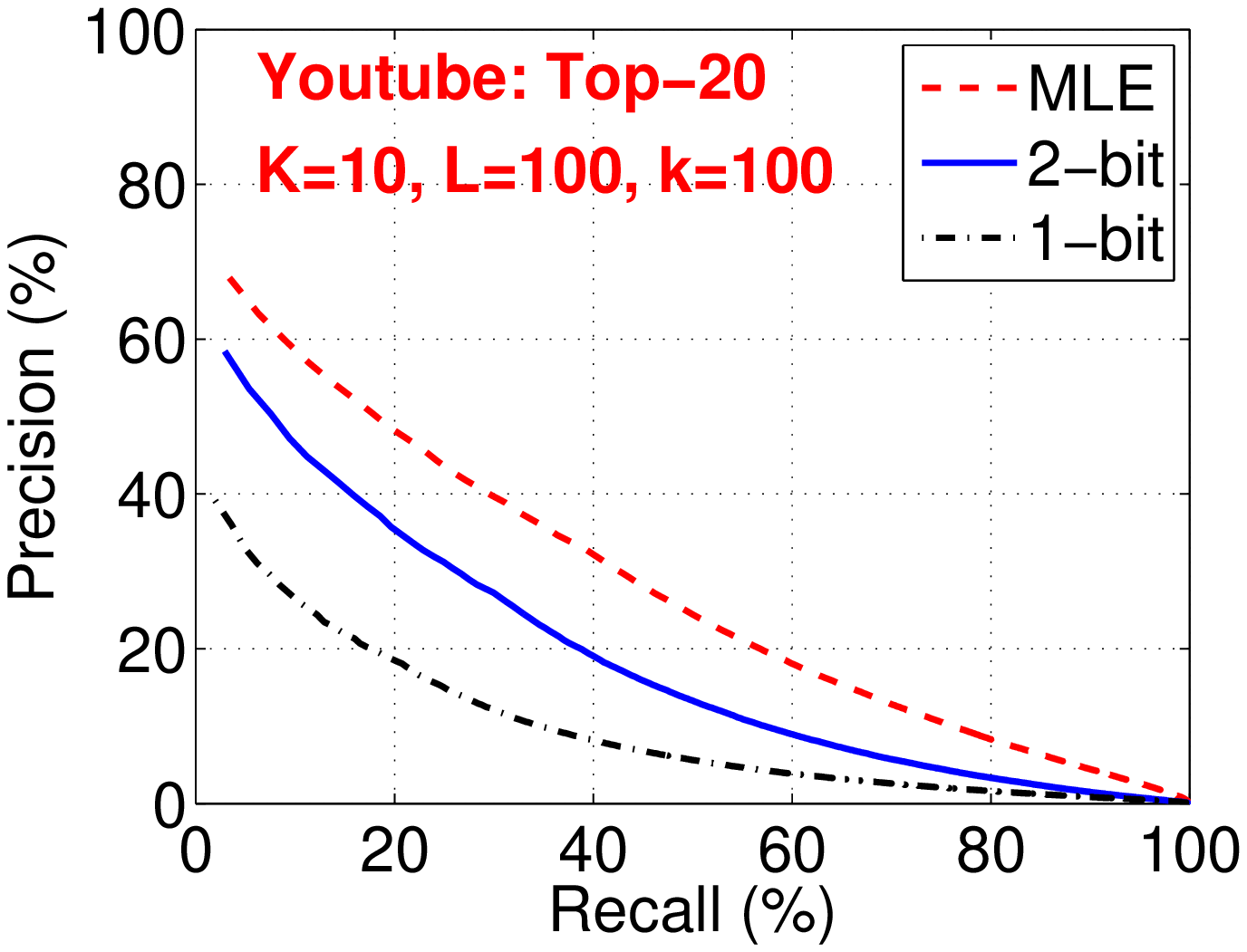}
\includegraphics[width = 1.7in]{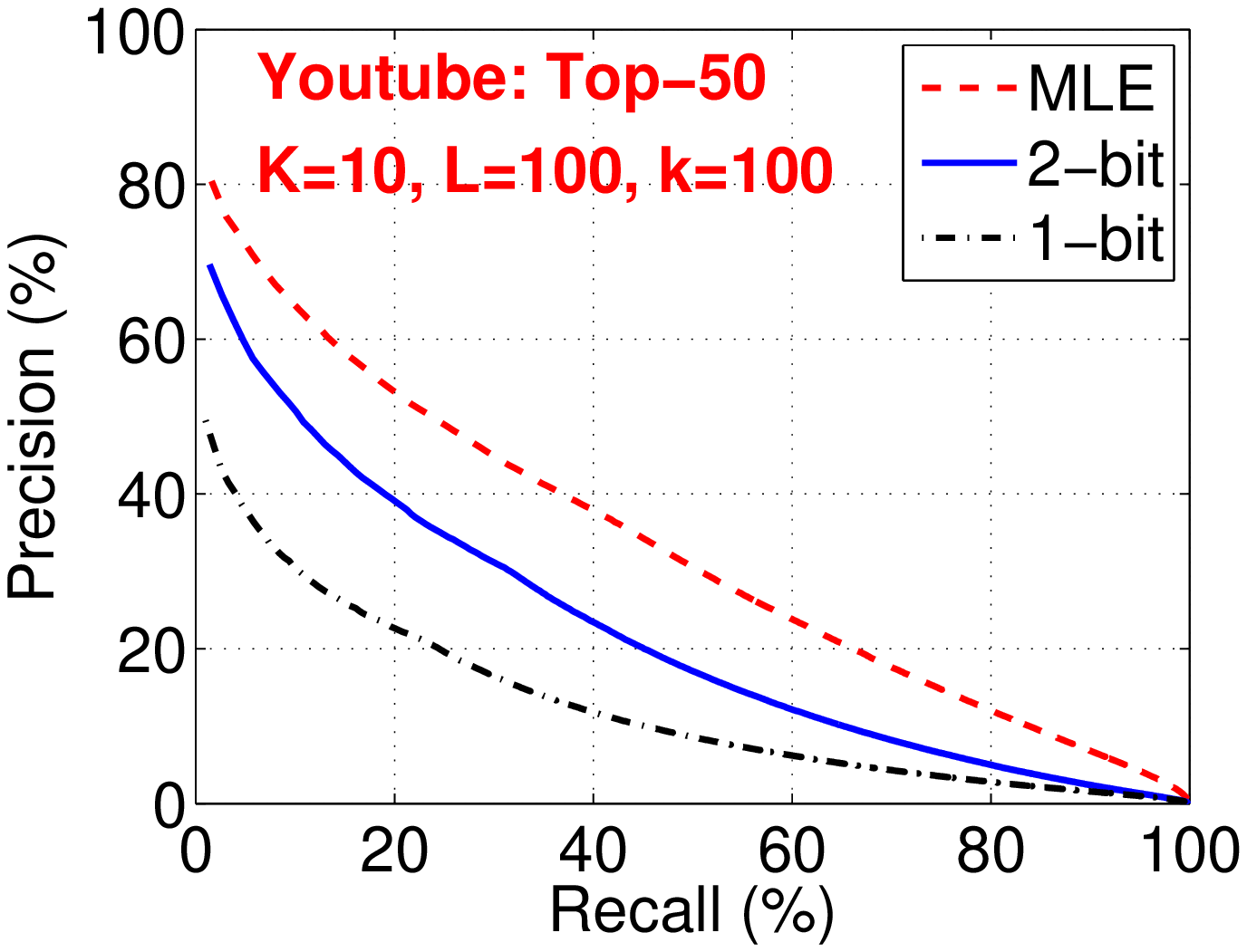}
\includegraphics[width = 1.7in]{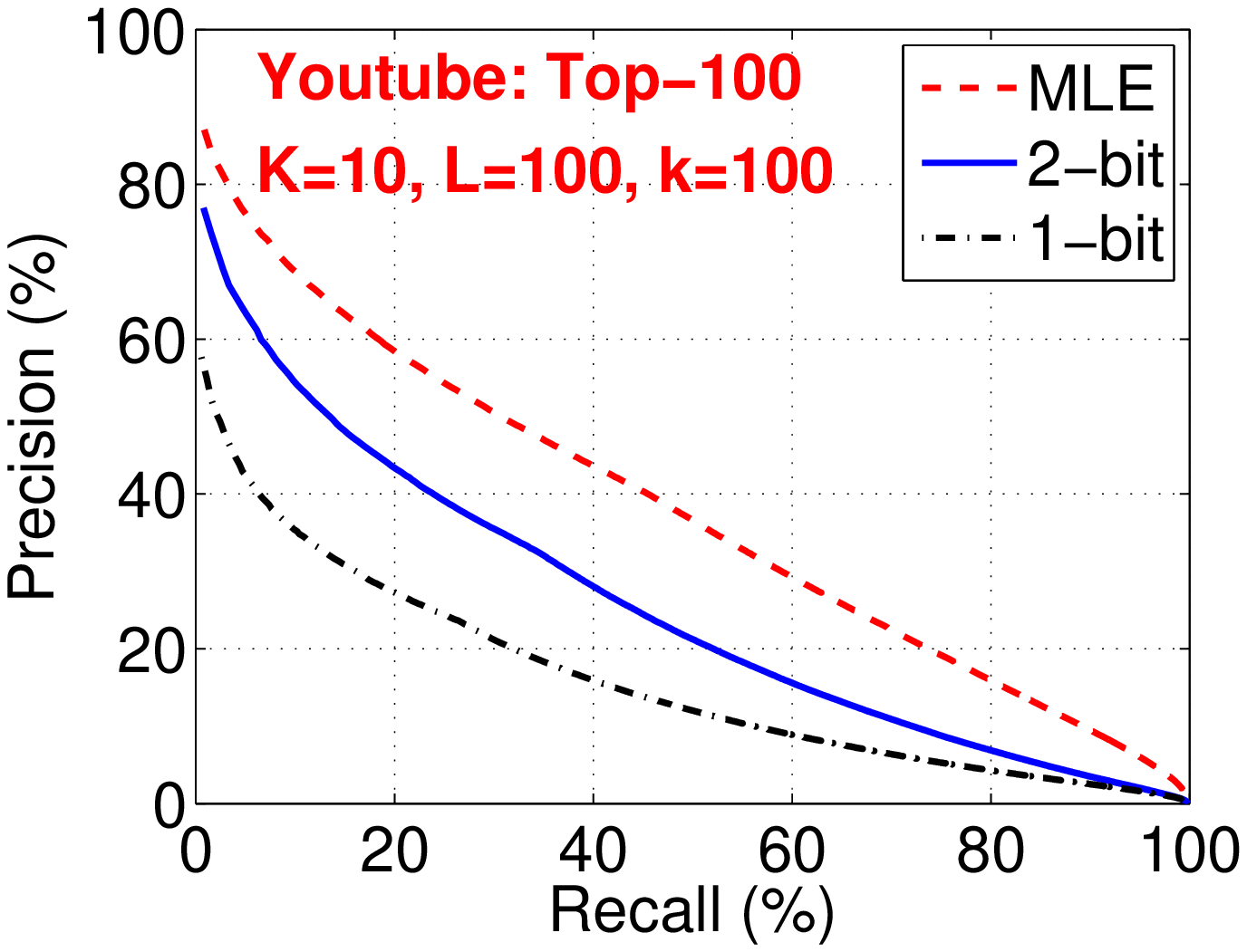}
}

\mbox{
\includegraphics[width = 1.7in]{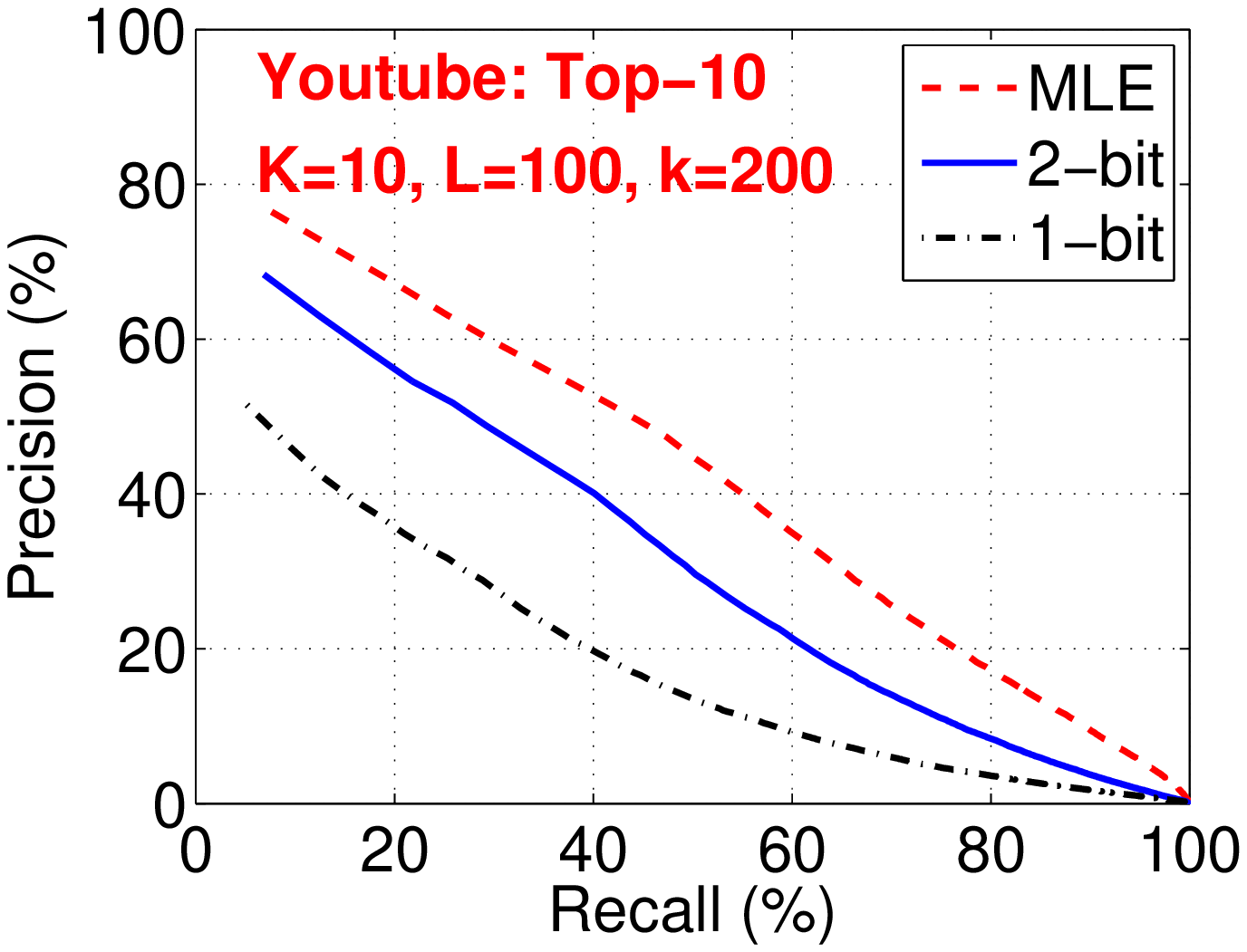}
\includegraphics[width = 1.7in]{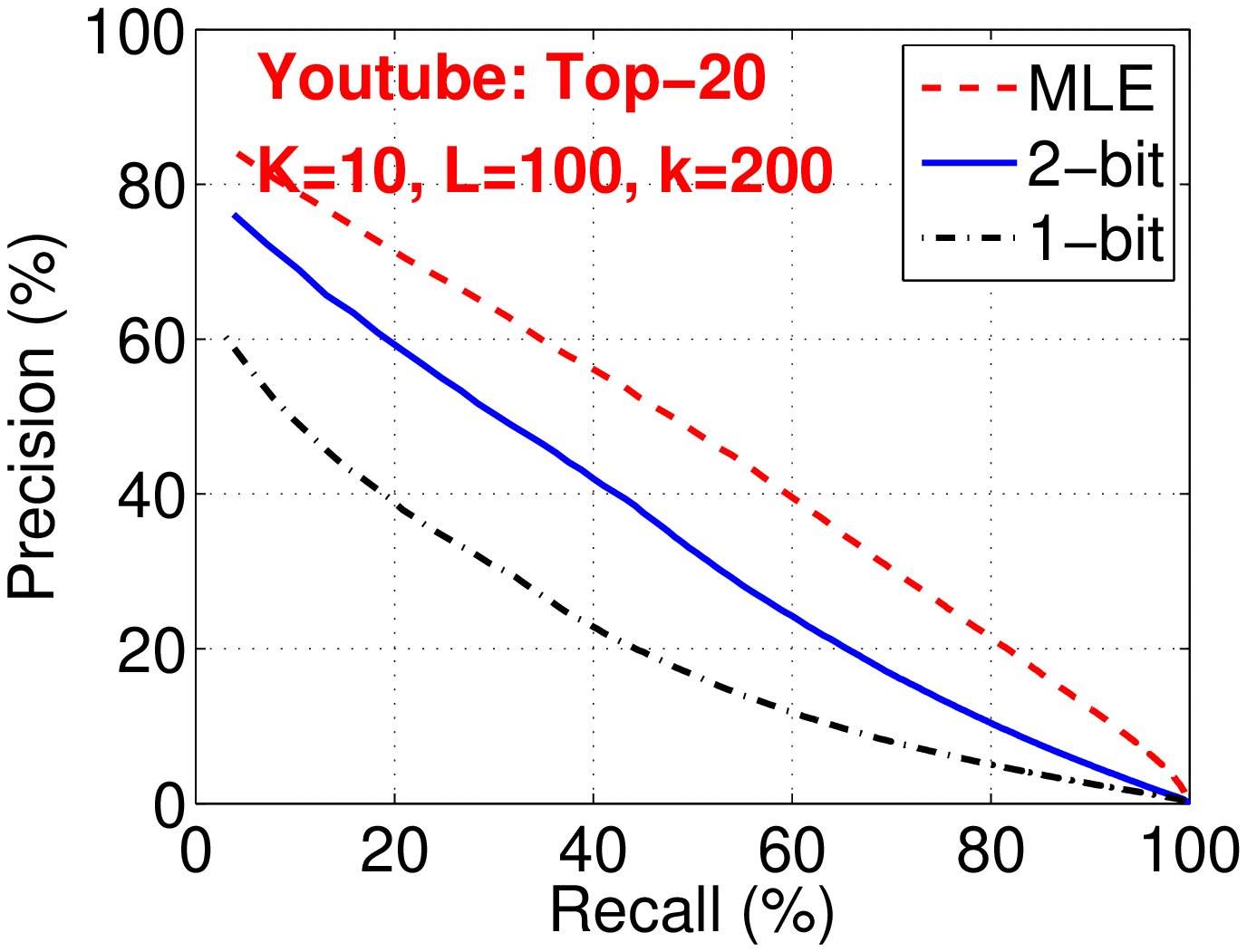}
\includegraphics[width = 1.7in]{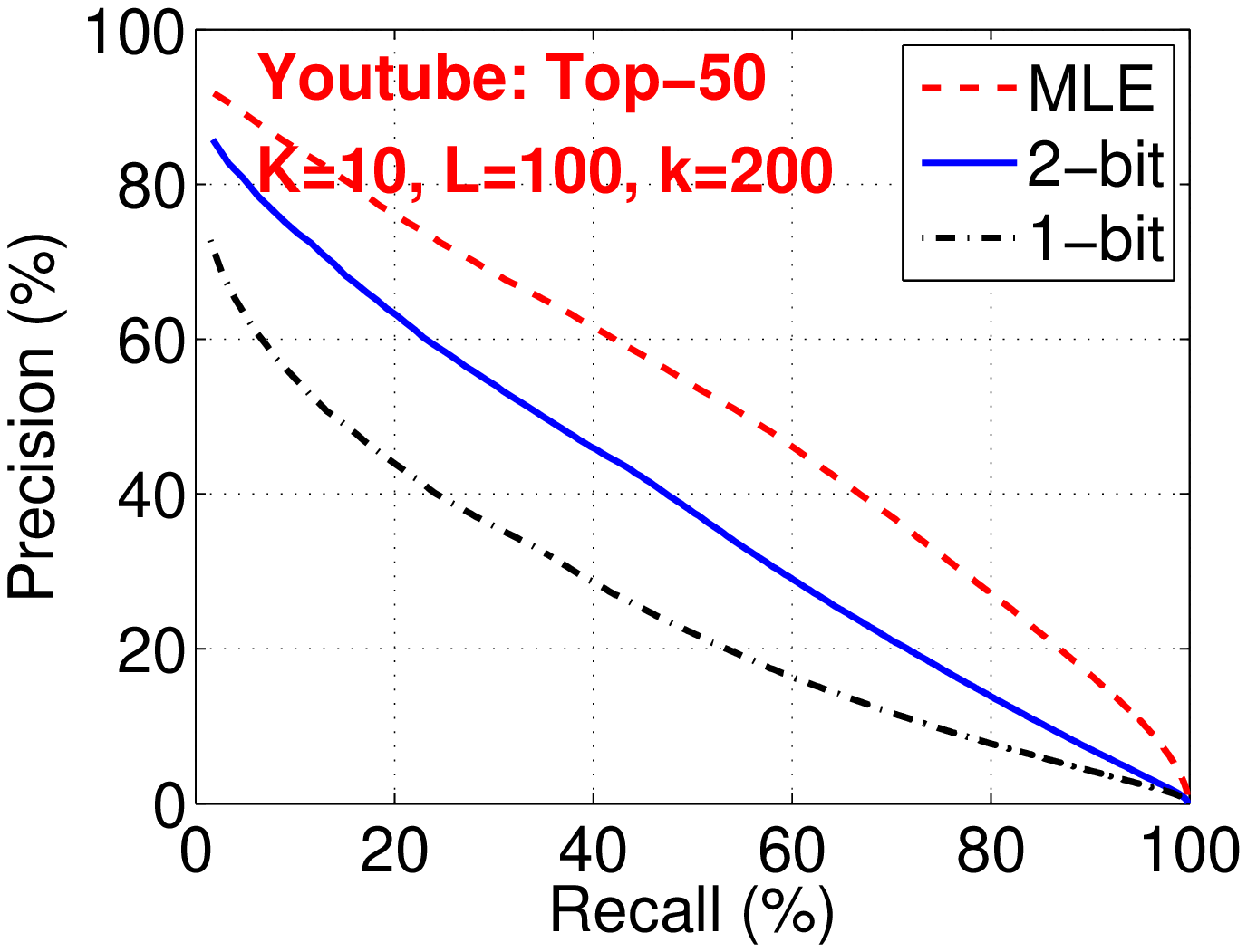}
\includegraphics[width = 1.7in]{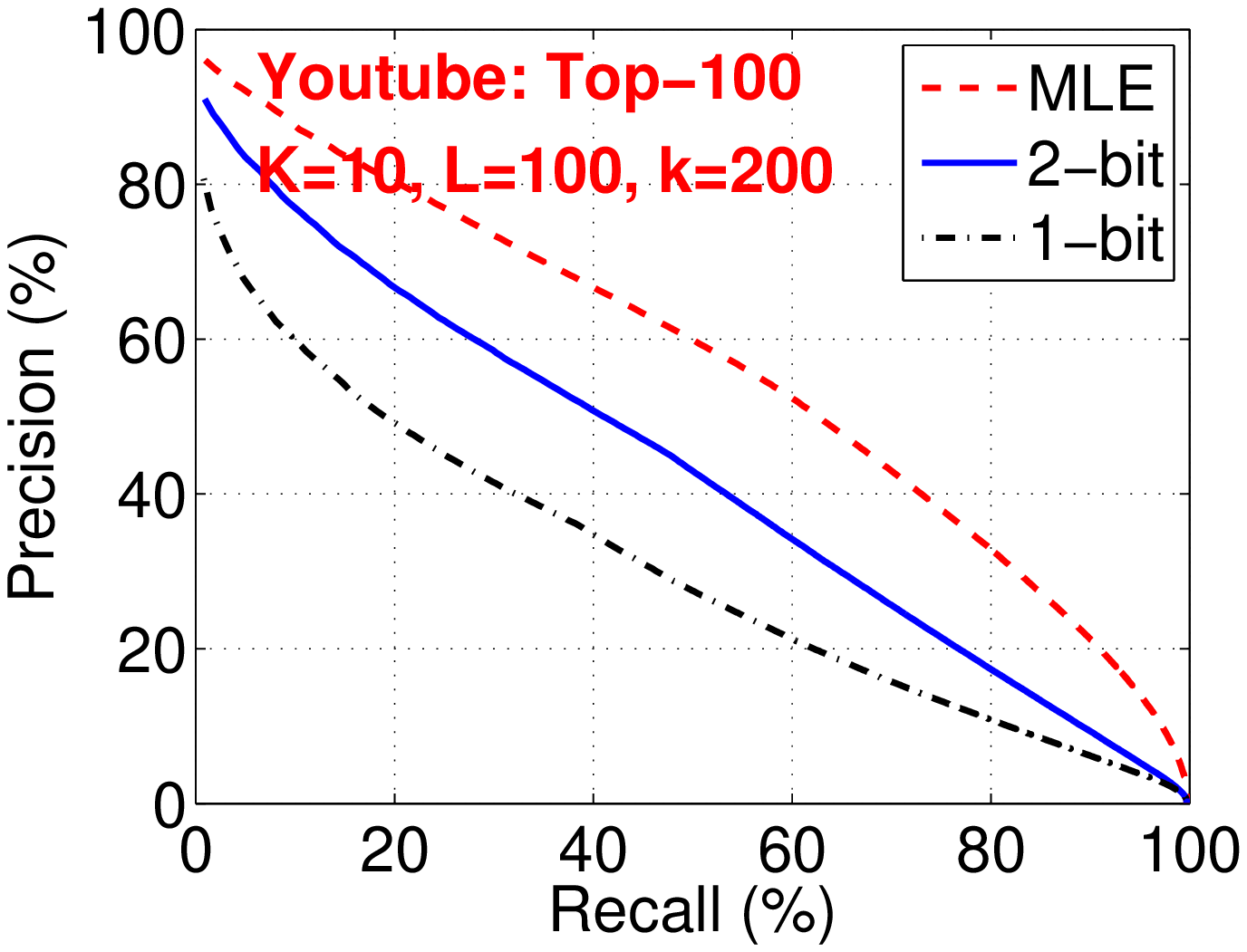}
}

\end{center}
\caption{\textbf{Youtube}: precision-recall curves (higher is better) for retrieving the top-10, -20, -50, -100 nearest neighbors using standard $(K,L)$-LSH scheme and 3 different estimators of similarities (for the retrieved data points). The Youtube dataset is a subset from the publicly available UCL-Youtube-Vision dataset. We use 97,934 data points for building hash tables and 5,000 data points for the query. The results are averaged over all the query points.  In the LSH experiments,  we fix $K=10$ and $L=50$ (upper two layers) and $L=100$ (bottom two layers). We estimate the similarities using two different sample sizes, for $k=100$ and $k=200$. We can see that for any combinations of parameters, the nonlinear MLE (labeled as ``MLE'') always substantially improves the 2-bit linear estimator (labeled as ``2-bit''), which substantially improves the 1-bit  estimator (labeled as ``1-bit''). }\label{fig_YoutubeK10}
\end{minipage}
\end{figure*}

\begin{figure*}[h!]
\begin{center}
\mbox
{
\includegraphics[width = 1.4in]{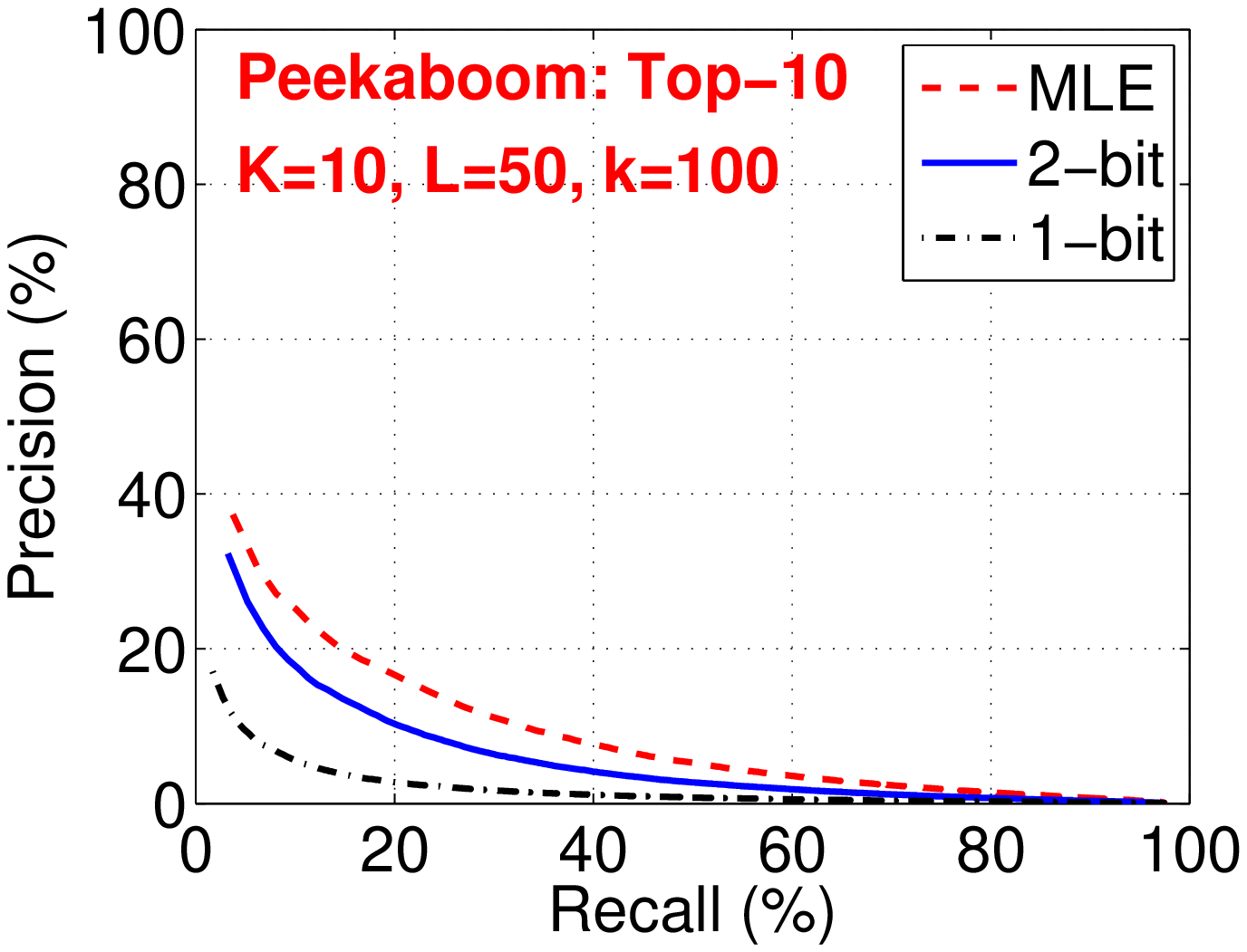}
\includegraphics[width = 1.4in]{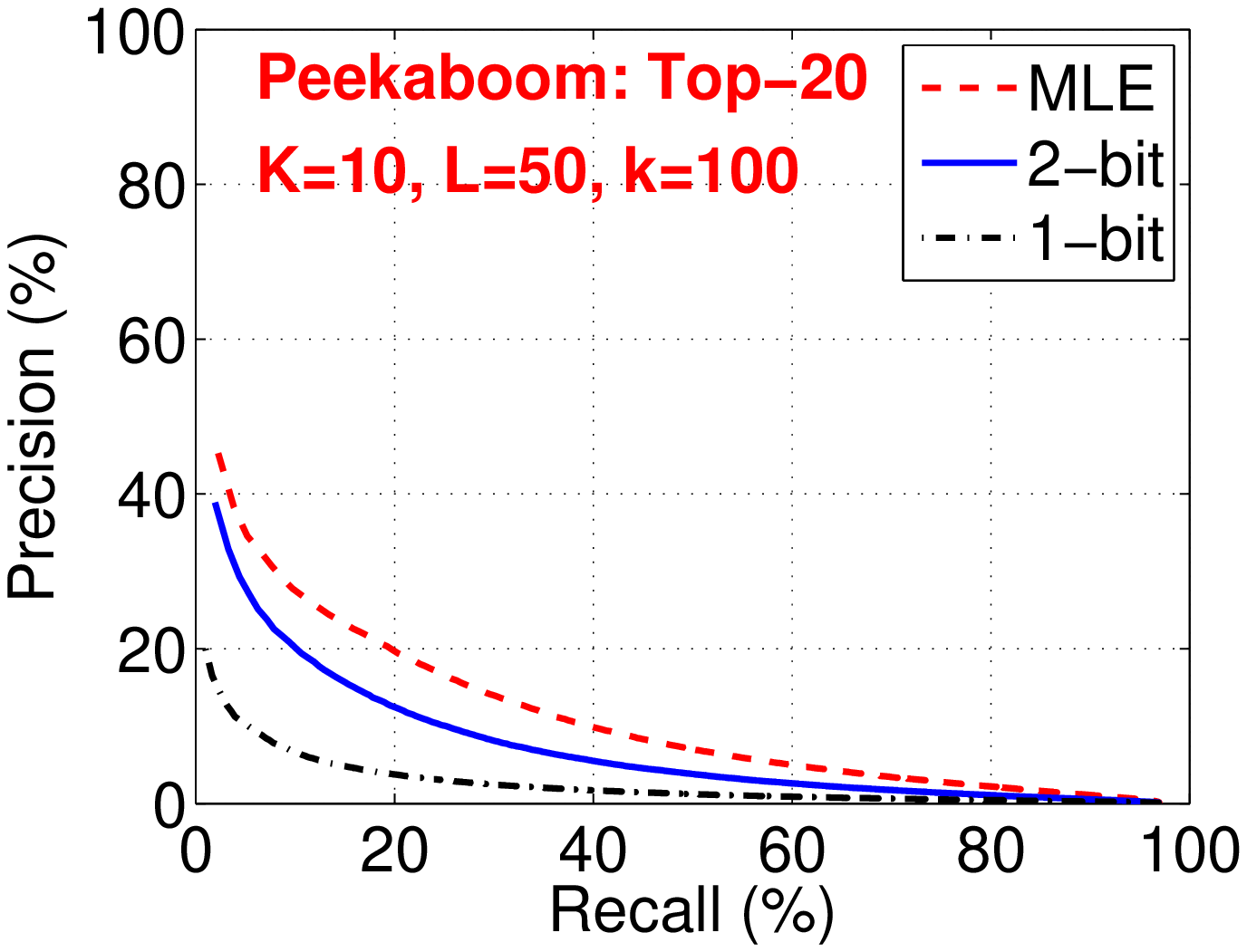}
\includegraphics[width = 1.4in]{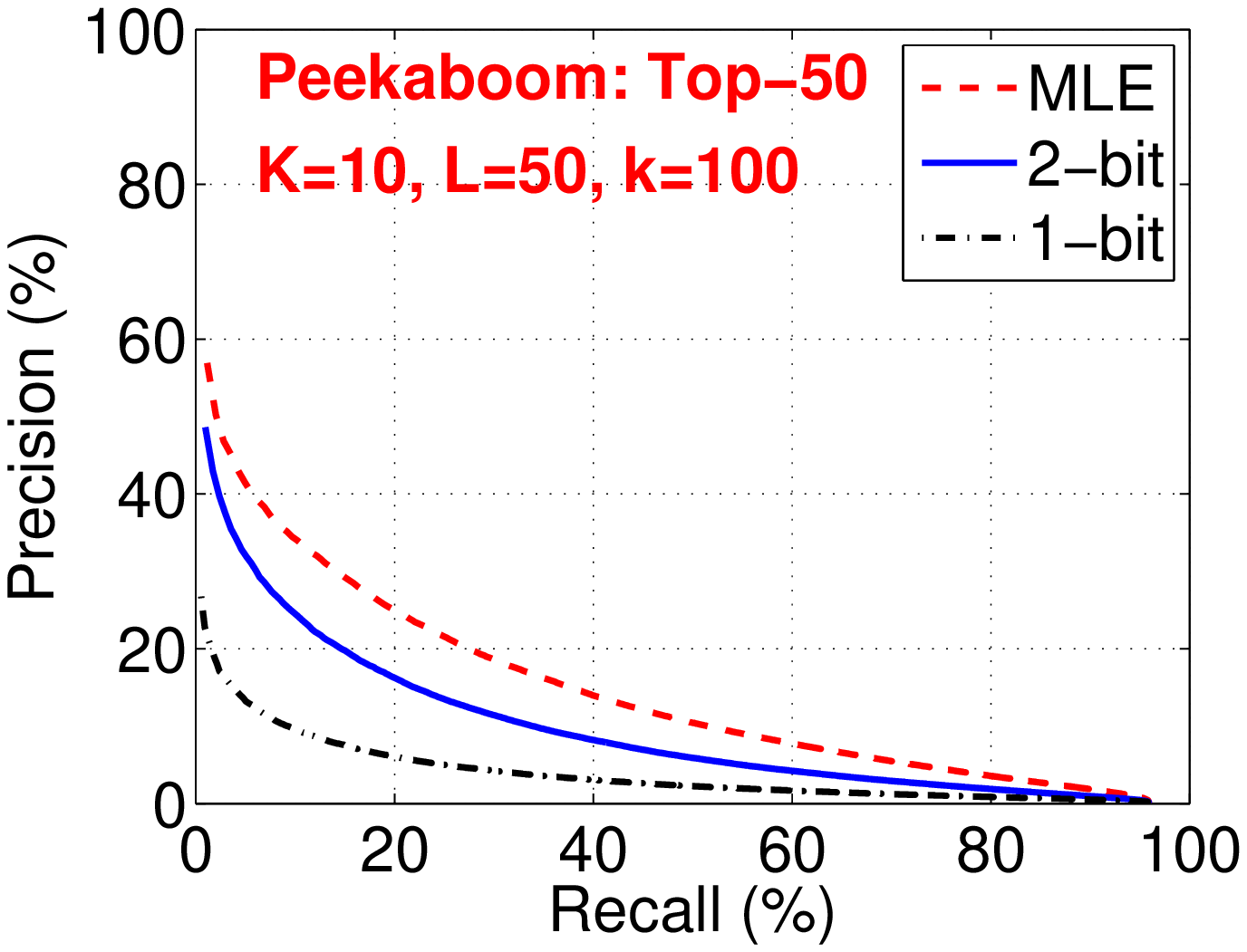}
\includegraphics[width = 1.4in]{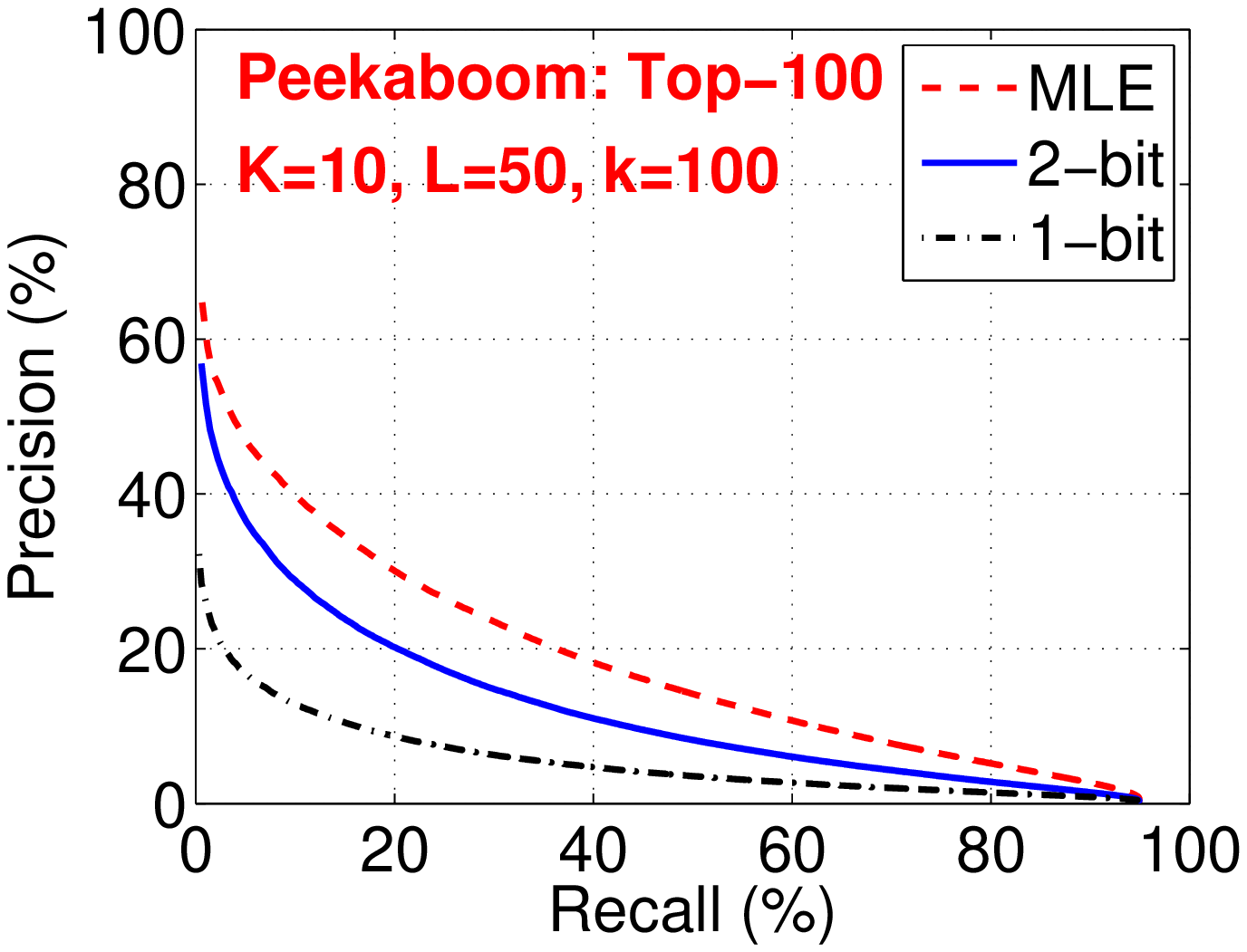}
}

\vspace{-0.025in}

\mbox{
\includegraphics[width = 1.4in]{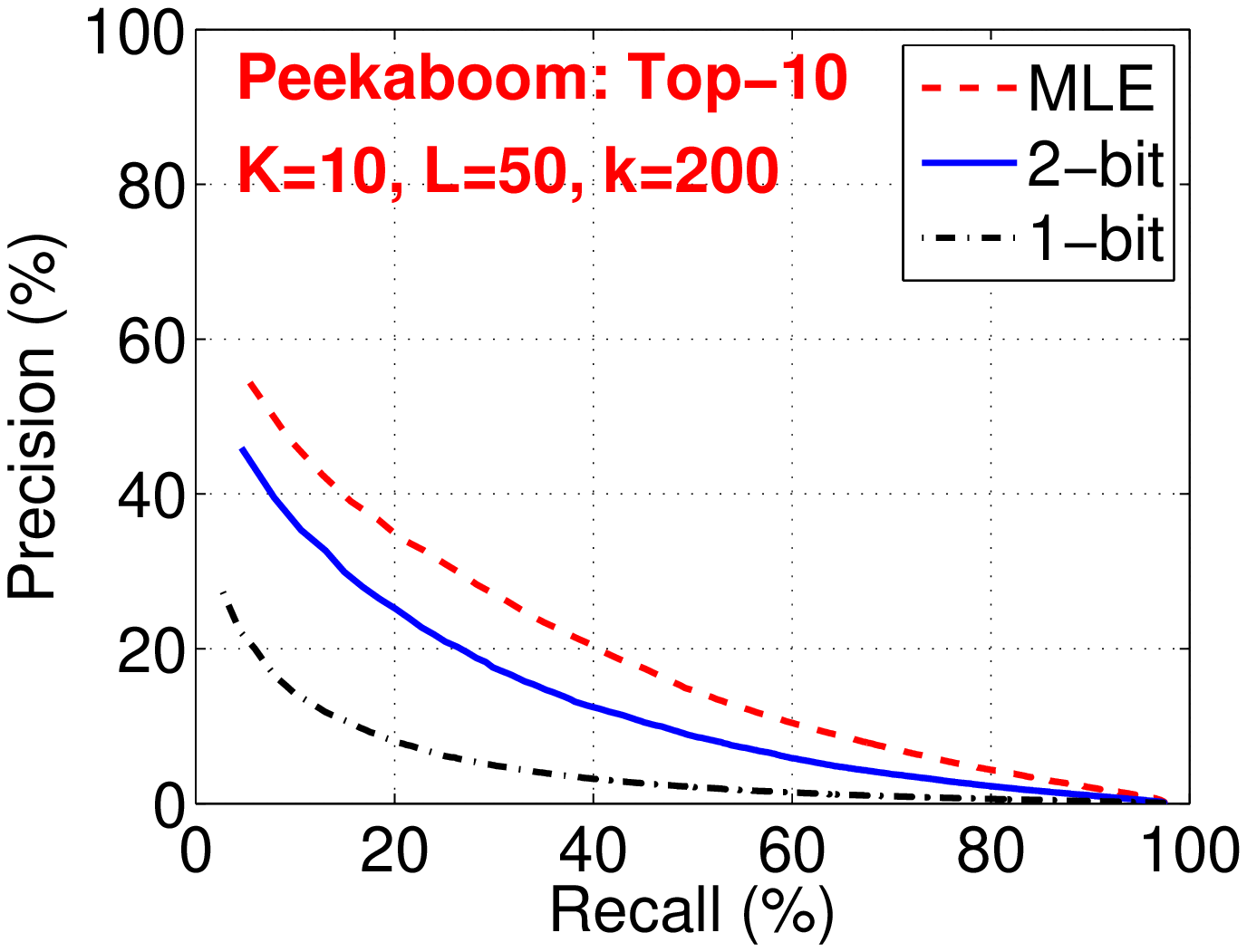}
\includegraphics[width = 1.4in]{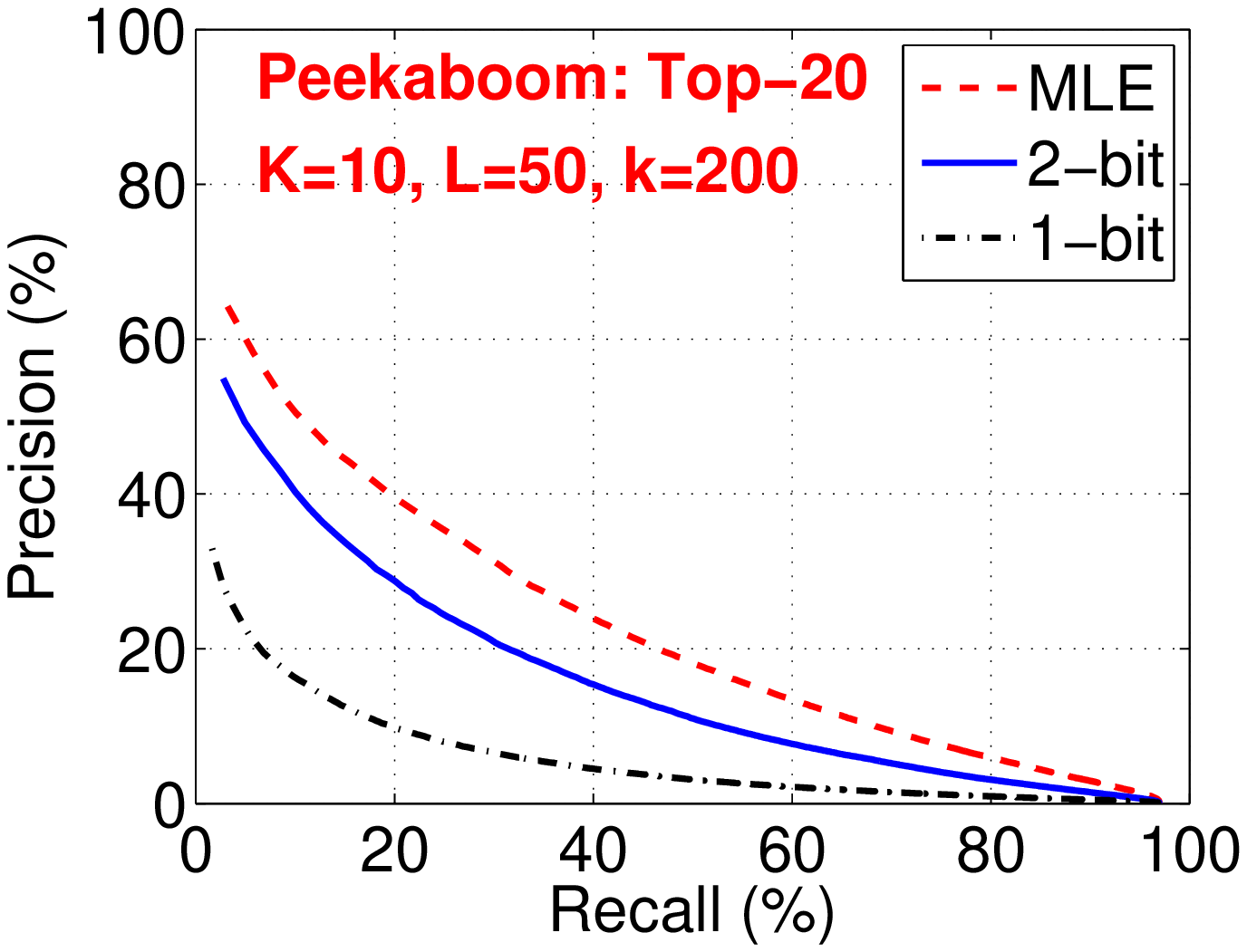}
\includegraphics[width = 1.4in]{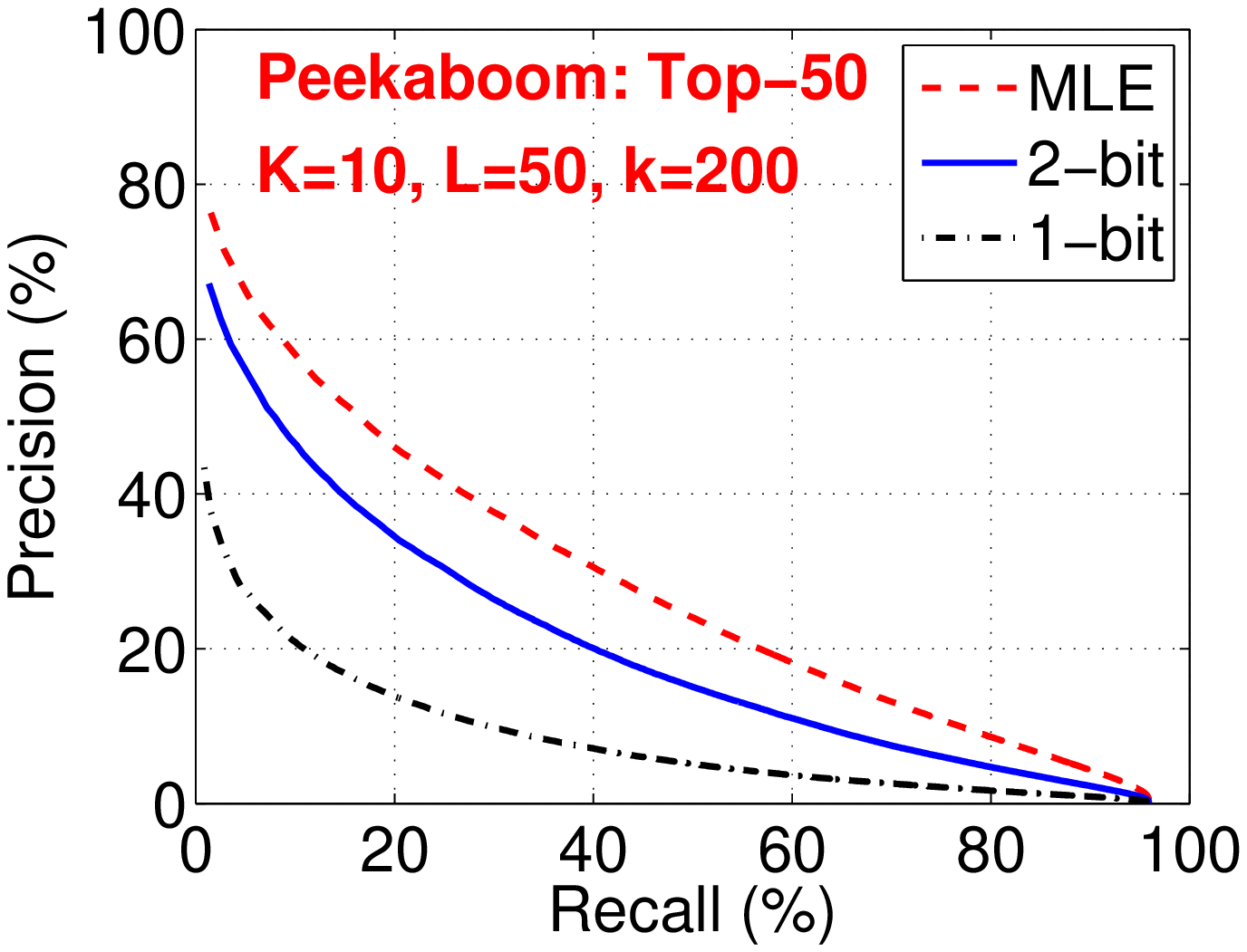}
\includegraphics[width = 1.4in]{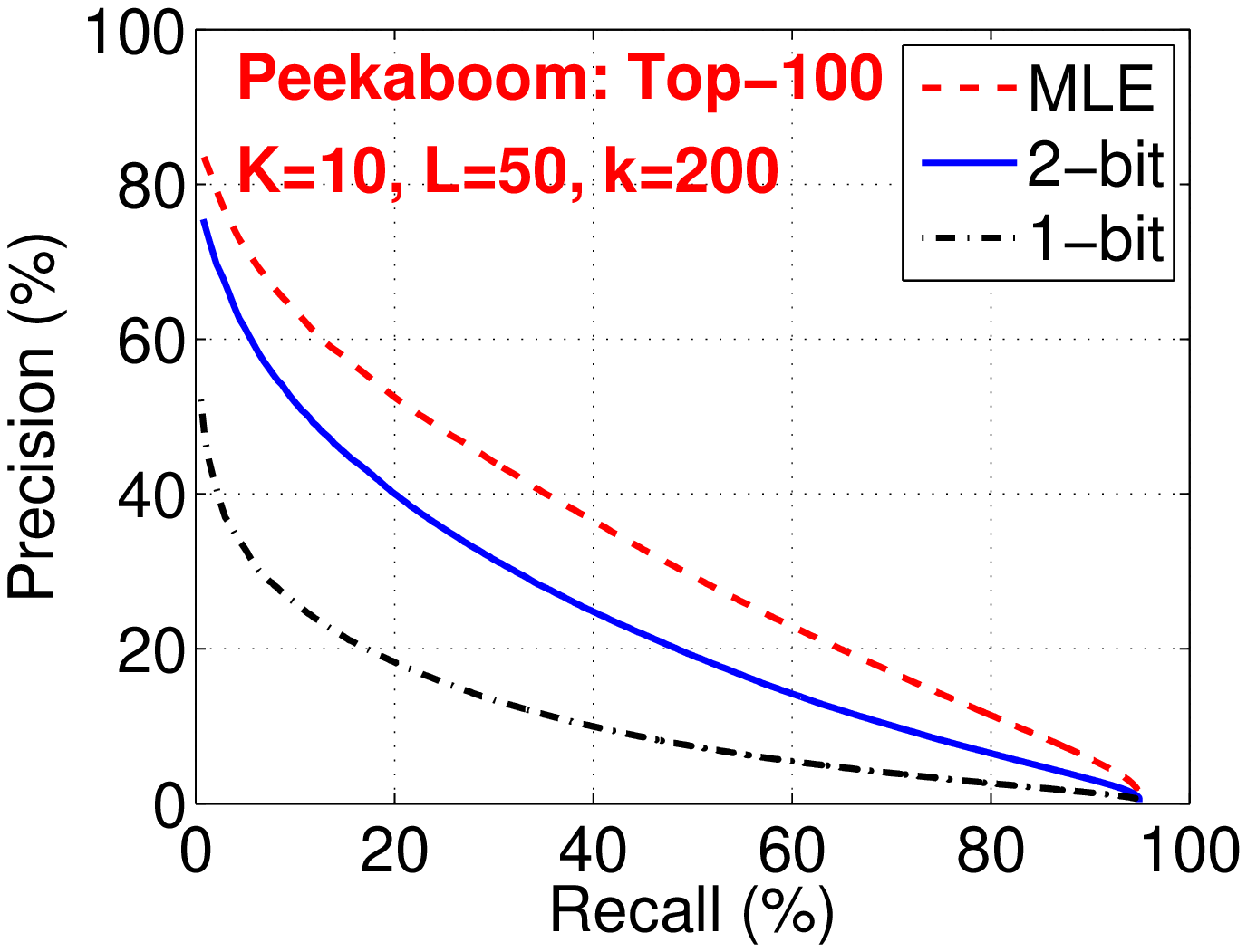}
}

\vspace{-0.025in}

\mbox
{
\includegraphics[width = 1.4in]{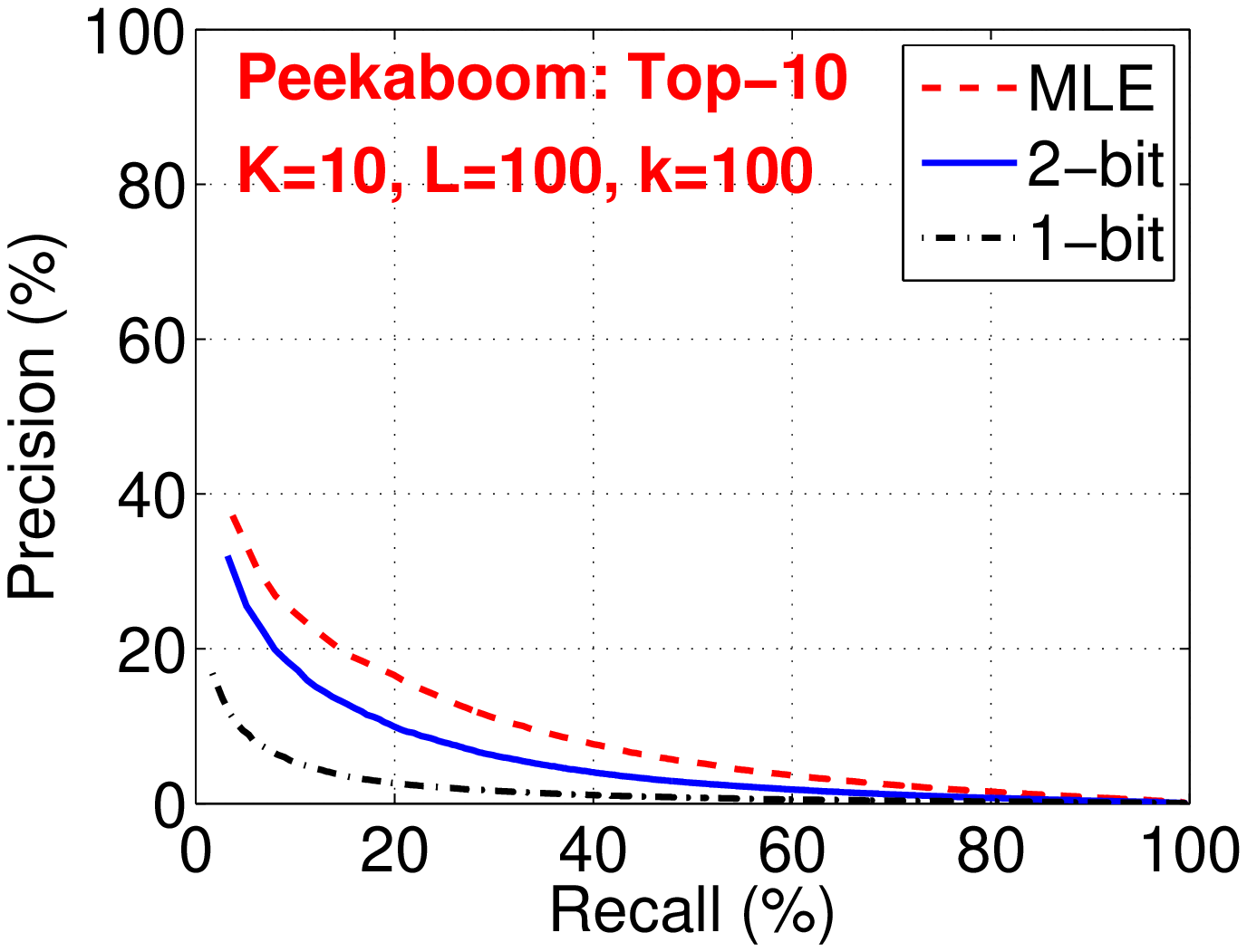}
\includegraphics[width = 1.4in]{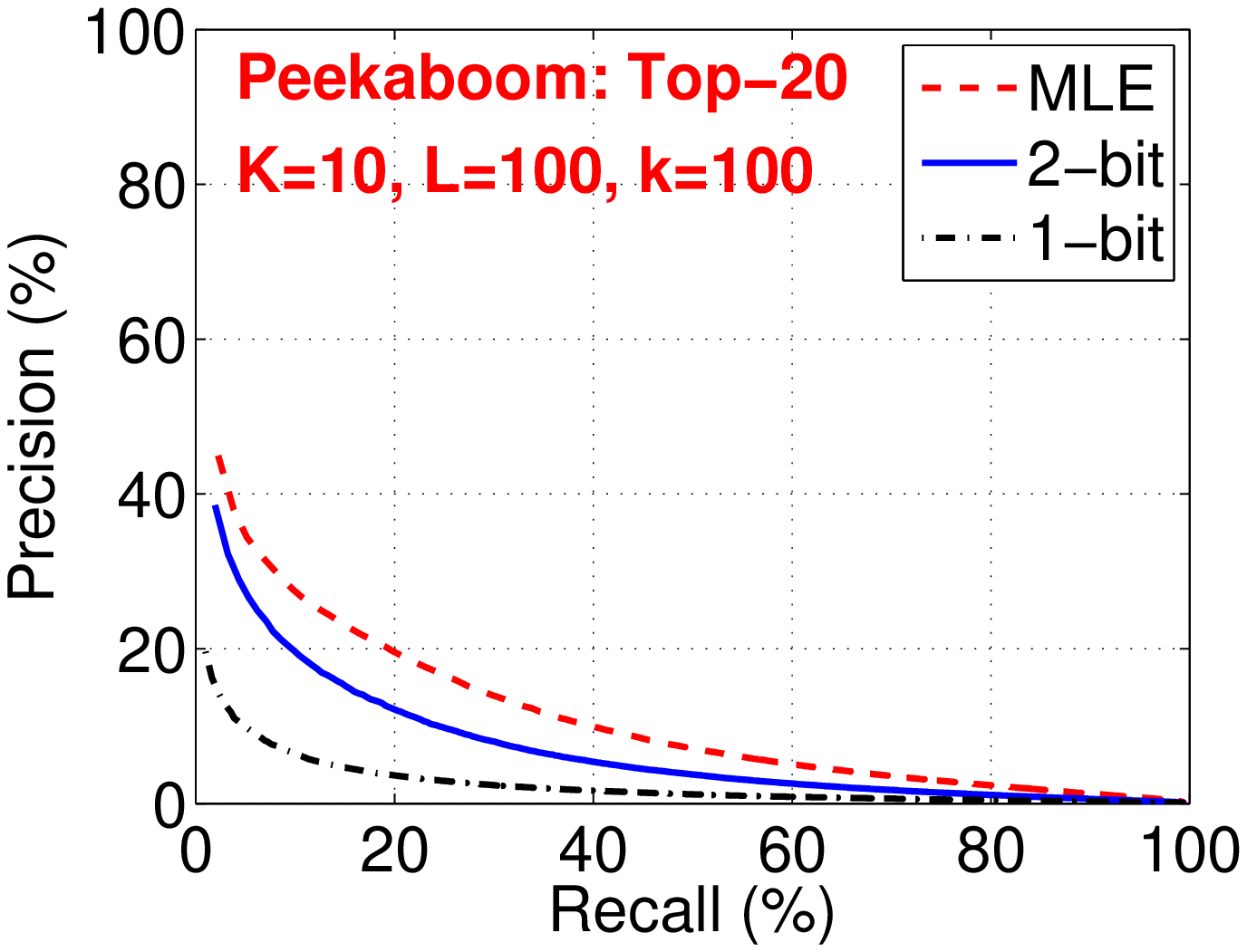}
\includegraphics[width = 1.4in]{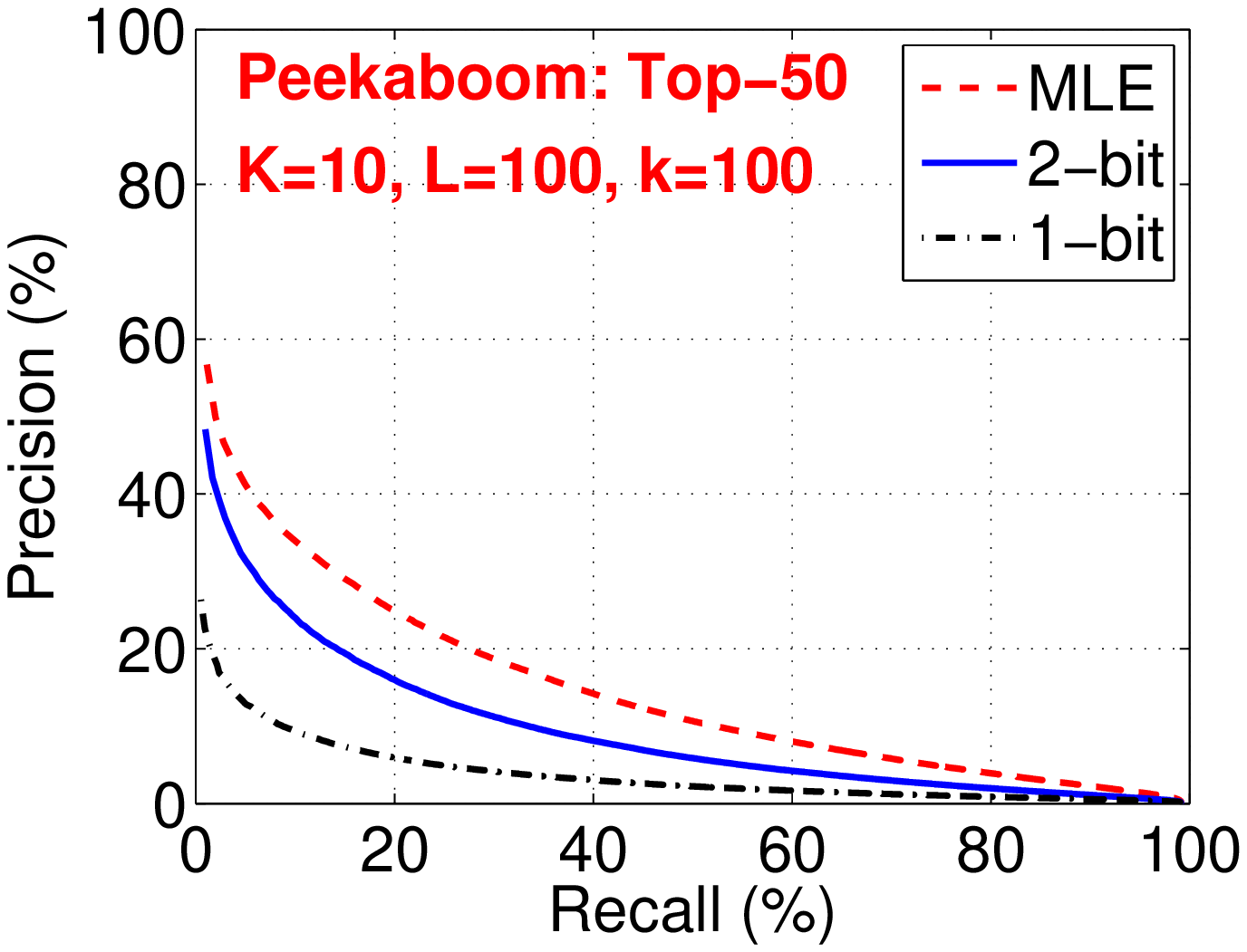}
\includegraphics[width = 1.4in]{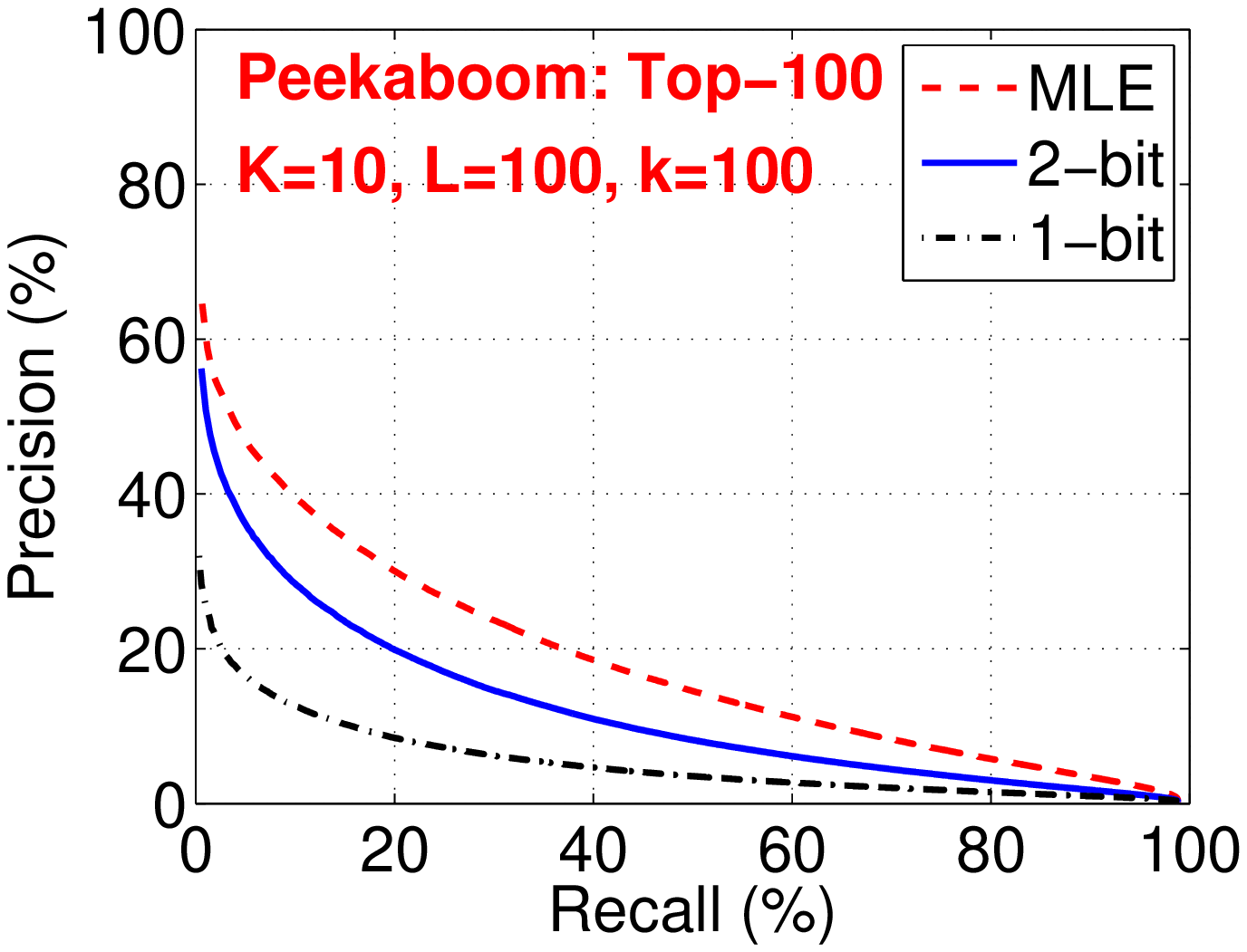}
}

\vspace{-0.025in}

\mbox{
\includegraphics[width = 1.4in]{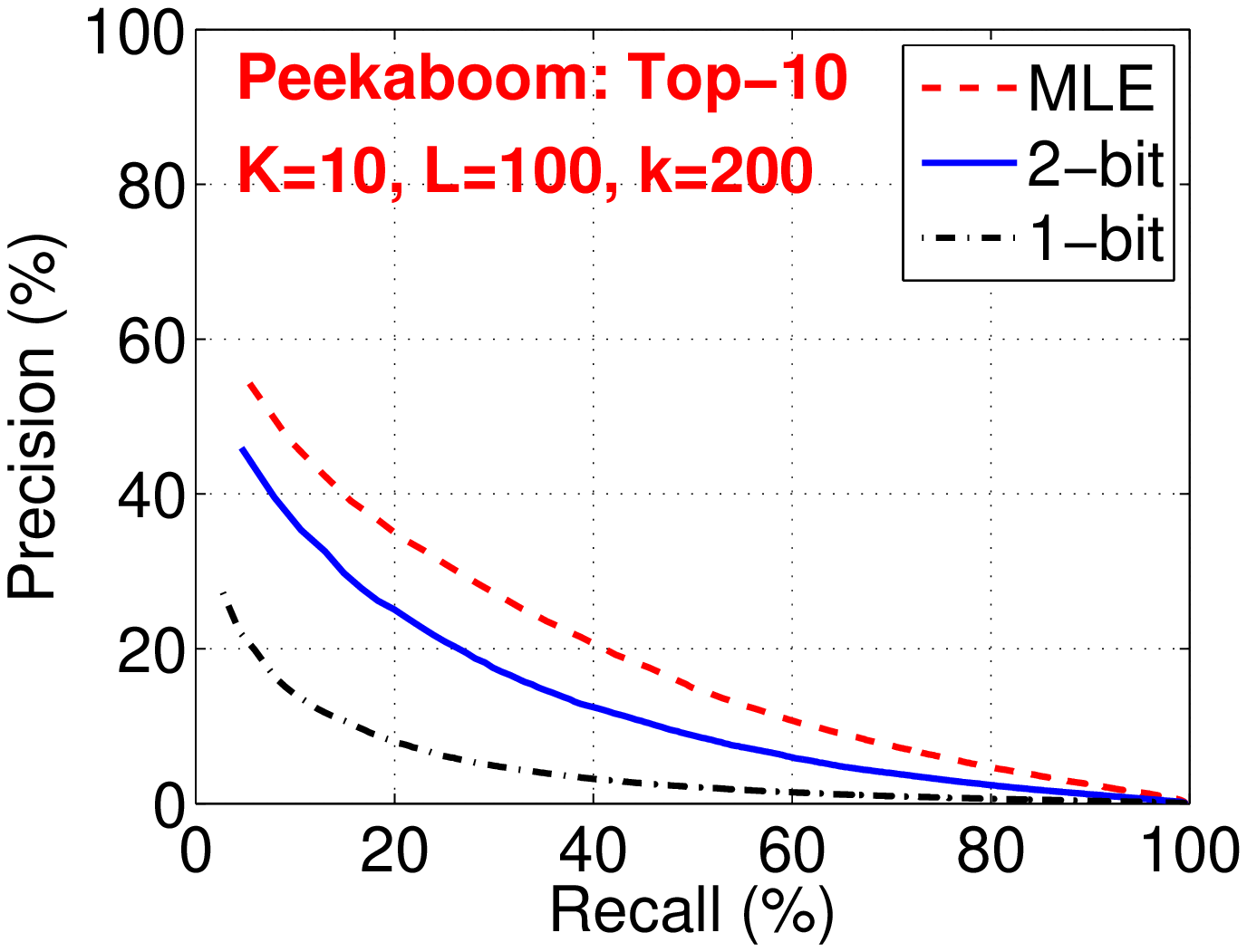}
\includegraphics[width = 1.4in]{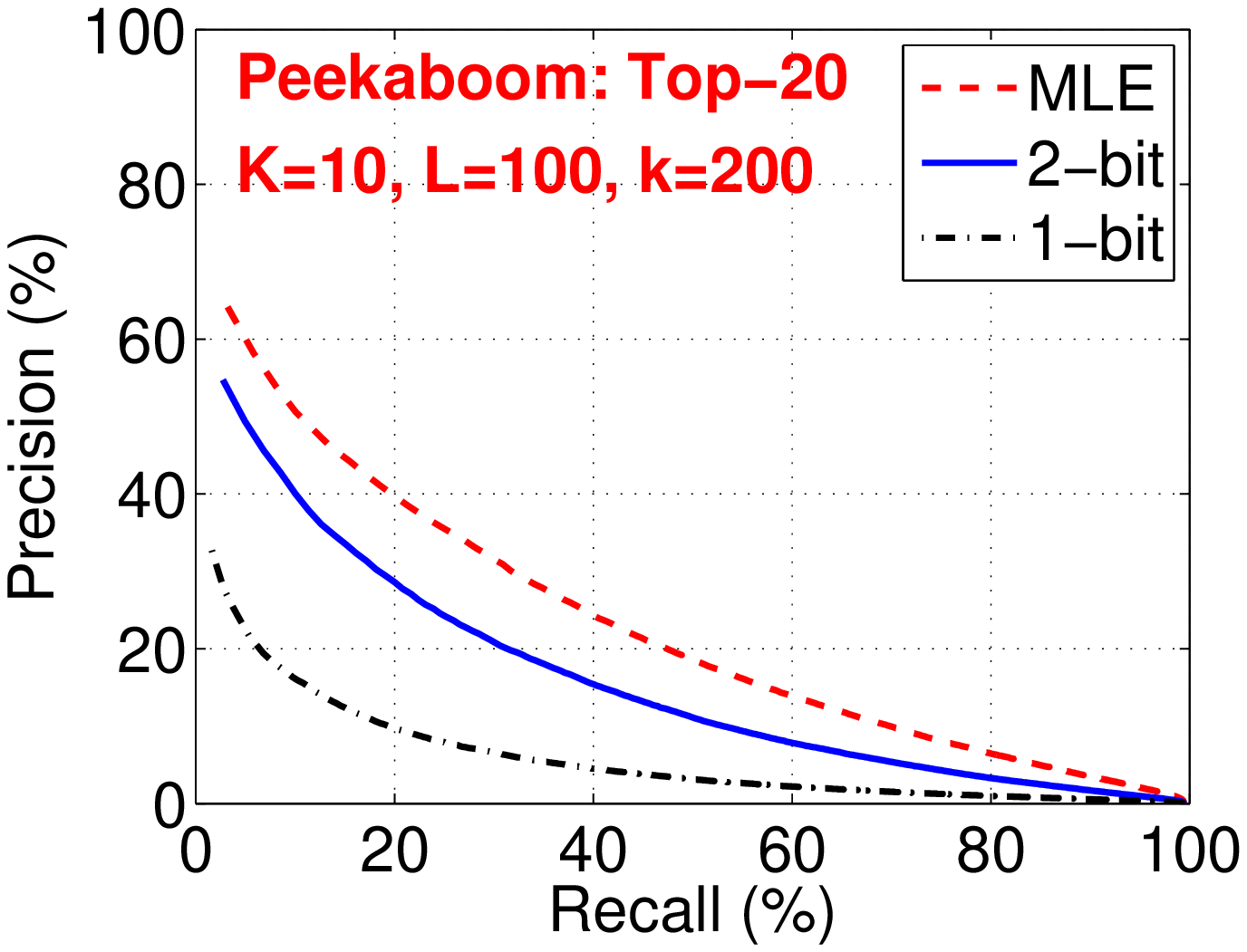}
\includegraphics[width = 1.4in]{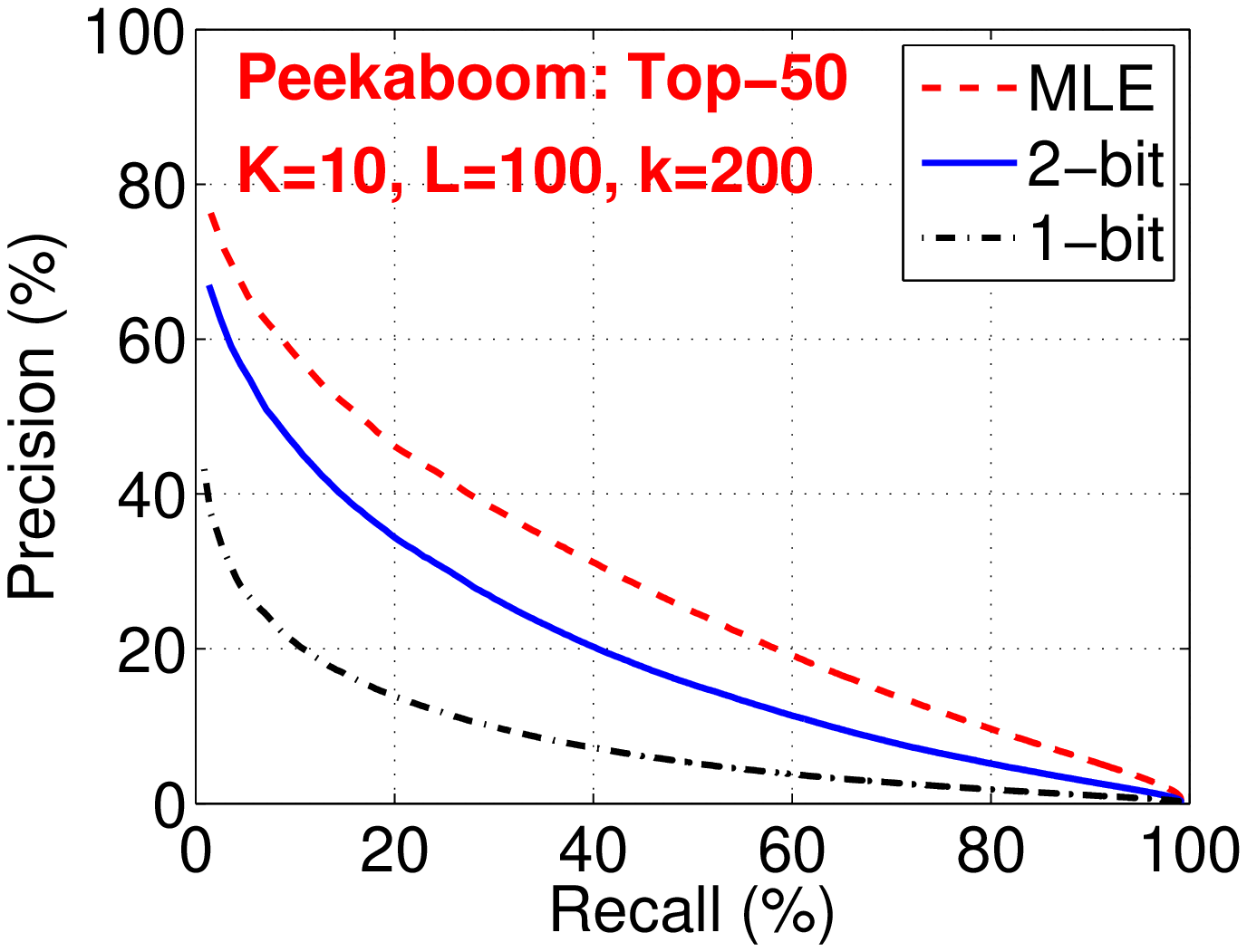}
\includegraphics[width = 1.4in]{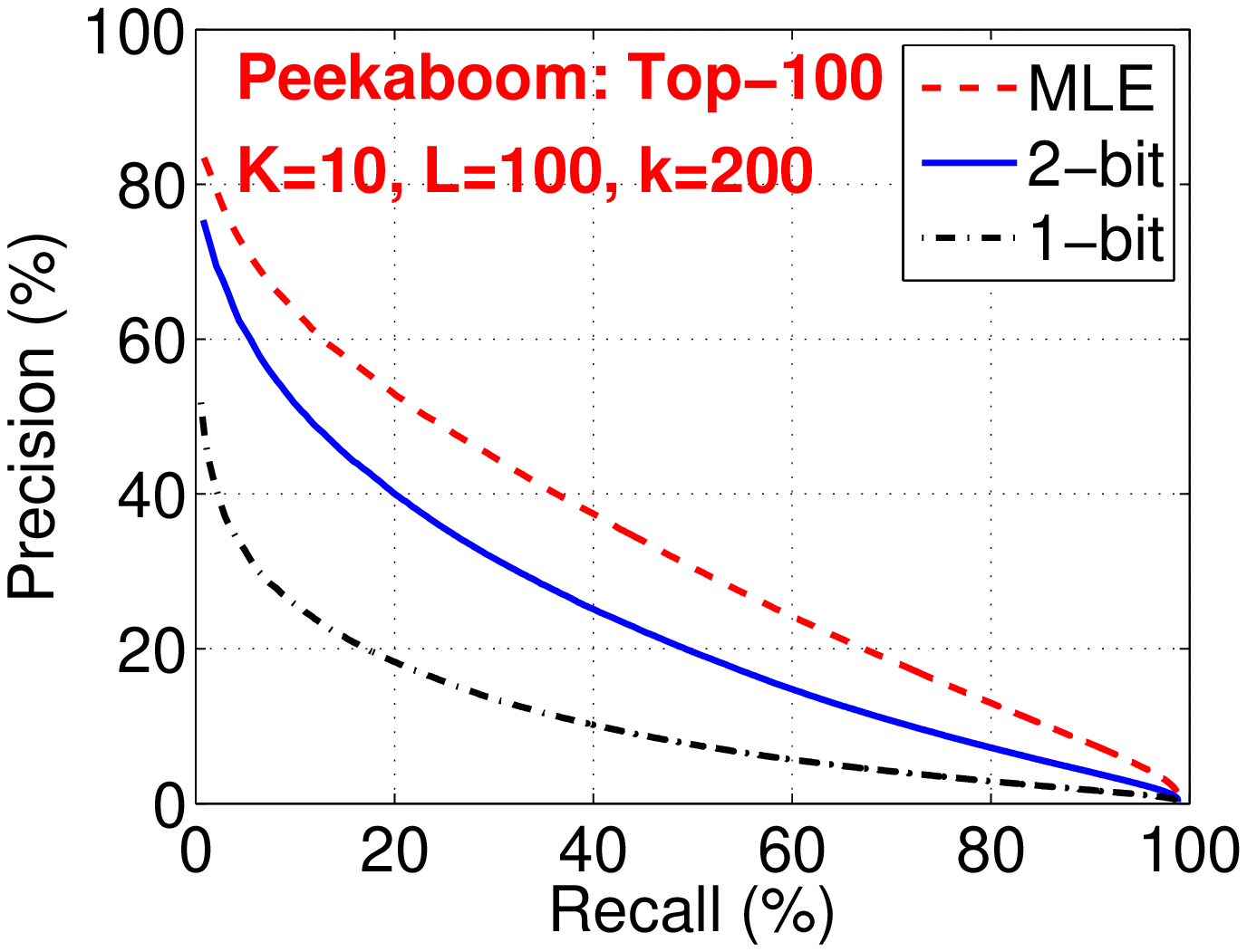}
}

\end{center}
\vspace{-0.25in}
\caption{\textbf{Peekaboom}: precision-recall curves (higher is better) for retrieving the top-10, -20, -50, -100 nearest neighbors using standard $(K,L)$-LSH scheme and 3 different estimators of similarities (for the retrieved data points). Peekaboom is a standard image retrieval dataset with 20,019 data points for building the tables and 2,000 data points for the query. }\label{fig_PeekaboomK10}\vspace{-0.14in}
\end{figure*}

\begin{figure*}[h!]
\begin{center}
\mbox
{
\includegraphics[width = 1.4in]{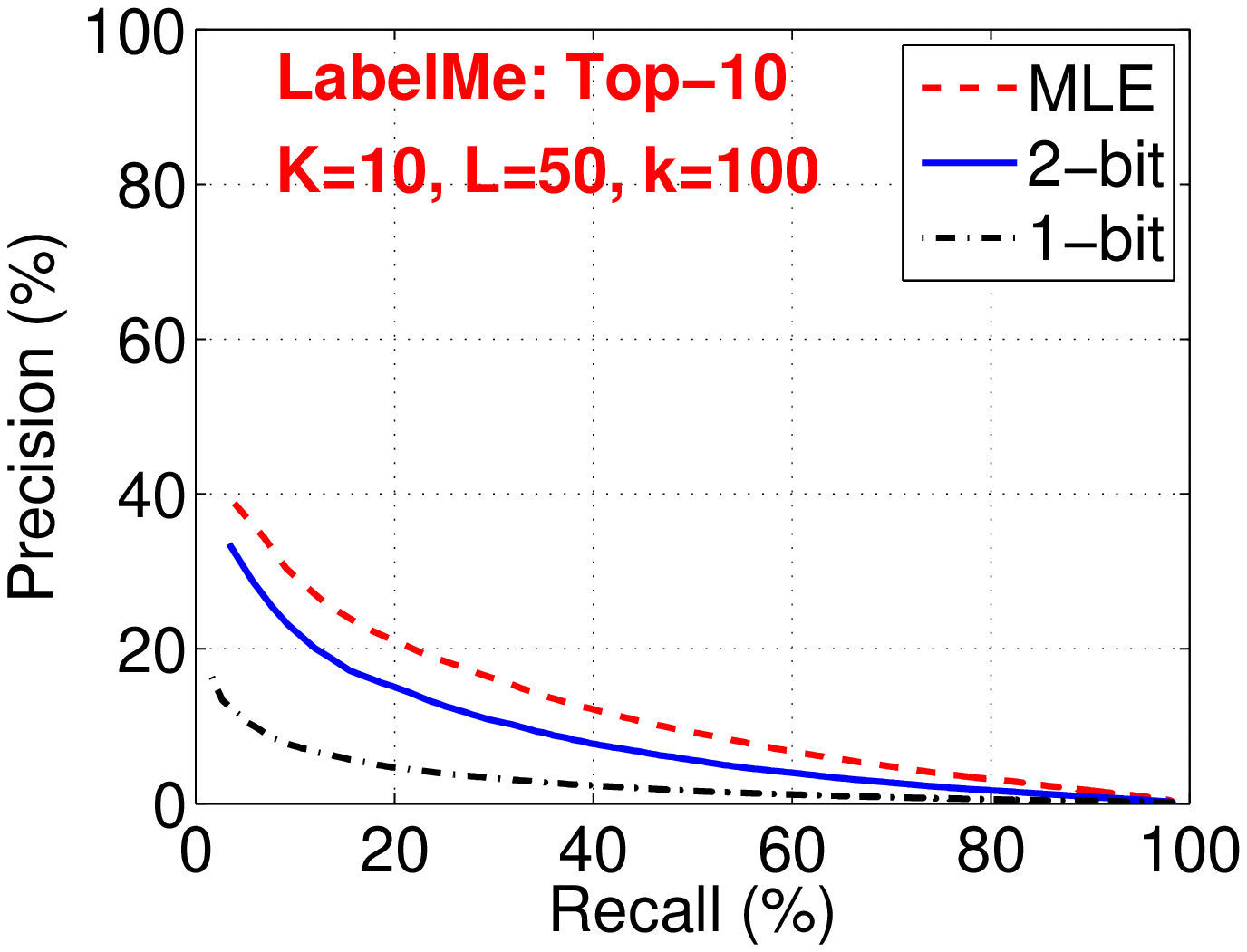}
\includegraphics[width = 1.4in]{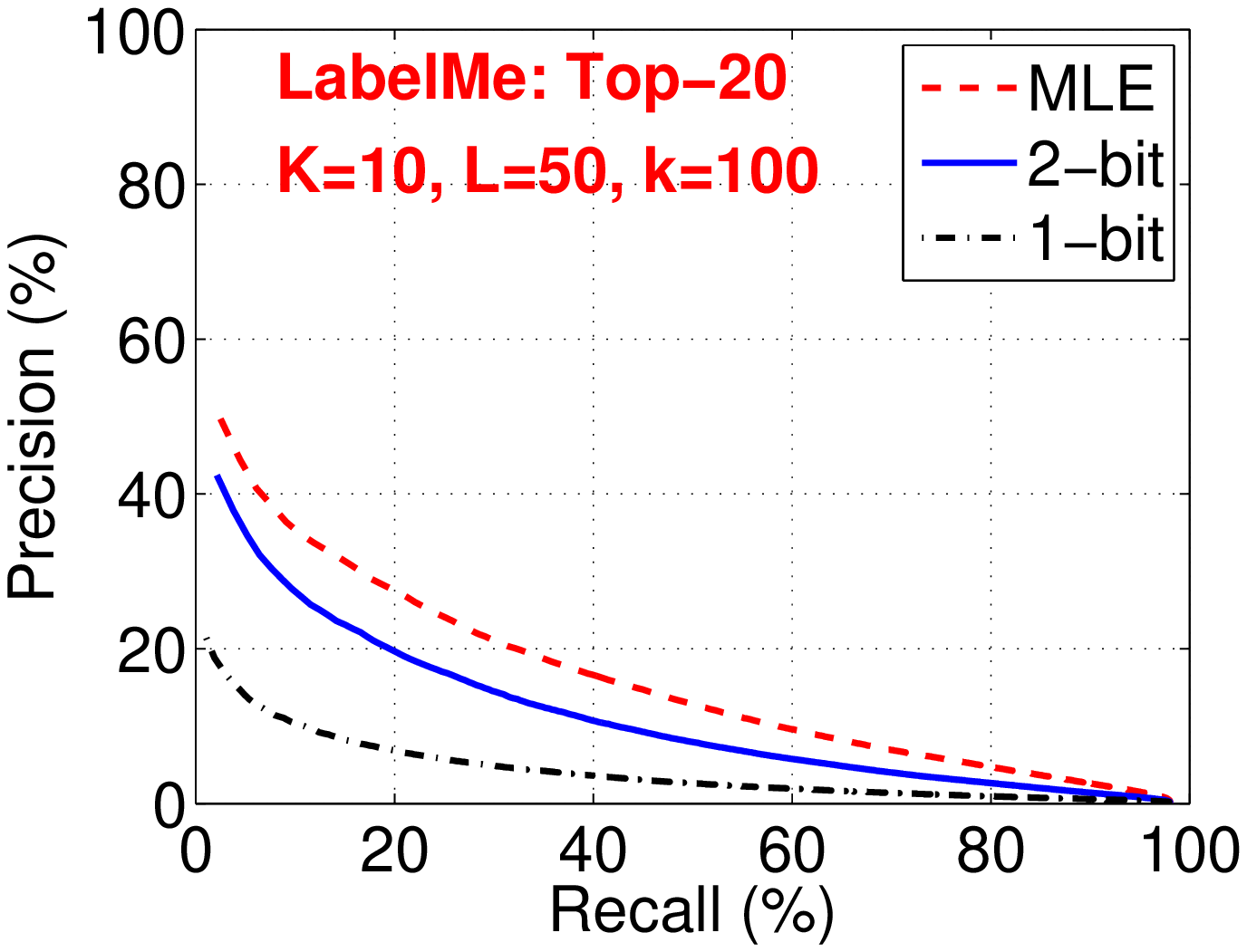}
\includegraphics[width = 1.4in]{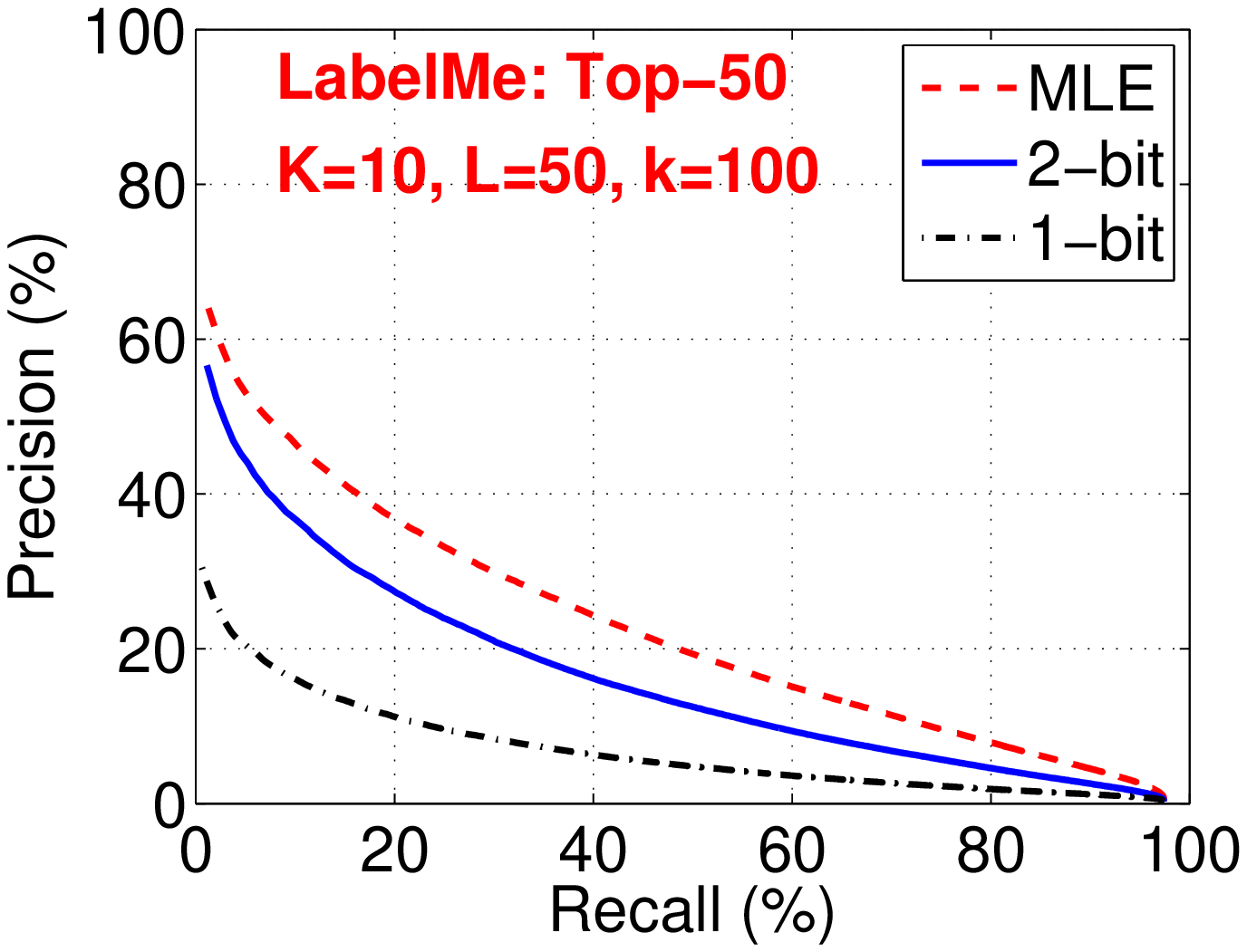}
\includegraphics[width = 1.4in]{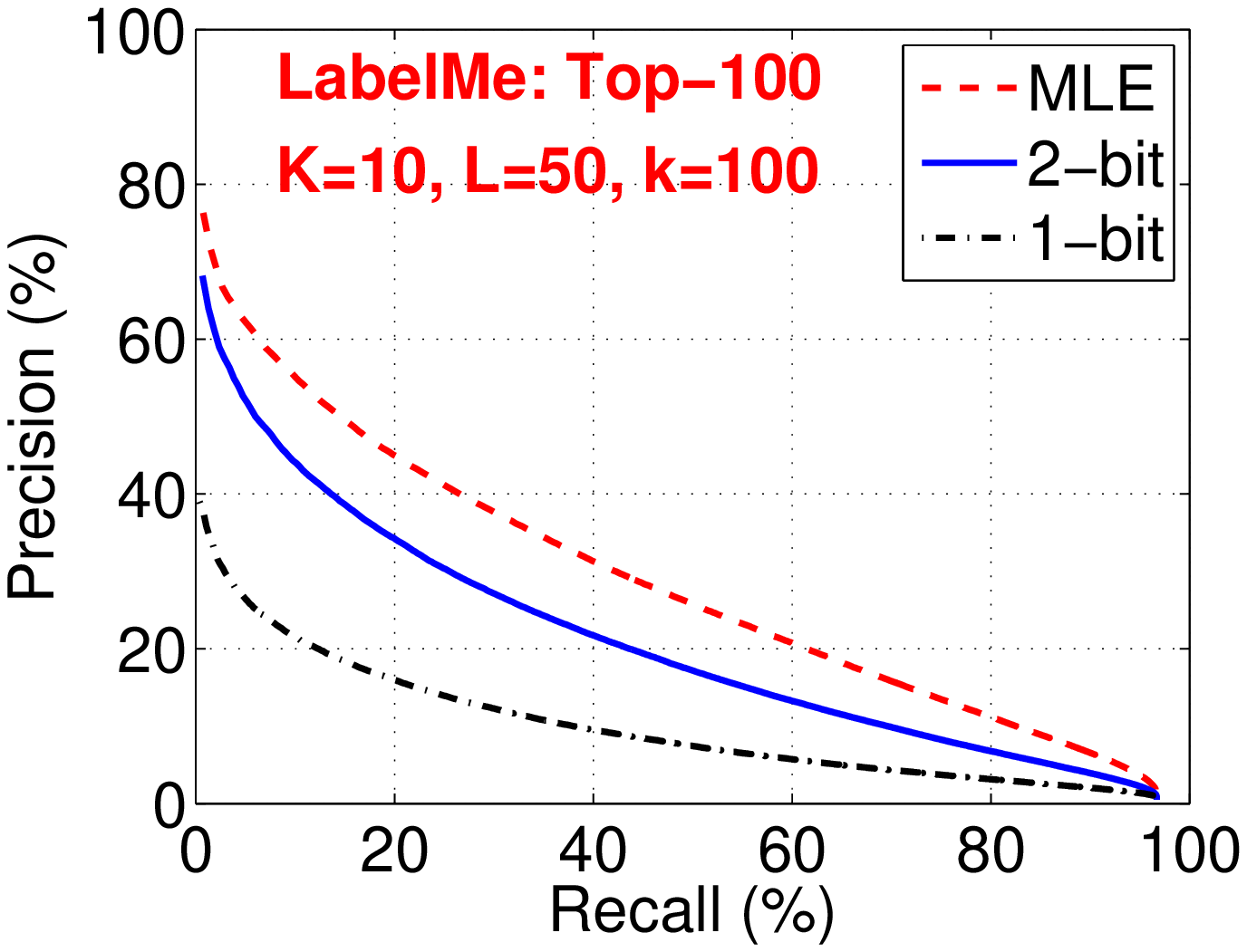}
}

\vspace{-0.025in}

\mbox{
\includegraphics[width = 1.4in]{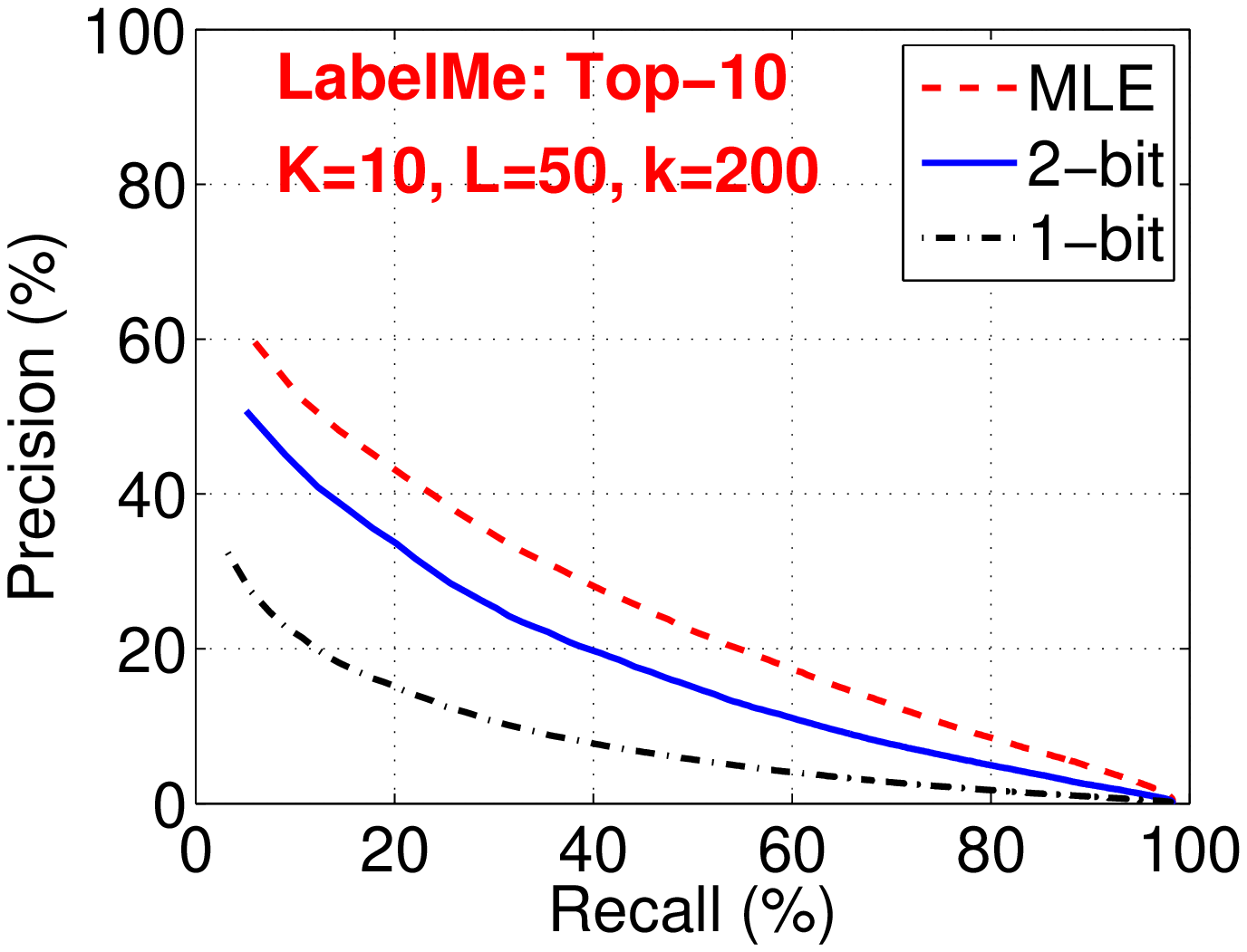}
\includegraphics[width = 1.4in]{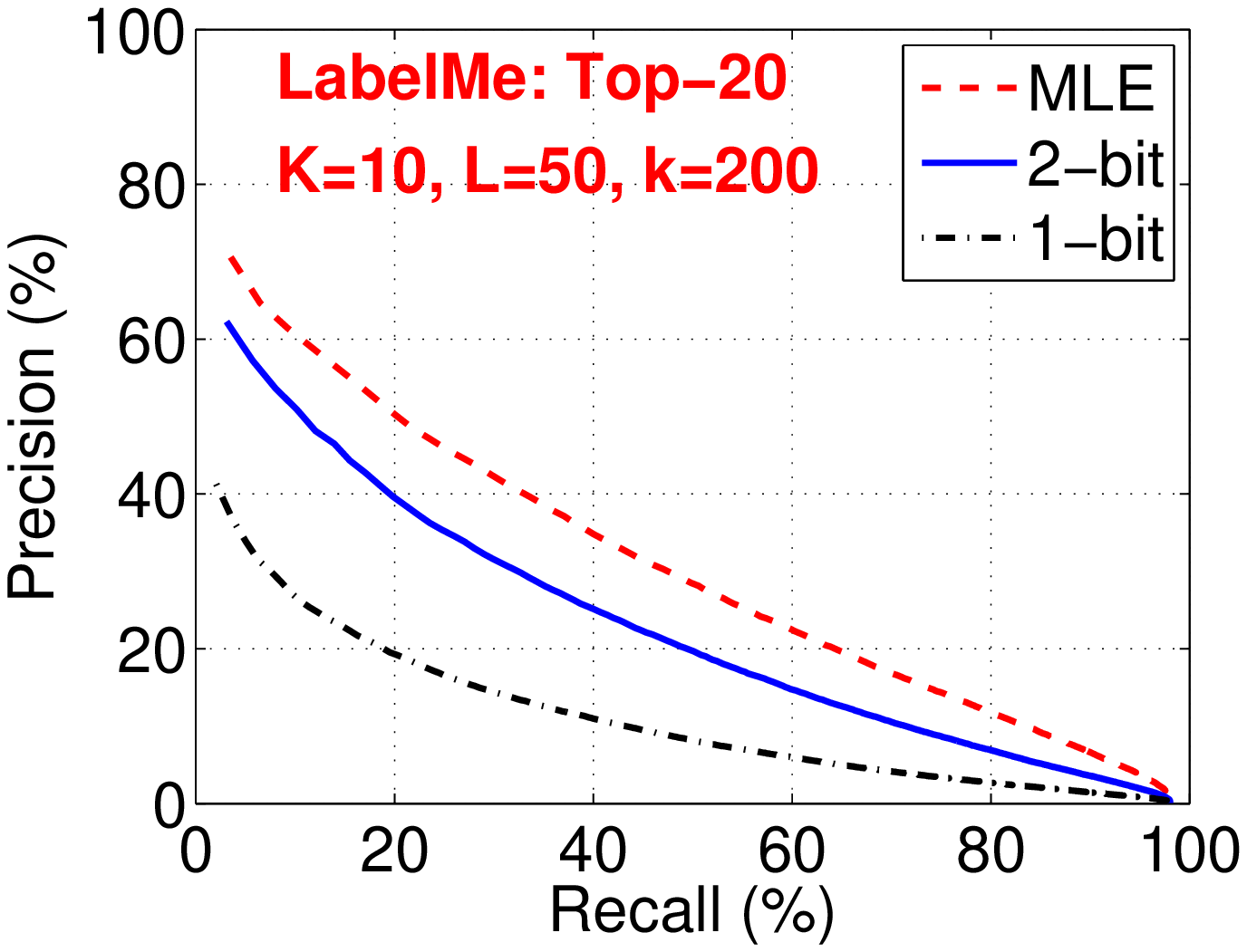}
\includegraphics[width = 1.4in]{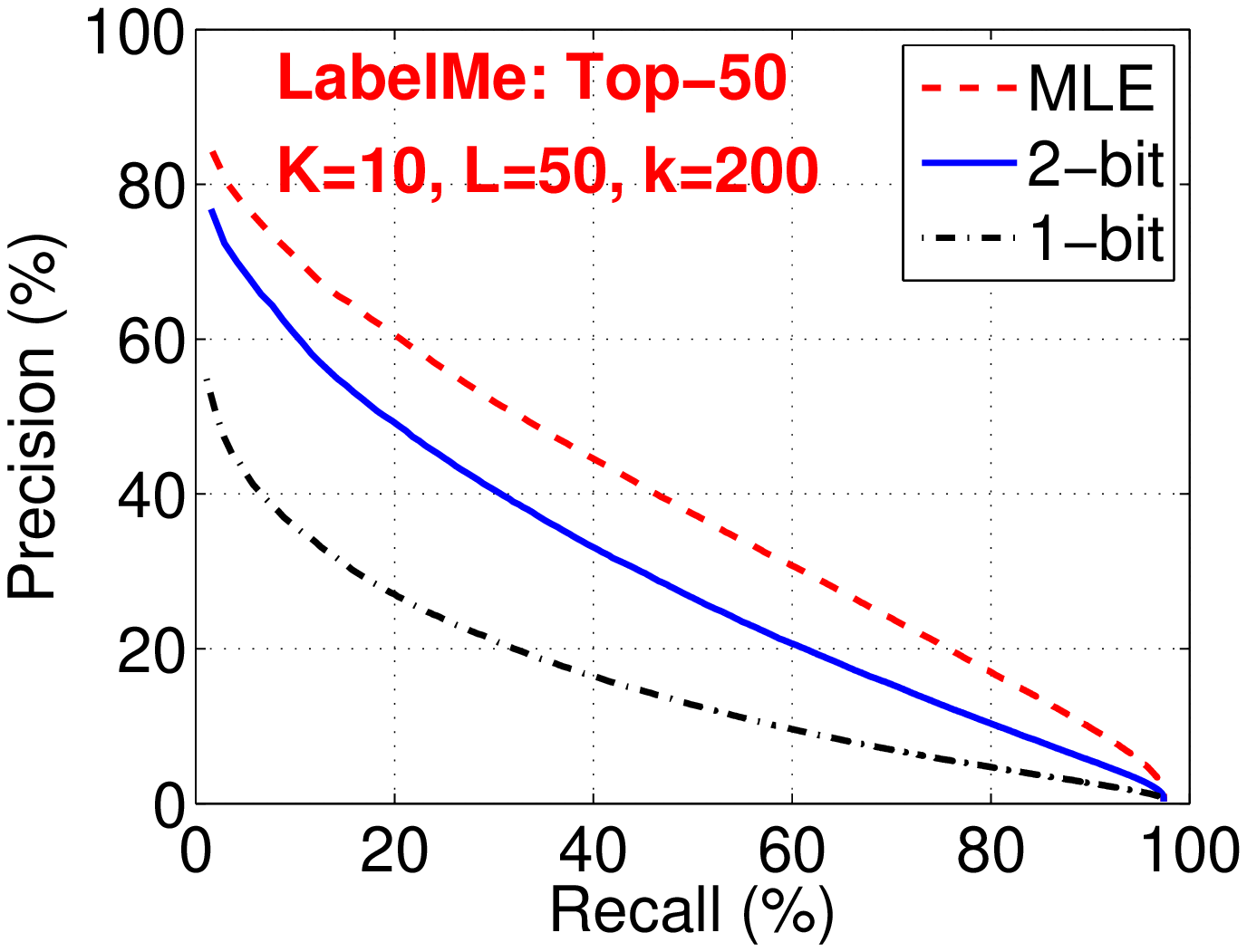}
\includegraphics[width = 1.4in]{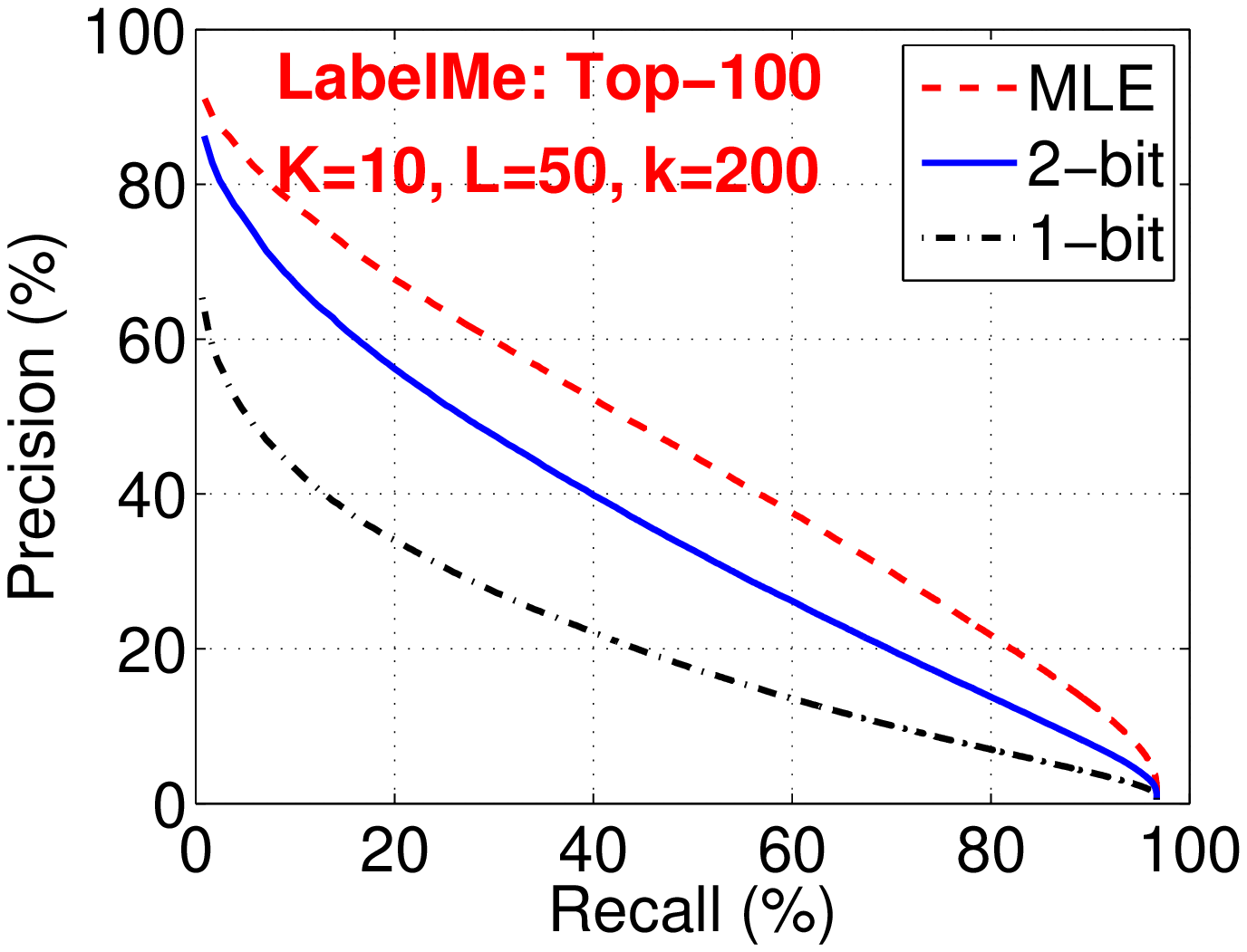}
}

\vspace{-0.025in}

\mbox
{
\includegraphics[width = 1.4in]{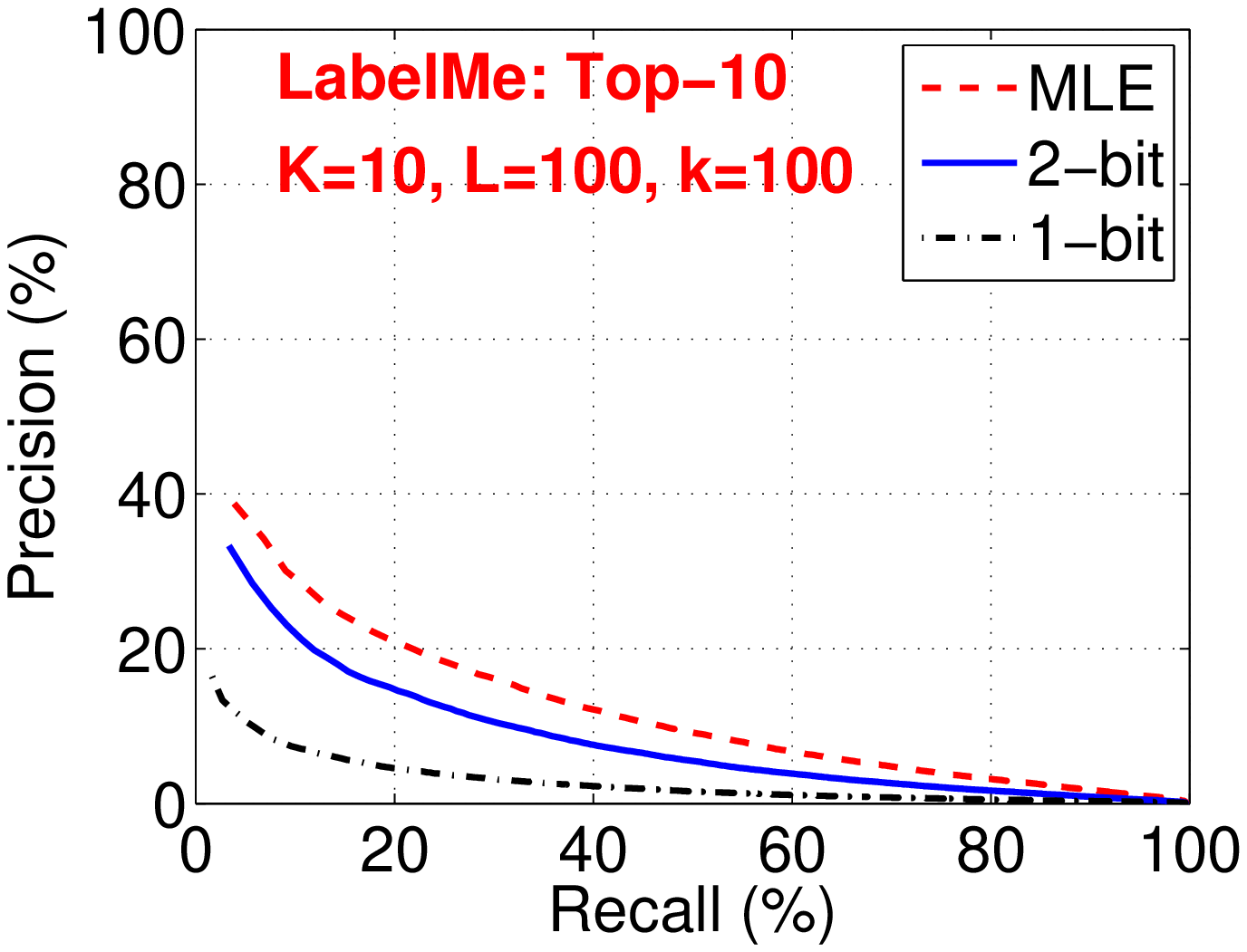}
\includegraphics[width = 1.4in]{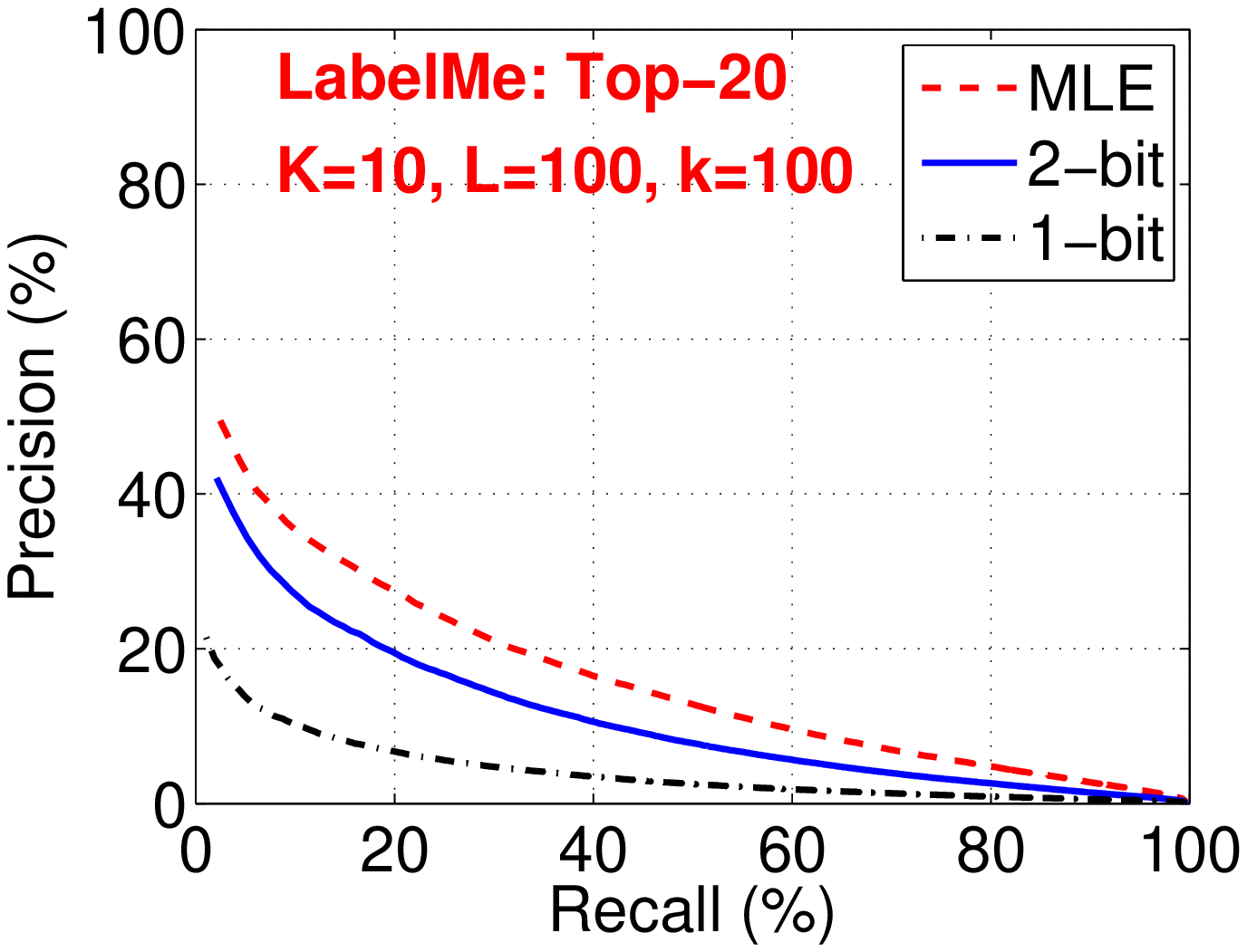}
\includegraphics[width = 1.4in]{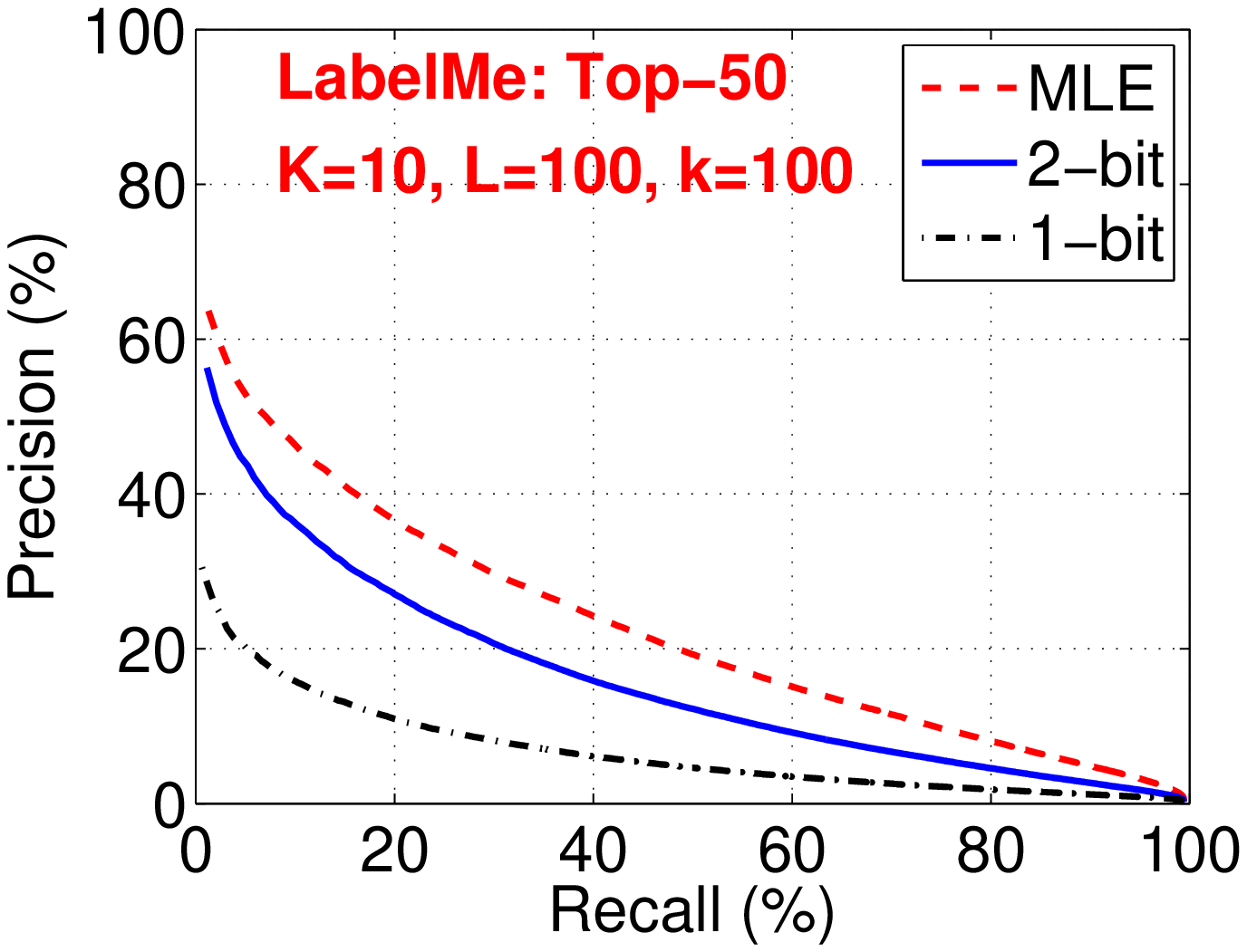}
\includegraphics[width = 1.4in]{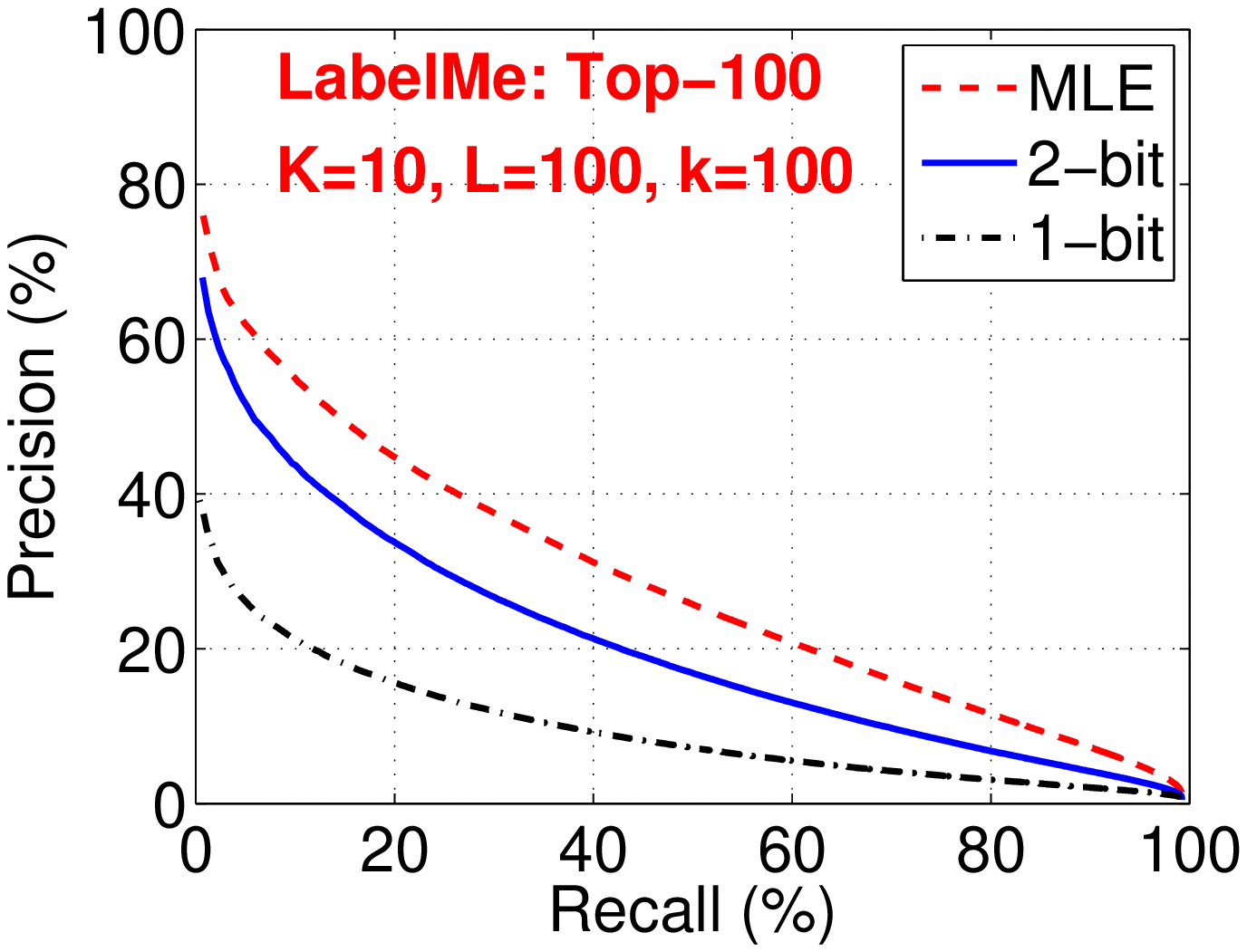}
}

\vspace{-0.025in}

\mbox{
\includegraphics[width = 1.4in]{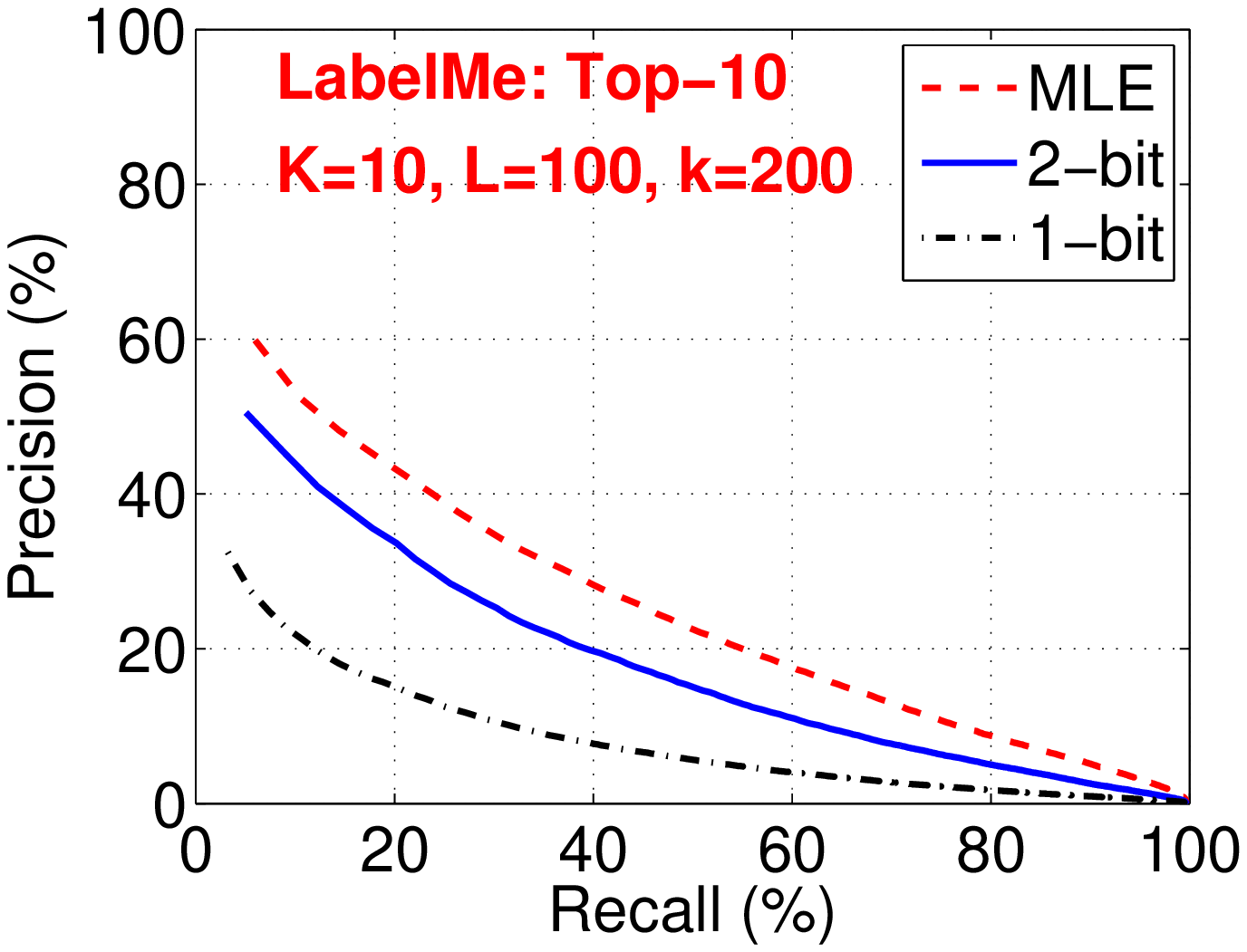}
\includegraphics[width = 1.4in]{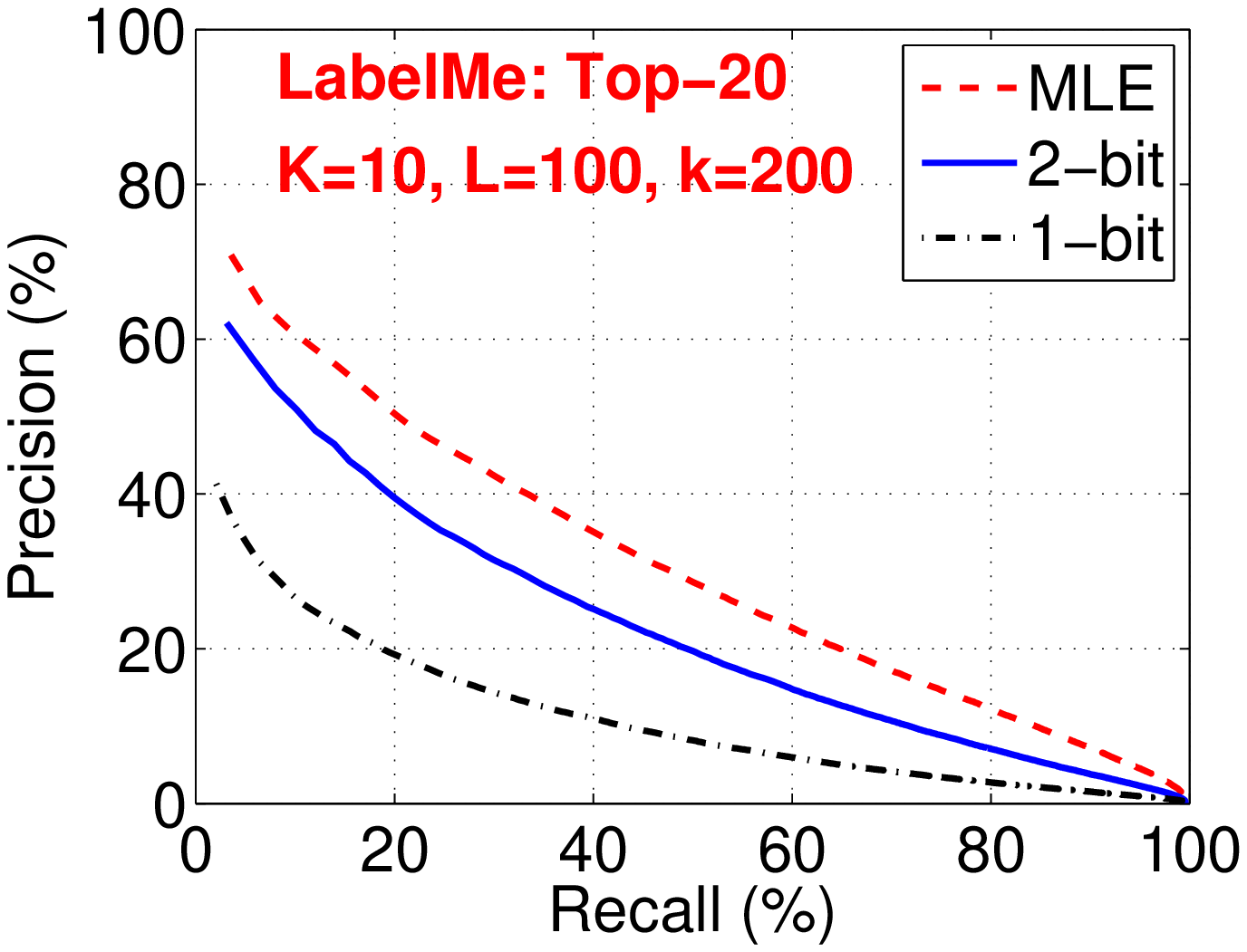}
\includegraphics[width = 1.4in]{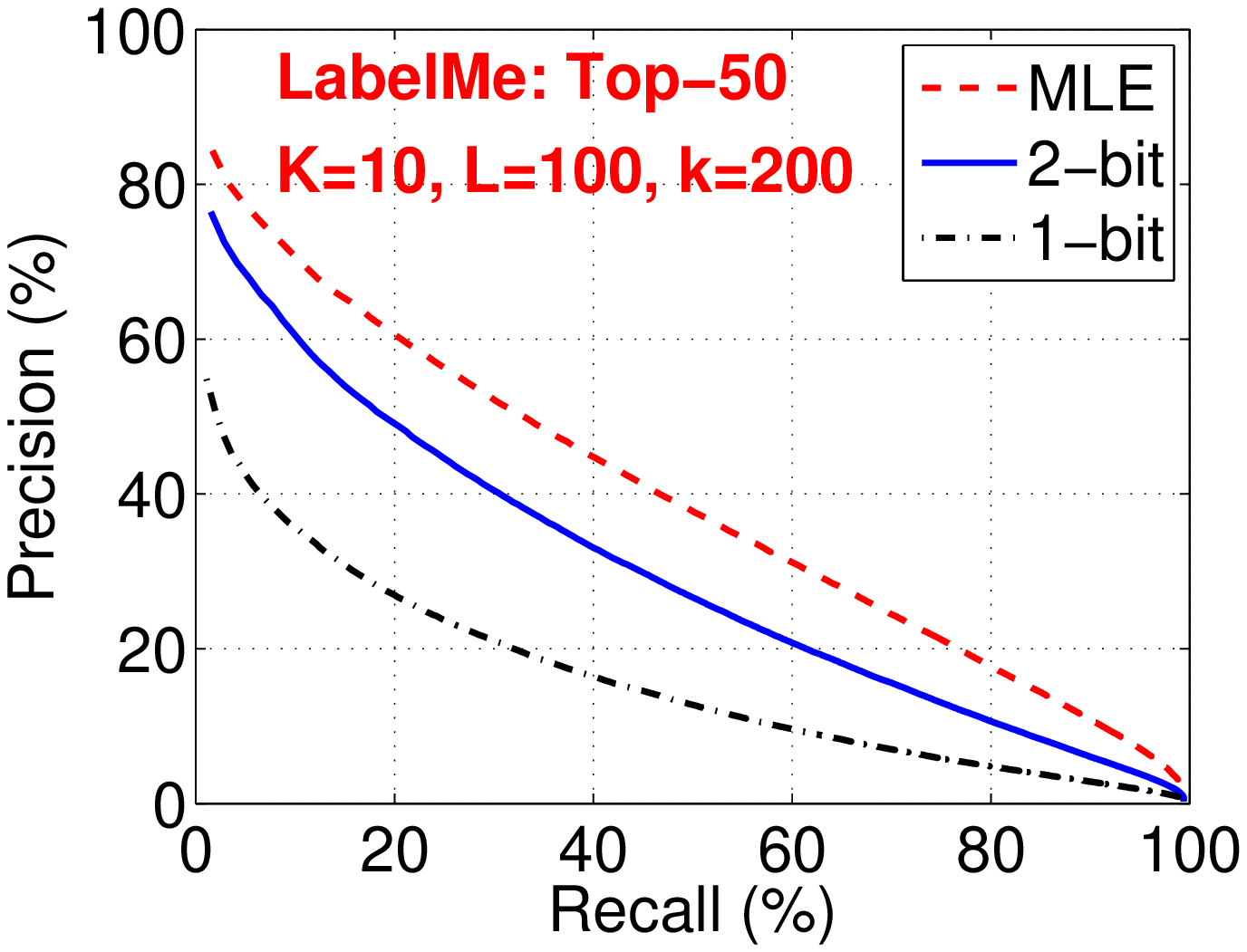}
\includegraphics[width = 1.4in]{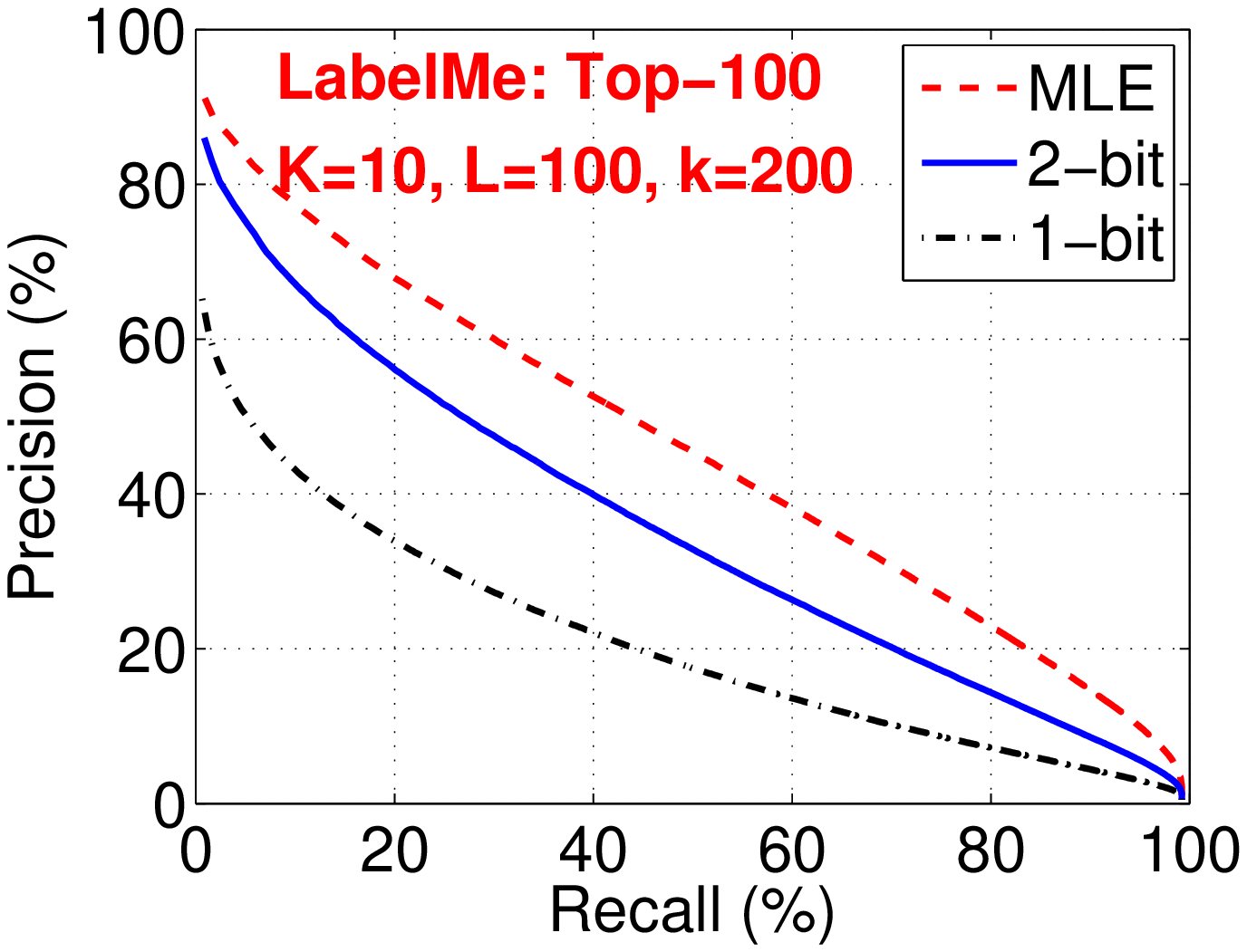}
}

\end{center}
\vspace{-0.25in}
\caption{\textbf{LabelMe}: precision-recall curves (higher is better.) for retrieving the top-10, -20, -50, -100 nearest neighbors using standard $(K,L)$-LSH scheme and 3 different estimators of similarities (for the retrieved data points).  LabelMe is a standard image retrieval dataset with 55,599 data points for building the tables and 1,998 data points for the query.
}\label{fig_LabelMeK10}\vspace{-0.15in}
\end{figure*}

As shown in Figure~\ref{fig_YoutubeK10}, Figure~\ref{fig_PeekaboomK10}, and Figure~\ref{fig_LabelMeK10}, in all our experiments, we see that the 2-bit MLE substantially improves the 2-bit linear estimator, which substantially improves the 1-bit estimator.

\clearpage\newpage

\newpage\clearpage
\section{Conclusion}
The method of random projections is a standard tool for many data processing applications which involve massive, high-dimensional datasets ( which are common in Web search and data mining). In the context of approximate near neighbor search by building hash tables, it is mandatary to quantize (code) the projected into integers. Prior to this work, there are two popular coding schemes: (i) an ``infinite-bit'' scheme~\cite{Proc:Datar_SCG04} by using uniform quantization with a random offset; and (ii) a ``1-bit'' scheme~\cite{Article:Goemans_JACM95,Proc:Charikar} by using the signs of the projected data. This paper  bridges these two strategies.\\

In this paper, we show that, for the purpose of building hash tables in the framework of LSH,  using uniform quantization   without the offset leads to improvement over the prior work~\cite{Proc:Datar_SCG04}. Our method only needs a small number of bits for coding each hashed value. Roughly speaking, when the target similarity is high (which is often interesting in practice), it is better to use 2 or 3 bits. But if the target similarity is not so high, 1 or 2 bits often suffice. Overall, we recommend the use of  a 2-bit scheme for LSH. Not surprisingly, as an additional benefit, using 2-bit scheme typically halves the preprocessing cost compared to using the 1-bit scheme.\\

For approximate near neighbor search, an important (and sometimes less well-discussed) step is the ``re-ranking'', which is needed in order to identify the truly similar data points among the large number of candidates retrieved from hash tables. This re-ranking step  requires a good estimator of the similarity, because storing the pre-computed all pairwise similarities is normally not feasible and computing the exact similarities on the fly can be time-consuming especially for high-dimensional data. In this paper, we propose the use of nonlinear estimators and we analyze the 2-bit case with details. Although the analysis appears sophisticated, the estimation procedure is computationally feasible and simple, for example, by tabulations. Compared to the standard 1-bit and 2-bit linear estimators, the proposed nonlinear estimator significantly improves the accuracy, both theoretically and empirically.\\

In summary, our paper advances the state-of-the-art of random projections in the context of approximate near neighbor search.

\vspace{-0.1in}
\bibliographystyle{plain}
\bibliography{../bib/IEEEabrv,../bib/mybibfile}

\end{document}